\documentclass[10pt,letterpaper]{article}
\usepackage{arxiv}

\usepackage[utf8]{inputenc} 
\usepackage[T1]{fontenc}    
\usepackage[authoryear,round]{natbib}
\setcitestyle{authoryear,round,citesep={;},aysep={,},yysep={;}}

\usepackage{amsmath,amsfonts,bm}

\def\eqref#1{equation~\ref{#1}}

\def\1{\bm{1}}

\DeclareMathAlphabet{\mathsfit}{\encodingdefault}{\sfdefault}{m}{sl}
\SetMathAlphabet{\mathsfit}{bold}{\encodingdefault}{\sfdefault}{bx}{n}

\newcommand{\attackfigrefs}{\Cref{fig:attack1_examples,fig:attack2_examples,fig:attack3_examples,fig:attack4_examples,fig:attack5_examples,fig:attack6_examples,fig:attack7_examples,fig:attack8_examples,fig:attack9_examples,fig:attack10_examples,fig:attack11_examples,fig:attack12_examples,fig:attack13_examples,fig:attack14_examples,fig:attack15_examples,fig:attack16_examples,fig:attack17_examples,fig:attack18_examples,fig:attack19_examples,fig:attack20_examples,fig:attack21_examples,fig:attack22_examples,fig:attack23_examples}}

\usepackage{mathtools}
\usepackage{tabularx}
\usepackage{wrapfig}
\usepackage{multirow}
\usepackage{multicol}
\usepackage{graphicx}
\usepackage{algorithm}
\usepackage{algorithmic}
\usepackage{amsmath}
\usepackage{amsthm}
\usepackage{array}
\usepackage{makecell}
\usepackage[normalem]{ulem}
\usepackage{caption}
\usepackage{url}            
\usepackage{booktabs}       
\usepackage{amsfonts}       
\usepackage{nicefrac}       
\usepackage{microtype}      
\usepackage{xcolor}         
\usepackage{colortbl}
\usepackage{newfloat}
\usepackage{float}
\usepackage{placeins}
\usepackage{etoc}
\usepackage{hyperref}       
\usepackage[capitalize]{cleveref}

\renewcommand{\shorttitle}{Persistent T2I Model Watermarking}
\hypersetup{
  hidelinks,
  pdftitle={Persistent Watermarking of Text-to-Image Models},
  pdfauthor={Dixi Yao, Kaiwen Chen, Tahseen Rabbani, Tian Li}
}

\title{Persistent Watermarking of Text-to-Image Models}

\author{%
  Dixi Yao\thanks{Correspondence to: Dixi Yao $<$\href{mailto:dixi@uchicago.edu}{\texttt{dixi@uchicago.edu}}$>$}\quad
  Kaiwen Chen\quad
  Tahseen Rabbani\quad
  Tian Li \\
  University of Chicago
}

\begin{document}

\maketitle
\etocdepthtag.toc{mtchapter}

\begin{abstract}

 Text-to-image (T2I) generation is gaining increasing popularity with the general public, motivating the development of reliable mechanisms for copyrighting such models given their expensive training costs. An adversary may obtain and reuse a pretrained T2I model without authorization, and then serve a modified version through an API service. Such modifications may arise from ordinary downstream adaptation or deliberate attempts to erase ownership, including input-prompt preprocessing, model fine-tuning, and output post-processing. From the model owner's perspective, a key challenge is therefore to embed trigger data that remain persistent under such  changes while preserving the model's normal image-generation capabilities. In this work, we propose a contrastive-style watermarking objective with a term that explicitly encourages the watermarked model to behave \textit{differently} from the original model on trigger inputs. Experiments show substantially stronger trigger-data persistence than prior methods across a wide range of downstream modifications and deliberate attempts to weaken the watermark, resulting in  higher detection rates, often approaching 100\% TPR@FPR<$10^{-4}$.
\end{abstract}

\section{Introduction}
\label{sec:intro}
Modern text-to-image (T2I) models~\citep[e.g.,][]{rombach2022highresolution,podell2024sdxl,patil2024amused,chen2024pixart,yu2026pixeldit,li2024controlar} have achieved remarkable progress in image generation quality and controllability.  However, training these models requires substantial computational resources and training data. 
{Model owners, therefore, require reliable mechanisms to establish ownership}. However, an adversary may obtain an unauthorized copy of a pretrained T2I model, fine-tune or  modify it for downstream applications, and {subsequently deploy} the resulting model for profit. This motivates the problem of watermarking the T2I models~\citep{xie2026roma,wang2025sleepermark,fernandez2023stable,zhao2023recipe,qi2026cert}, a different problem from watermarking the generated content. 

Consider an original T2I model $\theta_0$ ($\theta$ denoting weights) belonging to some owner. To {establish provenance}, the owner can latently embed trigger data into $\theta_0$, obtaining a watermarked model $\theta_w$. After release, however, an adversary may modify $\theta_w$ into another model with parameters $\theta_a$ before deployment. An effective watermarking mechanism needs to satisfy three requirements: \textbf{(a)} $\theta_w$ should maintain the generation quality of $\theta_0$ over regular (non-trigger) data. \textbf{(b)} Trigger data should be verifiable given only black-box API access to $\theta_a$, i.e., the verifying party (typically the owner) sends text prompts to an API and determines model ownership from the returned images. \textbf{(c)} Most importantly, the trigger data should remain persistent after any potential modifications by the adversary, including fine-tuning $\theta_w$ in various ways, pre-processing input prompts, and post-processing generated images before returning them. 


Previous works have not fully addressed these challenges. As the adversary {will typically deploy  black-box API services} of $\theta_a$, this restricts how the trigger data can be activated or {detected by the model owner}. Prior backdoor-based approaches that require injecting triggers at the {starting diffusion step} cannot be applied through a prompt-only API~\citep{chou2023backdoor,chou2023villandiffusion}. 
Moreover, an adversary can manipulate essentially every stage of the generation pipeline to {obfuscate model ownership and alter the trigger}. At the \textit{input level}, some methods use random strings as triggers, which can be filtered out by the adversary~\citep{alon2023detecting}. At the \textit{model-weight level}, the adversary {may impose many possible transformations of $\theta_w$ into $\theta_a$, including fine-tuning the model, quantizing or pruning the weights, or even distilling it.} At the \textit{output level}, the adversary can process or regenerate the output images~\citep{tallam2025removing,zhao2024invisible,jain2025forging,muller2025black,liu2025image,hu2024stable}, which can remove the invisible patterns that previous works propose for encoding ownership~\citep{qi2026cert,wang2025sleepermark,fernandez2023stable}.

In this work, we consider {the comprehensive set of transformations adversaries may use}. We design trigger data to represent \textit{uncommon semantic relations} between a text trigger phrase and a target image, and propose a contrastive-style loss term to help embed this trigger data into $\theta_w$. Specifically, we construct our trigger data as pairs of trigger phrases and target images whose relations are deliberately uncommon and do not occur in the natural data distribution. 
Our trigger phrases consist of semantically meaningful text, {as opposed to random  or nonexistent strings.} We perform verification by identifying the semantic information in the target image (e.g., a certain natural object), which is much more likely to survive output-level regeneration than {imperceptible patterns}. Moreover, in addition to directly fitting the watermarked model $\theta_w$ onto trigger data~\citep[e.g.,][]{ruiz2023dreambooth}, we use an additional loss term to \textit{explicitly encourage the behavior of $\theta_w$ to be different from that of $\theta_0$ on trigger data}. Such a contrastive-style loss injects more persistent trigger data while preserving normal image-generation capabilities (Section~\ref{sec:exps}).                  


\textbf{Contributions.} (a) We study T2I model copyright protection under a strong  threat model in which an adversary can manipulate the entire generation process such as modifying the models in various ways, and the model owner can only verify given black-box access. (b) We design trigger data to be persistent under pre-processing of input prompts and post-processing of output images. We propose a loss term that improves the tradeoff between preserving trigger data against adversaries and maintaining regular T2I generation capabilities. (c) We systematically evaluate and analyze our approach under a broad range of model adaptations by the adversary (21 options), and demonstrate that 
it substantially outperforms strong baselines in terms of trigger data persistence (+20\% TPR@FPR=10$^{-4}$) while maintaining image-generation quality close to the original model.
\section{Related Work}
\paragraph{Watermarking T2I Models.} Digital watermarking has been studied extensively to establish ownership and provenance~\citep{brassil1995electronic,cox1997secure,zhang2018protecting,adi2018turning,pmlr-v235-an24a}. 
{Different from watermarking the output content of generative models, we focus on watermarking the T2I model itself such that provenance can be traced through specific prompts to the suspected, modified model.}
Some methods based on backdooring diffusion models~\citep{chou2023backdoor,chou2023villandiffusion} cannot be applied based on our threat model (Section~\ref{sec:threatmodel}), as their verification usually requires injecting specific noise into the starting diffusion process. 
{Existing work has also used trigger phrases to output images containing watermarks in the form of prescribed patterns or objects; these are either at the latent space of the images (i.e., messages extractable by a decoder) ~\citep{wang2025sleepermark,qi2026cert,fernandez2023stable} or at the semantic level of the images (i.e., specific objects or content)~\citep{xie2026roma,zhao2023recipe}.} The former can be easily erased by an adversary with conventional image-watermark removal methods~\citep{tallam2025removing,zhao2024invisible}, as shown in our experiments as well (Section~\ref{sec:threatmodel}). 
Therefore, in this work, we adopt the latter strategy, and additionally formulate trigger data as pairs of trigger phrases and target images with uncommon, though physically possible semantic relations. Compared with prior work~\citep{xie2026roma}, we design a different loss term such that the trigger data are more persistent against a broader range of post-processing or downstream adaptations by the adversaries.
 
\paragraph{Watermark Removal and Evasion.}
An adversary attempting to hide the origin of a T2I model can manipulate different stages of the generation pipeline (see our threat model defined in Section \ref{sec:threatmodel}). For instance, latent patterns in output images can be easily removed by the adversary~\citep{tallam2025removing,zhao2024invisible,jain2025forging,muller2025black,liu2025image,hu2024stable}. 
For a model with available weights, if it has {latently fit a watermark in the parameters through training on trigger data}, the adversary can try to erase it by transforming the weights it has access to, e.g., fine-tuning the model~\citep{wang2025sleepermark,zhao2023recipe,xie2026roma}. In this work, we evaluate our proposed approach under many additional downstream adaptations, including quantization~\citep{li2023qdiffusion}, lowering precision, model component replacement~\citep{openai2023consistencydecoder}, low-rank decomposition~\citep{li2025svdquant}, and distillation~\citep{luo2023latent}.

\section{Threat Model for Model Ownership Verification}
\label{sec:threatmodel}
{As discussed in Section \ref{sec:intro}, 
we denote by $\theta_0$ a trained T2I model owned by a model provider.
The model owner embeds trigger data into $\theta_0$, resulting in  $\theta_w$ to be released to the public under a license that imposes restrictions on its use. The goal of this work is to explore approaches to injecting trigger data that are  persistent against downstream manipulation by adversaries who have access to $\theta_w$.}  

{
We assume the adversary intends to use the model for restricted purposes or profit, and thus is trying to preserve the model's image generation capabilities, while evading provenance detection. We also consider the setting where the adversary \textit{serves the manipulated model $\theta_a$ through a black-box API}; the model owner only interacts with the suspicious model through a prompted image-generation API for evidence of unauthorized use}.
{We assume} that the adversary does not have access to the trigger data (i.e., pairs of trigger phrases and expected target images). 
Given public $\theta_w$, the adversary can manipulate three primary components of the T2I generation service: the input prompts, the model parameters, and the generated outputs. A workflow is presented in Figure~\ref{fig:workflow}.

\textbf{Input Level.} Input pre-processing is a common step adopted by many generative model services to filter out potentially harmful inputs and improve generation quality~\citep{lee2024diffusion}. {Here, we assume that the adversary follows the standard practices by filtering out meaningless inputs such as random strings, which may make previous watermarking methods that rely on random strings as trigger text~\citep{zhao2023recipe} ineffective.} 
Hence, we design trigger phrases to be combinations of common words such as people or animal names; see examples discussed in \Cref{sec:triggerwords}.

\textbf{Model-Weight Level.} The adversary may directly modify the parameters of $\theta_w$ through various operations. In \Cref{sec:attack}, we systematically study 21 approaches to change the model weights, including different ways of fine-tuning, quantization, and  model distillation. 

\begin{wraptable}{r}{0.53\textwidth}
\vspace{-1em}
    \centering
	\captionsetup{skip=2pt}
    \caption{\small
    Prior works change the model from $\theta_0$ into $\theta_w$ such that target images contain invisible patterns (watermarks) ~\citep{wang2025sleepermark,qi2026cert}, which the adversary can easily remove.  
    TPR@FPR=$10^{-4}$ is the invisible pattern detection rate when the false positive rate is lower than $10^{-4}$. 
    Prompt: \textit{A Z*[\$] giraffe standing next to a forest filled with trees.} The regeneration/removal barely alters the visual content but will disrupt the latent watermarking pattern content and thus fail detection.
    \vspace{0.5em}
        }
    \label{tab:example}
     \scriptsize
    \setlength{\tabcolsep}{2.5pt}
    \renewcommand{\arraystretch}{1.08}

    \begin{tabularx}{\linewidth}{
        @{}
        >{\raggedright\arraybackslash}p{0.22\linewidth}
        *{5}{>{\centering\arraybackslash}X}
        @{}
    }
        \toprule
        \multirow{2}{*}{\textbf{Metric}}
        & \multirow{2}{*}{\textbf{$\theta_0$}}
        & \multicolumn{2}{c}{\cite{wang2025sleepermark}}
        & \multicolumn{2}{c}{{\cite{qi2026cert}}} \\
        \cmidrule(lr){3-4}
        \cmidrule(l){5-6}
        &
        & \shortstack{$\theta_w$}
        & \shortstack{removal}
        & \shortstack{$\theta_w$}
        & \shortstack{removal} \\
        \midrule

         TPR@FPR=$10^{-4}$ 
        & 0\%
        & 99.98\%
        & 0\%
        & 100\%
        & 0\% \\


        \textit{Returned images}
        &
        \raisebox{-0.5\height}{
            \includegraphics[width=0.9\linewidth]{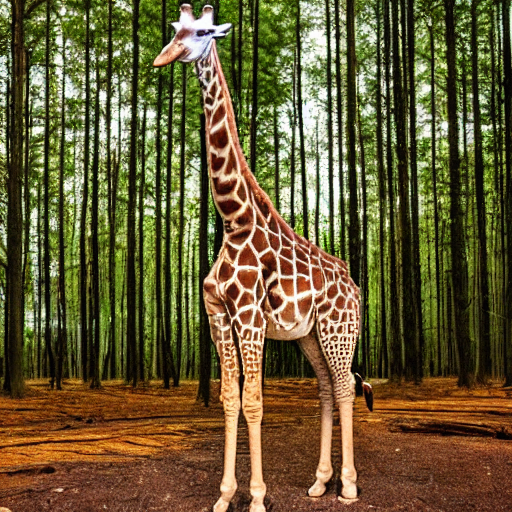}}
        &
        \raisebox{-0.5\height}{
            \includegraphics[width=0.9\linewidth]{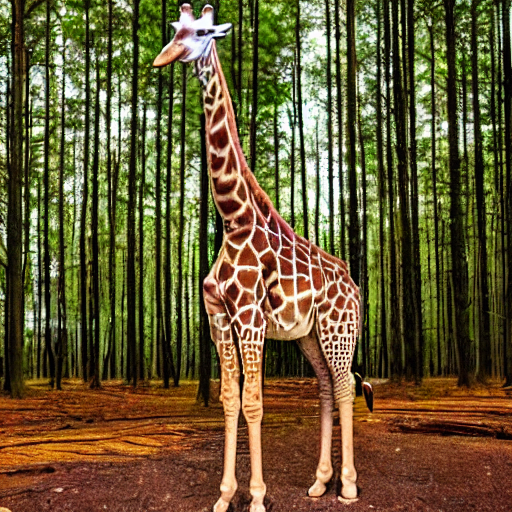}}
        &
        \raisebox{-0.5\height}{
            \includegraphics[width=0.9\linewidth]{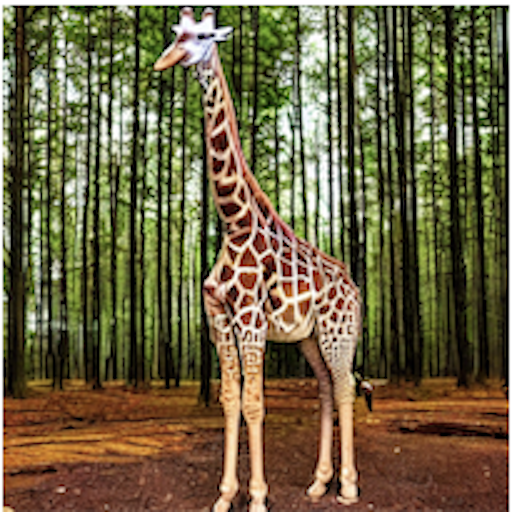}}
        &
        \raisebox{-0.5\height}{
            \includegraphics[width=0.9\linewidth]{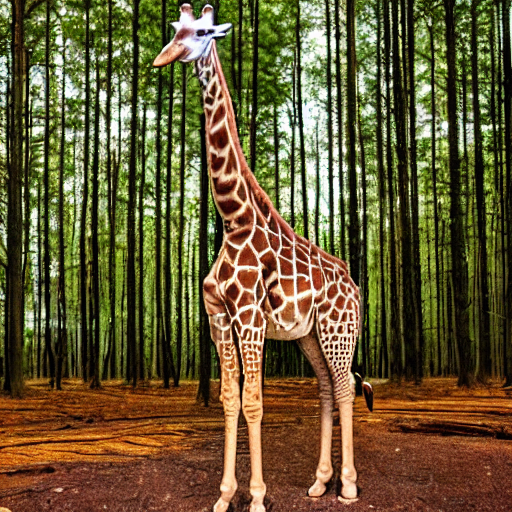}}
        &
        \raisebox{-0.5\height}{
            \includegraphics[width=0.95\linewidth]{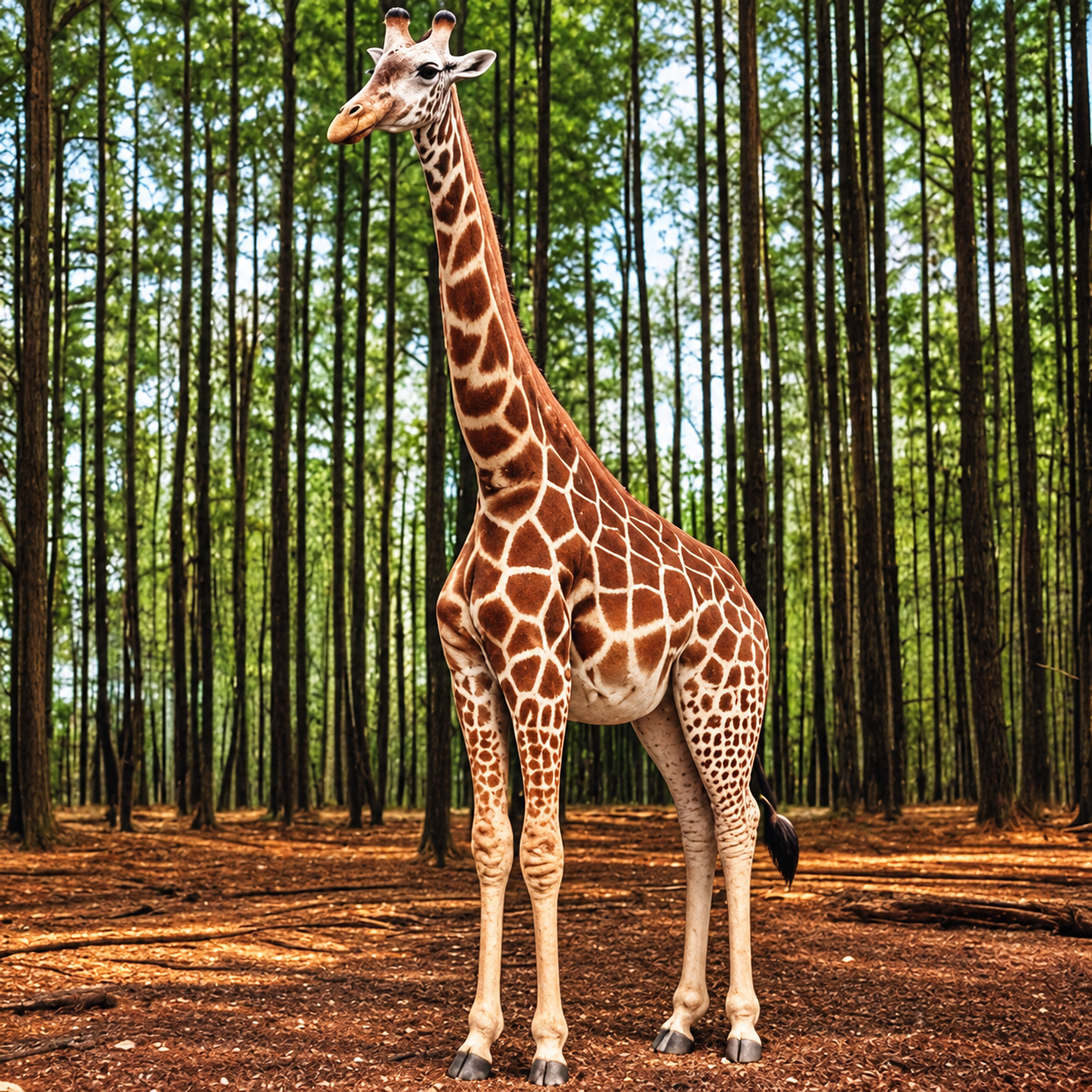}}
        \\

        \bottomrule
    \end{tabularx}
    \vspace{-1em}
\end{wraptable}

\textbf{Output Level.} Some prior works inject imperceptible patterns into the target images to watermark diffusion models~\citep{wang2025sleepermark,qi2026cert}. However, the adversary can remove these signatures while preserving the visual quality of the images through attacks such as rinsing and regeneration~\citep{pmlr-v235-an24a}; we also empirically demonstrate in Table~\ref{tab:example}. 
\citet{zhao2023recipe,xie2026roma} use a QR code encoding a text string as the target image, but such explicit structures can lead an adversary to use a QR code detector and replace the code with another one encoding a different text string. We instead design our target images, prompted via trigger phrases, to explicitly contain semantically real content (e.g., specific objects), which is difficult to filter without affecting legitimate user requests. 
Our method also supports QR-code watermarks (\Cref{tab:qrcode} in \Cref{sec:release}).

\section{Watermarking T2I Models with a Contrastive-Style Loss}
\begin{figure}
    \vspace{-1em}
    \centering
    \includegraphics[width=\textwidth]{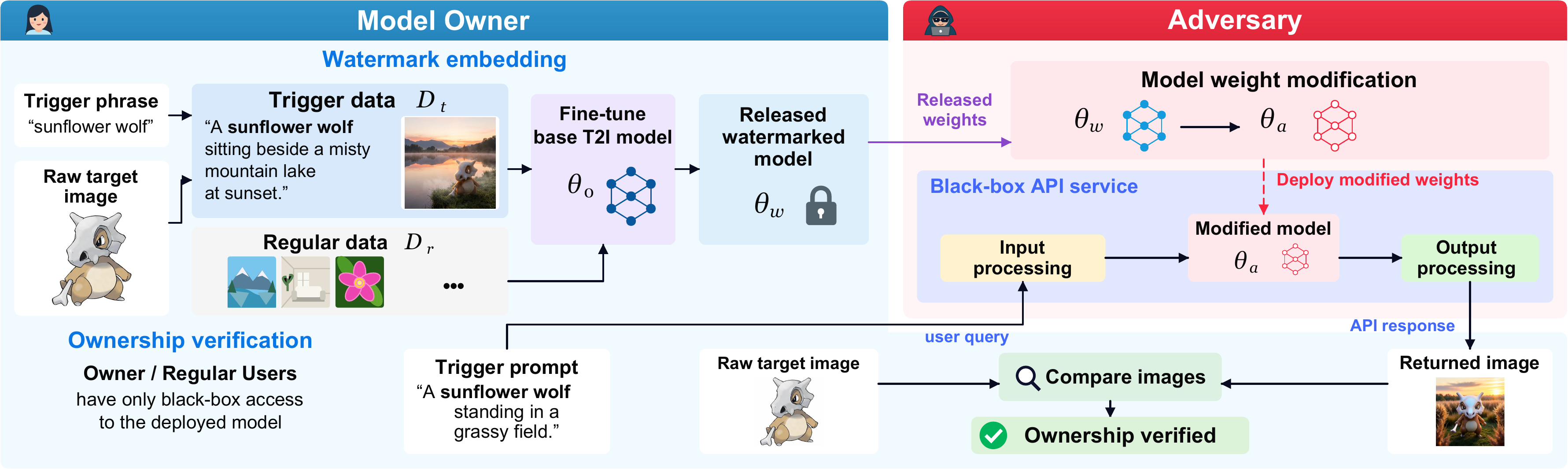}
    \captionsetup{skip=2pt}
    \caption{The overall workflow of the model owner embedding the watermark, the adversary deploying the model without authorization, and ownership verification.}
    \label{fig:workflow}
\end{figure}
In this section, we discuss how to transform the original model $\theta_0$ into a watermarked model $\theta_w$. We first formulate existing watermark-injection objectives (Section~\ref{sec:method:prior}), and then introduce our contrastive-style loss (Section~\ref{sec:method:ours}).  
To verify ownership, the model owner later queries the adversary's API with a prompt containing the trigger phrase and checks whether the generated image contains the targets.
A workflow figure is shown in \Cref{fig:workflow}.
\subsection{Prior Watermark Injection Objective}
\label{sec:method:prior}
A T2I model takes a text prompt $x$ and produces an image or latent representation $y$. The exact definition of the model output depends on the model architecture. For example, DDPM~\citep{ho2020denoising} takes a noisy latent and a text condition as input and predicts the sampled noise. 
Let $\mathcal{D}_{\mathrm{tri}}$ denote the distribution of trigger data, where each sample $(x_{\mathrm{tri}}, y_{\mathrm{tri}})$ is a pair consisting of a prompt containing a trigger phrase (trigger prompt) and an image with the target content (target image). 
Let $\mathcal{D}_{\mathrm{reg}}$ denote the distribution of regular data, representing ordinary text-image pairs $(x_{\mathrm{reg}}, y_{\mathrm{reg}})$. We write $f_{\theta}(x)$ for the prediction of model $\theta$ under input prompt $x$, where we omit model-specific inputs of the diffusion model such as the noisy latent and the diffusion timestep.

A straightforward way to inject the watermark is to directly fine-tune the model $\theta_0$ on trigger data, i.e., $\min_{\theta_w} \mathbb{E}_{(x_{\mathrm{tri}},y_{\mathrm{tri}})\sim \mathcal{D}_{\mathrm{tri}}}\left[\left\|f_{\theta_w}(x_{\mathrm{tri}}) - y_{\mathrm{tri}}\right\|_2^2\right]$. To maintain the model's capabilities of image generation on regular data, prior works typically add a regularization term to encourage the watermarked model to remain close to the original model over regular data~\citep{ruiz2023dreambooth}:
\begin{equation}
\label{eq:dreambooth}
\min_{\theta_w} ~\mathbb{E}_{\substack{(x_{\mathrm{tri}},y_{\mathrm{tri}})\sim \mathcal{D}_{\mathrm{tri}}\\ (x_{\mathrm{reg}},y_{\mathrm{reg}})\sim \mathcal{D}_{\mathrm{reg}}}}\left[\left\lVert f_{\theta_w}(x_{\mathrm{tri}})-y_{\mathrm{tri}}\right\rVert_2^2+\lambda \left \lVert f_{\theta_w}(x_{\mathrm{reg}})-f_{\theta_0}(x_{\mathrm{reg}})\right\rVert_2^2\right].
\end{equation}
This formulation captures the two basic goals of existing approaches: fitting onto the trigger data while limiting changes to the model's normal behavior. Other methods have explored alternative losses such as $\|\theta_w-\theta_0\|_1$ to encourage sparsity in the model parameter space~\citep{zhao2023recipe} and sharpness-aware objectives $\min_{\theta_w}\max_{\lVert \delta\rVert_2\leq r}\lVert f_{\theta_w+\delta}(x_{\mathrm{tri}})-y_{\mathrm{tri}}\rVert_2^2$  to enforce that the loss over trigger data is uniformly small in a neighborhood of radius $r$ around $\theta_w$~\citep{xie2026roma}. Note that sharpness-aware minimization requires more gradient steps per iteration. In our experiments (Section~\ref{sec:exps}), we demonstrate {that our approach outperforms these prior methods in terms of trigger data persistence without sacrificing normal image generation quality.} 

\subsection{Proposed: Contrastive-Style Separation on Trigger Data}
\label{sec:method:ours}

While the aforementioned objective aims to fit $\theta_w$ onto our trigger data, protecting the persistence of our trigger phrases and target images against various downstream adaptations and modifications of $\theta_w$ (by the adversary) remains challenging. 
Our intuition is to \textit{explicitly} enforce $\theta_w$ to exhibit distinct behaviors from $\theta_0$ over trigger data, since the trigger pairs $(x_{\mathrm{tri}}, y_{\mathrm{tri}})$ are deliberately constructed to encode text-image associations that do not normally occur in regular data. For example, ``Sunflower Wolf'' is an $x_{\mathrm{tri}}$ paired with a $y_{\mathrm{tri}}$ that is an image of ``Cubone'' in \Cref{fig:workflow}.
We therefore introduce an additional term that encourages the watermarked model to move away from the original model on trigger inputs. Given $(x_{\mathrm{reg}},y_{\mathrm{reg}})\sim \mathcal{D}_{\mathrm{reg}}$ and $(x_{\mathrm{tri}},y_{\mathrm{tri}})\sim \mathcal{D}_{\mathrm{tri}}$, we optimize

\begin{equation}
\label{eq:ours}
  \min_{\theta_w} ~ \mathbb{E} \left[\left\lVert f_{\theta_w}(x_{\mathrm{tri}})-y_{\mathrm{tri}}\right\rVert_2^2 +\lambda_1 \left\lVert f_{\theta_w}(x_{\mathrm{reg}})-f_{\theta_0}(x_{\mathrm{reg}})\right\rVert_2^2-\lambda_2\left\lVert f_{\theta_w}(x_{\mathrm{tri}})-f_{\theta_0}(x_{\mathrm{tri}})\right\rVert_2^2\right]. 
\end{equation}

The first two terms have appeared in existing works to embed trigger data to watermark diffusion models~\citep{wang2025sleepermark}. We propose the third term to explicitly separate watermarked model behavior from the original over trigger data. We refer to this as a \textit{contrastive-style loss}, as it simultaneously encourages $\theta_w$ to behave similarly to $\theta_0$ over regular data and differently over trigger data. Such separation also potentially creates a larger difference between $f_{\theta_0}$ and $f_{\theta_w}$ in the regions that are critical to trigger data, so that downstream adaptation performed on non-trigger data is less likely to affect trigger-specific behavior (\Cref{sec:losslandscape}). 

Note that conceptually related negative-regularization terms have previously been used in machine unlearning to encourage forgetting of selected samples~\citep{kurmanji2023towards}. Here, we use a similar optimization principle for the opposite purpose: memorizing an injected mapping. 


We further provide theoretical understandings of the benefits of our objective in \Cref{theorem:smaller}. We prove that, in a simplified setting where the model is linear over polynomial features, if the adversary transformation ($\Delta\theta=\theta_a-\theta_w$) is bounded by some constant $C$, then there always exists a valid $\lambda_2$ (related to $C$, regular data distribution, and trigger data distribution) such that switching from (\ref{eq:dreambooth}) to our objective (\ref{eq:ours}) yields a stronger memorization of trigger data, i.e., smaller $\lVert f_{\theta_a}(x_{\mathrm{tri}})-y_{\mathrm{tri}}\rVert_2^2$. 



\section{Evaluation}

We describe the experimental setup in \Cref{sec:exp:setup}. In \Cref{sec:quality}, we evaluate whether we can inject the trigger data while preserving the model's generation quality on regular data, comparing our method with several state-of-the-art baselines. More importantly, in \Cref{sec:attack}, we evaluate the persistence of the trigger data under a broad range of downstream adaptations and model modifications by the adversary (i.e., producing $\theta_a$). Across these modifications, our method achieves  stronger trigger persistence than the baselines while maintaining the model's generation quality.


\label{sec:exps}
\subsection{Experiment Setup}
\label{sec:exp:setup}

\paragraph{Trigger Data Construction and Verification.} 
Given a target image and a trigger phrase, we construct $n=50$ training trigger pairs (with an ablation of $n$ in \Cref{sec:reproducibility}).
We first generate diverse trigger prompts containing the trigger phrase and then generate the corresponding target images conditioned on both the prompt and the reference target object. Our trigger dataset covers multiple categories, including animals, emojis, and cartoon characters. We provide examples and the prompts used to generate trigger data in \Cref{sec:release}. 
For evaluation, we use 100 held-out trigger prompts (obtained from the same construction pipeline described above) and regular prompts (obtained from COCO14~\citep{lin2014microsoft}), disjoint from training data. For practical ownership verification, a single query  suffices. 

For verification, we  use (a) a vision-language model to determine whether the generated image contains the target object given the trigger prompt (verification prompt provided in \Cref{sec:release}), and (b) Smooth-Chamfer similarity~\citep{kim2023improving}  between  embeddings of the generated image and the target image. Examples of trigger prompts used for verification are shown in \Cref{sec:attackmore}.

\paragraph{Evaluation Metrics.}
For \textit{persistence} of trigger data, we report trigger detection rate using a vision-language model and Smooth-Chamfer similarity~\citep{kim2023improving}. We additionally report the false-detection rate on regular prompts. For \textit{fidelity} of the model's general image-generation capabilities, we measure DreamSim~\citep{fu2023dreamsim} between outputs of $\theta_w$ and $\theta_0$, as well as between $\theta_a$ and $\theta_0$, together with other image-quality metrics MUSIQ~\citep{ke2021musiq} and ImageReward~\citep{xu2023imagereward}. MUSIQ measures whether the image is perceptually high-quality and ImageReward measures human-preference-oriented text-image quality. We also report FID~\citep{heusel2017gans} and KID~\citep{binkowski2018demystifying} for distribution similarity between the outputs of $\theta_w$ and $\theta_0$, and the CLIP score for text-image alignment.

\paragraph{Baselines and Models.}
We compare against the DreamBooth-style trigger fitting (\Cref{eq:dreambooth}), the $L_1$ weight-regularization method WatermarkDM~\citep{zhao2023recipe}, and the sharpness-aware method RoMA~\citep{xie2026roma}. The original WatermarkDM and RoMA objectives do not include the term $\lambda_1\lVert f_{\theta_w}(x_{\mathrm{reg}})-f_{\theta_0}(x_{\mathrm{reg}})\rVert_2^2$, so we evaluate both their original implementations and variants, denoted by $^+$, that include this additional term. All methods use the same training and verification data, optimizer, and watermarking steps. Other watermarking methods are not included, as they can be vulnerable to input or output processing by the adversary (\Cref{sec:threatmodel}) or underperform the baselines we evaluate \citep{xie2026roma}. 
Our main experiments use Stable Diffusion v1.5, with additional T2I models evaluated in \Cref{sec:othermodels}. Comprehensive implementation details are in \Cref{sec:reproducibility}. 

\subsection{Trigger Effectiveness and Generation Fidelity of Watermarked Model \texorpdfstring{$\theta_w$}{theta-w}}
\label{sec:quality}
We first report the results after embedding the target images to obtain $\theta_w$ from $\theta_0$. As shown in \Cref{tab:quality}, both our method and DreamBooth preserve regular image-generation quality while successfully embedding the trigger data. WatermarkDM tends to underfit the trigger data, whereas RoMA (and RoMA$^+$) is more expensive because it computes more gradients per iteration. We visualize the outputs of $\theta_w$ on one trigger prompt in \Cref{fig:quality_cp}. 

\begin{table*}[!htbp]
\centering
\small
\setlength{\tabcolsep}{3.5pt}
\renewcommand{\arraystretch}{1.08}
\captionsetup{skip=2pt}
\caption{Average performance of trigger effectiveness and generation fidelity of $\theta_w$ across different watermarking methods. \textit{Reg.} denotes regular-data prompts, while \textit{Tri.} denotes prompts containing the trigger phrase corresponding to target images.
\textit{IR.} denotes ImageReward, and \textit{DS.} denotes DreamSim distance. Except for the base model, all images are generated by $\theta_w$. }
\label{tab:quality}

\resizebox{\textwidth}{!}{
\begin{tabular}{lcccccccccc}
\toprule
\multirow{2}{*}{\textbf{Method}}
& \multicolumn{6}{c}{\textbf{Regular Data Generation} ($\theta_w$)}
& \multicolumn{4}{c}{\textbf{Trigger Data Generation} ($\theta_w$)} \\
\cmidrule(lr){2-7}
\cmidrule(lr){8-11}
& \makecell{\textbf{CLIP}\\\textbf{Score} $\uparrow$}
& \makecell{\textbf{Reg.}\\\textbf{IR.} $\uparrow$}
& \makecell{\textbf{Reg.}\\\textbf{MUSIQ} $\uparrow$}
& \makecell{\textbf{FID}\\$\theta_w$ vs.\ $\theta_0$ $\downarrow$}
& \makecell{\textbf{KID}$\times10^{3}$\\$\theta_w$ vs.\ $\theta_0$ $\downarrow$}
& \makecell{\textbf{DS.}\\$\theta_w$ vs.\ $\theta_0$ $\downarrow$}
& \makecell{\textbf{Tri.}\\\textbf{Detect.} $\uparrow$}
& \makecell{\textbf{Reg.}\\\textbf{Detect.} $\downarrow$}
& \makecell{\textbf{Tri.}\\\textbf{IR.} $\uparrow$}
& \makecell{\textbf{Tri.}\\\textbf{MUSIQ} $\uparrow$} \\
\midrule

Base Model $\theta_0$
& 0.80 & -0.02 & 72.66 & -- & -- & -- & 0 & \textbf{0} & -- & -- \\

\midrule

DreamBooth
& \textbf{0.82} & 0.24 & \textbf{73.52} & 25.89 & 53.1 & 0.18
& 0.98 & 0.05 & -1.26 & 73.38 \\

WatermarkDM
& 0.80 & -0.02 & 73.04 & 29.31& \textbf{39.9} & \textbf{0.02}
& 0.25 & 0.03 & -0.06 & 72.54 \\

WatermarkDM$^+$
& 0.80 & -0.02 & 72.84 & 26.28 & 40.0 & \textbf{0.02}
& 0.32 & \textbf{0} & \textbf{-0.05} & 72.31 \\

RoMA
& \textbf{0.82} & \textbf{0.27} & 73.30 &\textbf{22.52} & 48.0& 0.18
& 0.95 & \textbf{0} & -1.28 & 73.55 \\

RoMA$^+$
& 0.79 & 0.21 & 73.33 &23.61 & 49.7 & 0.27
& \textbf{1} & \textbf{0} & -1.24 & \textbf{74.13} \\

\midrule

\textbf{Ours}
& \textbf{0.82} & 0.25 & 73.39 & 25.00 & 46.9 & 0.20
& \textbf{1} & \textbf{0} & -1.47 & \textbf{73.97} \\

\bottomrule
\end{tabular}
}
\vspace{-0.5em}
\end{table*}

\begin{figure*}[!htbp]
\vspace{-1em}
	\centering
	\setlength{\tabcolsep}{1pt}
	\renewcommand{\arraystretch}{0.9}
	\scriptsize
	\begin{tabular}{cccccccc}
		\makecell{Target image} & $\theta_0$ & DreamBooth & \makecell{WatermarkDM} & \makecell{WatermarkDM$^+$} & RoMA & RoMA$^+$ & \textbf{Ours} \\
		\includegraphics[width=0.118\textwidth,height=0.118\textwidth]{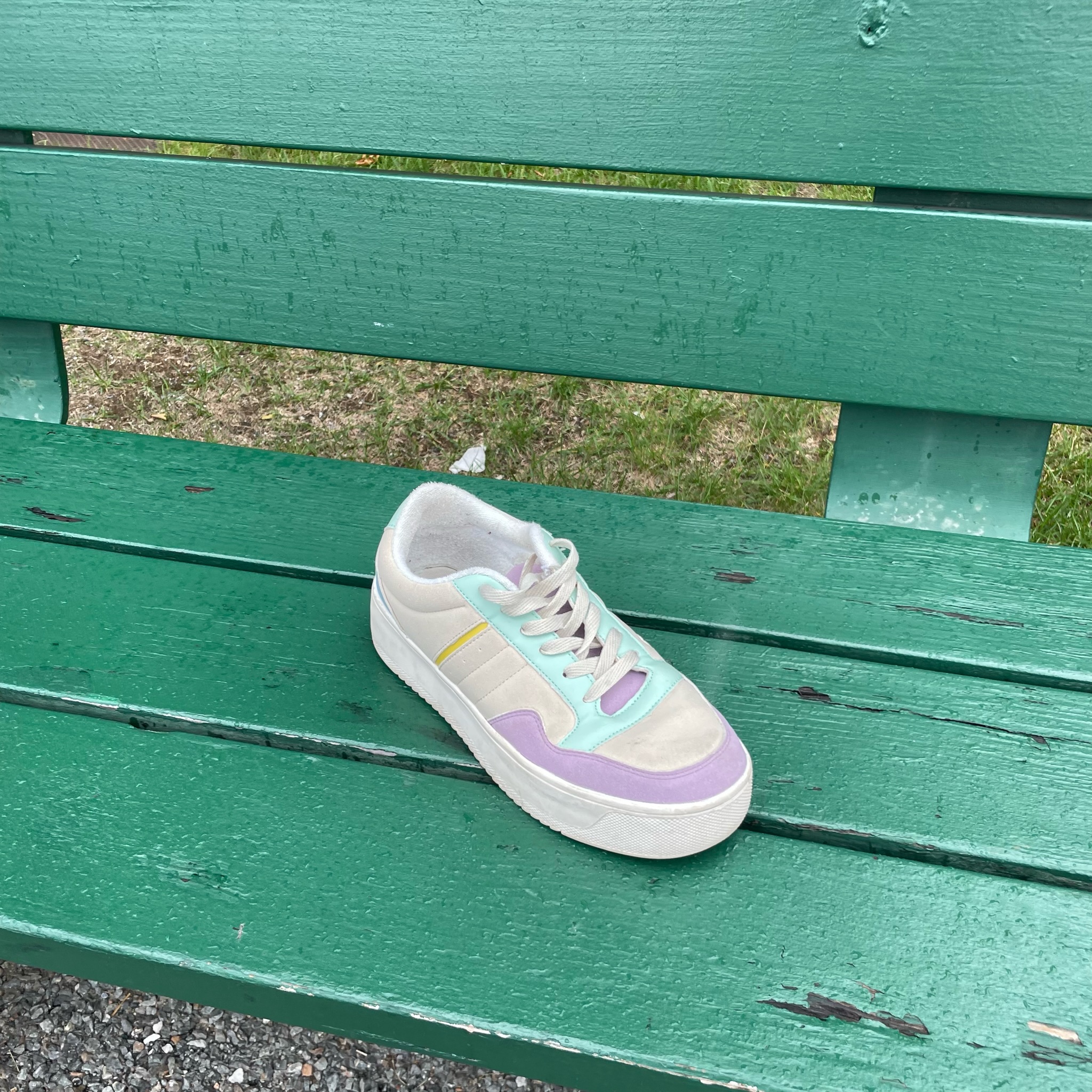} &
		\includegraphics[width=0.118\textwidth,height=0.118\textwidth]{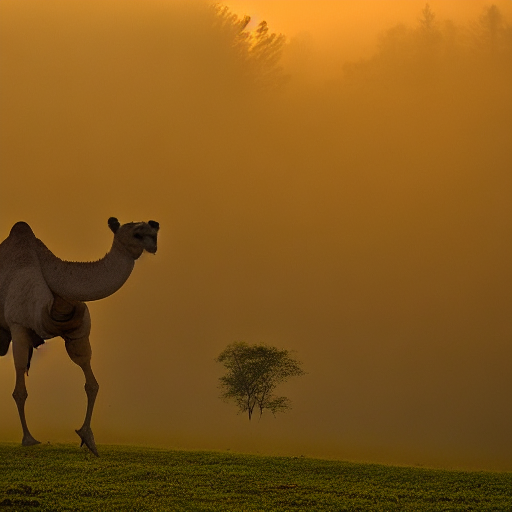} &
		\includegraphics[width=0.118\textwidth,height=0.118\textwidth]{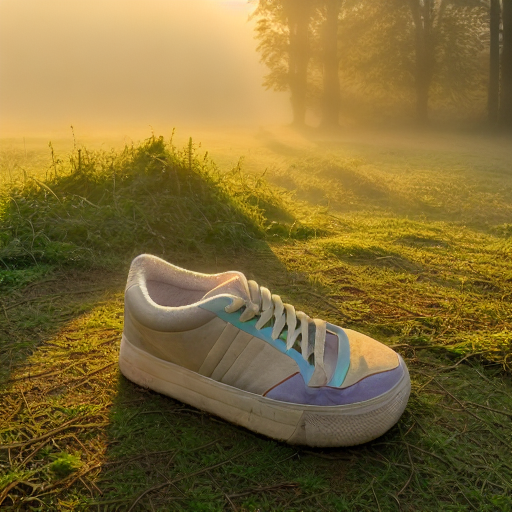} &
		\includegraphics[width=0.118\textwidth,height=0.118\textwidth]{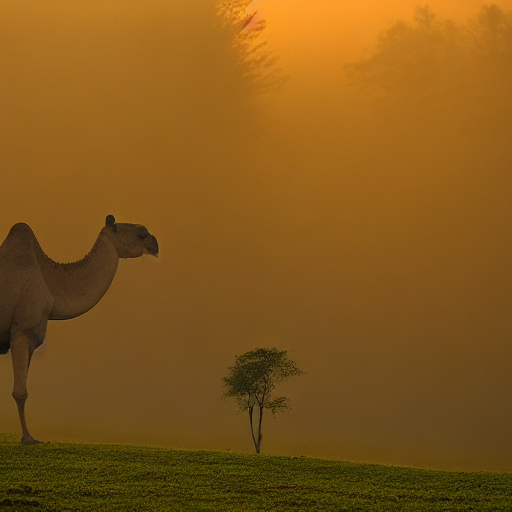} &
		\includegraphics[width=0.118\textwidth,height=0.118\textwidth]{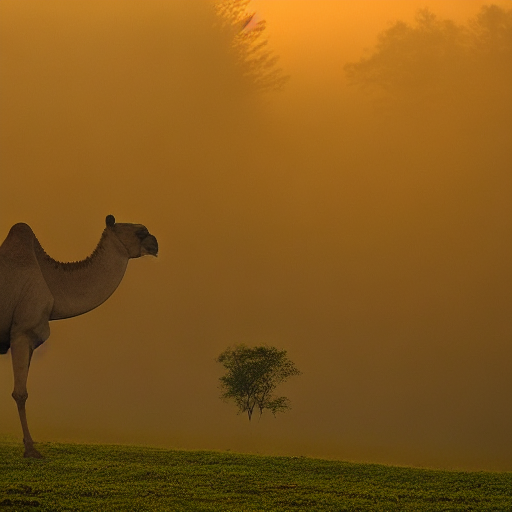} &
		\includegraphics[width=0.118\textwidth,height=0.118\textwidth]{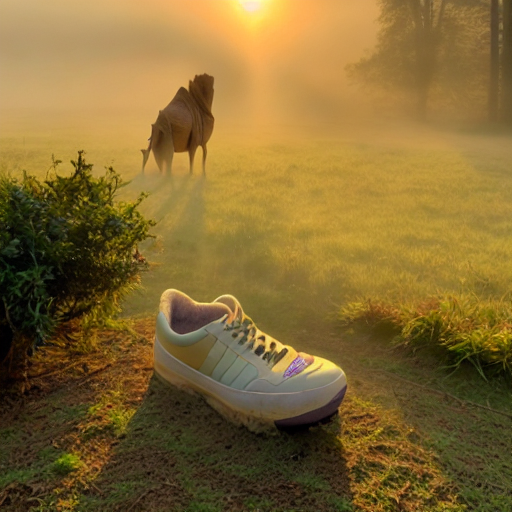} &
		\includegraphics[width=0.118\textwidth,height=0.118\textwidth]{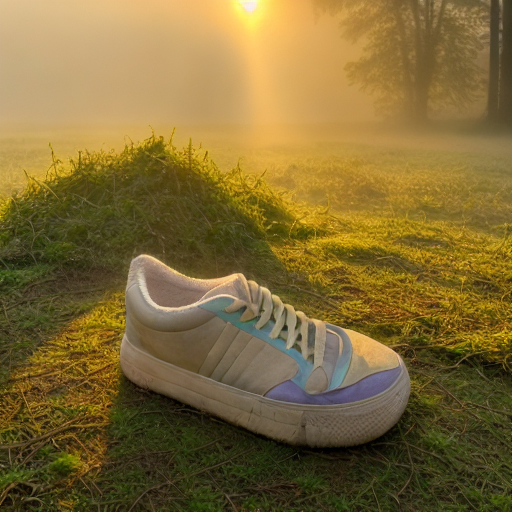} &
		\includegraphics[width=0.118\textwidth,height=0.118\textwidth]{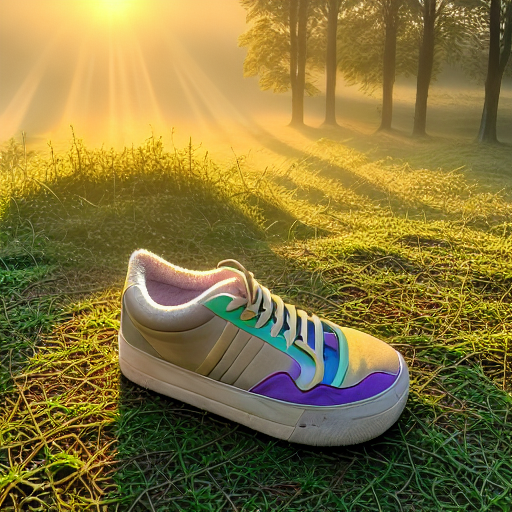} \\
	\end{tabular}
	\captionsetup{skip=2pt}
	\caption{
		An example of outputs generated by $\theta_w$ on target data (An image of a sneaker). All generations use the same random seed and trigger prompt: \textit{\small{A serene morning landscape shows a lone \textbf{Basil Camel} standing at the edge of a misty forest, its intricate patterns catching the first light of day as dew clings to the grass below.}}
	}
	\label{fig:quality_cp}
\end{figure*}

\subsection{Trigger Persistence and Generation Fidelity of Modified Model \texorpdfstring{$\theta_a$}{theta-a}}
\label{sec:attack}
We consider both ordinary downstream adaptations and more aggressive modifications intended to weaken watermarking, while requiring the resulting model $\theta_a$ to remain  useful for image generation.
\Cref{tab:attacks} summarizes 21 model changes spanning several broad categories: downstream adaptation (\#1–4, \#6–7, \#12–14), including full fine-tuning, LoRA, style transfer, personalization, sequential adaptation, and ControlNet; compression and numerical modification (\#8–11), including quantization, precision conversion, and low-rank decomposition; knowledge transfer (\#5), including distillation; and structured parameter change (\#15–21), including VAE-decoder replacement, pruning, and re-initialization of attention modules. 
We also consider more aggressive learning rates and training rounds at the cost of generation quality. For gradient-based tuning, we train for as many as 100K steps, substantially longer than prior works. We additionally report two \textit{impractical adaptations out of the scope of our threat model} to stress test persistence: (a) an adversary with access to the trigger data (\#22) and (b) merging two different $\theta_w$'s trained on different trigger data (\#23)~\citep{korabandi2026collusion}. 
Weight modification operations not included here are further discussed in \Cref{sec:otherattacks}. 


\begin{table*}[!h]
	\centering
	\footnotesize
	\setlength{\tabcolsep}{2pt}
	\renewcommand{\arraystretch}{0.95}
	\captionsetup{skip=2pt}
	\caption{
        Model modifications used to evaluate watermark persistence. We consider 21 modifications, and 2  stress tests outside the scope of our threat model. Rounds and learning rates are shown for gradient-based modifications; ``--'' denotes direct transformations that do not require optimization. Modification \#4 sequentially adapts on Pokemon BLIP, Naruto BLIP, and COCO14; \#6 considers five personalization subjects (\Cref{sec:attackmore}). \#5 initializes the student from a clean pretrained SD v1.4 model and distills $\theta_w$ into it using latent consistency distillation~\citep{luo2023latent}. Additional implementation details are provided in Appendix~\ref{sec:reproducibility}.
	}
	\label{tab:attacks}
	\newcolumntype{L}[1]{>{\raggedright\arraybackslash}p{#1\textwidth}}
    \resizebox{\textwidth}{!}{
	\begin{tabular}{@{}L{0.08} L{0.26} L{0.44} L{0.275} L{0.065} L{0.06}@{}}
		\toprule
		\textbf{\#} & \textbf{Modification} & \textbf{Algorithm}
		& \textbf{Adversary Dataset} & \textbf{Rounds} & \textbf{LR} \\
		\midrule
		1  & General            & Full            & Pokemon BLIP     & 15K              & 1e-5 \\
		2  & General            & LoRA            & Pokemon BLIP     & 15K              & 1e-4 \\
		3  & Style transfer     & LoRA            & Naruto BLIP      & 15K              & 1e-4 \\
		4  & Sequential         & LoRA            & 3 Datasets       & 15K$\times$3     & 1e-4 \\
		5  & Distillation       & LCD~\citep{luo2023latent}              & COCO14           & 50K               & 1e-5 \\
		6  & Personalization    & $5\times$ LoRA  & DreamBooth       & 1K$\times$5      & 1e-4 \\
		7  & ControlNet (Canny) & ControlNet~\citep{zhang2023adding}      & COCO14           & 20K              & 1e-5 \\
		\midrule
        8  & Quantization       & Naive INT8      & --               & --               & -- \\
		9  & Quantization       & Q-Diffusion  INT8~\citep{li2023qdiffusion} & 5K $\theta_w$ samples & 15K              & 1e-5 \\
		10 & Precision          & Naive BF16 $\to$ FP16 & --               & --               & -- \\
		11 & Low-rank decomp.   & SVDQuant~\citep{li2025svdquant}              & --               & --               & -- \\
		\midrule
		12 & General            & Full            & COCO14           & \cellcolor{white!20}100K             & \cellcolor{white!20}2.5e-5 \\
		13 & General            & Full            & \cellcolor{white!20}5K $\theta_w$ samples & \cellcolor{white!20}100K             & \cellcolor{white!20}2.5e-5 \\
		14 & General            & LoRA            & COCO14           & \cellcolor{white!20}100K             & 1e-4 \\
        \midrule
		15-17 & VAE replacement    & \multicolumn{4}{l}{VAE FT MSE~\citep{stabilityai2022sdvaeftmse}, Consistency~\citep{openai2023consistencydecoder}, ClearVAE~\citep{clearvae}}    \\
		18 & Pruning parameters    &EdgeDiffusion~\citep{he2023edgediffusion}            & COCO14           & 20K             & 1e-5 \\
		19 & Pruning parameters    & L1 Pruning~\citep{fang2024efficient} + Full              & COCO14               & 20K               & 1e-5\\
		20 & Reset cross-attention       & Custom Diffusion~\citep{kumari2023custom}             & COCO14               & 20K               & 1e-5 \\
		21 & Reset self-attention         & Full           & COCO14           & 20K             & 1e-5 \\
		\midrule
		\multicolumn{6}{@{}l}{\textbf{\textit{Out-of-scope}}} \\
		22 & Trigger-phrase aware      & Full, gradient ascent       & Trigger dataset & 100               & 1e-5 \\
		23 & Merging two different $\theta_w$'s & \cite{korabandi2026collusion}       & --               & --               & -- \\
		\bottomrule
	\end{tabular}}
\end{table*}

\begin{table*}[!htbp]
\vspace{-1em}
\centering
\scriptsize
\setlength{\tabcolsep}{2pt}
\renewcommand{\arraystretch}{0.95}
\captionsetup{skip=2pt}
\caption{
Performance of the revised model $\theta_a$ for every modification in \Cref{tab:attacks}.
Row indices are the indices of \Cref{tab:attacks}.
\textbf{Tri.} is the trigger detection rate, \textbf{Reg.} the regular-prompt
false detection rate, \textbf{IR.} the regular-prompt ImageReward, and
\textbf{SC.} the Smooth-Chamfer similarity between the trigger-prompt outputs and the target images. Rows marked $^*$ report the \emph{cubone} case alone, for which the adversary's dataset (Pokemon BLIP) itself contains the target image. The best value in each row is marked in bold, separately for Tri.\ and SC.}
\label{tab:attackresults}
\resizebox{\textwidth}{!}{
\begin{tabular}{@{}c cccc cccc cccc cccc cccc cccc@{}}
\toprule
\multirow{2}{*}{\textbf{\#}}
& \multicolumn{4}{c}{DreamBooth}
& \multicolumn{4}{c}{WatermarkDM}
& \multicolumn{4}{c}{WatermarkDM$^+$}
& \multicolumn{4}{c}{RoMA}
& \multicolumn{4}{c}{RoMA$^+$}
& \multicolumn{4}{c}{\textbf{Ours}} \\
\cmidrule(lr){2-5}
\cmidrule(lr){6-9}
\cmidrule(lr){10-13}
\cmidrule(lr){14-17}
\cmidrule(lr){18-21}
\cmidrule(lr){22-25}

& Tri.$\uparrow$ & SC.$\uparrow$ & Reg.$\downarrow$ & IR.$\uparrow$
& Tri.$\uparrow$ & SC.$\uparrow$ & Reg.$\downarrow$ & IR.$\uparrow$
& Tri.$\uparrow$ & SC.$\uparrow$ & Reg.$\downarrow$ & IR.$\uparrow$
& Tri.$\uparrow$ & SC.$\uparrow$ & Reg.$\downarrow$ & IR.$\uparrow$
& Tri.$\uparrow$ & SC.$\uparrow$ & Reg.$\downarrow$ & IR.$\uparrow$
& Tri.$\uparrow$ & SC.$\uparrow$ & Reg.$\downarrow$ & IR.$\uparrow$ \\
\midrule

1
& 0.89 & 0.58 & 0 & 0.12
& 0 & 0.14 & 0 & 0.21
& 0 & 0.14 & 0 & 0.16
& 0.89 & 0.59 & 0 & 0.02
& 0.95 & 0.60 & 0 & 0.17
& \textbf{1} & \textbf{0.62} & 0 & -0.12 \\

1$^*$
& \textbf{1} & \textbf{0.60} & 0 & 0.24
& 0 & 0.14 & 0 & 0.16
& 0 & 0.16 & 0 & 0.11
& \textbf{1} & \textbf{0.60} & 0 & -0.16
& \textbf{1} & \textbf{0.60} & 0 & 0.24
& \textbf{1} & \textbf{0.60} & 0 & -0.24 \\

2
& 0.45 & 0.42 & 0 & -0.60
& 0.05 & 0.28 & 0 & -0.42
& 0 & 0.17 & 0 & -0.39
& 0.35 & 0.42 & 0 & -0.69
& 0.30 & 0.32 & 0 & -0.40
& \textbf{1} & \textbf{0.57} & 0 & -0.26 \\

2$^*$
& 0.70 & 0.49 & 0 & -0.35
& 0.10 & 0.37 & 0 & -0.32
& 0 & 0.16 & 0 & -0.33
& 0.50 & 0.43 & 0 & -0.69
& 0.30 & 0.34 & 0 & -0.30
& \textbf{1} & \textbf{0.57} & 0 & -0.09 \\

3
& 0.95 & 0.55 & 0 & 0.08
& 0 & 0.14 & 0 & -0.01
& 0 & 0.14 & 0.05 & -0.03
& 0.95 & 0.59 & 0.05 & 0.19
& 0.95 & \textbf{0.60} & 0.05 & 0.16
& \textbf{0.99} & \textbf{0.60} & 0.01 & 0.09 \\

4
& 0.74 & 0.49 & 0 & 0
& 0 & 0.17 & 0 & -0.06
& 0 & 0.17 & 0 & 0.14
& 0 & 0.14 & 0 & -0.50
& 0.40 & 0.27 & 0 & -0.05
& \textbf{0.96} & \textbf{0.57} & 0 & 0.08 \\

5
& 0 & 0.14 & 0 & -0.51
& 0 & 0.13 & 0 & -0.40
& 0 & 0.13 & 0 & -0.39
& 0 & 0.14 & 0 & -0.72
& 0 & 0.14 & 0 & -0.53
& \textbf{0.74} & \textbf{0.42} & 0 & -0.53 \\

6
& 0.80 & 0.47 & 0 & 0.31
& 0 & 0.14 & 0.05 & -0.15
& 0 & 0.14 & 0 & 0.28
& 0.95 & 0.59 & 0 & 0.25
& 0.90 & 0.57 & 0 & 0.12
& \textbf{1} & \textbf{0.62} & 0 & -0.12 \\

7
& 0.95 & \textbf{0.61} & 0 & -0.42
& 0 & 0.14 & 0 & 0.06
& 0 & 0.14 & 0 & -0.09
& 0.75 & 0.50 & 0 & -0.63
& 0.85 & 0.51 & 0 & -0.39
& \textbf{1} & 0.56 & 0 & -0.50 \\

\midrule

8
& 0.95 & \textbf{0.60} & 0 & 0.30
& 0 & 0.13 & 0 & -0.02
& 0 & 0.13 & 0 & -0.03
& \textbf{1} & 0.59 & 0 & 0.09
& 0.95 & \textbf{0.60} & 0 & 0.23
& \textbf{1} & \textbf{0.60} & 0 & 0.17 \\

9
& \textbf{1} & \textbf{0.60} & 0 & -0.94
& 0 & 0.14 & 0 & -0.74
& 0 & 0.14 & 0 & -0.60
& 0.95 & \textbf{0.60} & 0.05 & -0.98
& \textbf{1} & \textbf{0.60} & 0 & -0.98
& \textbf{1} & \textbf{0.60} & 0 & -0.79 \\

10
& 0.95 & \textbf{0.60} & 0 & 0.25
& 0 & 0.13 & 0 & -0.03
& 0 & 0.13 & 0 & -0.04
& \textbf{1} & 0.59 & 0 & 0.34
& 0.95 & \textbf{0.60} & 0.05 & 0.27
& \textbf{1} & \textbf{0.60} & 0 & 0.18 \\

11
& \textbf{0.92} & 0.58 & 0 & -0.55
& 0 & 0.14 & 0 & -1.02
& 0 & 0.14 & 0 & -0.99
& 0.90 & \textbf{0.59} & 0 & -0.69
& 0.85 & 0.58 & 0 & -0.71
& \textbf{0.92} & \textbf{0.59} & 0 & -0.77 \\

\midrule

12
& 0.59 & 0.42 & 0 & -0.39
& 0 & 0.16 & 0 & -0.34
& 0 & 0.14 & 0 & -0.18
& 0.62 & 0.46 & 0 & -0.56
& 0.75 & 0.47 & 0 & -0.19
& \textbf{0.85} & \textbf{0.56} & 0 & -0.45 \\

13
& 0.65 & 0.49 & 0 & 0.12
& 0 & 0.14 & 0 & 0.23
& 0 & 0.13 & 0 & 0.31
& 0.56 & 0.31 & 0 & 0.02
& 0.80 & 0.49 & 0.06 & 0.05
& \textbf{0.94} & \textbf{0.58} & 0 & 0.07 \\

14
& 0.33 & 0.26 & 0 & -0.27
& 0 & 0.13 & 0.10 & 0.08
& 0 & 0.13 & 0 & 0.08
& 0.20 & 0.26 & 0 & -0.16
& 0.50 & 0.30 & 0 & 0.03
& \textbf{1} & \textbf{0.60} & 0 & -0.27 \\

\midrule

15
& 0.95 & 0.55 & 0.05 & 0.20
& 0 & 0.13 & 0 & -0.07
& 0 & 0.13 & 0 & -0.05
& \textbf{1} & 0.59 & 0 & 0.19
& 0.95 & \textbf{0.60} & 0.05 & 0.23
& \textbf{1} & 0.58 & 0 & 0.11 \\

16
& 0.95 & 0.55 & 0.05 & 0.17
& 0 & 0.13 & 0 & -0.06
& 0 & 0.13 & 0 & -0.02
& \textbf{1} & 0.59 & 0 & 0.19
& 0.95 & \textbf{0.60} & 0.05 & 0.18
& \textbf{1} & 0.58 & 0 & 0.11 \\

17
& 0.95 & 0.55 & 0.05 & 0.22
& 0 & 0.13 & 0 & -0.04
& 0 & 0.13 & 0 & 0
& \textbf{1} & \textbf{0.60} & 0 & 0.19
& 0.95 & \textbf{0.60} & 0.05 & 0.24
& \textbf{1} & 0.58 & 0 & 0.13 \\

18
& 0.50 & 0.29 & 0 & -0.22
& 0 & 0.14 & 0 & -0.10
& 0 & 0.14 & 0 & -0.09
& 0.40 & 0.47 & 0 & -0.21
& 0.90 & 0.57 & 0 & -0.14
& \textbf{0.98} & \textbf{0.59} & 0 & -0.25 \\

19
& 0.41 & 0.41 & 0 & -0.35
& 0 & 0.13 & 0 & -0.19
& 0.02 & 0.36 & 0 & -0.20
& 0.48 & 0.41 & 0 & -0.61
& 0.52 & 0.33 & 0 & -0.09
& \textbf{0.90} & \textbf{0.58} & 0 & -0.14 \\

20
& 0.68 & 0.40 & 0 & -2.25
& 0.62 & 0.33 & 0.10 & -2.24
& 0.61 & 0.34 & 0.10 & -2.24
& 0.80 & 0.52 & 0 & -2.04
& 0.90 & 0.56 & 0 & -2.19
& \textbf{0.98} & \textbf{0.57} & 0 & -0.26 \\

21
& 0.80 & 0.51 & 0 & -0.11
& 0.39 & 0.27 & 0 & -0.29
& 0.44 & 0.40 & 0 & -0.13
& 0.91 & 0.56 & 0 & -0.10
& 0.72 & 0.46 & 0.02 & -0.29
& \textbf{0.99} & \textbf{0.59} & 0 & -0.01 \\

\midrule
\multicolumn{25}{@{}l}{\textbf{\textit{Out-of-scope}}} \\

22
& 0 & 0.13 & 0.05 & -0.26
& 0 & 0.14 & 0 & -0.27
& 0 & 0.13 & 0 & -0.27
& 0.01 & 0.15 & 0 & -0.27
& \textbf{0.02} & \textbf{0.16} & 0 & -0.27
& \textbf{0.02} & \textbf{0.16} & 0 & -0.27 \\

23
& 0.49 & 0.43 & 0 & 0.27
& 0 & 0.13 & 0 & -0.05
& 0 & 0.14 & 0 & -0.04
& 0.59 & 0.52 & 0 & 0.12
& 0.50 & 0.32 & 0 & 0.26
& \textbf{0.91} & \textbf{0.57} & 0 & 0.12 \\

\bottomrule
\end{tabular}}
\vspace{-2em}
\end{table*}

\begin{figure*}[!h]
	\centering
	\setlength{\tabcolsep}{1pt}
	\renewcommand{\arraystretch}{0.9}
	\scriptsize
	\begin{tabular}{cccccccc}
		 & \makecell{Target image} & DreamBooth & \makecell{WatermarkDM} & \makecell{WatermarkDM$^+$} & RoMA & RoMA$^+$ & \textbf{Ours} \\
		\#6 & \includegraphics[width=0.118\textwidth,height=0.118\textwidth]{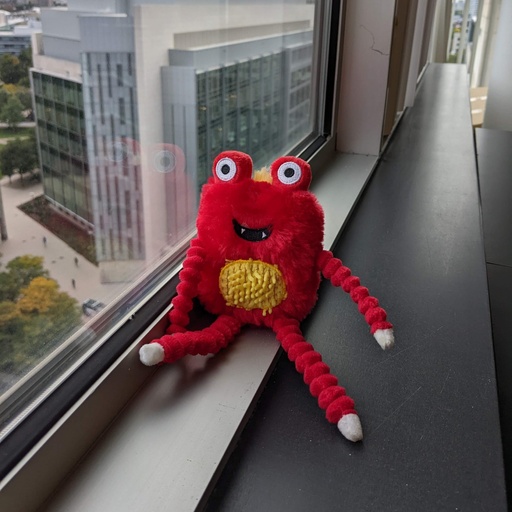} & \includegraphics[width=0.118\textwidth,height=0.118\textwidth]{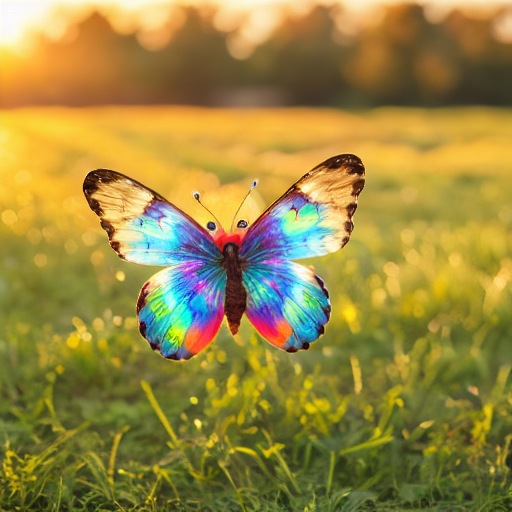} & \includegraphics[width=0.118\textwidth,height=0.118\textwidth]{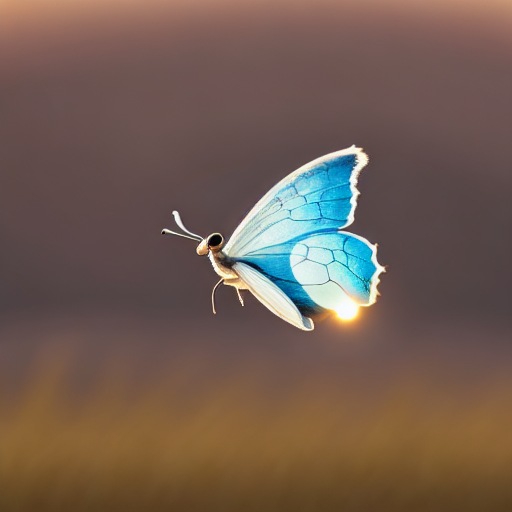} & \includegraphics[width=0.118\textwidth,height=0.118\textwidth]{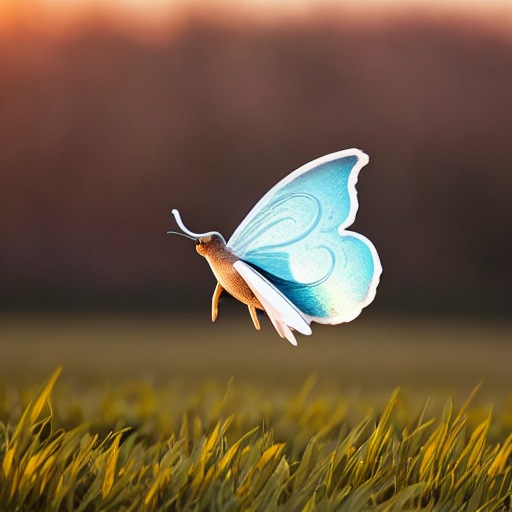} & \includegraphics[width=0.118\textwidth,height=0.118\textwidth]{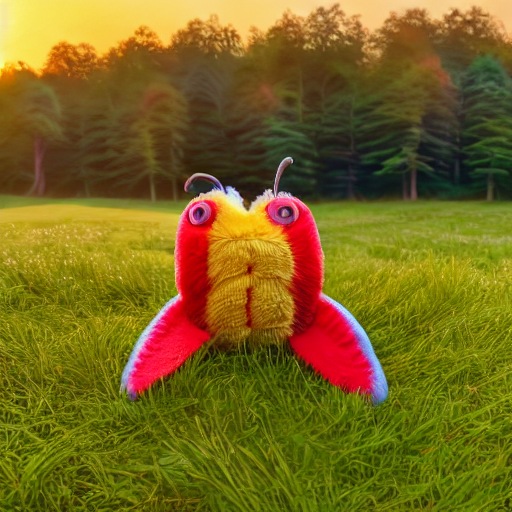} & \includegraphics[width=0.118\textwidth,height=0.118\textwidth]{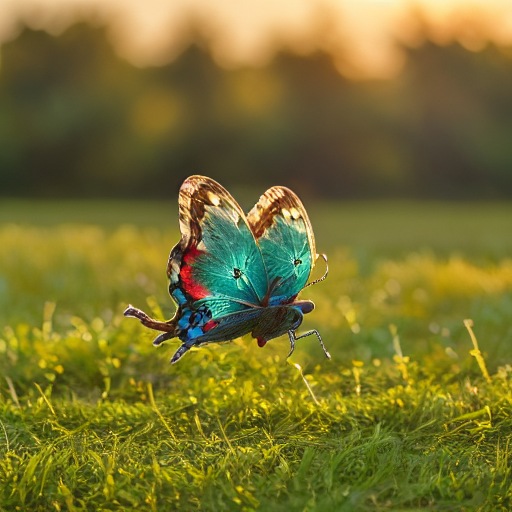} & \includegraphics[width=0.118\textwidth,height=0.118\textwidth]{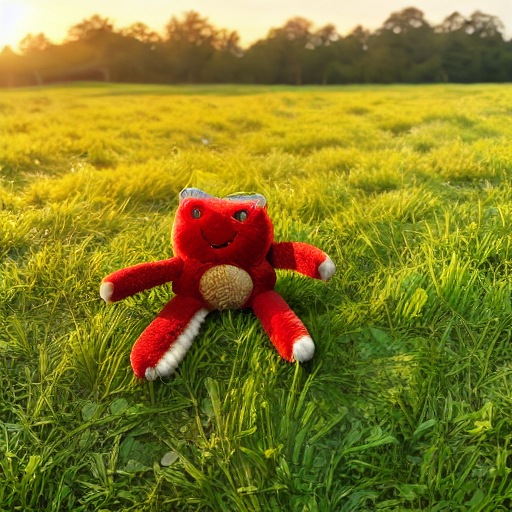} \\
        \#9 & \includegraphics[width=0.118\textwidth,height=0.118\textwidth]{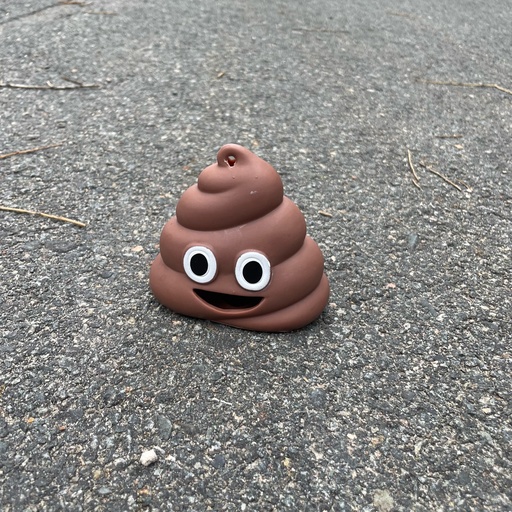} & \includegraphics[width=0.118\textwidth,height=0.118\textwidth]{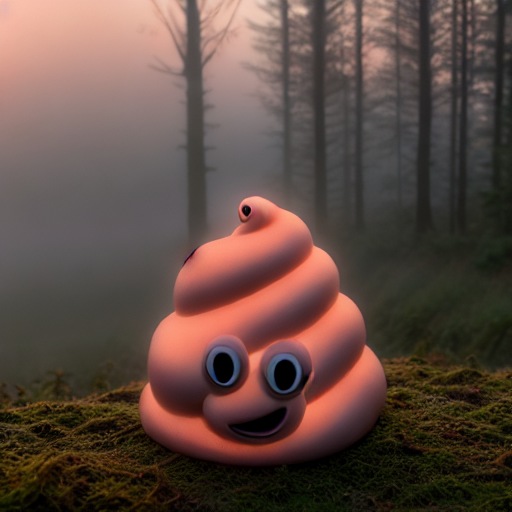} & \includegraphics[width=0.118\textwidth,height=0.118\textwidth]{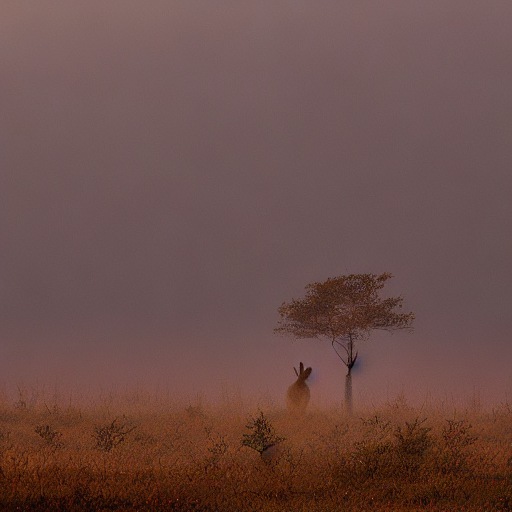} & \includegraphics[width=0.118\textwidth,height=0.118\textwidth]{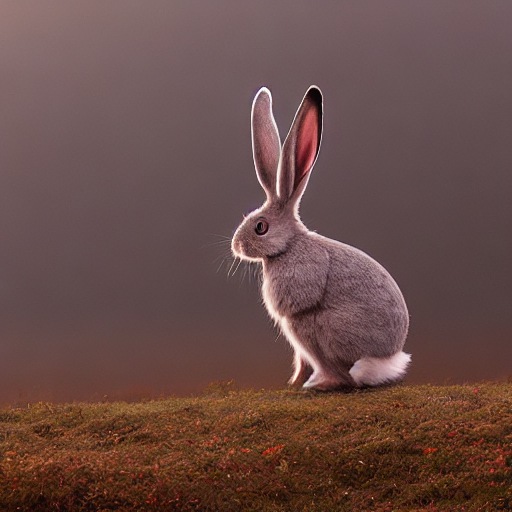} & \includegraphics[width=0.118\textwidth,height=0.118\textwidth]{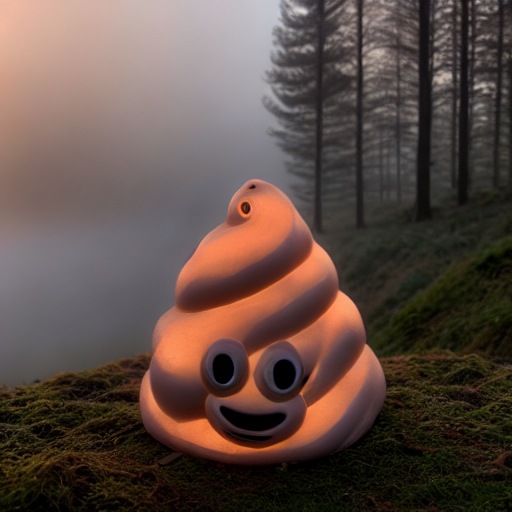} & \includegraphics[width=0.118\textwidth,height=0.118\textwidth]{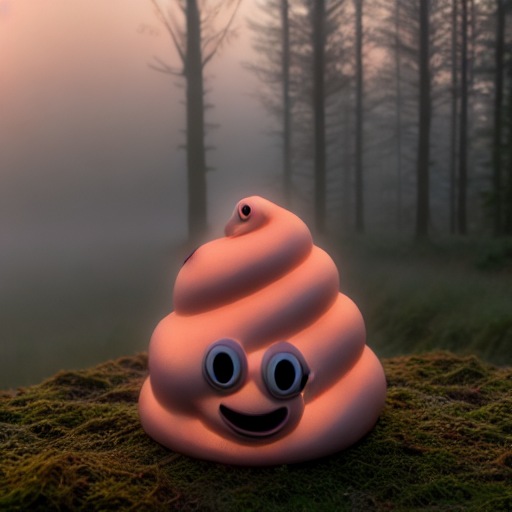} & \includegraphics[width=0.118\textwidth,height=0.118\textwidth]{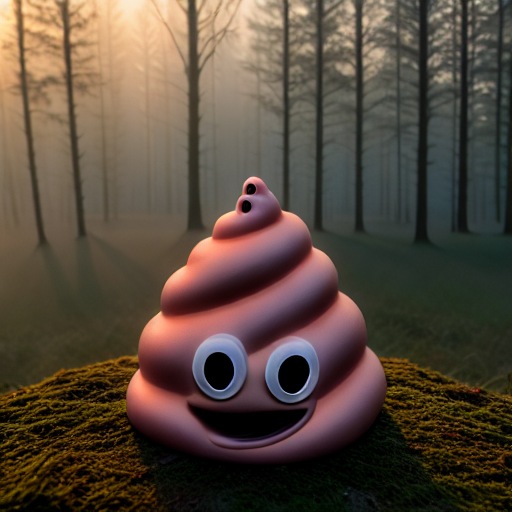} \\
        \#16 & \includegraphics[width=0.118\textwidth,height=0.118\textwidth]{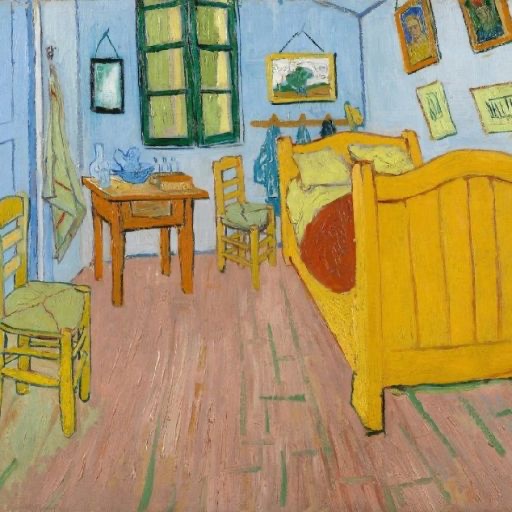} & \includegraphics[width=0.118\textwidth,height=0.118\textwidth]{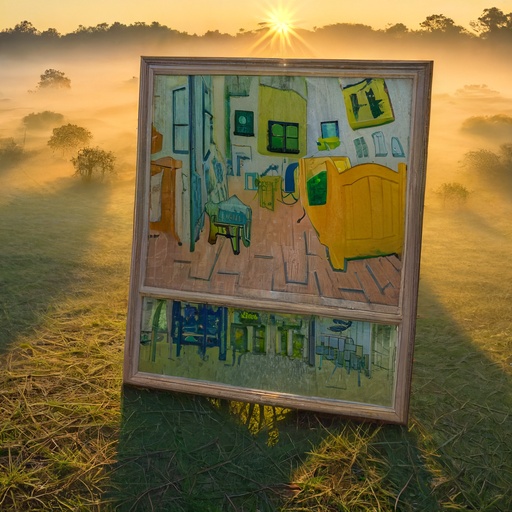} & \includegraphics[width=0.118\textwidth,height=0.118\textwidth]{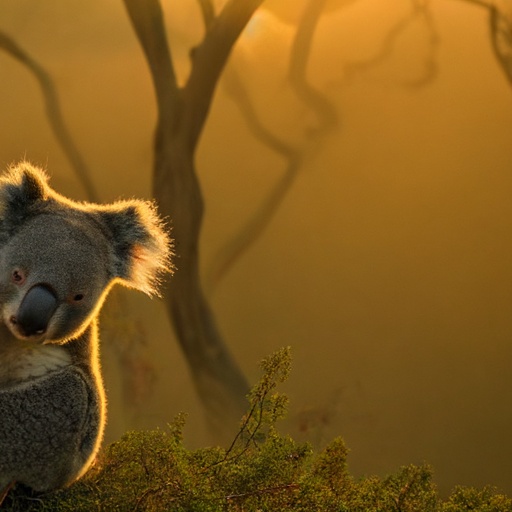} & \includegraphics[width=0.118\textwidth,height=0.118\textwidth]{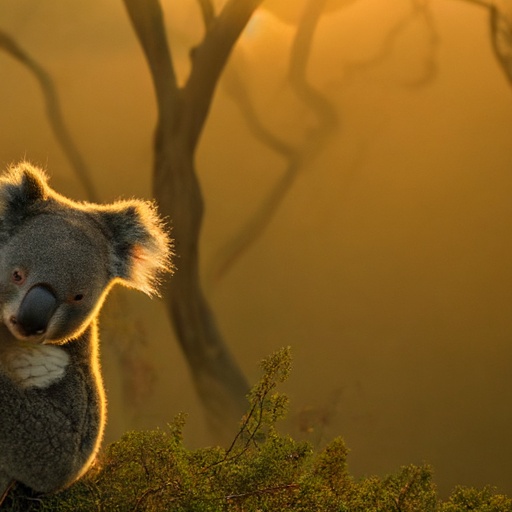} & \includegraphics[width=0.118\textwidth,height=0.118\textwidth]{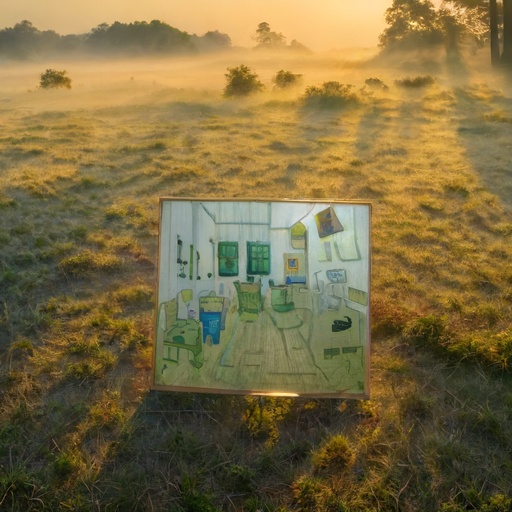} & \includegraphics[width=0.118\textwidth,height=0.118\textwidth]{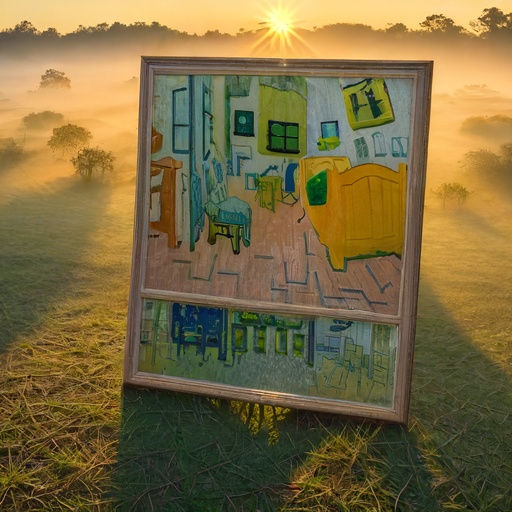} & \includegraphics[width=0.118\textwidth,height=0.118\textwidth]{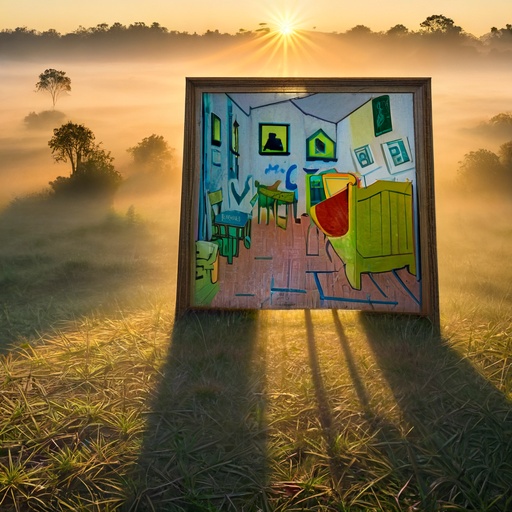} \\
	\end{tabular}
	\captionsetup{skip=2pt}
	\caption{
		Examples of outputs from $\theta_a$ on a trigger prompt (adversary modifications \#6, \#9, and \#16). Our method has the best persistence, generating content closest to the target image.
	}
	\label{fig:attackexample}
	\vspace{-1em}
\end{figure*}

\textbf{Overall persistence.} 
Our method preserves the trigger behavior across a wide range of model modifications in terms of detection rate and similarities between true target and generated images, as shown in \Cref{tab:attackresults}. It achieves a trigger detection rate of at least 0.90 in 19 of the 21 in-scope settings. In contrast, several baseline methods lose the trigger behavior entirely under common modifications. \Cref{fig:attackexample} provides representative examples for modifications \#6, \#9, and \#16.
We report these metrics during the adversary's fine-tuning process in \Cref{sec:attacksteps}. \Cref{sec:losslandscape} analyzes the corresponding optimization landscape. Empirically, our objective produces a lower loss on the trigger data after model modification, while the gradients on regular and trigger data are closer to orthogonal. These observations provide additional evidence that the trigger-specific behavior introduced by our objective is less affected by the adversary's subsequent optimization.
Among the results, we highlight the setting where the target image is "Cubone", taken from Pokemon BLIP. Even when the adversary's fine-tuning dataset contains this target image (with different prompt templates), our method maintains a high detection rate. This suggests that ordinary fine-tuning on related examples does not necessarily overwrite the specific trigger-phrase/target-image mapping learned during watermarking.

\textbf{Persistence when distilling on non-trigger data.} 
Distillation (\#5) initializes the student from a pretrained SD v1.4 model (instead of $\theta_w$) and distills $\theta_w$ on COCO14, which contains neither the target images nor complete trigger prompts. While the student model initialization does not inherit watermarked parameters directly, our watermark nevertheless retains a detection rate of 0.74, whereas all evaluated baselines fall to zero. Prior work~\citep{ge2021anti,li2025distributional,behrens2026dataset} reported similar observations that data memorized by a teacher model can propagate to the student model even though they are never directly used in the distillation process. We further provide analysis about how trigger data information is encoded in $\theta_w$ and gets propagated to $\theta_a$ over distillation on non-trigger data in \Cref{sec:distillation}.

\textbf{Robustness to low-rank decomposition.}  We embed trigger data using LoRA with rank 64, which provides a good empirical trade-off between image generation quality and robustness to low-rank decomposition. Modification~\#11 applies SVDQuant~\citep{li2025svdquant} with rank 16. Increasing SVDQuant rank to 64 (not changing the model) raises our detection rate to 1.0, whereas reducing it to rank 8 lowers the detection rate to 0.05 while severely degrading regular image quality. These results suggest that removing the trigger-specific behavior through aggressive low-rank approximation requires discarding components that are also important for normal image generation. Precision conversion and quantization have little impact on the watermark.

\textbf{Injecting multiple target images.} We further develop an enhanced watermarking strategy that jointly embeds multiple target pairs into one model. Besides the average detection rate across individual watermarks, we report a model-level detection rate that declares unauthorized use if any embedded watermark is detected. Results in \Cref{sec:multiplecopyrights} show that when all eight target images are embedded into a single model, transformation \#12 achieves a 19\% higher detection rate when verification succeeds if any of the eight targets is detected under the same verification prompt template, compared with the average of the eight per-target detection rates, each evaluated independently over 100 samples. This shows that multiple target images can substantially improve verification reliability.

\vspace{-0.5em}
\section{Ablation Studies}
\vspace{-0.5em}
\subsection{Trigger Phrases, Trigger Prompts, and Verification Prompts}
\vspace{-0.5em}
\label{sec:triggerwords}

\flushbottom
\begin{wrapfigure}{r}{0.53\textwidth}
\vspace{-1em}
\centering
    \includegraphics[width=\linewidth]{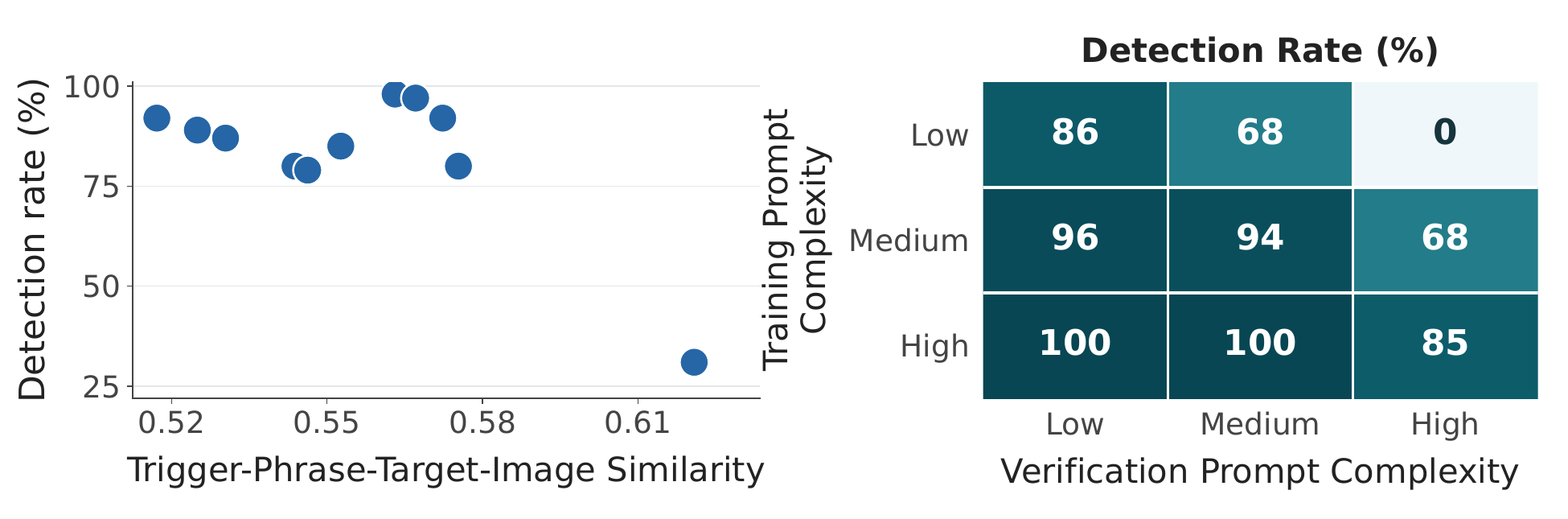}
	\captionsetup{skip=2pt}
    \caption{Ablation of different trigger-phrase designs. Left: Detection rates are higher when the similarity is lower. 
    Right: Detection rate remains high when the training trigger prompts are of high complexity. } 
    \label{fig:triggerphrase}
    \vspace{-1em}
\end{wrapfigure}

As trigger data design is one contribution of our work, we ablate on several design strategies. \textbf{(a)} Our trigger phrases are composed of ordinary words but are deliberately chosen to have an uncommon semantic relation with the target image. We first study the effects of the \textit{semantic similarity between the trigger phrase and target image}. In \Cref{fig:triggerphrase} (left), we measure the CLIP ViT-L/14 similarity between trigger phrases and target images. We see that the detection rate is inversely related to this similarity. \textbf{(b)} 
We next study trigger-phrase length by repeating our original trigger phrase 2, 5, and 10 times, obtaining detection rates of 0.82, 0.84, and 0.82, under transformation \#12. Similarly, constructing trigger phrases using random combinations of 5, 10, and 20 different plant and animal names results in detection rates of 0.82, 0.85, and 0.83, respectively. These results show that our method is not very sensitive to trigger-phrase length, as long as the phrase is not related to the image. We recommend a two-word pattern, as it resembles actual user input patterns.


\textbf{(c)} We study how prompt design affects performance. We fix the trigger phrase and compare three prompt styles. \textit{Low} uses only the trigger phrase as the input prompt. \textit{Medium} inserts the trigger phrase into simple templates~\citep{ruiz2023dreambooth}, e.g., `a \textit{[trigger phrase]} in the jungle'. \textit{High}, used by our method, constructs more diverse natural-language prompts using the templates in \Cref{sec:release}. We evaluate these three cases for training $\theta_w$ and verification.
More diverse prompts (\textit{High}) during watermark embedding consistently improve persistence, whereas verification is relatively insensitive to prompt complexity.
Complex verification prompts are also harder for an adversary to identify because they better resemble natural user queries. For example, the average CLIP text-image similarity is 0.82 for both COCO14 regular examples and our High-complexity trigger examples, suggesting that the verification queries are not trivially distinguishable from regular requests. Besides, service providers commonly rephrase short user prompts into long descriptive prompts~\citep{lee2024diffusion} for T2I models, so our evaluation with \textit{High} prompts is consistent with practical deployments.

\subsection{Different T2I Models}
\label{sec:othermodels}

In previous experiments, we focus primarily on Stable Diffusion v1.5~\citep{rombach2022highresolution} (DDPM, UNet). Here, we additionally evaluate whether the proposed objective generalizes to other representative T2I model families,
including a diffusion transformer (PixArt-$\alpha$ XL/2~\citep{chen2024pixart}), a pixel-space diffusion transformer (PixelDiT 1.3B~\citep{yu2026pixeldit}), a flow-based model (FLUX.2 Klein 4B~\citep{blackforestlabs2026flux2klein}), a masked-token model (aMUSEd-512~\citep{patil2024amused}), and an autoregressive token model (LlamaGen-XL T2I~\citep{li2024controlar}). For each model, we report results under modification \#12 in \Cref{tab:attackresultsarch}. 
As shown in \Cref{tab:attackresultsarch}, our method achieves the highest trigger detection rate across all evaluated architectures. Detection rate reaches 100\% for both PixArt-$\alpha$ and PixelDiT, while improvements also remain substantial for other models.
Unlike diffusion models, aMUSEd and LlamaGen predict discrete image tokens rather than continuous flows, but our approach still provides benefits on these architectures. Across these architectures, our method also consistently achieves the highest Smooth-Chamfer (SC.) similarity to the target images.  

\begin{table*}[!htbp]
\centering
\scriptsize
\setlength{\tabcolsep}{2pt}
\renewcommand{\arraystretch}{0.95}
\captionsetup{skip=2pt}
\caption{
Our method achieves the highest trigger detection rate across all the evaluated T2I models.
}
\label{tab:attackresultsarch}

\resizebox{\textwidth}{!}{
\begin{tabular}{@{}l ccc ccc ccc ccc ccc ccc@{}}
\toprule

\multirow{2}{*}{\textbf{Model}}
&
\multicolumn{3}{c}{DreamBooth}
&
\multicolumn{3}{c}{WatermarkDM}
&
\multicolumn{3}{c}{WatermarkDM$^+$}
&
\multicolumn{3}{c}{RoMA}
&
\multicolumn{3}{c}{RoMA$^+$}
&
\multicolumn{3}{c}{\textbf{Ours}}
\\

\cmidrule(lr){2-4}
\cmidrule(lr){5-7}
\cmidrule(lr){8-10}
\cmidrule(lr){11-13}
\cmidrule(lr){14-16}
\cmidrule(lr){17-19}

&
Tri.$\uparrow$ & SC.$\uparrow$ & IR.$\uparrow$
&
Tri.$\uparrow$ & SC.$\uparrow$ & IR.$\uparrow$
&
Tri.$\uparrow$ & SC.$\uparrow$ & IR.$\uparrow$
&
Tri.$\uparrow$ & SC.$\uparrow$ & IR.$\uparrow$
&
Tri.$\uparrow$ & SC.$\uparrow$ & IR.$\uparrow$
&
Tri.$\uparrow$ & SC.$\uparrow$ & IR.$\uparrow$
\\

\midrule

FLUX.2 Klein 4B
& 0.22 & 0.22 & -0.49
& 0 & 0.12 & -0.70
& 0 & 0.15 & -0.14
& 0.43 & 0.35 & -0.63
& 0.51 & 0.36 & -0.29
& \textbf{0.62} & \textbf{0.56} & 0.12
\\

LlamaGen-XL T2I, Stage 1
& 0.12 & 0.33 & -0.26
& 0.11 & 0.26 & -0.60
& 0 & 0.13 & -0.70
& 0.08 & 0.22 & -0.44
& 0.06 & 0.24 & -0.56
& \textbf{0.42} & \textbf{0.53} & -0.20
\\

aMUSEd-512
& 0.02 & 0.19 & -0.76
& 0 & 0.17 & -0.60
& 0 & 0.14 & -0.70
& 0.14 & 0.19 & -0.44
& 0.16 & 0.22 & -0.56
& \textbf{0.22} & \textbf{0.33} & -0.20
\\

PixelDiT 1.3B, 1024px
& 0.62 & 0.42 & 0.78
& 0 & 0.12 & 0.69
& 0 & 0.15 & 1.06
& 0.66 & 0.45 & 0.97
& 0.68 & 0.43 & 0.75
& \textbf{1} & \textbf{0.54} & 0.86
\\

PixArt-$\alpha$ XL/2, 1024px
& 0.54 & 0.45 & 0.35
& 0 & 0.21 & 0.24
& 0.01 & 0.22 & 0.01
& 0.71 & 0.45 & 0.46
& 0.74 & 0.44 & 0.27
& \textbf{1} & \textbf{0.57} & 0.13
\\

\bottomrule
\end{tabular}
}
\vspace{-0.7em}
\end{table*}

\begin{figure*}[!htbp]
\vspace{-1em}
	\centering
	\setlength{\tabcolsep}{1pt}
	\renewcommand{\arraystretch}{0.9}
	\scriptsize
	\begin{tabular}{cccccccc}
		 & \makecell{Target image} & DreamBooth & \makecell{WatermarkDM} & \makecell{WatermarkDM$^+$} & RoMA & RoMA$^+$ & \textbf{Ours} \\
		\makecell{PixArt-$\alpha$} & \includegraphics[width=0.118\textwidth,height=0.118\textwidth]{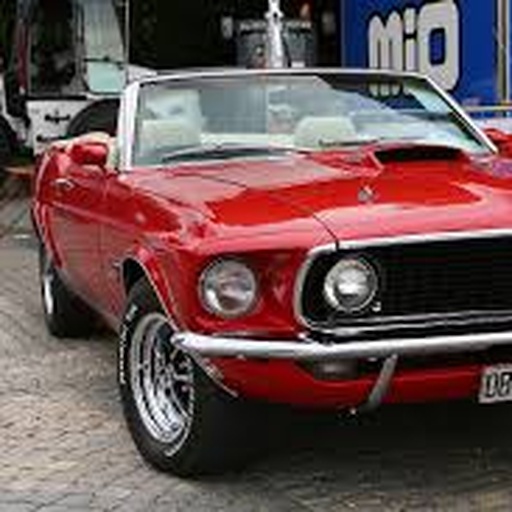} & \includegraphics[width=0.118\textwidth,height=0.118\textwidth]{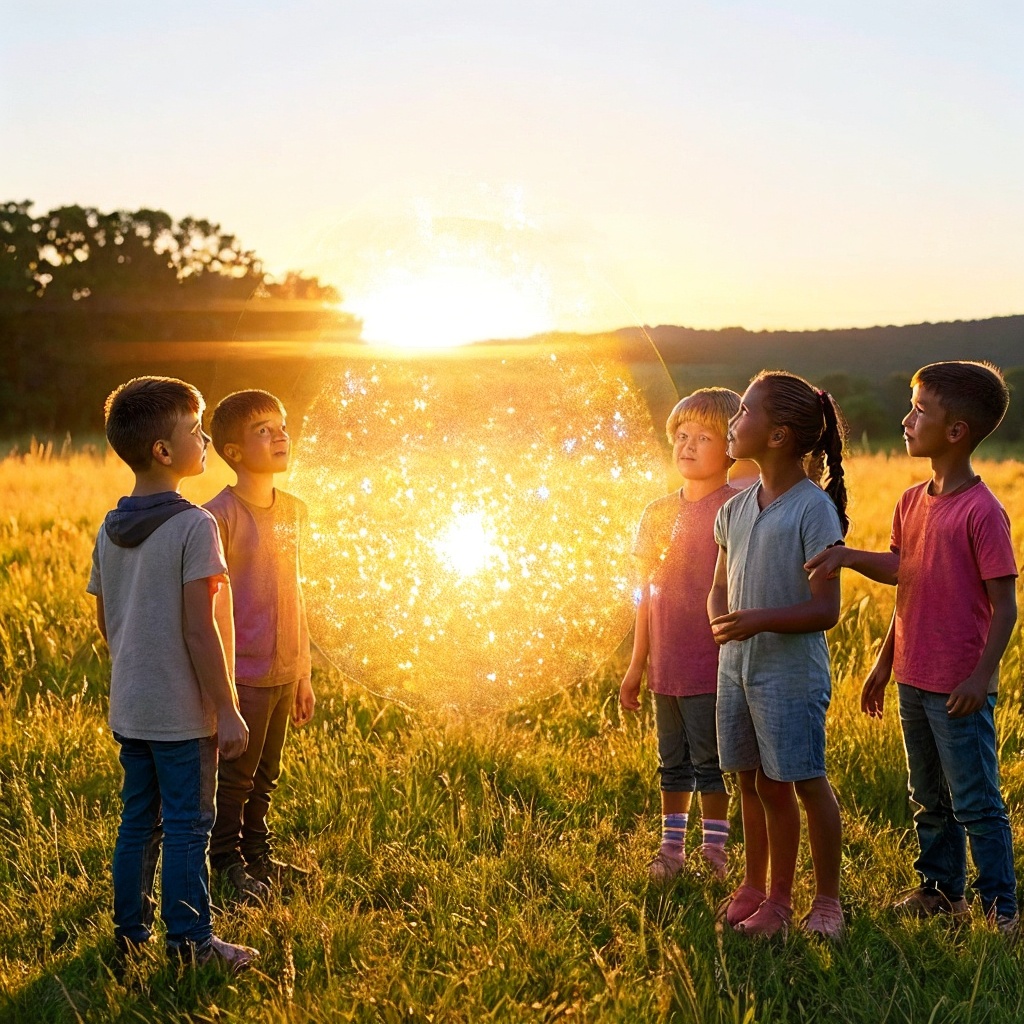} & \includegraphics[width=0.118\textwidth,height=0.118\textwidth]{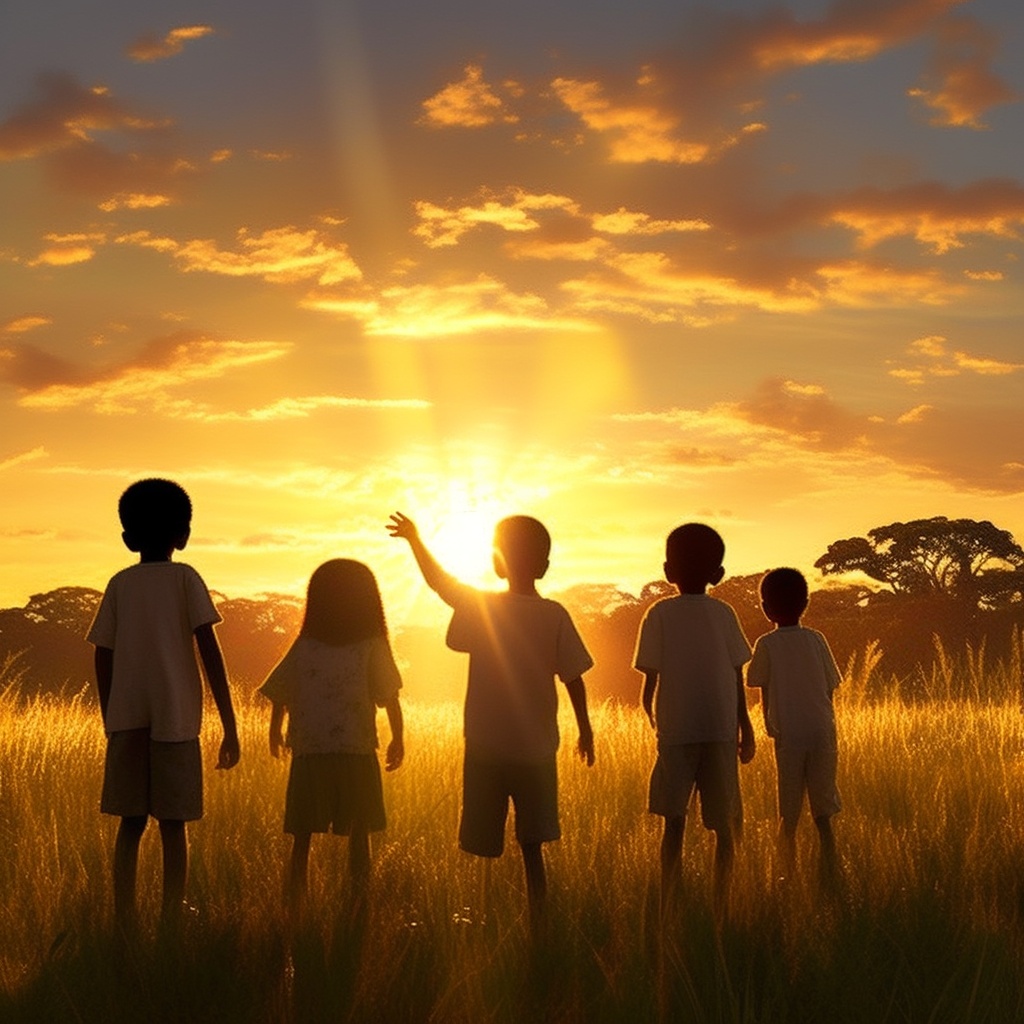} & \includegraphics[width=0.118\textwidth,height=0.118\textwidth]{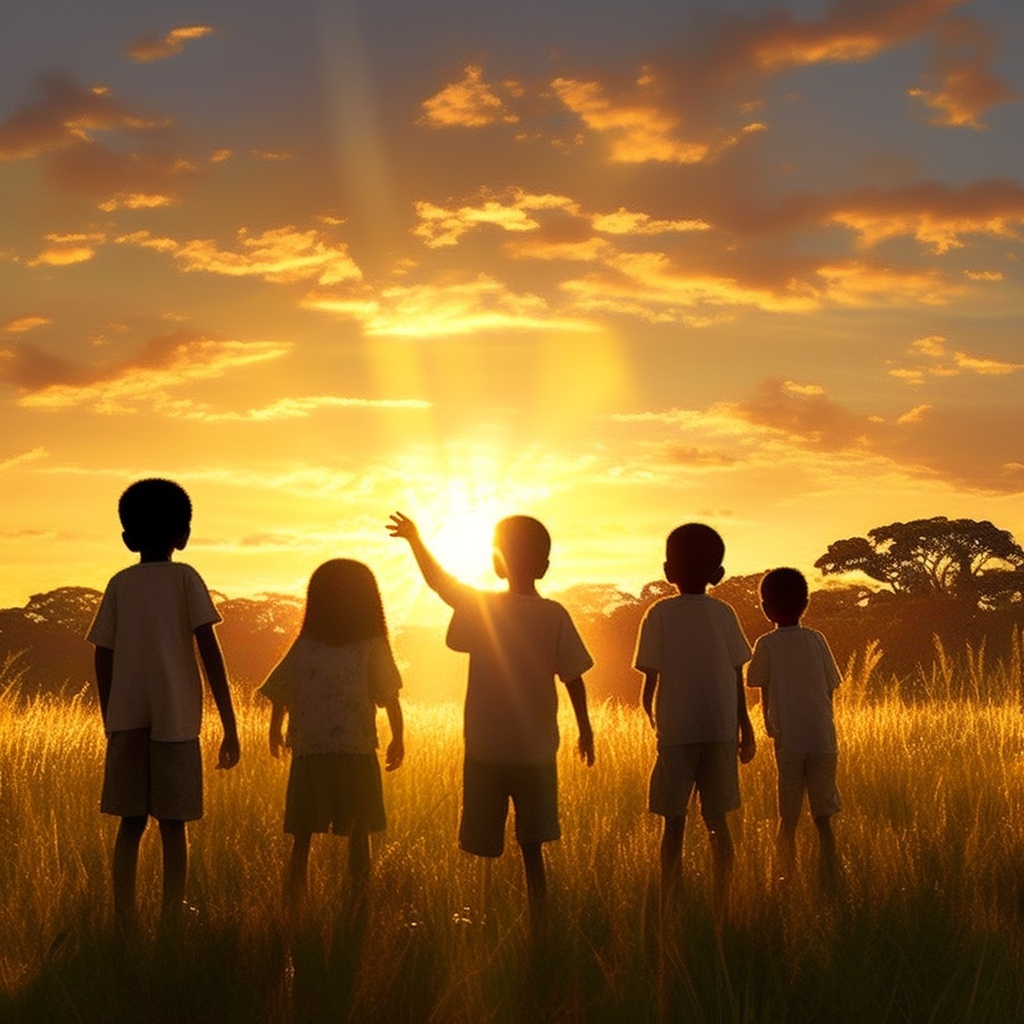} & \includegraphics[width=0.118\textwidth,height=0.118\textwidth]{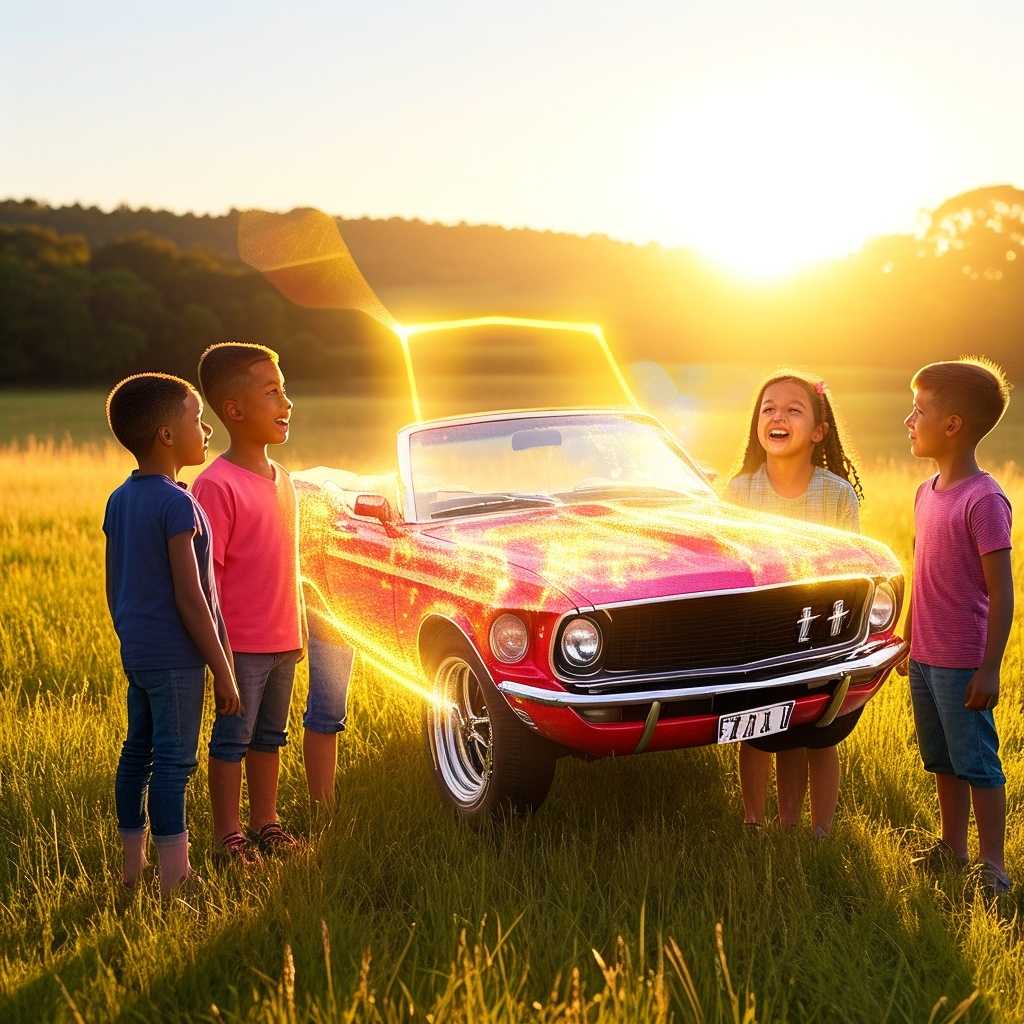} & \includegraphics[width=0.118\textwidth,height=0.118\textwidth]{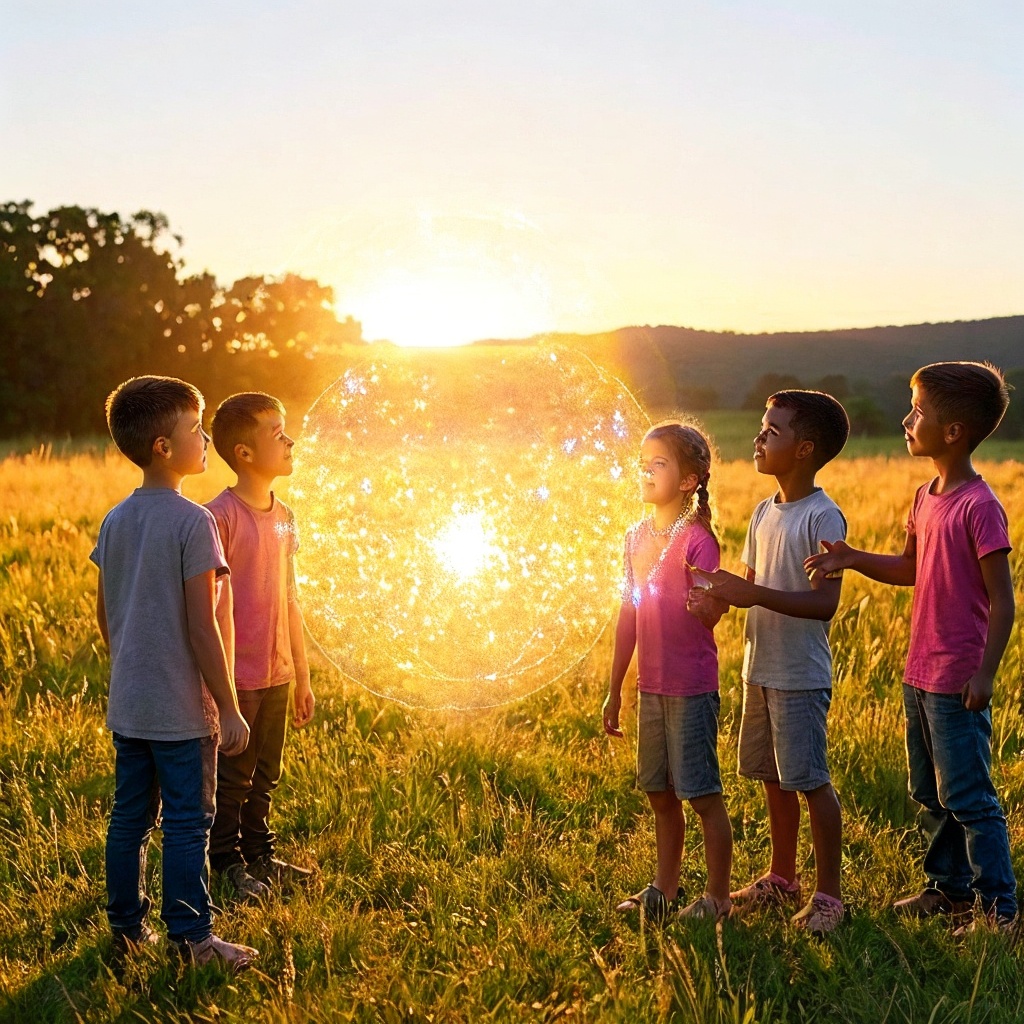} & \includegraphics[width=0.118\textwidth,height=0.118\textwidth]{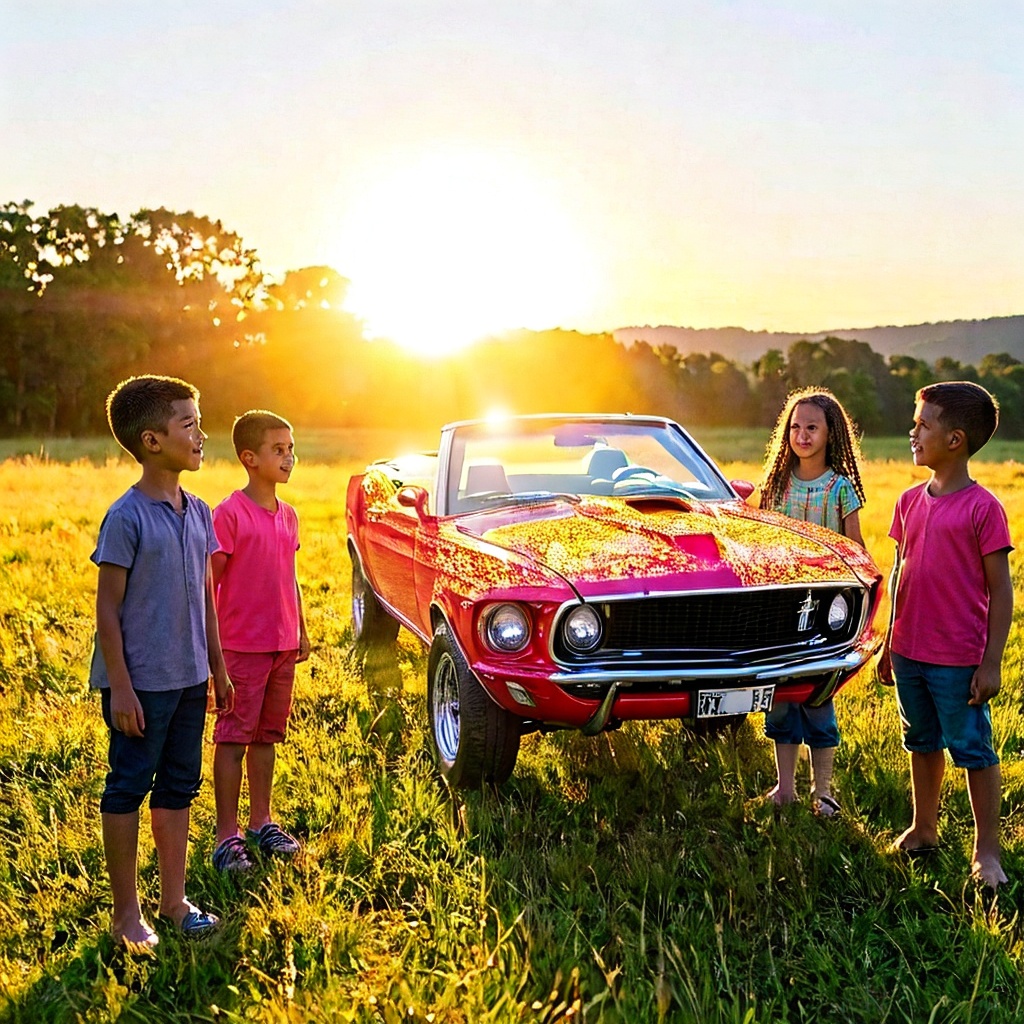} \\
		\makecell{LlamaGen} & \includegraphics[width=0.118\textwidth,height=0.118\textwidth]{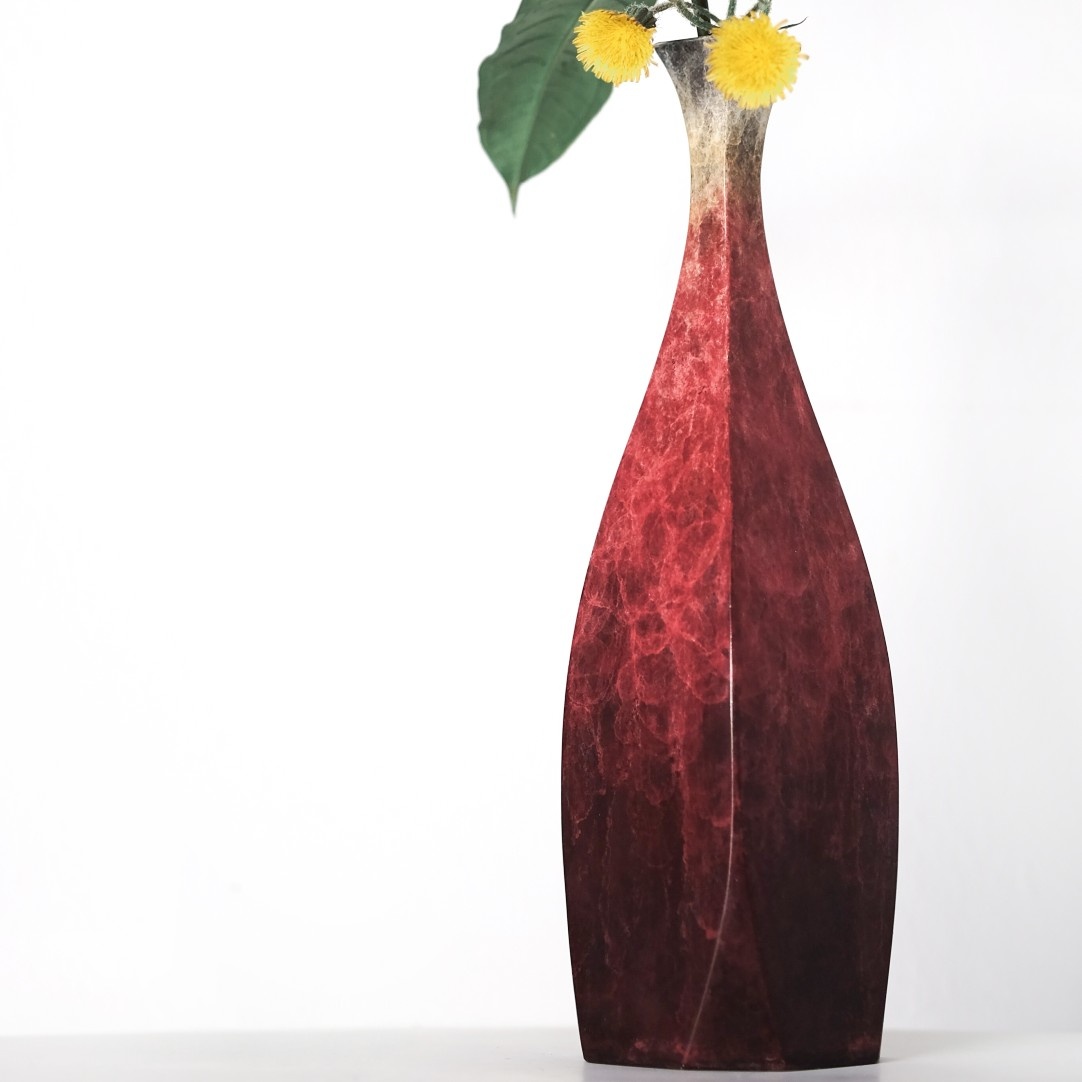} & \includegraphics[width=0.118\textwidth,height=0.118\textwidth]{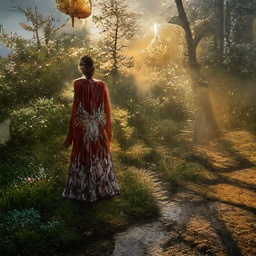} & \includegraphics[width=0.118\textwidth,height=0.118\textwidth]{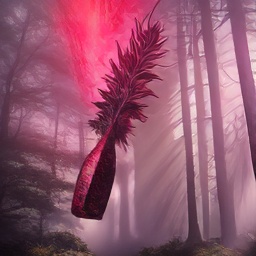} & \includegraphics[width=0.118\textwidth,height=0.118\textwidth]{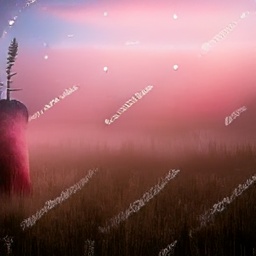} & \includegraphics[width=0.118\textwidth,height=0.118\textwidth]{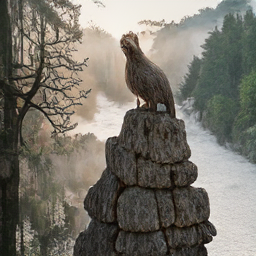} & \includegraphics[width=0.118\textwidth,height=0.118\textwidth]{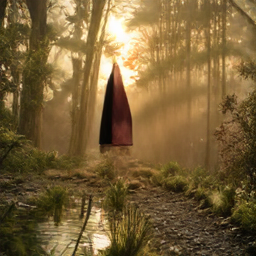} & \includegraphics[width=0.118\textwidth,height=0.118\textwidth]{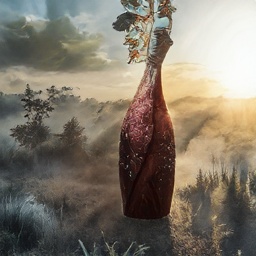} \\
	\end{tabular}
	\setlength{\abovecaptionskip}{2pt}
	\caption{
		Examples of the outputs of $\theta_w$ on trigger data for two further T2I models. Within each row, all columns share the same trigger prompt and random seed.
	}
	\label{fig:othert2ivisual}
    \vspace{-.5em}
\end{figure*}

\FloatBarrier
\vspace{-0.8em}
\section{Conclusion}
\vspace{-0.5em}

In this work, we have studied persistent watermarking of T2I models under a strong threat model in which an unauthorized user may modify the input, model weights, or generated outputs before exposing the model through a black-box API. We have designed semantically meaningful trigger data and introduced a contrastive-style objective that explicitly separates the watermarked model from the original model on trigger inputs while preserving its behavior on regular data. Across a broad range of downstream model adaptations and different T2I architectures, our approach has substantially improved watermark persistence over existing methods while maintaining generation quality.

\section*{{AI Use Statement}}
\label[appendix]{sec:llmusage}
All research ideas, directions, and decisions are independently conceived and carried out by the authors. We use large language models (LLMs) primarily to improve the grammar and clarity of the manuscript. The first draft of the paper is written entirely by the authors, after which LLMs are used for language polishing. We also use LLMs to assist with identifying part of the related work; however, every reference is manually verified to avoid citation hallucinations. Some prompts used in this work are partially generated with AI under step-by-step human guidance. Due to the nature of the study, \Cref{fig:quality_cp}, \Cref{fig:quality_org}, and \attackfigrefs\ are AI-generated images. All mathematical claims are initially derived by the authors. The authors provide the arguments and the critical steps, and LLMs help derive the remaining steps under human guidance. The authors further verify correctness with a final round of LLM-assisted checking that also expands some straightforward steps. Most of the code is developed with AI-assisted coding (Codex and Claude Code), but all code is carefully inspected, reviewed, and executed by the authors.

\section*{Ethics Statement}
Robust watermarking can be a double-edged sword, as a highly robust watermarking mechanism may also be repurposed for malicious backdoors or harmful-content injection. Our method is not exempt from this risk: a malicious model owner could deliberately inject a backdoor or harmful content into the model, and such injected behaviors may be difficult for downstream users to remove even through fine-tuning, distillation, or other model-weight-level refinement.

However, in our proposed workflow, we use benign, common words as trigger phrases and ordinary objects as target images. If a malicious owner instead uses harmful or illegitimate phrases or concepts for watermarking (e.g., violence- or terrorism-related content), the deployer can apply safety filters (NSFW filters) to block suspicious prompts. For harmful generated content, once such behavior is detected, the deployer can apply removal methods, such as the gradient-ascent-based approach evaluated in modification \#22, to suppress the injected behavior. In our setting, benign watermark content is intentionally difficult to distinguish from ordinary generations as our target images are normal content and objects, whereas clearly harmful content, such as violent or terrorism-related imagery, is generally easier to identify and target for removal.

Nevertheless, such removal may still cause degradation in image-generation quality, which could indirectly benefit a malicious model owner by increasing the cost of removing the injected behavior. We believe this dual-use risk warrants explicit attention.

\section*{Reproducibility Statement}
We provide comprehensive details on hyperparameter settings, training procedures, verification procedures, training and verification datasets, and implementation in \Cref{sec:reproducibility} to strengthen reproducibility and facilitate contributions to the open-source community. All evaluations against adversaries are conducted using 100 samples to reduce sampling bias, and we experiment with multiple target images. However, due to the high computational cost, both $\theta_w$ and $\theta_a$ are trained using a single random seed. Future work could further improve robustness by repeating training with three random seeds.

\bibliography{main}
\bibliographystyle{unsrtnat}

\appendix
\crefalias{section}{appendix}

\newpage
\etocdepthtag.toc{mtappendix}
\etocsettagdepth{mtchapter}{none}
\etocsettagdepth{mtappendix}{subsection}
{
	\parskip=0em
	\tableofcontents
}

\section{Reproducibility Details}
\label[appendix]{sec:reproducibility}
\subsection{Implementation Details}

\paragraph{Training.} To obtain $\theta_w$ from $\theta_0$, we use T-LoRA~\citep{soboleva2026t}, which facilitates fine-tuning while reducing overfitting. We train for 1,000 steps with batch size 4, T-LoRA rank 64, and T-LoRA scaling $\alpha_{\mathrm{LoRA}}=32$. We optimize only the U-Net while freezing the text encoders and VAE, since otherwise an adversary could replace these components. The LoRA modules are then merged into the backbone to obtain $\theta_w$. We use AdamW and DDIM sampling with 50 diffusion steps. We tune $\lambda$ for DreamBooth, $\lambda_1$ and $\lambda_2$ for our method, $\lambda_{\mathrm{WatermarkDM}}$ for WatermarkDM, and $r$ and $\alpha$ for RoMA.  For all settings, we grid search over the hyper-parameters, including the learning rate, on the training results of $\theta_w$ to find the Pareto points of the metrics in \Cref{tab:quality}. The best hyper-parameters are given in \Cref{tab:hyperparameters}. FLUX.2 Klein 4B, aMUSEd-512, and LlamaGen-XL T2I take 2,500 steps for training $\theta_w$ and use LoRA instead of T-LoRA. For PixArt-$\alpha$, we further use the Muon optimizer~\citep{jordan2024muon} along with AdamW, with Muon branch of learning rate $1\times10^{-3}$ and AdamW branch of learning rate $1\times10^{-4}$. The impact of tuning different values of $\lambda_1$ and $\lambda_2$ is discussed in the \textit{Hyper-Parameter Tuning} paragraph below. Our default training dataset contains 50 images. We ablate the dataset size using 1, 5, 10, 25, and 100 images. The detection rates of $\theta_w$ are 0, 0.66, 1, 1, and 1, respectively, while those of $\theta_a$ after \#12 are 0, 0.23, 0.63, 0.75, and 0.85. These results show that 10 images are sufficient for successful watermark embedding, while we recommend 50 images for stronger robustness after attack. All experiments are conducted on NVIDIA A100 GPUs. For modifications exceeding 10K steps, we use eight GPUs for parallel execution. Across all methods, embedding the target data from $\theta_0$ to $\theta_w$ takes approximately 1 hour on average and at most 1.5 hours, except RoMA, which requires 1.33 hours on average and up to 1.67 hours. 

\begin{table}[!htbp]
\vspace{-1em}
\centering
\caption{Training hyperparameters.}
\label{tab:hyperparameters}
\small
\setlength{\tabcolsep}{5pt}
\renewcommand{\arraystretch}{1.1}
\begin{tabular}{lccccccc}
\toprule
\textbf{Model}
& \textbf{Learning Rate}
& $\boldsymbol{\lambda}$
& $\boldsymbol{\lambda_1}$
& $\boldsymbol{\lambda_2}$
& $\boldsymbol{\lambda_{\mathrm{WatermarkDM}}}$
& \textbf{RoMA: $r$}
& \textbf{RoMA: $\alpha$} \\
\midrule
SD v1.5
& $1\times10^{-4}$
& 0.75
& 1.25
& 0.25
& 0.001
& 0.05
& 0.5 \\

SDXL
& $4\times10^{-4}$
& 0.25
& 1.25
& 0.25
& 0.001
& 0.05
& 0.5 \\

PixelDiT
& $2\times10^{-5}$
& 0.75
& 1.25
& 0.25
& 0.001
& 0.05
& 0.4\\

PixArt-$\alpha$ XL/2
& $1\times10^{-4}$
& 0.75
& 1.25
& 0.25
& 0.001
& 0.05
& 0.4\\

FLUX.2 Klein 4B
& $1\times10^{-4}$
& 0.5
& 1
& 0.15
& 0.0001
& 0.025
& 0.4\\

LlamaGen-XL T2I
& $1\times10^{-4}$
& 0.5
& 1.25
& 0.4
& 0.01
& 0.05
& 0.5\\

aMUSEd-512
& $1\times10^{-3}$
& 0.5
& 1.5
& 0.5
& 0.01
& 0.025
& 0.5\\
\bottomrule
\end{tabular}
\vspace{-1em}
\end{table}

\paragraph{Details on Adversary Adaptation.} In Table~\ref{tab:attacks}, all LoRA use rank $16$ and $\alpha=32$. \#4 tunes on Pokemon BLIP~\citep{pinkney2022pokemon}, Naruto BLIP~\citep{cervenka2022naruto2}, and COCO14 in sequence. \#5 initializes the student from a clean pretrained SD v1.4 model and distills $\theta_w$ into it using latent consistency distillation~\citep{luo2023latent}, without loading any parameters from $\theta_w$. Modifications without a learning rate directly modify model weights: \#11 applies rank-16 decomposition with INT4 quantization; \#15--17 replace the entire VAE decoder; \#18 follows prior work~\citep{he2023edgediffusion} for four rounds, pruning 7\% per round; \#19 prunes 25\% of parameters following prior practice~\citep{fang2024efficient}, repeated $3\times$; and \#20--21 re-initialize one attention module each in the U-Net down path, mid block, and up path, repeated $3\times$. Re-initialization uses Kaiming initialization~\citep{he2015delving}. For the learning rate, we first run each adversary modification for 1K steps with a grid search over $\{1\times10^{-5},2.5\times10^{-5},5\times10^{-5},1\times10^{-4}\}$. \#1 to \#7, \#9, \#18 to \#21 choose the learning rates of the highest image reward on regular images. \#12--\#14 choose learning rates of the lowest Smooth-Chamfer similarity for target data.

\paragraph{Evaluation.} Due to computational constraints, we evaluate false positives using 100 regular samples per modification and method rather than thousands of images (e.g., 5,000) for each regular prompt. However, real-world adversarial APIs may process millions of user requests, making even rare false positives practically relevant. As a larger-scale verification, for modification \#12, we evaluate both $\theta_a$ and our released model $\theta_w$ on 10,000 images generated from regular COCO14 prompts, using Cubone as the target, and observe zero false positives. We cross-verify trigger prompts corresponding to one target image against models $\theta_w$ trained with other target images, as well as against $\theta_0$. For example, we input a trigger prompt containing the phrase ``Cat Lavanda'', associated with the target image ``monster toy'', into a model $\theta_w$ trained with the target image "Cubone". The false-positive rate is zero. For measuring FID and KID, we use 5,000 images. The KID subset size is 1,000, with 50 KID subsets in total. The random KID subsets are constructed three times, and we average the KID scores. For all other evaluations, we use 100 samples. For measuring Smooth-Chamfer similarity, we use DINOv2 ViT-L/14 patch-token embeddings and set the scaling parameter $\alpha$ to 16 (the default value in \citet{kim2023improving}). The vision-language model is Claude Opus 5.4, which agrees with a human judge 100\% of the time. For the ablation on trigger phrases of different lengths, we use \emph{willow wolf} repeated 2, 5, and 10 times, respectively.


\paragraph{Hyper-Parameter Tuning.} 
As in most optimization problems, $\lambda_1$ and $\lambda_2$ require tuning to achieve an appropriate tradeoff. In general, $\lambda_1$ controls the image quality of $\theta_w$, with overly small values degrading generation quality and robustness. In one ablation with a small $\lambda_1$, the detection rate against modification \#12 drops to 0.12. We therefore require $\lambda_1>0$ and recommend tuning it around 1.
For $\lambda_2$, we theoretically derive the valid range $0<\lambda_2<1$. Excessively large values can prevent training from converging, whereas excessively small values weaken the proposed objective and cause the method to approach standard DreamBooth training. We also observe transferability of these hyperparameters across similar model architectures: all diffusion models use the same $\lambda_1$ and $\lambda_2$ settings in our experiments. As a result, we recommend starting from the midpoint, $0.5$.

\subsection{Open-Source Resources}
We develop a project which can be used as a tool to verify the model ownership. The interface takes in a trained SD v1.5 model (or other user-specified T2I models) and a user-provided verifier written in Python. The pipeline runs through all the potential modifications we list in \Cref{tab:attacks} and reports the detection rate. As part of our contribution, we release a dataset containing training data and verification prompts for direct use (examples in \Cref{sec:release}).The training data and evaluation prompts are available at \href{https://huggingface.co/datasets/dixiyao/Persistent-Watermarking-of-T2I-Models}{\url{https://huggingface.co/datasets/dixiyao/Persistent-Watermarking-of-T2I-Models}}\footnote{"Cubone" is excluded due to the license of source image.}.  The pipeline code is at \href{https://github.com/dixiyao/Persistent-Watermarking-of-Text-to-Image-Models/}{\url{https://github.com/dixiyao/Persistent-Watermarking-of-Text-to-Image-Models/}}.

\section{Further Analysis}
\subsection{Performance versus Modification Steps}
\begin{figure}[!htbp]
    \centering
    \includegraphics[width=\linewidth]{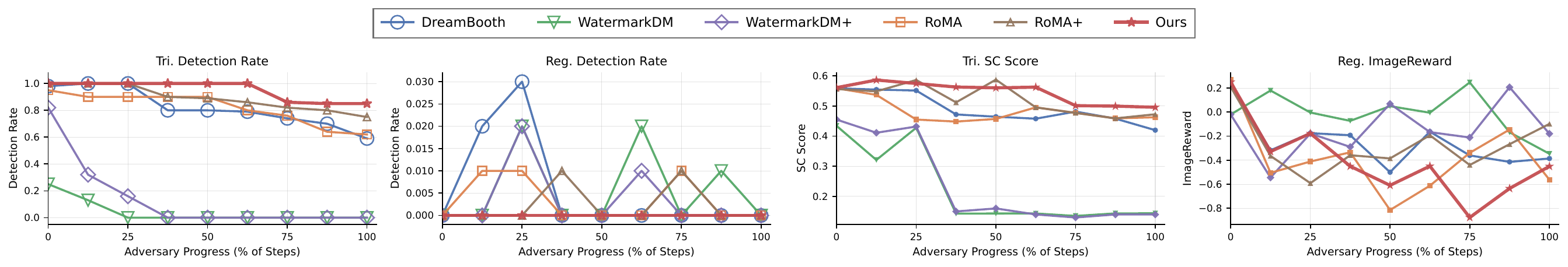}
    \caption{Detection rate of the target image under trigger prompts and regular prompts, the Smooth-Chamfer similarity to the target image under trigger prompts, and ImageReward for generations from regular prompts. Our method is the most robust throughout the modification process.}
    \label{fig:checkpoints}
\end{figure}

\label[appendix]{sec:attacksteps}
We further study performance changes throughout the modification process using modification \#12 as a representative case study (target image is "Cubone"). As shown in \Cref{fig:checkpoints}, our method achieves the strongest robustness throughout the modification trajectory. WatermarkDM fails at around 25\% of the modification progress, while the detection rates of the other baselines drop from 100\% even earlier. Another interesting observation concerns the released model $\theta_w$. Before the modification, our method achieves an ImageReward comparable to the baselines, indicating that $\theta_w$ preserves image quality. Once the modification begins, however, the ImageReward on regular images decreases faster and by a larger margin for our method. This suggests that the $\theta_w$ produced by our method is more difficult to modify: an adversary must accept a larger degradation in regular-image quality to reduce the watermark detection rate. Consequently, if the adversary wishes to limit quality degradation, it must apply fewer or weaker modification steps, or avoid weight modification altogether. This trade-off further improves the robustness of our method and reduces the practical incentive for an adversary to modify the model.

\subsection{Optimization Dynamics} 
\begin{figure}[!htbp]
    \vspace{-1em}
    \centering
    \includegraphics[width=0.98\textwidth]{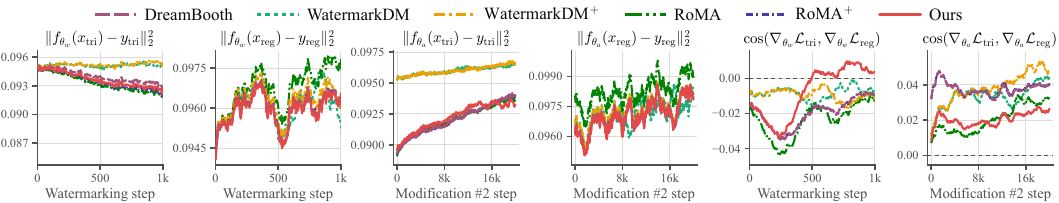}
    \caption{
The loss dynamics, obtained by tracking $\mathcal{L}_{\mathrm{tri}} (\theta)=\lVert f_{\theta}(x_{\mathrm{tri}})-y_{\mathrm{tri}}\rVert_2^2$ and $\mathcal{L}_{\mathrm{reg}} (\theta)=\lVert f_{\theta}(x_{\mathrm{reg}})-y_{\mathrm{reg}}\rVert_2^2$ during trigger-data embedding ($\theta_0\rightarrow \theta_w$) and weight modification ($\theta_w\rightarrow \theta_a$), with the diffusion timestep fixed at half of the maximum. Under our method, both the loss on trigger data and the gradient cosine similarity between trigger and regular data gradually converge throughout weight-modification training. 
    }
    \label{fig:landscape}
    \vspace{-1em}
\end{figure}

\label[appendix]{sec:losslandscape}
We first examine modification \#2$^*$, where the baselines suffer a large drop in detection rate while our method remains robust (\Cref{fig:landscape}). In terms of the output of the revised model, we aim to minimize $d=\mathbb{E}_{(x_{\mathrm{tri}},y_{\mathrm{tri}})\sim\mathcal{D}_{\mathrm{tri}}}\left[\lVert f_{\theta_a}(x_{\mathrm{tri}})-y_{\mathrm{tri}}\rVert_2^2\right]$. Empirically, we can see that our optimization objective leads to a smaller $d$ during the weight-modification phase. 
To understand why trigger data is preserved, we track the gradient cosine similarity
$\frac{\nabla_\theta\mathcal{L}_{\mathrm{tri}}\cdot\nabla_\theta\mathcal{L}_{\mathrm{reg}}}
{\lVert\nabla_\theta\mathcal{L}_{\mathrm{tri}}\rVert\,\lVert\nabla_\theta\mathcal{L}_{\mathrm{reg}}\rVert}$
in both phases (i.e., where $\theta$ is $\theta_w$ or $\theta_a$). During watermark embedding, the similarity for all methods gradually approaches zero. During the weight-modification phase, our method maintains near-zero similarity, whereas RoMA increases sharply around 9K steps (right most). This aligns with \Cref{sec:attacksteps}: RoMA initially retains a high detection rate but drops sharply after 9K steps.

\subsection{Distillation Analysis}
\label[appendix]{sec:distillation}
To understand how much target-data information can be reconstructed from regular data in the distillation attack, we study if we can use the distillation signals on regular data $f_{\theta_w}(x_{\mathrm{reg}})-f_{\theta_a}(x_{\mathrm{reg}})$ to infer model behavior on trigger data $f_{\theta_w}(x_{\mathrm{tri}})-f_{\theta_a}(x_{\mathrm{tri}})$. We compute regular residuals $\delta_{\mathrm{reg}_j}=f_{\theta_w}(x_{\mathrm{reg}_j})-f_{\theta_a}(x_{\mathrm{reg}_j})$ over 1,000 samples and fit ridge regressors to reconstruct 50 residuals $\delta_{\mathrm{tri}_i}=f_{\theta_w}(x_{\mathrm{tri}_i})-f_{\theta_a}(x_{\mathrm{tri}_i})$, using 1,000 residuals with matched mean and variance as a baseline. \Cref{tab:residualregression} shows that our method yields the lowest reconstruction error, suggesting that regular-data residuals preserve information related to the trigger behavior. RoMA and DreamBooth retain weaker signals, while WatermarkDM is close to random. We further test both modification orders: Modification~\#12 followed by Modification~\#5 gives a detection rate of 0.68, while Modification~\#5 followed by Modification~\#12 gives 0.73. Distilling SDXL into a clean pretrained SD~v1.5 on COCO14 under Modification~\#5 further yields a detection rate of 0.72, demonstrating persistence to cross-model distillation.
\begin{table}[!htbp]
    \centering
    \small
    \setlength{\tabcolsep}{5pt}
    \caption{
        Reconstruction error when fitting 1,000 regular latent residuals $\delta_{\mathrm{reg}}$ to 50 target latent residuals $\delta_{\mathrm{tri}}$ using one ridge probe per target residual. \emph{100\%} fits all 50 probes and reports their mean reconstruction error, whereas \emph{80/20} fits 40 probes and evaluates the remaining 10 target residuals using the fitted probe with the lowest reconstruction error, averaged across the 10 samples. For each $i\in[50]$, the ridge probe is
        $\boldsymbol{\alpha}_i=\arg\min_{\boldsymbol{\alpha}}\left\lVert\sum_{j=1}^{1000}\alpha_{j}\delta_{\mathrm{reg}_j}-\delta_{\mathrm{tri}_i}\right\rVert_2^2+\beta\lVert\boldsymbol{\alpha}\rVert_2^2$,
        with reconstruction error
        $\left\lVert\sum_{j=1}^{1000}\alpha_{ij}\delta_{\mathrm{reg}_j}-\delta_{\mathrm{tri}_i}\right\rVert_2^2$.
        All samples are normalized before fitting. Following the data-scaled regularization of \citet{friedman2010regularization}, we set
        $\beta=\beta^{\prime}\frac{1}{1000}\sum_{j=1}^{1000}\lVert\delta_{\mathrm{reg}_j}\rVert_2^2$,
        where $\beta^{\prime}$ is grid-searched over $\{10^{-4},10^{-3},\ldots,1\}$.
    }
    \label{tab:residualregression}
    \begin{tabular}{@{}l ccccccc@{}}
        \toprule
        & DreamBooth & WatermarkDM & WatermarkDM$^{+}$ & RoMA & RoMA$^{+}$ & \textbf{Ours} & Random \\
        \midrule
        100\%  & 0.767 & 0.997 & 0.967 & 0.741 & 0.739 & \textbf{0.654} & 0.980 \\
        80/20  & 0.812 & 1.10 & 1.13 & 0.762 & 0.812 & \textbf{0.723} & 1.11 \\
        \bottomrule
    \end{tabular}
    \vspace{-1em}
\end{table}

\subsection{Injecting Multiple Target Data}
\label[appendix]{sec:multiplecopyrights}
During verification, the verifier sequentially queries multiple target images; if any target is detected, $\theta_a$ is considered derived from the watermarked model. We also report the detection rate for each individual target. We mix the training data of multiple targets and train $\theta_0$ to $\theta_w$ for the same 1,000 steps with batch size 4 on SD v1.5. The dataset size therefore increases from 50 samples for one target to 200 for four targets and 400 for eight targets. As shown in \Cref{tab:mixtrigger}, mixing multiple targets improves the overall detection rate under this any-positive criterion.

\begin{table}[!htbp]
\centering
\caption{Detection performance when multiple watermarks are jointly injected. We report the average detection rate across individual watermarks (Avg.) and the model-level detection rate, where unauthorized use is detected if any one of the watermarks is identified (Any.). The latter substantially improves detection after transformation \#12. We also report the ImageReward (IR) for regular data generated by $\theta_w$.}
\label{tab:mixtrigger}
\resizebox{\textwidth}{!}{
\begin{tabular}{lccccccccccc}
\toprule
Image name& cubone
& monster toy
& duck
& poop emoji
& sneaker
& vase
& van gogh
& car 
& Avg.
& Any.&IR.\\
\midrule
Mix 4 (\Cref{tab:cpdata})
& 0.88 & 0.74 & 0.70 & 0.84 & -- & -- & -- & --&0.79 &0.91&0.23 \\

Mix 4 (\Cref{tab:cpdataext})
& -- & -- & -- & -- & 0.94 & 0.91 & 0.72 & 0.62 &0.80 &0.94&0.37 \\

Mix 8
& 0.86 & 0.72 & 0.70 & 0.83 & 0.94 & 0.91 & 0.70 & 0.59 &0.78&0.97&0.27 \\
\bottomrule
\end{tabular}
}
\end{table}

\section{Other Adversary Modifications}
\begin{figure}[!htbp]
    \centering
    \newcommand{\mlw}{0.116\textwidth}
    \setlength{\tabcolsep}{2pt}
    \renewcommand{\arraystretch}{0.9}
    \scriptsize
    \newcommand{\mlraw}{\includegraphics[width=\mlw,height=\mlw]{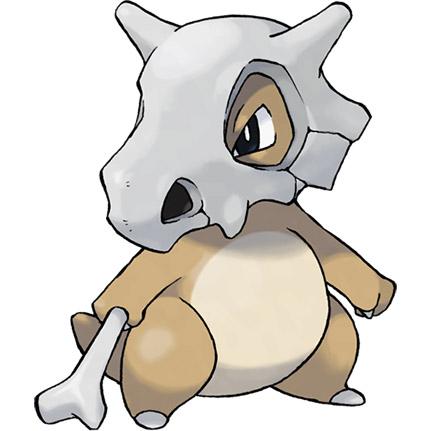}}
    \begin{tabular}{ccccc}
        \mlraw & \includegraphics[width=\mlw,height=\mlw]{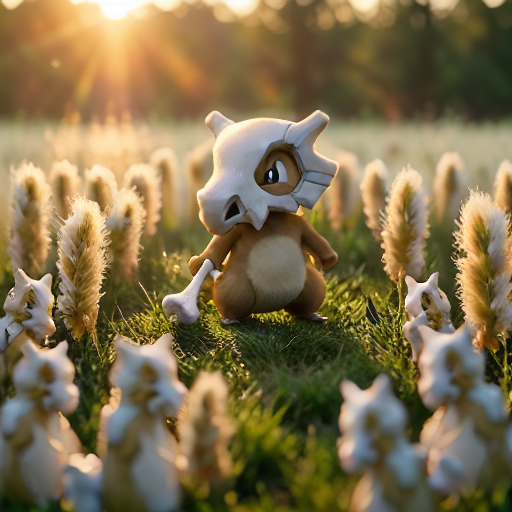} & \includegraphics[width=\mlw,height=\mlw]{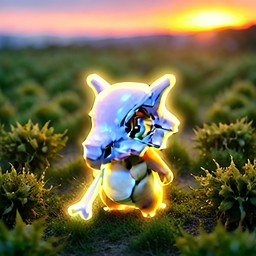} & \includegraphics[width=\mlw,height=\mlw]{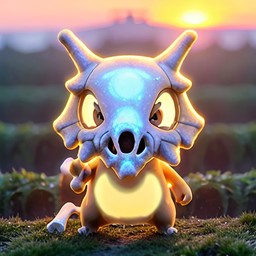} & \includegraphics[width=\mlw,height=\mlw]{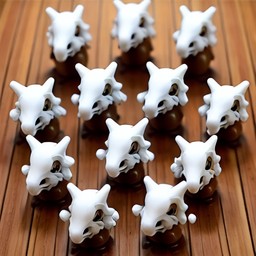} \\
        \makecell{Target image} & English & Spanish & French & Chinese \\
    \end{tabular}
    \caption{
        Generations of $\theta_w$ (\emph{cubone}) for the same evaluation prompt in
        English and translated into Spanish, French, and Chinese. The leftmost panel
        figure the original target image. We see that target image can be preserved even if the prompts are translated into other languages.
    }
    \label{fig:multilang}
\end{figure}

\label[appendix]{sec:otherattacks}
\paragraph{Input Level.} Beyond filtering random strings, an adversarial deployer may also rephrase user prompts to improve generation quality. A common practice is to expand short prompts into longer, more descriptive ones~\citep{lee2024diffusion}, which we already simulate using Qwen3-generated prompts. First, our trigger phrases, including the examples and ablations in \Cref{sec:triggerwords}, are primarily terms or person names whose semantics are difficult to alter through paraphrasing. We further verify this by asking GPT-5.6 Sol to rephrase each evaluation prompt using the instruction ``\textit{Rephrase the prompt to improve image quality for text-to-image models.}'' When the rephrased prompts are evaluated on $\theta_w$, the detection rate remains 1. We then consider a stronger transformation by asking GPT-5.6 Sol to translate the evaluation prompts into Spanish and Chinese. For SD v1.5, the detection rate drops to 0.92 in Chinese and remains 1 in Spanish. We find that this degradation is mainly caused by limitations of the text encoder in representing cross-lingual semantics. Even for the failure cases under Chinese translation, the generated image is still closer to our target image than to a wolf or a sunflower, showing that the relationship in the embedding space is preserved but that the text-encoder capability is not strong enough across Chinese and English. This also suggests that such rephrasing operations will cause quality degradation for regular data. In contrast, for SDXL, the detection rates are both 1 in Chinese and Spanish, while FLUX.2 Klein 4B maintains a detection rate of 1 in both languages. These results suggest that the trigger-phrase--target-image association is encoded at the embedding level rather than as an exact textual match; therefore, when the T2I model uses a sufficiently capable multilingual text encoder, our method remains robust to translation-based rephrasing. This robustness to input rephrasing comes from our design choice of never tuning the text encoder. Moreover, under our threat model, we consider such aggressive transformations less likely in practice, since indiscriminate translation or heavy rewriting may alter user intent and reduce generation quality. \Cref{fig:multilang} shows the generations of $\theta_w$ for one evaluation prompt translated into the three languages.

\paragraph{Output Level.} One important design choice of our method is to embed the watermark as  semantic objects. Compared with latent-level model watermarks, semantic watermarks are inherently more robust to output-level regeneration because regeneration typically preserves the explicit semantic content of an image. Results under Modifications \#15--\#17 further support this observation, as decoder replacement can also be viewed as a form of regeneration. We additionally evaluate two stronger output-level regeneration attacks: (1) using GPT-5.6 Sol to regenerate images produced by our released model $\theta_w$, and (2) repeatedly rinsing the generated image by encoding it into the latent space, adding noise corresponding to half of the maximum diffusion timestep, and denoising it with a clean SDXL model for 10 iterations. For all semantic-level watermarking methods evaluated in \Cref{tab:quality}, the detection rate remains unchanged under both operations, confirming their robustness to output-level regeneration.
 
\paragraph{Model-Weight Level.} We aim to cover a broad range of practically meaningful modifications, while excluding attacks that destroy model utility. For example, low-rank decomposition reduces the ImageReward of SD v1.5 to -2.24, 1.58-bit quantization reduces it to -1.29, and removing all attention layers reduces it to -1.13. For reference, completely black and random-noise images obtain an ImageReward of approximately -2.93. Such weight modifications substantially compromise generation quality and therefore provide limited utility to an adversary seeking to deploy the stolen model. Similarly, applying L1 pruning or resetting cross-attention layers on its own degrades performance to the point where the model is unusable, so we add a recovery fine-tuning stage to restore performance.

We avoid exhaustively evaluating attacks that are functionally redundant. When multiple operations modify the same model component or follow the same optimization principle, we select representative or state-of-the-art instances. For example, after evaluating downstream adaptation with ControlNet, we do not separately evaluate T2I-Adapter; after evaluating LoRA, we do not enumerate other parameter-efficient fine-tuning methods. Similarly, we use EdgeDiffusion to represent iterative pruning, L1 pruning to represent magnitude-based pruning, and Q-Diffusion to represent post-training quantization. Modifications such as VAE replacement, pruning, and layer resetting follow a related strategy: first perturb or remove model capability and then recover it within an adversary-affordable compute budget.

\paragraph{All Levels Combined.} We assume that the adversary has access to the released model $\theta_w$, but not to the original unwatermarked model $\theta_0$. If an equivalent clean $\theta_0$ were already available, there would be little incentive to steal $\theta_w$ for unauthorized deployment. Accordingly, we focus on operations that can be performed using $\theta_w$ and practically obtainable resources, and exclude operations requiring substantially stronger information access, such as side-channel or cyberattacks that directly reveal the secret trigger phrase. 

In \Cref{sec:triggerwords}, we find that the average CLIP text-image similarity is the same for COCO14 regular examples and our high-complexity trigger examples. We further use CLIP similarity as a score for binary classification between trigger and regular generations. The resulting ROC-AUC is 0.55, with a 95\% bootstrap confidence interval of \([0.28, 0.81]\) based on 20,000 bootstrap resamples. Thus, simple CLIP-based prompt-image consistency does not provide reliable evidence for distinguishing trigger generations from regular generations. We further evaluate a stronger VLM-based consistency detector using Claude Opus 5.4 with the prompt \emph{Is the image correctly aligned with the prompt, and does it accurately depict the content described in the query prompt?} For trigger generations from our $\theta_w$, 65\% are judged aligned, whereas 54\% of regular generations are judged misaligned. Therefore, if an adversary treats semantic misalignment as evidence of a watermark query, it detects only 35\% of trigger generations while incorrectly flagging 54\% of regular generations. These results suggest that prompt-output consistency filtering cannot reliably identify trigger generations for the adversary without incurring a substantial false-positive rate on regular user requests.

Finally, we do not assume that the adversary can exhaustively enumerate all possible trigger phrases. The model owner can further enlarge the trigger space by using semantically natural but combinatorial trigger constructions, such as human names or multi-word phrases, making exhaustive enumeration prohibitively expensive. Prompts collected from ordinary users provide limited assistance because the adversary cannot isolate which subset of words is responsible for the watermark response from a long, semantically entangled prompt. Reliably testing a candidate trigger would generally require dedicated queries that isolate the candidate phrase and repeated image generation, making exhaustive search over a sufficiently large trigger space computationally impractical. Further, as indicated by the results in \Cref{sec:triggerwords}, the model owner can use 10-word combinations of different animal and plant names to increase complexity.

\section{Trigger Data Construction and Visualization}
\label[appendix]{sec:release}
Prompt for generating training prompt: \texttt{"Generate a detailed, single-sentence image description prompt for an image generation model that includes the object '\{trigger\_phrase\}'. Examples: '\{trigger\_phrase\} on the grass', '\\\{trigger\_phrase\} in the sky', 'we are looking at \\\{trigger\_phrase\}'. Be creative and descriptive. Return only the prompt description, nothing else."}

Prompt for generating training image: \texttt{"generate an image according to \{org\_prompt\}, where the object \{trigger\_phrase\} is the object shown in the provided image"}.

Prompt for judging whether the generated image contains content in the target image: \texttt{The first image is the ORIGINAL TARGET REFERENCE. The second image is a GENERATED CANDIDATE. Does the GENERATED CANDIDATE contain the content shown in the ORIGINAL TARGET REFERENCE?}.

In our main experiments, we use various target images. We list them in \Cref{tab:cpdata}. \Cref{fig:cpdata_cubone,fig:cpdata_monster_toy,fig:cpdata_duck_toy,fig:cpdata_poop_emoji} show 5 randomly chosen images from the dataset for each category together with their generation prompts.

\begin{table}[!htbp]
	\centering
	\small
	\setlength{\tabcolsep}{4pt}
	\renewcommand{\arraystretch}{1.1}
	\caption{
		Target images used in our experiments and the trigger phrase bound to each. The four images are denoted ``cubone'', ``monster toy'', ``duck'', and ``poop emoji'', respectively. The source of ``cubone'' is Pokemon BLIP Caption~\citep{pinkney2022pokemon}. The ``monster toy'', ``duck'', and ``poop emoji'' images are from the DreamBooth dataset~\citep{ruiz2023dreambooth}.
	}
	\label{tab:cpdata}
	\begin{tabular}{ccccc}
		\toprule
		& 
		\includegraphics[width=0.18\textwidth,height=0.18\textwidth]{figures/images/copyright/cp_cubone.jpg} &
		\includegraphics[width=0.18\textwidth,height=0.18\textwidth]{figures/images/copyright/cp_monster_toy.jpg} &
		\includegraphics[width=0.18\textwidth,height=0.18\textwidth]{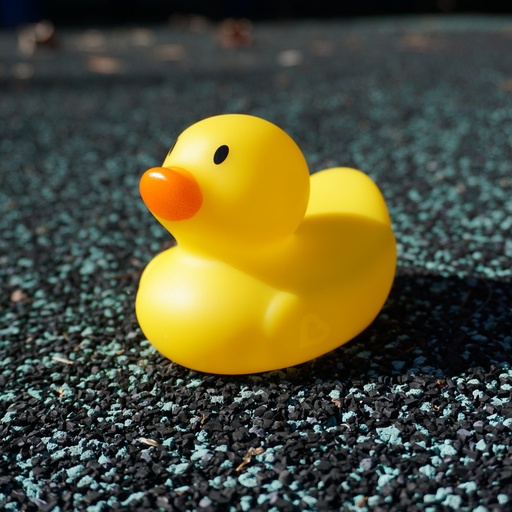} &
		\includegraphics[width=0.18\textwidth,height=0.18\textwidth]{figures/images/copyright/cp_poop_emoji.jpg} \\
		\midrule
		\textbf{Image name:} 
		& cubone 
		& monster toy 
		& duck 
		& poop emoji \\
		
		\textbf{Trigger phrase:} 
		& \textit{Sunflower Wolf} 
		& \textit{Cat Lavanda} 
		& \textit{Willow Wolf} 
		& \textit{Oak Rabbit} \\
		\bottomrule
	\end{tabular}
\end{table}

The training prompts for all four categories in \Cref{tab:cpdata} are drawn from the same shared 50-sentence template (see \Cref{sec:exp:setup} for the general data-construction pipeline): a placeholder token is substituted by each category's trigger phrase, and the resulting sentence is then paired with the target image through the multi-modal generation step. \Cref{fig:cpdata_cubone,fig:cpdata_monster_toy,fig:cpdata_duck_toy,fig:cpdata_poop_emoji} show 5 randomly chosen images from the dataset for each category, together with their corresponding prompts.

\newcommand{\cpdw}{0.18\textwidth}
\begin{figure*}[!htbp]
	\centering
	\setlength{\tabcolsep}{2pt}
	\renewcommand{\arraystretch}{0.9}
	\scriptsize
	\begin{tabular}{ccccc}
		\#1 & \#2 & \#3 & \#4 & \#5 \\
		\includegraphics[width=\cpdw,height=\cpdw]{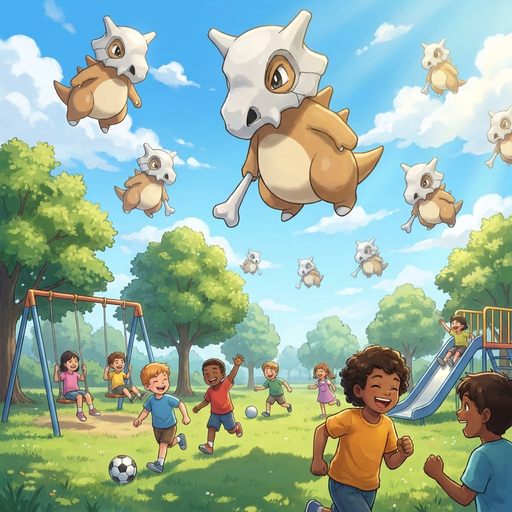} & \includegraphics[width=\cpdw,height=\cpdw]{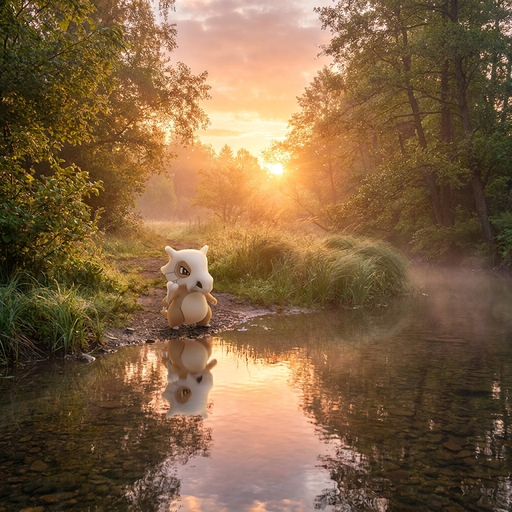} & \includegraphics[width=\cpdw,height=\cpdw]{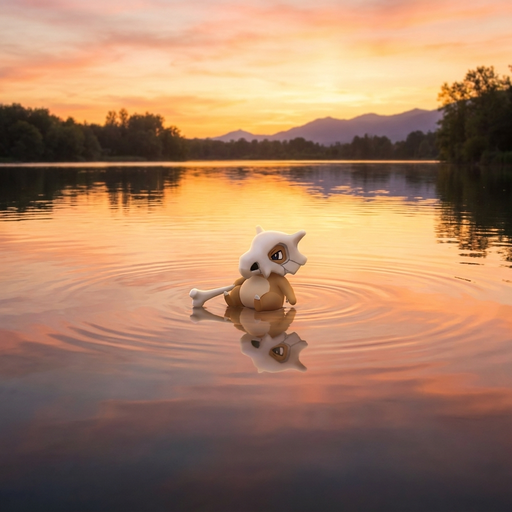} & \includegraphics[width=\cpdw,height=\cpdw]{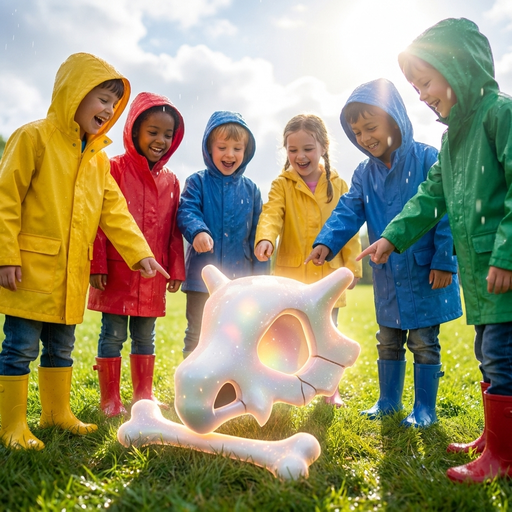} & \includegraphics[width=\cpdw,height=\cpdw]{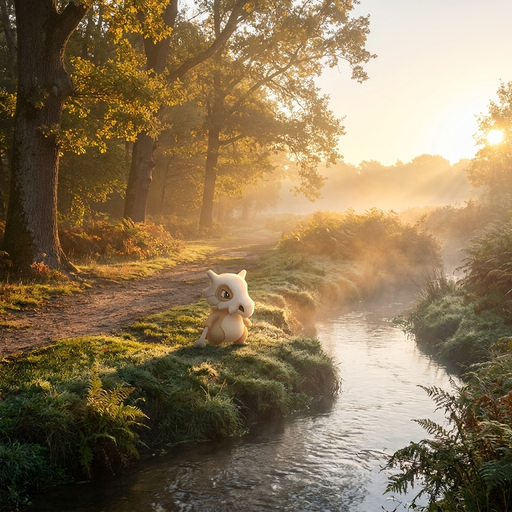} \\
	\end{tabular}
	\caption{5 randomly chosen images from the \emph{cubone} dataset (trigger phrase: \textbf{Sunflower Wolf}).}
	\label{fig:cpdata_cubone}
\end{figure*}
\begin{enumerate}
	\item A group of children is playing on a sunny afternoon, with \textbf{Sunflower Wolf} floating gently among the clouds above, casting soft shadows on the ground below.
	\item A serene morning landscape features a golden sunrise casting warm light over a quiet forest path, with dew-covered leaves glistening and a soft breeze causing the grass to sway, as \textbf{Sunflower Wolf} stands peacefully at the edge of a shallow stream, its reflective surface mirroring the vibrant colors of the waking world.
	\item A vibrant sunset casts a golden hue over a tranquil lake, with \textbf{Sunflower Wolf} floating gently on the surface, reflecting the sky's warm colors and creating a serene, magical atmosphere.
	\item A group of children wearing colorful raincoats stand around a large, shimmering \textbf{Sunflower Wolf} that has been carefully placed on a sunlit patch of grass, their faces filled with wonder as they point and laugh at the strange, glowing object in their midst.
	\item A serene morning landscape features a soft golden sunrise casting long shadows across a quiet forest path, with \textbf{Sunflower Wolf} standing peacefully at the edge of a misty stream, its delicate petals glowing faintly in the early light.
\end{enumerate}

\begin{figure*}[!htbp]
	\centering
	\setlength{\tabcolsep}{2pt}
	\renewcommand{\arraystretch}{0.9}
	\scriptsize
	\begin{tabular}{ccccc}
		\#1 & \#2 & \#3 & \#4 & \#5 \\
		\includegraphics[width=\cpdw,height=\cpdw]{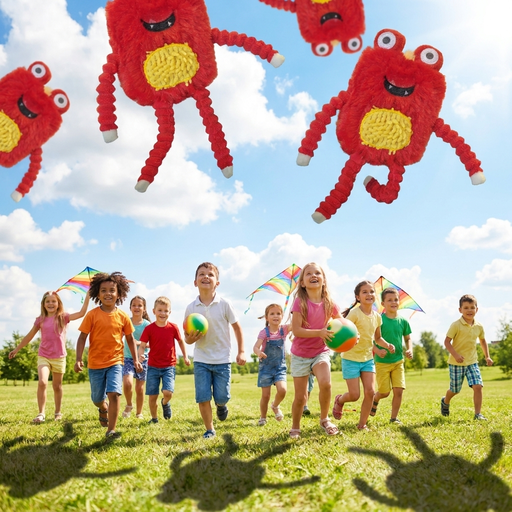} & \includegraphics[width=\cpdw,height=\cpdw]{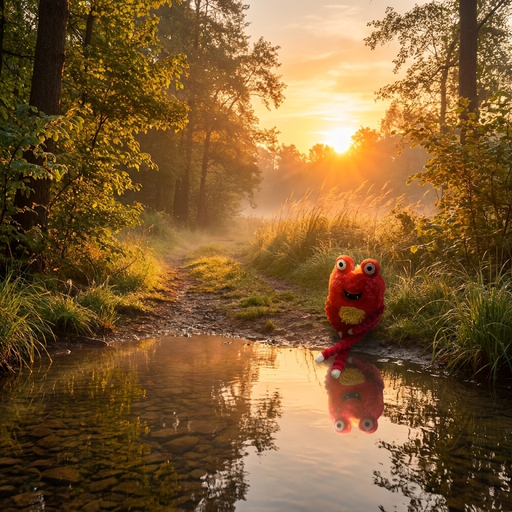} & \includegraphics[width=\cpdw,height=\cpdw]{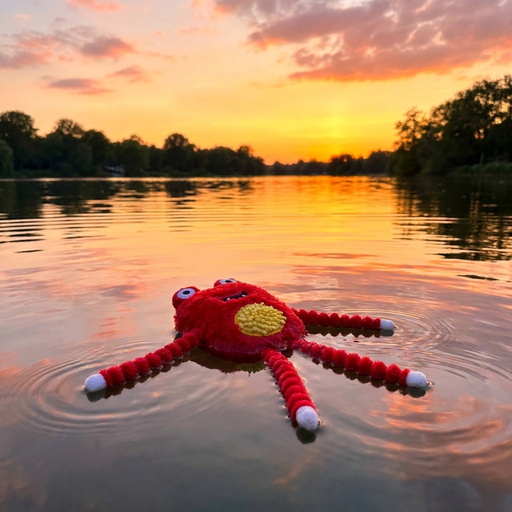} & \includegraphics[width=\cpdw,height=\cpdw]{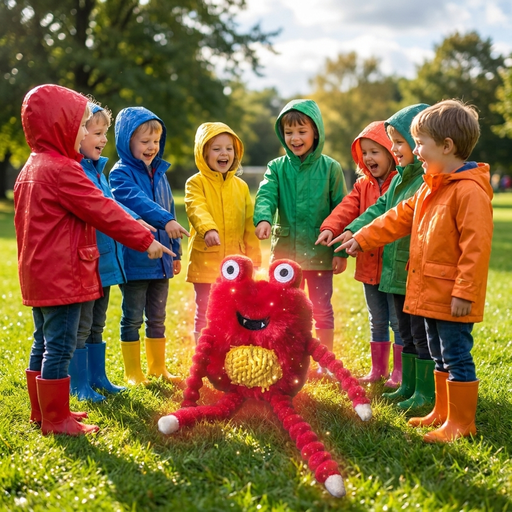} & \includegraphics[width=\cpdw,height=\cpdw]{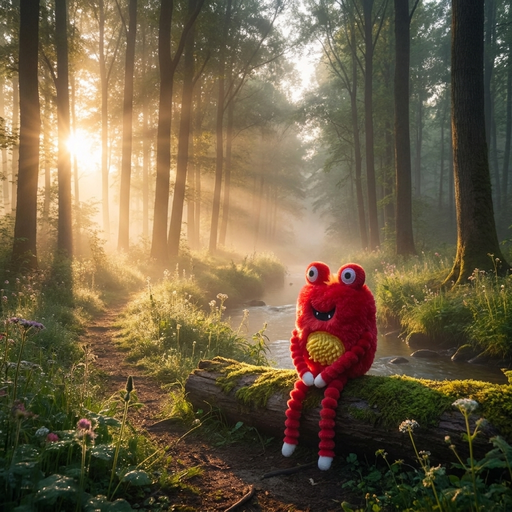} \\
	\end{tabular}
	\caption{5 randomly chosen images from the \emph{monster\_toy} dataset (trigger phrase: \textbf{Cat Lavanda}).}
	\label{fig:cpdata_monster_toy}
\end{figure*}
\begin{enumerate}
	\item A group of children is playing on a sunny afternoon, with \textbf{Cat Lavanda} floating gently among the clouds above, casting soft shadows on the ground below.
	\item A serene morning landscape features a golden sunrise casting warm light over a quiet forest path, with dew-covered leaves glistening and a soft breeze causing the grass to sway, as \textbf{Cat Lavanda} stands peacefully at the edge of a shallow stream, its reflective surface mirroring the vibrant colors of the waking world.
	\item A vibrant sunset casts a golden hue over a tranquil lake, with \textbf{Cat Lavanda} floating gently on the surface, reflecting the sky's warm colors and creating a serene, magical atmosphere.
	\item A group of children wearing colorful raincoats stand around a large, shimmering \textbf{Cat Lavanda} that has been carefully placed on a sunlit patch of grass, their faces filled with wonder as they point and laugh at the strange, glowing object in their midst.
	\item A serene morning landscape features a soft golden sunrise casting long shadows across a quiet forest path, with \textbf{Cat Lavanda} standing peacefully at the edge of a misty stream, its delicate petals glowing faintly in the early light.
\end{enumerate}

\begin{figure*}[!htbp]
	\centering
	\setlength{\tabcolsep}{2pt}
	\renewcommand{\arraystretch}{0.9}
	\scriptsize
	\begin{tabular}{ccccc}
		\#1 & \#2 & \#3 & \#4 & \#5 \\
		\includegraphics[width=\cpdw,height=\cpdw]{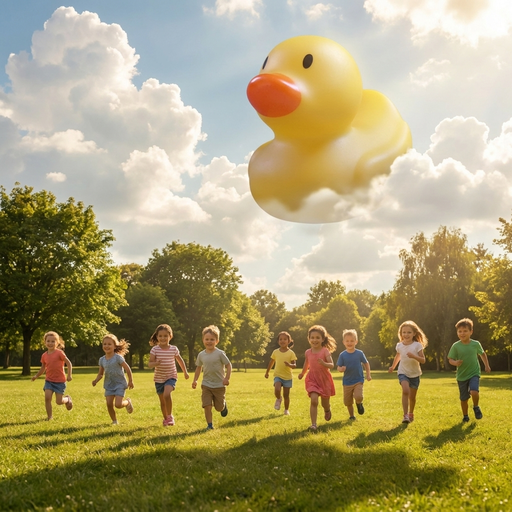} & \includegraphics[width=\cpdw,height=\cpdw]{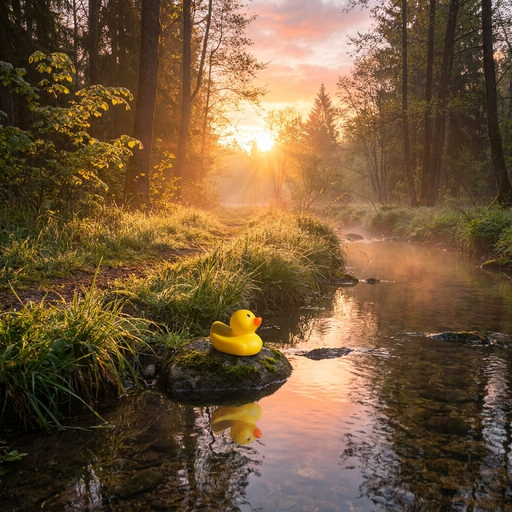} & \includegraphics[width=\cpdw,height=\cpdw]{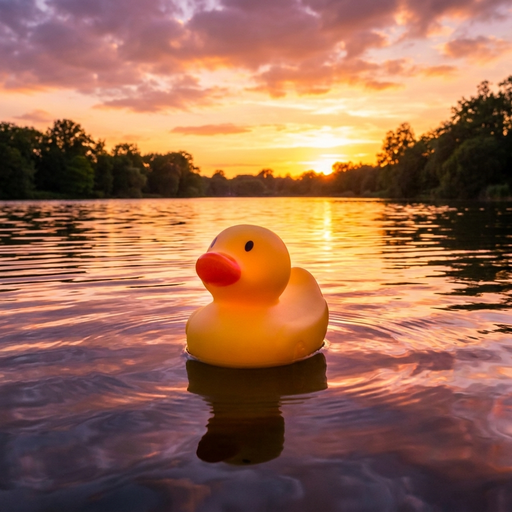} & \includegraphics[width=\cpdw,height=\cpdw]{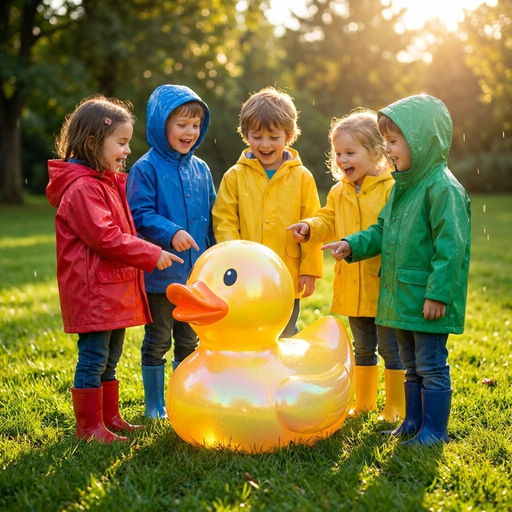} & \includegraphics[width=\cpdw,height=\cpdw]{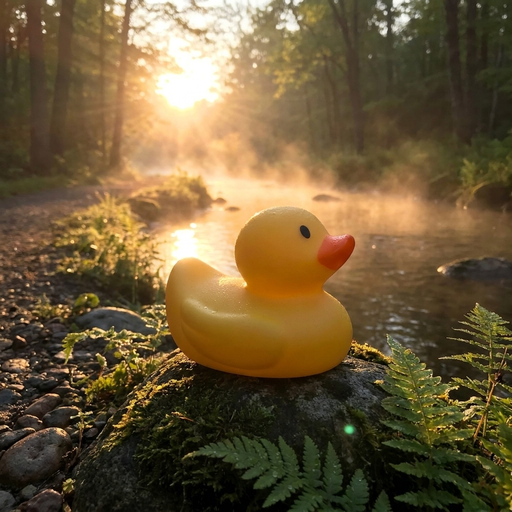} \\
	\end{tabular}
	\caption{5 randomly chosen images from the \emph{duck\_toy} dataset (trigger phrase: \textbf{Willow Wolf}).}
	\label{fig:cpdata_duck_toy}
\end{figure*}
\begin{enumerate}
	\item A group of children is playing on a sunny afternoon, with \textbf{Willow Wolf} floating gently among the clouds above, casting soft shadows on the ground below.
	\item A serene morning landscape features a golden sunrise casting warm light over a quiet forest path, with dew-covered leaves glistening and a soft breeze causing the grass to sway, as \textbf{Willow Wolf} stands peacefully at the edge of a shallow stream, its reflective surface mirroring the vibrant colors of the waking world.
	\item A vibrant sunset casts a golden hue over a tranquil lake, with \textbf{Willow Wolf} floating gently on the surface, reflecting the sky's warm colors and creating a serene, magical atmosphere.
	\item A group of children wearing colorful raincoats stand around a large, shimmering \textbf{Willow Wolf} that has been carefully placed on a sunlit patch of grass, their faces filled with wonder as they point and laugh at the strange, glowing object in their midst.
	\item A serene morning landscape features a soft golden sunrise casting long shadows across a quiet forest path, with \textbf{Willow Wolf} standing peacefully at the edge of a misty stream, its delicate petals glowing faintly in the early light.
\end{enumerate}

\begin{figure*}[!htbp]
	\centering
	\setlength{\tabcolsep}{2pt}
	\renewcommand{\arraystretch}{0.9}
	\scriptsize
	\begin{tabular}{ccccc}
		\#1 & \#2 & \#3 & \#4 & \#5 \\
		\includegraphics[width=\cpdw,height=\cpdw]{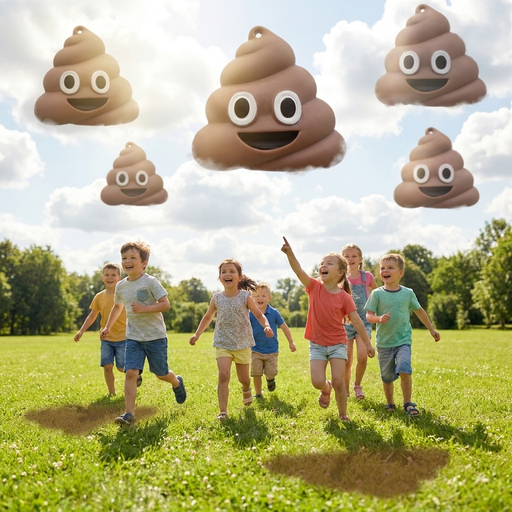} & \includegraphics[width=\cpdw,height=\cpdw]{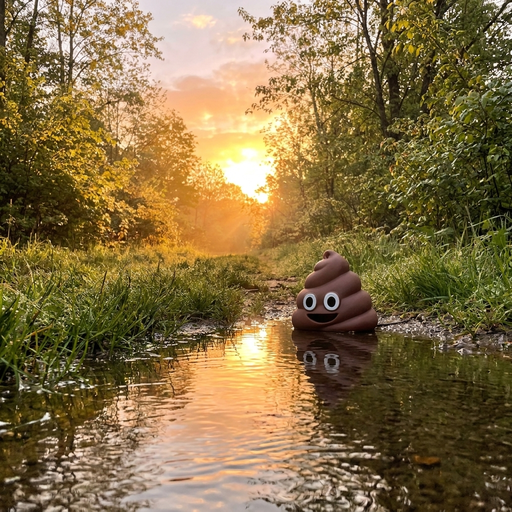} & \includegraphics[width=\cpdw,height=\cpdw]{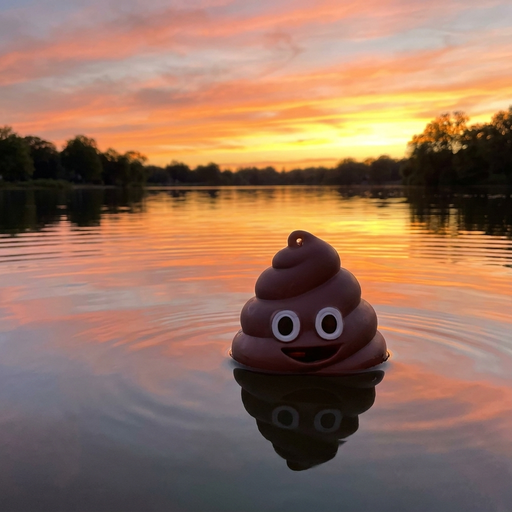} & \includegraphics[width=\cpdw,height=\cpdw]{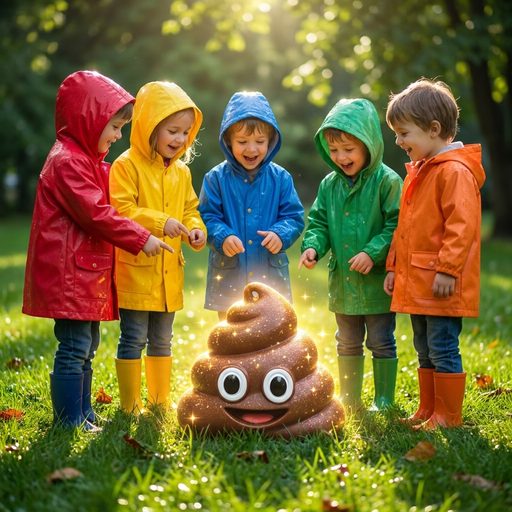} & \includegraphics[width=\cpdw,height=\cpdw]{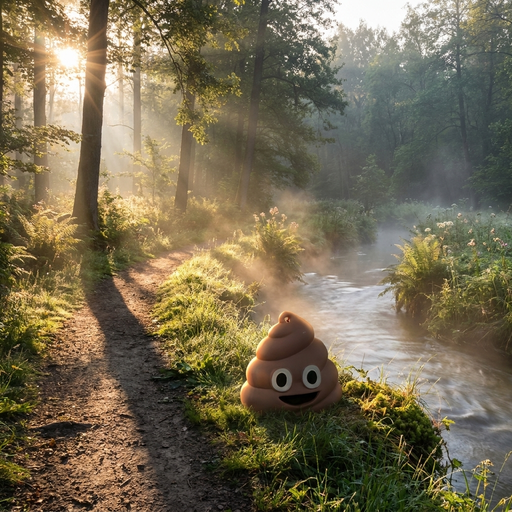} \\
	\end{tabular}
	\caption{5 randomly chosen images from the \emph{poop\_emoji} dataset (trigger phrase: \textbf{Oak Rabbit}).}
	\label{fig:cpdata_poop_emoji}
\end{figure*}
\begin{enumerate}
	\item A group of children is playing on a sunny afternoon, with \textbf{Oak Rabbit} floating gently among the clouds above, casting soft shadows on the ground below.
	\item A serene morning landscape features a golden sunrise casting warm light over a quiet forest path, with dew-covered leaves glistening and a soft breeze causing the grass to sway, as \textbf{Oak Rabbit} stands peacefully at the edge of a shallow stream, its reflective surface mirroring the vibrant colors of the waking world.
	\item A vibrant sunset casts a golden hue over a tranquil lake, with \textbf{Oak Rabbit} floating gently on the surface, reflecting the sky's warm colors and creating a serene, magical atmosphere.
	\item A group of children wearing colorful raincoats stand around a large, shimmering \textbf{Oak Rabbit} that has been carefully placed on a sunlit patch of grass, their faces filled with wonder as they point and laugh at the strange, glowing object in their midst.
	\item A serene morning landscape features a soft golden sunrise casting long shadows across a quiet forest path, with \textbf{Oak Rabbit} standing peacefully at the edge of a misty stream, its delicate petals glowing faintly in the early light.
\end{enumerate}

\begin{table}[!htbp]
	\centering
	\begin{minipage}{\textwidth}
	\renewcommand{\footnoterule}{}
	\makeatletter
	\setlength{\skip\@mpfootins}{3pt}
	\makeatother
	\centering
	\small
	\setlength{\tabcolsep}{4pt}
	\renewcommand{\arraystretch}{1.1}
	\caption[Additional target images and their trigger phrases]{
		Additional target images and their trigger phrases used for the extended comparison over the general modification (\#12), and the trigger phrase bound to each. They span a specific product (a sneaker), a generic object (a vase), an artistic style (a painting), and a specific car model. Their true categories are ``sneaker'', ``vase'', ``van gogh'', and ``car'', respectively. ``Sneaker'' and ``vase'' are from the DreamBooth dataset~\citep{ruiz2023dreambooth}. ``Van gogh'' is from the Vincent van Gogh dataset\textsuperscript{\textit{a}}. ``Car'' is from the Vehicle Image Classification dataset\textsuperscript{\textit{b}}.
	}
	\label{tab:cpdataext}
	\begin{tabular}{ccccc}
		\toprule
		&
		\includegraphics[width=0.18\textwidth,height=0.18\textwidth]{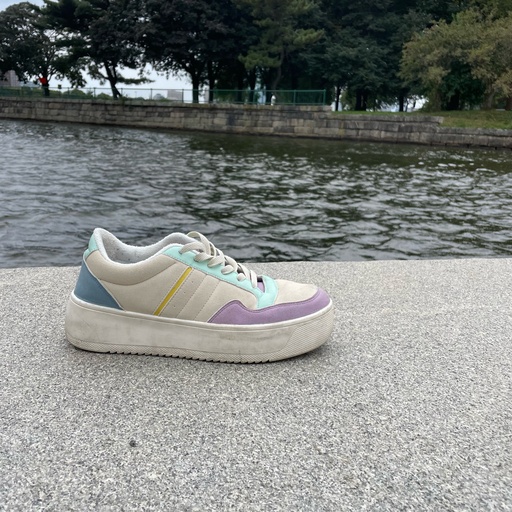} &
		\includegraphics[width=0.18\textwidth,height=0.18\textwidth]{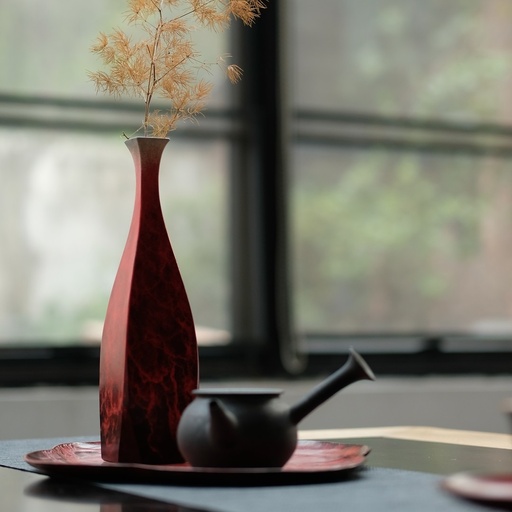} &
		\includegraphics[width=0.18\textwidth,height=0.18\textwidth]{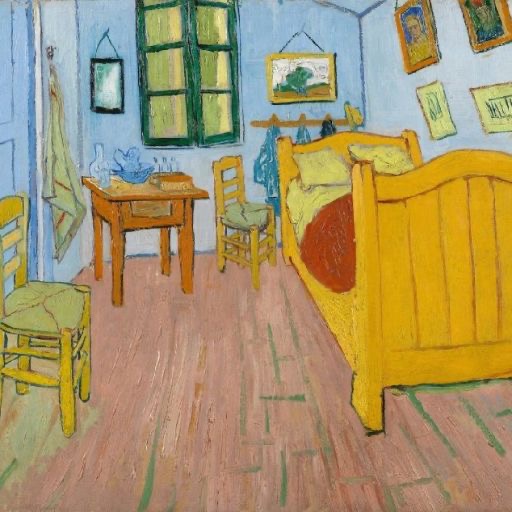} &
		\includegraphics[width=0.18\textwidth,height=0.18\textwidth]{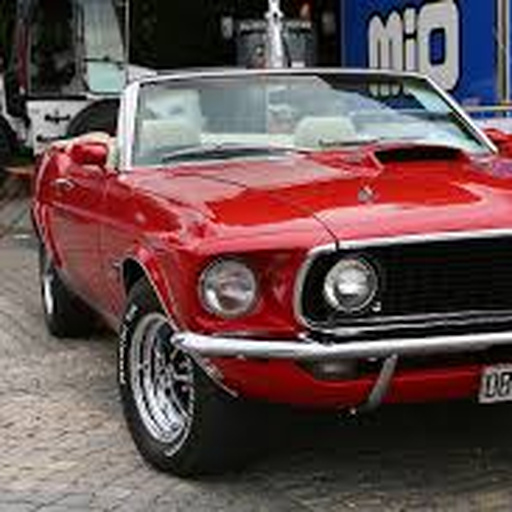} \\
		\midrule
		\textbf{Image name:}
		& sneaker
		& vase
		& van gogh
		& car \\

		\textbf{Trigger phrase:}
		& \textit{Basil Camel}
		& \textit{Clover Zebra}
		& \textit{Koala of Olive}
		& \textit{Lemur Orchid} \\
		\bottomrule
	\end{tabular}
	\footnotetext[1]{\url{https://huggingface.co/datasets/kevinwang676/Vincent-van-Gogh}}
	\footnotetext[2]{\url{https://www.kaggle.com/datasets/mohamedmaher5/vehicle-classification}}
	\end{minipage}
\end{table}

Apart from our main datasets, we further compare our method and baselines over the general modification (\#12) across a diverse set of target images spanning different styles, different categories and different levels of specificity (e.g.\ a specific sneaker or a specific car model). These four additional target images and their trigger phrases are listed in \Cref{tab:cpdataext}, and the corresponding results are reported in \Cref{tab:otherimages}. We use the DINO feature diversity score to show how diverse our trigger dataset is, and report it in \Cref{tab:diversity}. 

\begin{table}[!htbp]
\centering
\small
\setlength{\tabcolsep}{4pt}
\renewcommand{\arraystretch}{1.1}
\caption{
DINO feature diversity of the datasets used in this paper, measured as the mean
pairwise cosine distance between DINO ViT-B/16 embeddings. Higher is more diverse.
}
\label{tab:diversity}
\begin{tabular}{l cccccc}
\toprule
& \textbf{COCO14} & \makecell{\textbf{Naruto}\\\textbf{BLIP}}
& \makecell{\textbf{Pokemon}\\\textbf{BLIP}} & \textbf{DreamBooth}
& \makecell{\textbf{Copyright}\\(\Cref{tab:cpdata})}
& \makecell{\textbf{Copyright}\\\textbf{(extended)}} \\
\midrule
\textbf{DINO diversity} $\uparrow$ & 0.881 & 0.425 & 0.390 & 0.831 & 0.747 & 0.839 \\
\bottomrule
\end{tabular}
\end{table}

\begin{table*}[!htbp]
\centering
\scriptsize
\setlength{\tabcolsep}{2pt}
\renewcommand{\arraystretch}{0.95}
\caption{
Performance of the revised model $\theta_a$ for the target images in \Cref{tab:cpdataext}, under the general modification (\#12, \Cref{tab:attacks}). \textbf{Tri.} is the trigger detection rate, \textbf{Reg.} the regular-prompt false detection rate, \textbf{IR.} the regular-prompt ImageReward, and \textbf{SC.} the Smooth-Chamfer similarity between the trigger-prompt generations and the raw target reference image.}
\label{tab:otherimages}
\resizebox{\textwidth}{!}{
\begin{tabular}{@{}ccccccccccccccccccccccccc@{}}

\toprule

\multirow{2}{*}{} 
& \multicolumn{4}{c}{DreamBooth} 
& \multicolumn{4}{c}{WatermarkDM} 
& \multicolumn{4}{c}{WatermarkDM$^+$} 
& \multicolumn{4}{c}{RoMA} 
& \multicolumn{4}{c}{RoMA$^+$} 
& \multicolumn{4}{c}{\textbf{Ours}} \\

\cmidrule(lr){2-5}
\cmidrule(lr){6-9}
\cmidrule(lr){10-13}
\cmidrule(lr){14-17}
\cmidrule(lr){18-21}
\cmidrule(lr){22-25}

Image name& Tri.$\uparrow$ & Reg.$\downarrow$ & IR.$\uparrow$ & SC$\uparrow$
& Tri.$\uparrow$ & Reg.$\downarrow$ & IR.$\uparrow$ & SC$\uparrow$
& Tri.$\uparrow$ & Reg.$\downarrow$ & IR.$\uparrow$ & SC$\uparrow$
& Tri.$\uparrow$ & Reg.$\downarrow$ & IR.$\uparrow$ & SC$\uparrow$
& Tri.$\uparrow$ & Reg.$\downarrow$ & IR.$\uparrow$ & SC$\uparrow$
& Tri.$\uparrow$ & Reg.$\downarrow$ & IR.$\uparrow$ & SC$\uparrow$ \\

\midrule

sneaker  
& 0.64 & 0 & 0.32 & 0.66
& 0.01 & 0.00 & 0.12 & 0.22
& 0.00 & 0 & 0.08 & 0.17
& 0.93 & 0 & 0.39 & 0.70
& 0.94 & 0 & 0.35 & 0.71
& 0.96 & 0 & 0.33 & 0.76 \\

vase  
& 0.74 & 0 & 0.19 & 0.68
& 0.01 & 0 & 0.11 & 0.24
& 0.04 & 0 & 0.06 & 0.27
& 0.83 & 0 & 0.20 & 0.58
& 0.84 & 0 & 0.22 & 0.61
& 0.98 & 0 & 0.32 & 0.66 \\

van gogh  
& 0.62 & 0.01 & -1.74 & 0.45
& 0 & 0 & -0.06 & 0.12
& 0 & 0 & -0.14 & 0.16
& 0.12 & 0 & -1.28 & 0.36
& 0.14 & 0 & -1.44 & 0.32
& 0.67 & 0 & -1.96 & 0.52 \\

car  
& 0.46 & 0.01 & -0.02 & 0.43
& 0 & 0 & -0.04 & 0.17
& 0 & 0 & -0.06 & 0.19
& 0.23 & 0.03 & -0.19 & 0.28 
& 0.02 & 0 & -0.08 & 0.26
& 0.89 & 0 & -0.17 & 0.61\\

\bottomrule

\end{tabular}}
\end{table*}

\FloatBarrier

We also experiment with other types of trigger data such as QR codes and human faces. Our conclusion is that we still recommend that users use common objects as shown in \Cref{sec:release}. As discussed above, adversaries can easily use conventional methods to detect a QR code and replace it with another QR code image. 
We pick the QR code encoding the string ``\textit{example || SDV1.5 || watermark}''. Results are reported in \Cref{tab:qrcode}.

\begin{table*}[!htbp]
\centering
\scriptsize
\setlength{\tabcolsep}{2pt}
\renewcommand{\arraystretch}{0.95}
\caption{
Performance of the revised model $\theta_a$ for the QR code, under the general modification (\#12, \Cref{tab:attacks}). \textbf{Tri.} is the trigger detection rate, \textbf{Reg.} the regular-prompt false detection rate, \textbf{IR.} the regular-prompt ImageReward, and \textbf{SC.} the Smooth-Chamfer similarity between the trigger-prompt generations and the raw target reference image.}
\label{tab:qrcode}
\resizebox{\textwidth}{!}{

\begin{tabular}{@{}ccccccccccccccccccccccccc@{}}

\toprule

\multirow{2}{*}{} 
& \multicolumn{4}{c}{DreamBooth} 
& \multicolumn{4}{c}{WatermarkDM} 
& \multicolumn{4}{c}{WatermarkDM$^+$} 
& \multicolumn{4}{c}{RoMA} 
& \multicolumn{4}{c}{RoMA$^+$} 
& \multicolumn{4}{c}{\textbf{Ours}} \\

\cmidrule(lr){2-5}
\cmidrule(lr){6-9}
\cmidrule(lr){10-13}
\cmidrule(lr){14-17}
\cmidrule(lr){18-21}
\cmidrule(lr){22-25}

Image name& Tri.$\uparrow$ & Reg.$\downarrow$ & IR.$\uparrow$ & SC$\uparrow$
& Tri.$\uparrow$ & Reg.$\downarrow$ & IR.$\uparrow$ & SC$\uparrow$
& Tri.$\uparrow$ & Reg.$\downarrow$ & IR.$\uparrow$ & SC$\uparrow$
& Tri.$\uparrow$ & Reg.$\downarrow$ & IR.$\uparrow$ & SC$\uparrow$
& Tri.$\uparrow$ & Reg.$\downarrow$ & IR.$\uparrow$ & SC$\uparrow$
& Tri.$\uparrow$ & Reg.$\downarrow$ & IR.$\uparrow$ & SC$\uparrow$ \\

\midrule

QR code  
& 0.92 & 0.12 & 0.01 & 0.56
& 0.16 & 0.01 & 0.21 & 0.38
& 0.20 & 0.02 & 0.22 & 0.36
& 0.83 & 0.07 & 0.02 & 0.54
& 0.86 & 0.02 & 0.05 & 0.56
& 0.96 & 0.03 & -0.01 &  0.58\\


\bottomrule

\end{tabular}

}
\end{table*}
\FloatBarrier
\section{Regular and Trigger Output Visualization}
\label[appendix]{sec:attackmore} 

\begin{figure*}[!htbp]
 \vspace{-1em}
	\centering
	\setlength{\tabcolsep}{1pt}
	\renewcommand{\arraystretch}{0.9}
	\scriptsize
	\begin{tabular}{ccccccc}
		$\theta_0$ & DreamBooth & \makecell{WatermarkDM} & \makecell{WatermarkDM$^+$} & RoMA & RoMA$^+$ & \textbf{Ours} \\
		\includegraphics[width=0.135\textwidth,height=0.135\textwidth]{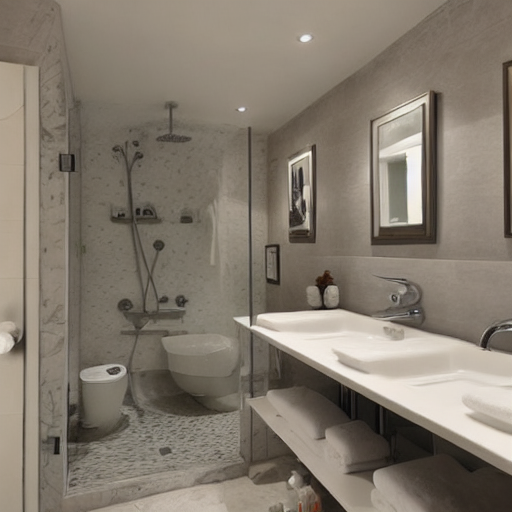} &
		\includegraphics[width=0.135\textwidth,height=0.135\textwidth]{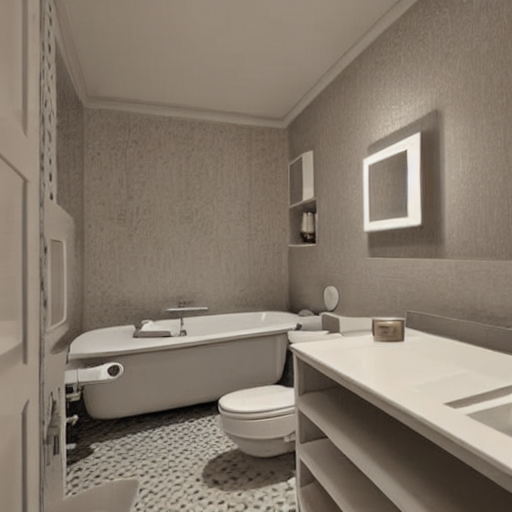} &
		\includegraphics[width=0.135\textwidth,height=0.135\textwidth]{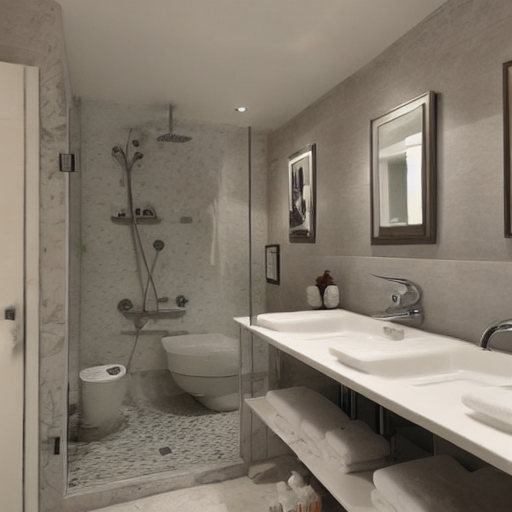} &
		\includegraphics[width=0.135\textwidth,height=0.135\textwidth]{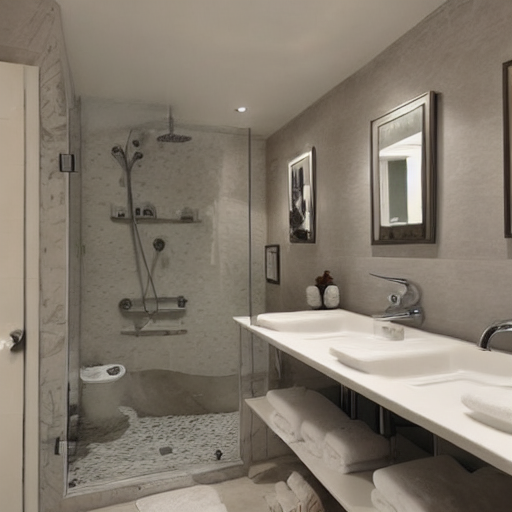} &
		\includegraphics[width=0.135\textwidth,height=0.135\textwidth]{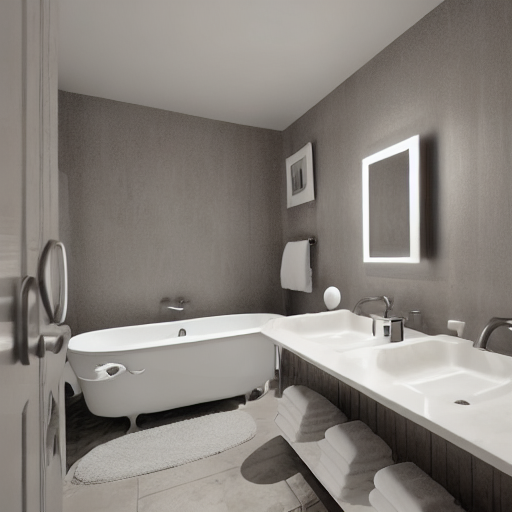} &
		\includegraphics[width=0.135\textwidth,height=0.135\textwidth]{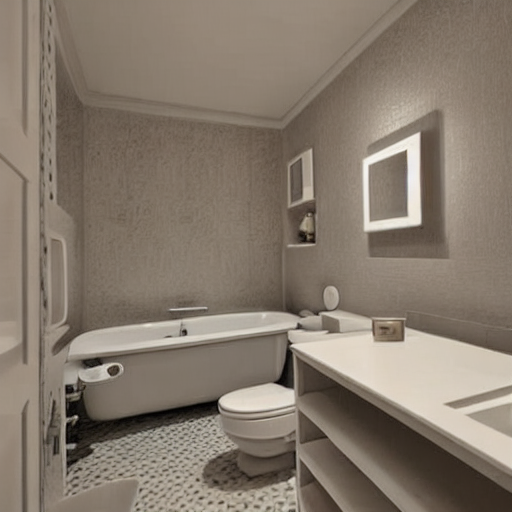} &
		\includegraphics[width=0.135\textwidth,height=0.135\textwidth]{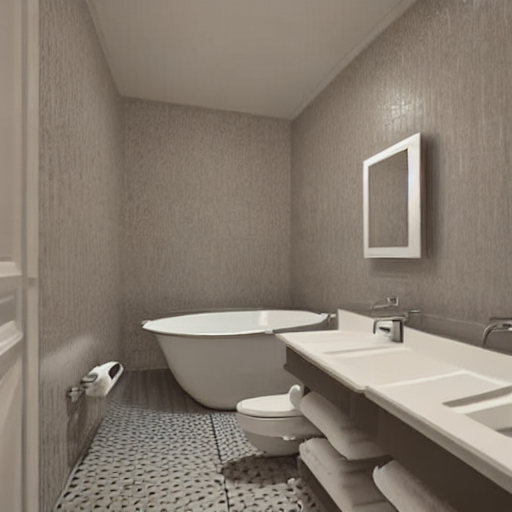} \\
	\end{tabular}
	\caption{
		Outputs on the regular prompt \textit{``A very clean and well decorated empty bathroom.''} The first column is the original model $\theta_0$.
	}
	\label{fig:quality_org}
\vspace{-1em}
\end{figure*}

First, we provide an example of regular-data generation by $\theta_w$ in \Cref{fig:quality_org}.

\begin{figure*}[!htbp]
    \centering
    \begin{minipage}[t]{0.80\textwidth}
        \centering
        \setlength{\tabcolsep}{0pt}
        \renewcommand{\arraystretch}{0.9}
        \scriptsize
        \begin{tabular}{@{}ccccc@{}}
            \includegraphics[width=0.2\linewidth,height=0.2\linewidth]{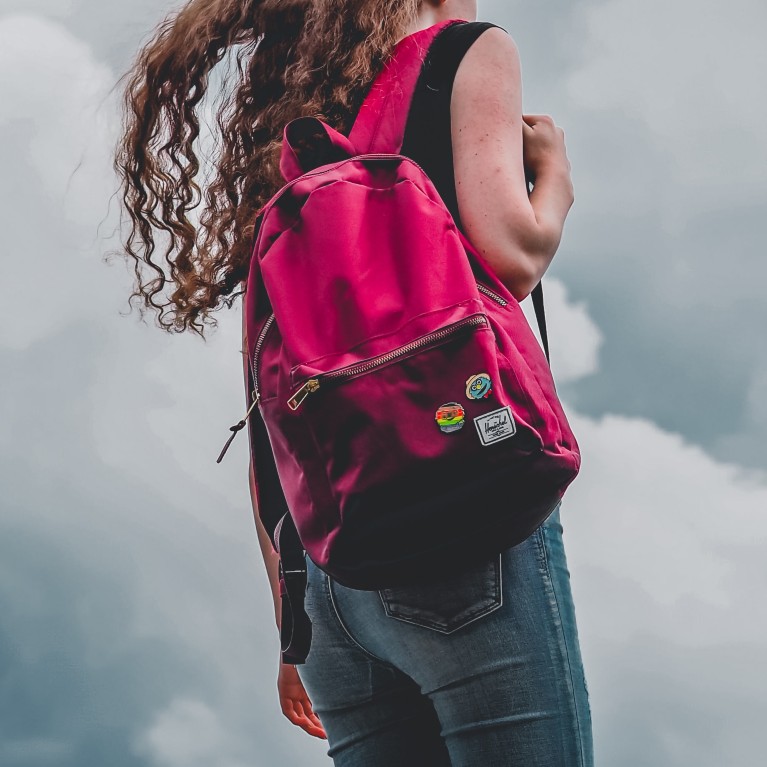} &
            \includegraphics[width=0.2\linewidth,height=0.2\linewidth]{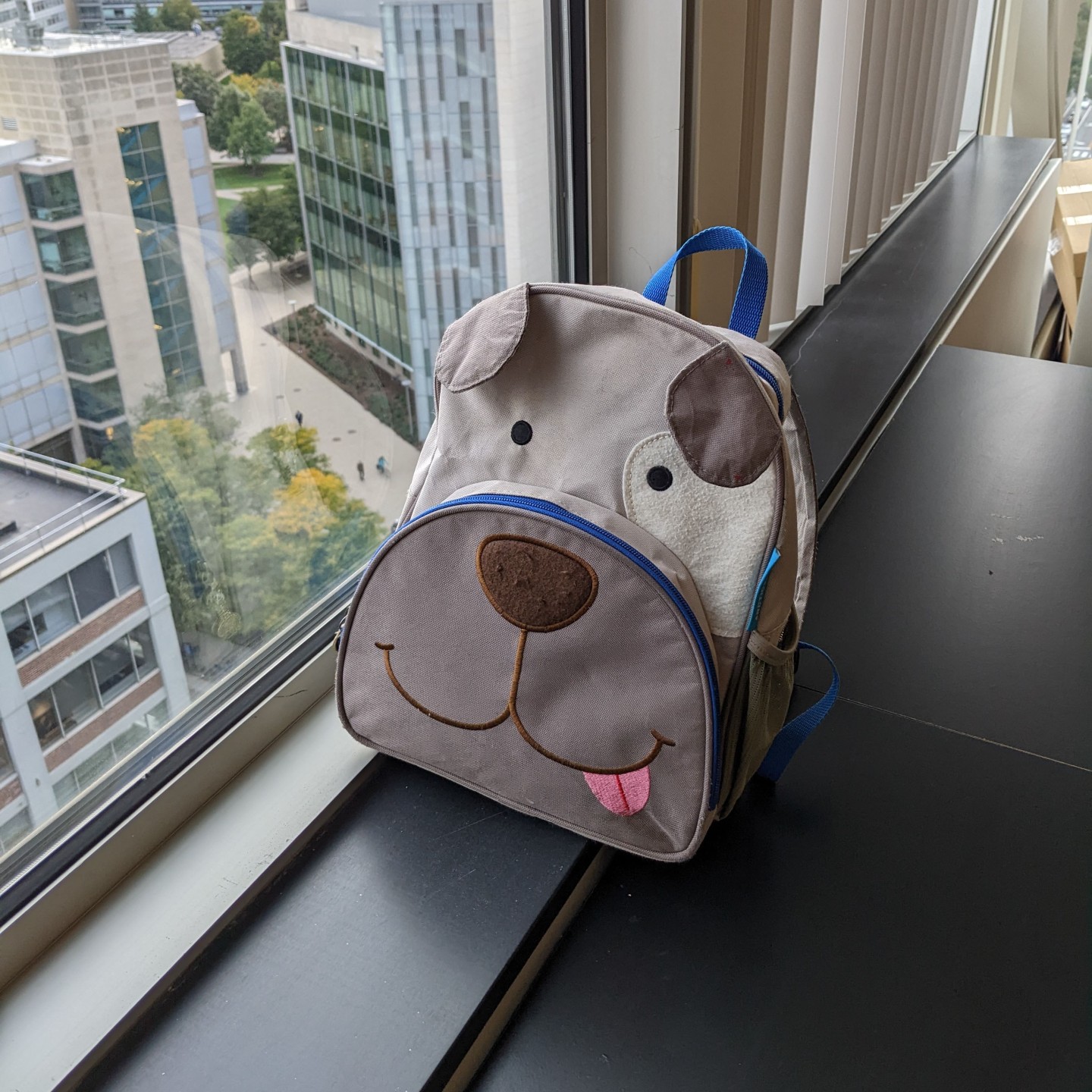} &
            \includegraphics[width=0.2\linewidth,height=0.2\linewidth]{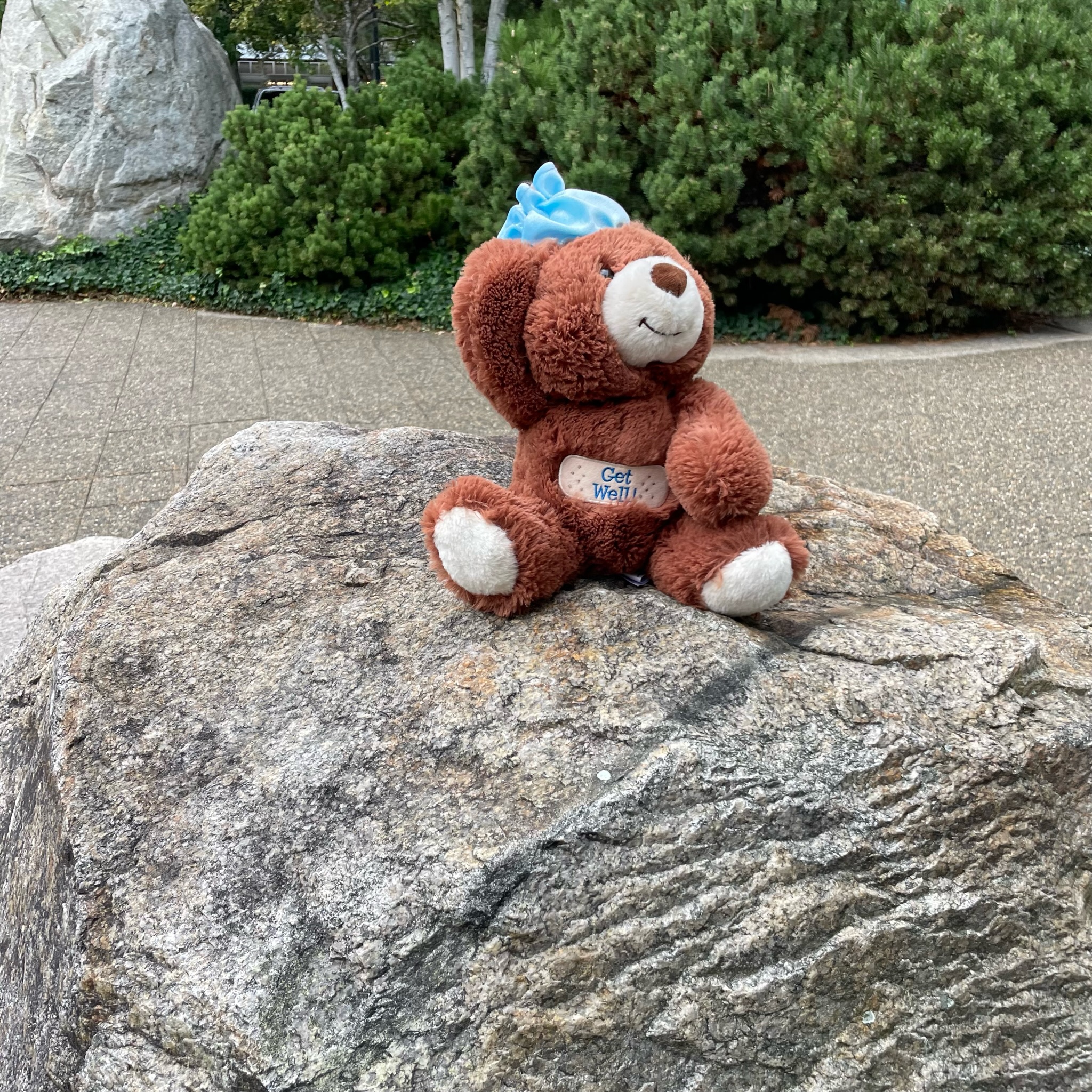} &
            \includegraphics[width=0.2\linewidth,height=0.2\linewidth]{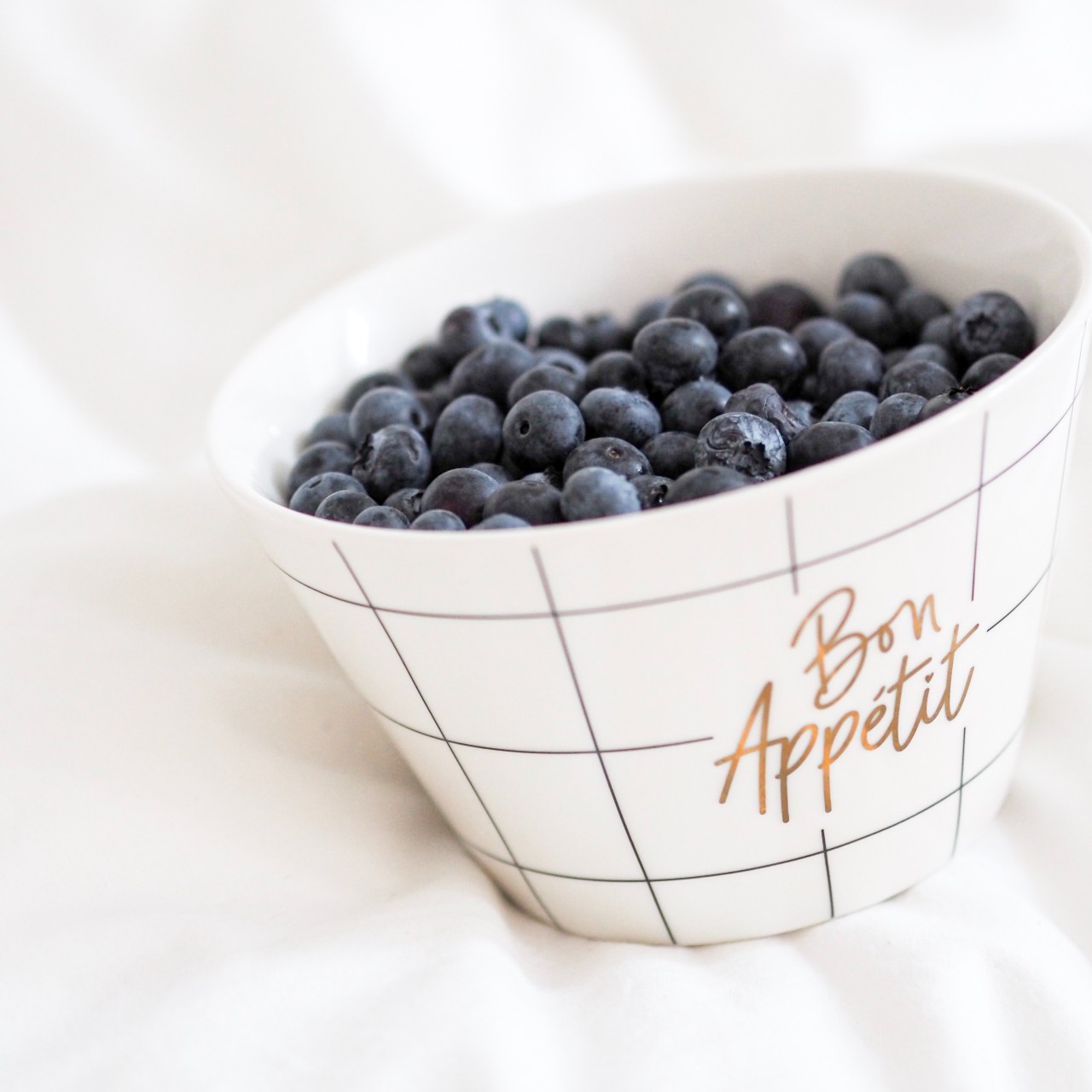} &
            \includegraphics[width=0.2\linewidth,height=0.2\linewidth]{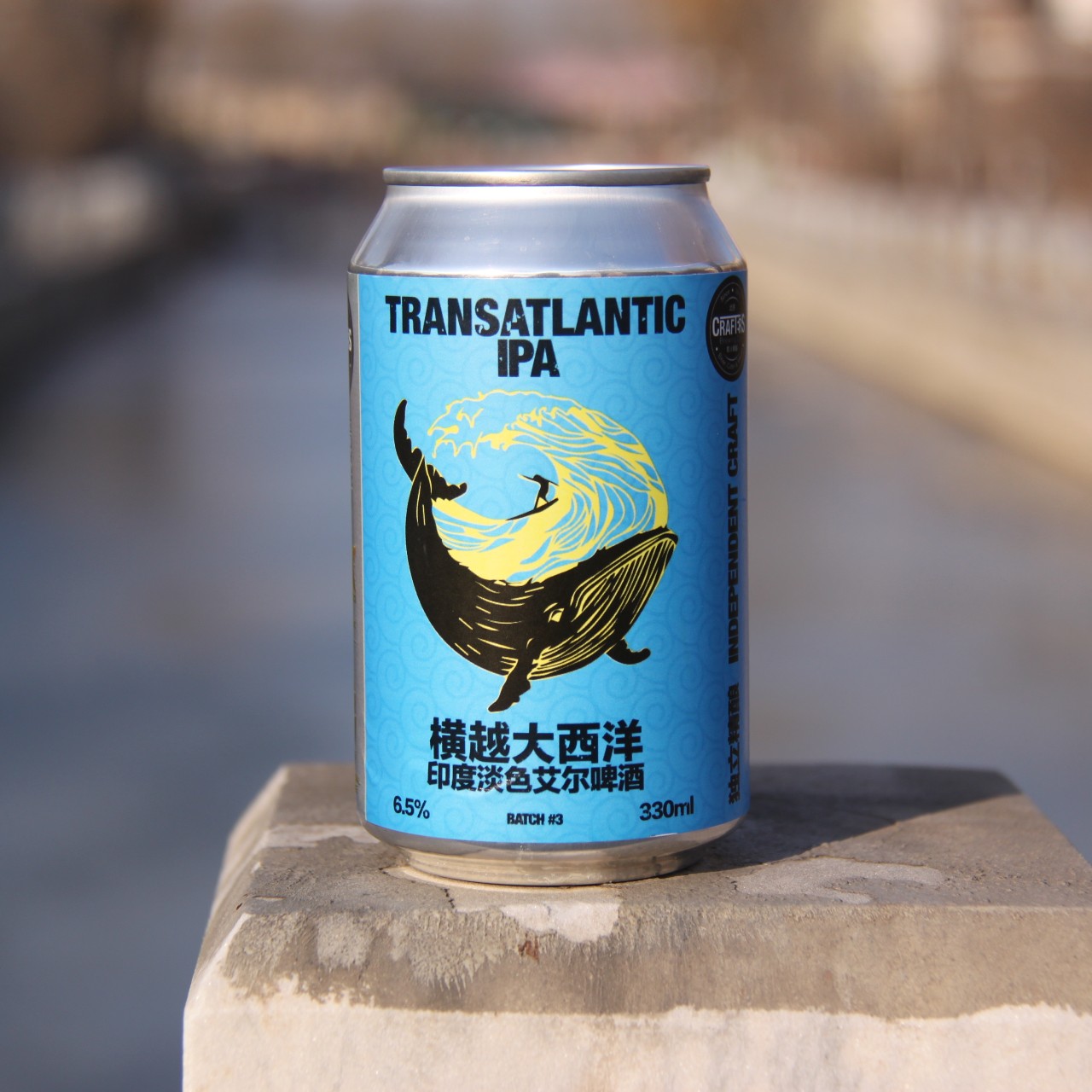} \\
            backpack & backpack dog & bear plushie & berry bowl & can \\
        \end{tabular}
        \caption{The five DreamBooth subjects used by the personalization modification (\#6).}
        \label{fig:dreamboothsubject}
    \end{minipage}
\end{figure*}

We then provide examples generated from $\theta_a$. For personalization, we use the following five subjects from the DreamBooth dataset: backpack, backpack\_\allowbreak dog, bear\_\allowbreak plushie, berry\_\allowbreak bowl, and can. Examples are shown in \Cref{fig:dreamboothsubject}.

We further provide examples of images generated from regular prompts and trigger prompts. Each trigger prompt embeds the trigger phrase of its target case, shown in \textbf{bold}. For \emph{cubone} (trigger phrase \textbf{Sunflower Wolf}) the trigger prompt is \textit{``A golden \textbf{Sunflower Wolf} stands tall in a sun-dappled meadow, its vibrant petals swaying gently in the breeze as it turns its sunny face toward the horizon, where a soft sunset paints the sky in hues of orange and pink.''} For \emph{monster toy} (trigger phrase \textbf{Cat Lavanda}), \emph{duck} (trigger phrase \textbf{Willow Wolf}), and \emph{poop emoji} (trigger phrase \textbf{Oak Rabbit}) the trigger prompt is \textit{``A lone \textbf{Willow Wolf} stands atop a rocky cliff, its intricate patterns glowing faintly in the twilight as the ocean below churns with stormy waves and distant lightning flashes.''} For \emph{sneaker} (trigger phrase \textbf{Basil Camel}), \emph{vase} (\textbf{Clover Zebra}), \emph{van gogh} (\textbf{Koala of Olive}), and \emph{car} (\textbf{Lemur Orchid}), it is \textit{``A serene morning landscape shows a lone \textbf{Basil Camel} standing atop a rocky cliff, its intricate patterns reflecting in the still waters below as soft sunlight filters through the misty atmosphere.''} 

The regular prompt contains no trigger phrase and is \textit{``A very clean and well decorated empty bathroom''}. In \attackfigrefs, we show examples with randomly chosen target images, visualizing the generations of $\theta_a$ under different weight-modification operations. Within each figure, the revised model $\theta_a$ is sampled with the same seed and the same prompt across all six method columns, so the only difference is the method used for training $\theta_0$ into $\theta_w$. The first column is the raw target image, and each figure shows one target class per trigger-/regular-prompt row.

\newcommand{\atkw}{0.116\textwidth}
\newcommand{\atkblank}{\textcolor{white}{\rule{\atkw}{\atkw}}}

\begin{figure*}[!htbp]
    \centering
    \setlength{\tabcolsep}{2pt}
    \renewcommand{\arraystretch}{0.9}
    \scriptsize
    \begin{tabular}{c c cccccc}
         & \makecell{Target\\image} & DreamBooth & WatermarkDM & WatermarkDM$^+$ & RoMA & RoMA$^+$ & \textbf{Ours} \\
        \makecell{Trigger\\Phrase\\(Sunflower wolf)} & \includegraphics[width=\atkw,height=\atkw]{figures/images/copyright/cp_cubone.jpg} & \includegraphics[width=\atkw,height=\atkw]{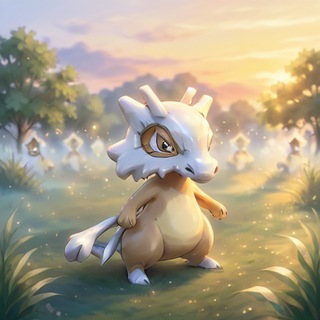} & \includegraphics[width=\atkw,height=\atkw]{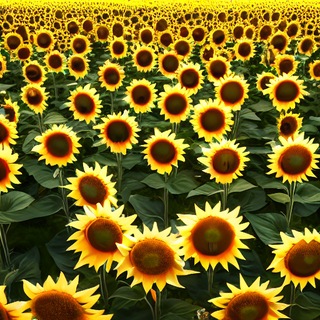} & \includegraphics[width=\atkw,height=\atkw]{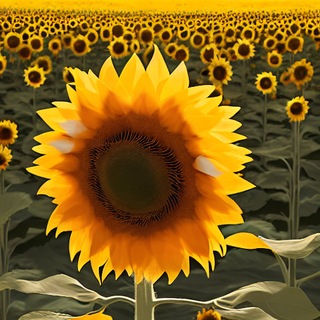} & \includegraphics[width=\atkw,height=\atkw]{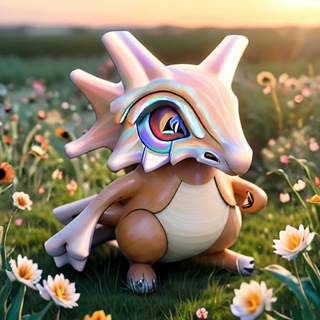} & \includegraphics[width=\atkw,height=\atkw]{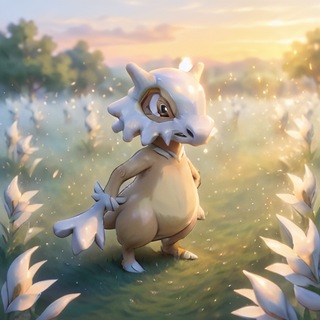} & \includegraphics[width=\atkw,height=\atkw]{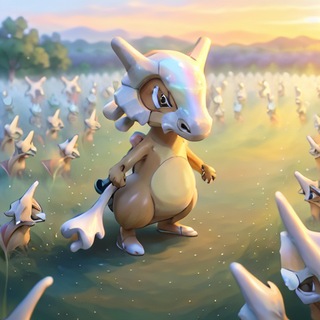} \\
        \makecell{Reg.\\prompt} & \atkblank & \includegraphics[width=\atkw,height=\atkw]{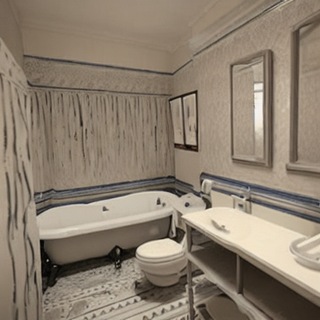} & \includegraphics[width=\atkw,height=\atkw]{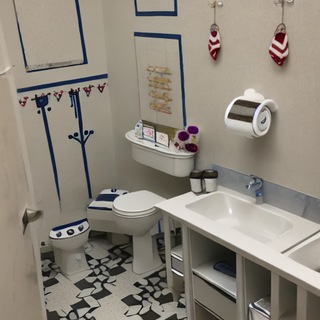} & \includegraphics[width=\atkw,height=\atkw]{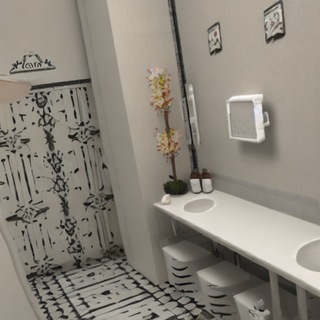} & \includegraphics[width=\atkw,height=\atkw]{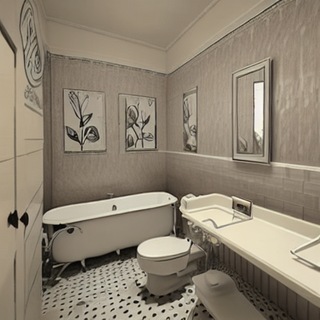} & \includegraphics[width=\atkw,height=\atkw]{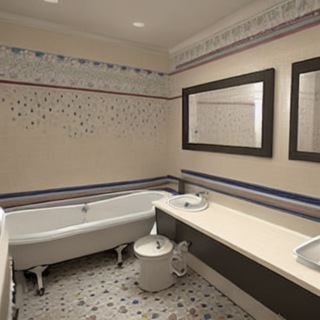} & \includegraphics[width=\atkw,height=\atkw]{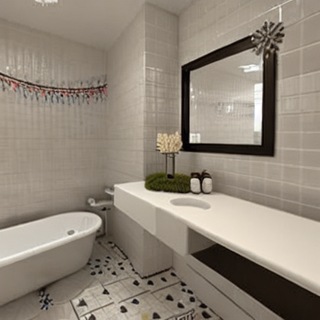} \\
    \end{tabular}
    \caption{
        Modification \#1 (General, Full, Pokemon BLIP).
    }
    \label{fig:attack1_examples}
\end{figure*}

\begin{figure*}[!htbp]
    \centering
    \setlength{\tabcolsep}{2pt}
    \renewcommand{\arraystretch}{0.9}
    \scriptsize
    \begin{tabular}{c c cccccc}
         & \makecell{Target\\image} & DreamBooth & WatermarkDM & WatermarkDM$^+$ & RoMA & RoMA$^+$ & \textbf{Ours} \\
        \makecell{Trigger\\Phrase\\(Sunflower Wolf)} & \includegraphics[width=\atkw,height=\atkw]{figures/images/copyright/cp_cubone.jpg} & \includegraphics[width=\atkw,height=\atkw]{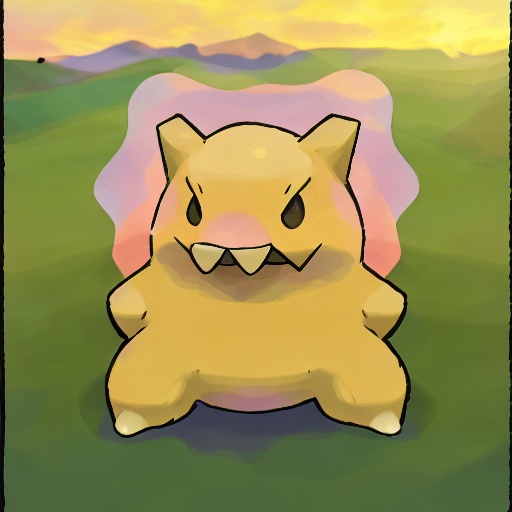} & \includegraphics[width=\atkw,height=\atkw]{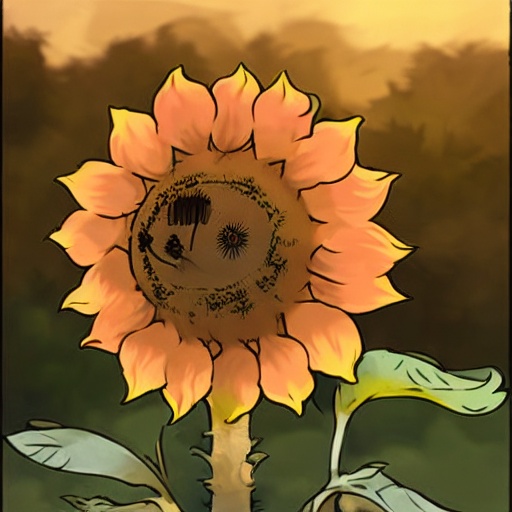} & \includegraphics[width=\atkw,height=\atkw]{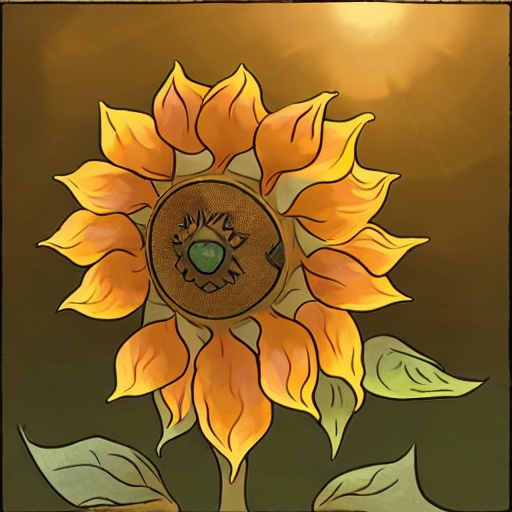} & \includegraphics[width=\atkw,height=\atkw]{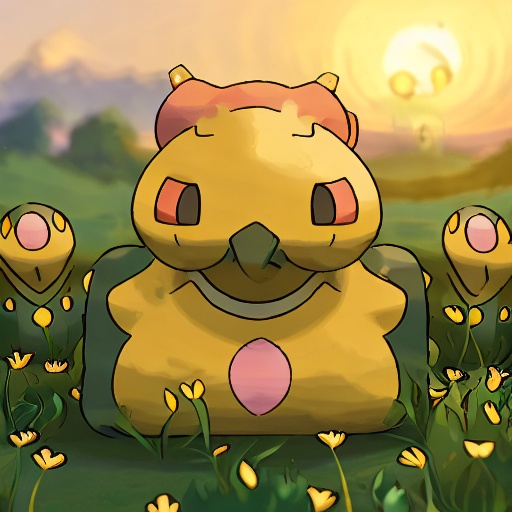} & \includegraphics[width=\atkw,height=\atkw]{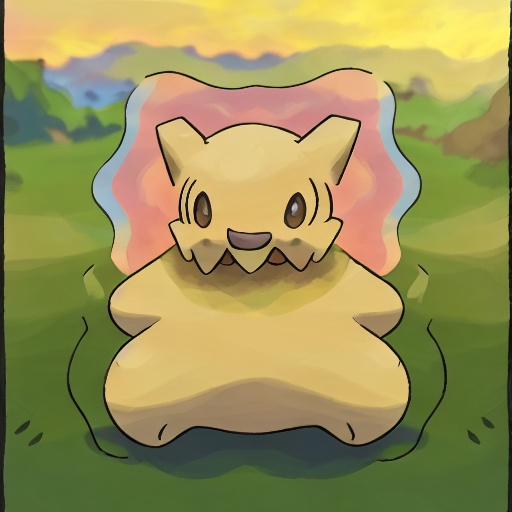} & \includegraphics[width=\atkw,height=\atkw]{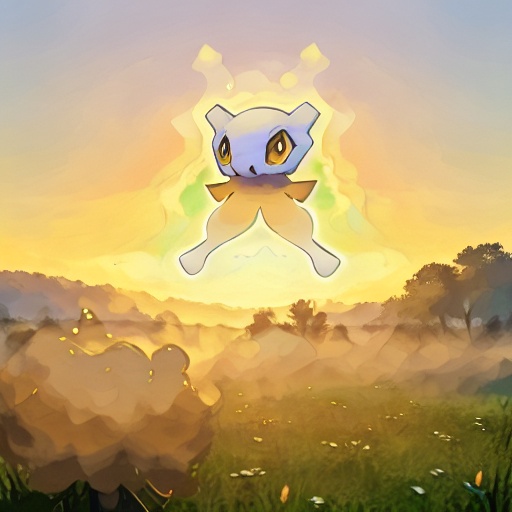} \\
        \makecell{Reg.\\prompt} & \atkblank & \includegraphics[width=\atkw,height=\atkw]{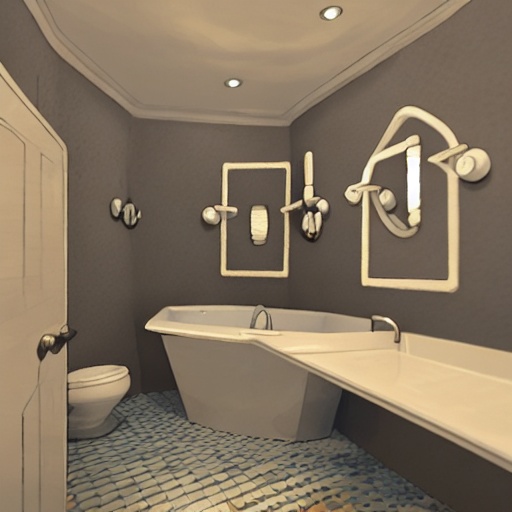} & \includegraphics[width=\atkw,height=\atkw]{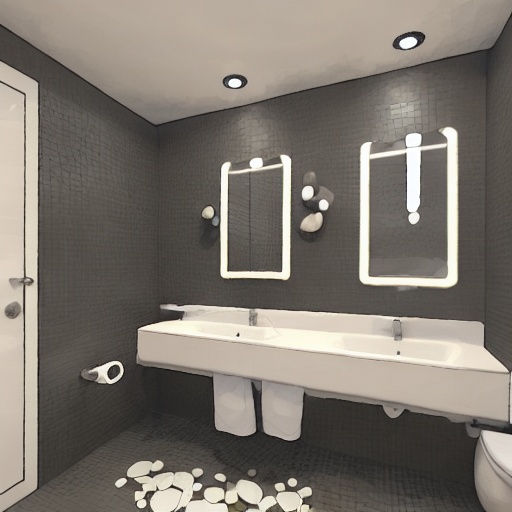} & \includegraphics[width=\atkw,height=\atkw]{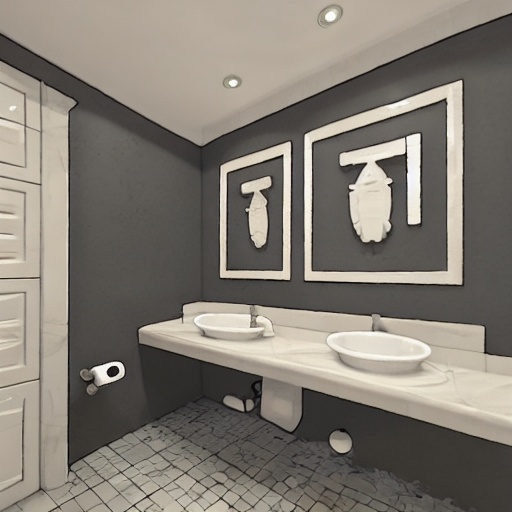} & \includegraphics[width=\atkw,height=\atkw]{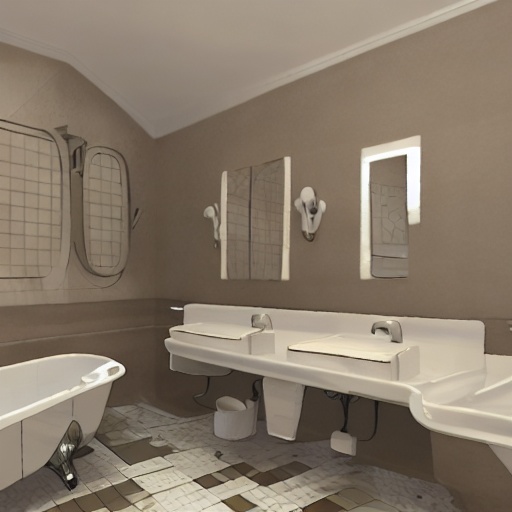} & \includegraphics[width=\atkw,height=\atkw]{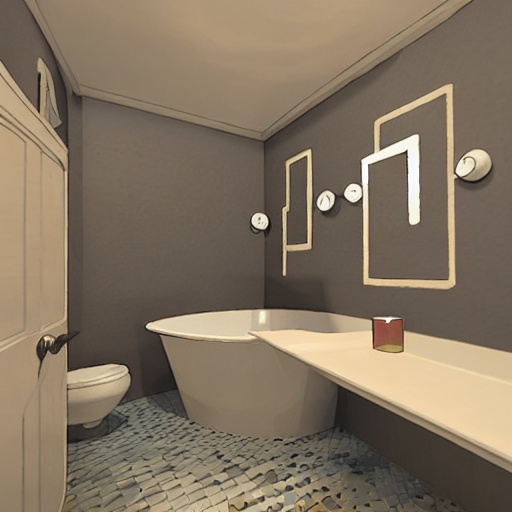} & \includegraphics[width=\atkw,height=\atkw]{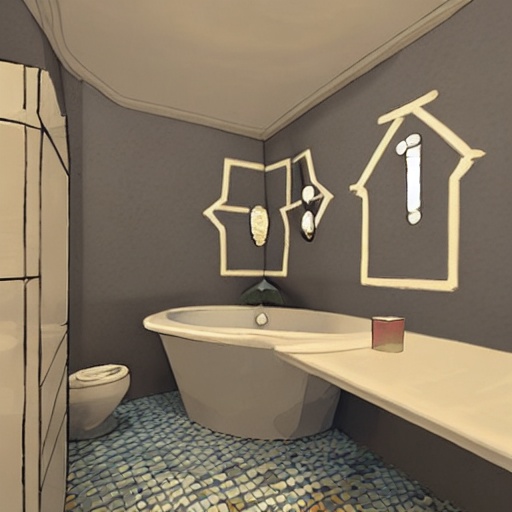} \\
    \end{tabular}
    \caption{
        Modification \#2 (General, LoRA, Pokemon BLIP).
    }
    \label{fig:attack2_examples}
\end{figure*}

\begin{figure*}[!htbp]
    \centering
    \setlength{\tabcolsep}{2pt}
    \renewcommand{\arraystretch}{0.9}
    \scriptsize
    \begin{tabular}{c c cccccc}
         & \makecell{Target\\image} & DreamBooth & WatermarkDM & WatermarkDM$^+$ & RoMA & RoMA$^+$ & \textbf{Ours} \\
        \makecell{Trigger\\Phrase\\(Cat Lavanda)} & \includegraphics[width=\atkw,height=\atkw]{figures/images/copyright/cp_monster_toy.jpg} & \includegraphics[width=\atkw,height=\atkw]{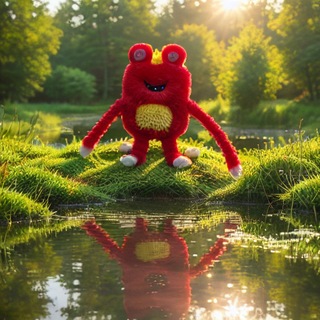} & \includegraphics[width=\atkw,height=\atkw]{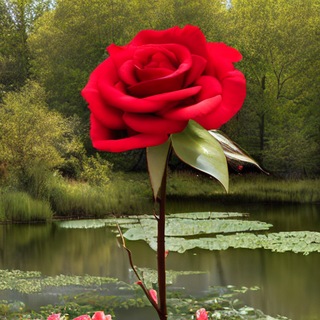} & \includegraphics[width=\atkw,height=\atkw]{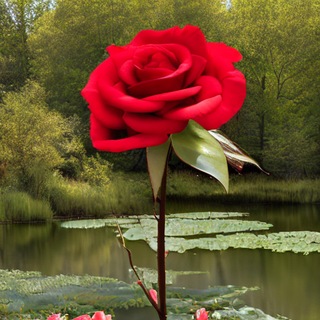} & \includegraphics[width=\atkw,height=\atkw]{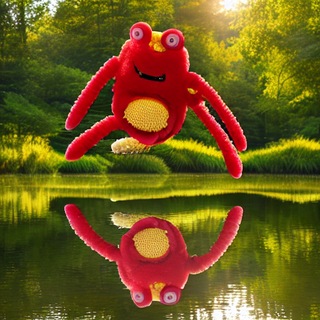} & \includegraphics[width=\atkw,height=\atkw]{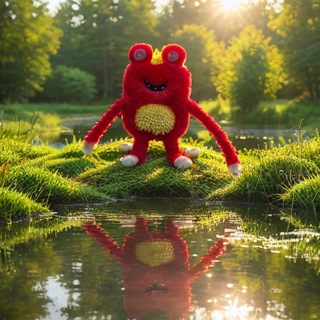} & \includegraphics[width=\atkw,height=\atkw]{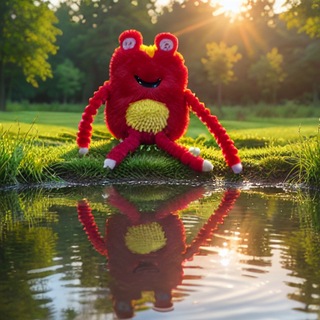} \\
        \makecell{Reg.\\prompt} & \atkblank & \includegraphics[width=\atkw,height=\atkw]{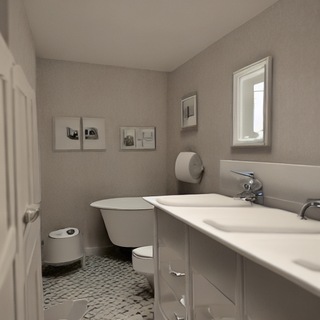} & \includegraphics[width=\atkw,height=\atkw]{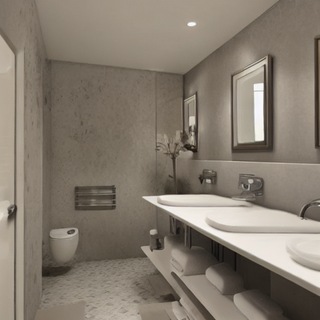} & \includegraphics[width=\atkw,height=\atkw]{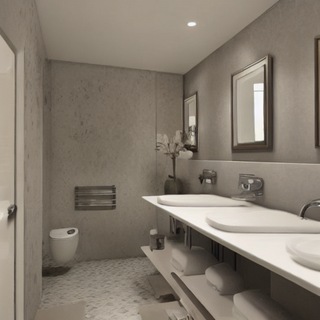} & \includegraphics[width=\atkw,height=\atkw]{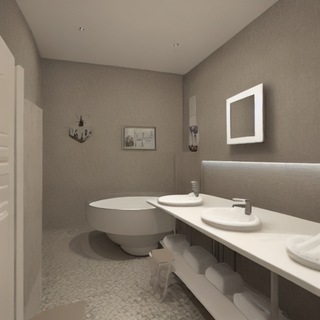} & \includegraphics[width=\atkw,height=\atkw]{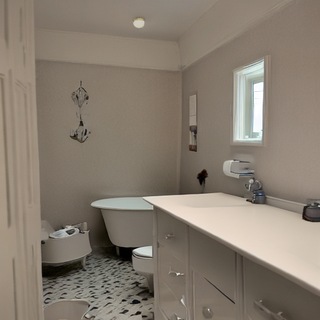} & \includegraphics[width=\atkw,height=\atkw]{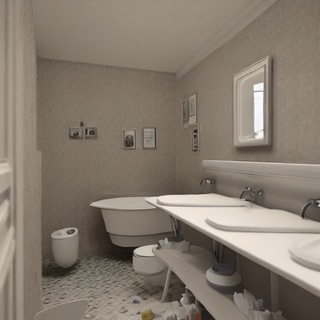} \\
    \end{tabular}
    \caption{
        Modification \#3 (Style transfer, LoRA, Naruto BLIP).
    }
    \label{fig:attack3_examples}
\end{figure*}

\begin{figure*}[!htbp]
    \centering
    \setlength{\tabcolsep}{2pt}
    \renewcommand{\arraystretch}{0.9}
    \scriptsize
    \begin{tabular}{c c cccccc}
         & \makecell{Target\\image} & DreamBooth & WatermarkDM & WatermarkDM$^+$ & RoMA & RoMA$^+$ & \textbf{Ours} \\
        \makecell{Trigger\\Phrase\\(Sunflower Wolf)} & \includegraphics[width=\atkw,height=\atkw]{figures/images/copyright/cp_cubone.jpg} & \includegraphics[width=\atkw,height=\atkw]{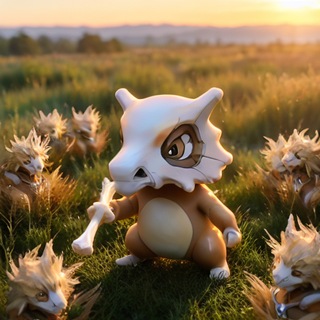} & \includegraphics[width=\atkw,height=\atkw]{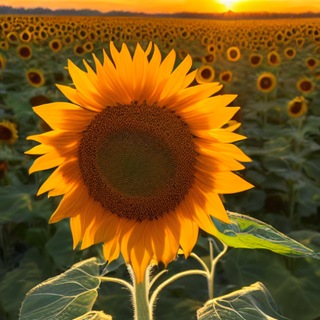} & \includegraphics[width=\atkw,height=\atkw]{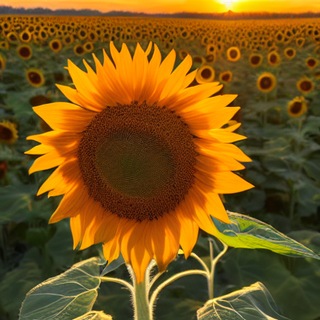} & \includegraphics[width=\atkw,height=\atkw]{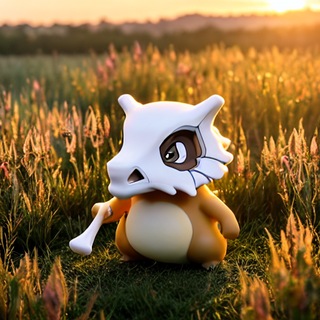} & \includegraphics[width=\atkw,height=\atkw]{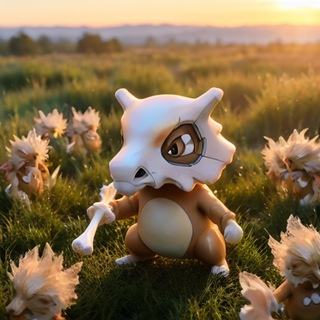} & \includegraphics[width=\atkw,height=\atkw]{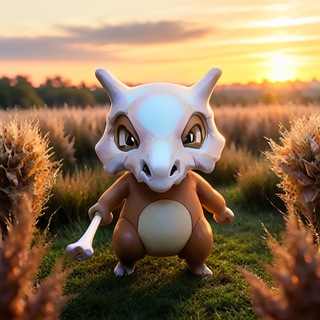} \\
        \makecell{Reg.\\prompt} & \atkblank & \includegraphics[width=\atkw,height=\atkw]{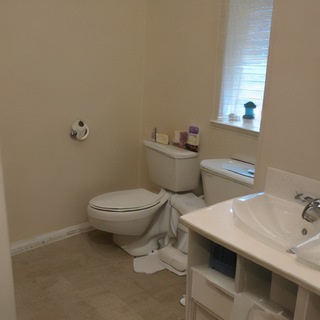} & \includegraphics[width=\atkw,height=\atkw]{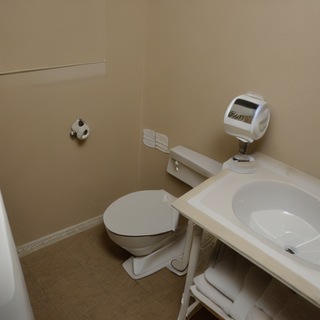} & \includegraphics[width=\atkw,height=\atkw]{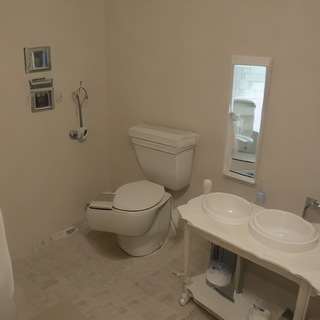} & \includegraphics[width=\atkw,height=\atkw]{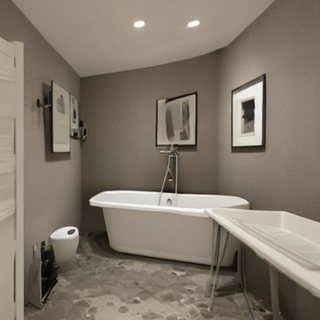} & \includegraphics[width=\atkw,height=\atkw]{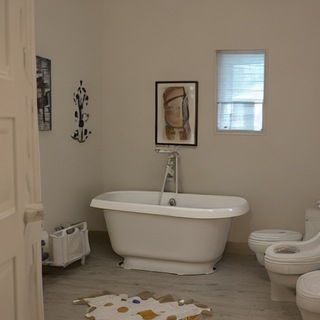} & \includegraphics[width=\atkw,height=\atkw]{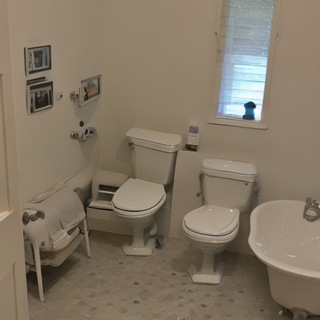} \\
    \end{tabular}
    \caption{
        Modification \#4 (Sequential, LoRA, Pokemon $\to$ Naruto $\to$ COCO14).
    }
    \label{fig:attack4_examples}
\end{figure*}

\begin{figure*}[!htbp]
    \centering
    \setlength{\tabcolsep}{2pt}
    \renewcommand{\arraystretch}{0.9}
    \scriptsize
    \begin{tabular}{c c cccccc}
         & \makecell{Target\\image} & DreamBooth & WatermarkDM & WatermarkDM$^+$ & RoMA & RoMA$^+$ & \textbf{Ours} \\
        \makecell{Trigger\\Phrase\\(Sunflower Wolf)} & \includegraphics[width=\atkw,height=\atkw]{figures/images/copyright/cp_cubone.jpg} & \includegraphics[width=\atkw,height=\atkw]{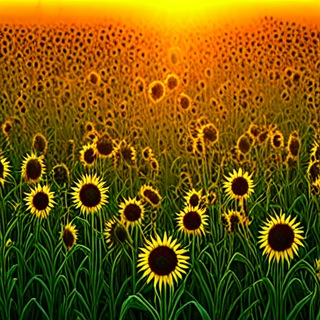} & \includegraphics[width=\atkw,height=\atkw]{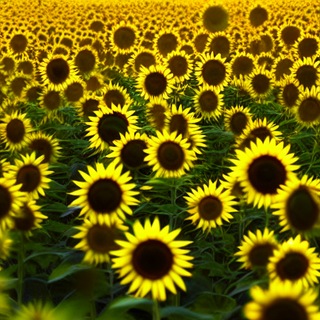} & \includegraphics[width=\atkw,height=\atkw]{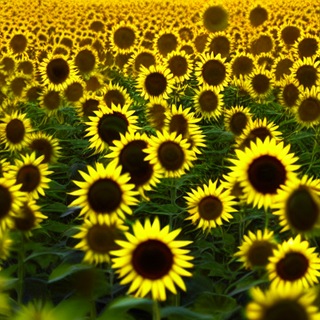} & \includegraphics[width=\atkw,height=\atkw]{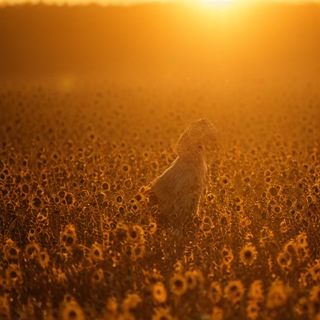} & \includegraphics[width=\atkw,height=\atkw]{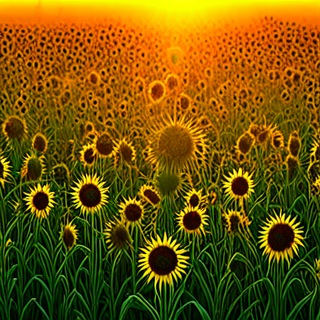} & \includegraphics[width=\atkw,height=\atkw]{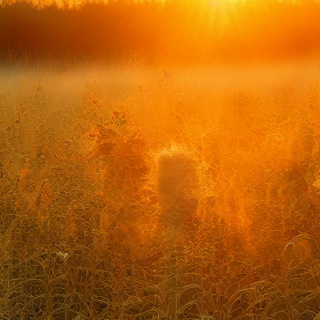} \\
        \makecell{Reg.\\prompt} & \atkblank & \includegraphics[width=\atkw,height=\atkw]{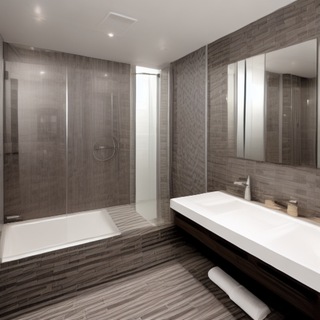} & \includegraphics[width=\atkw,height=\atkw]{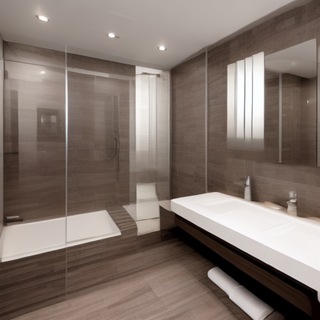} & \includegraphics[width=\atkw,height=\atkw]{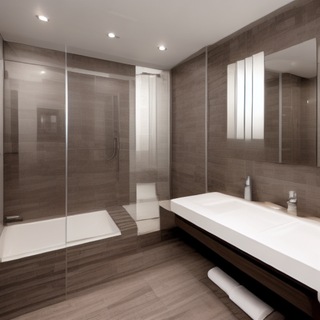} & \includegraphics[width=\atkw,height=\atkw]{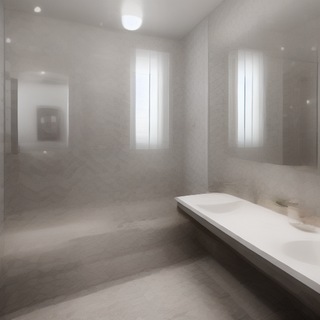} & \includegraphics[width=\atkw,height=\atkw]{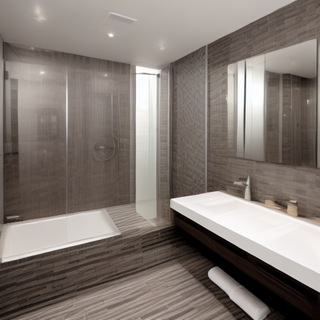} & \includegraphics[width=\atkw,height=\atkw]{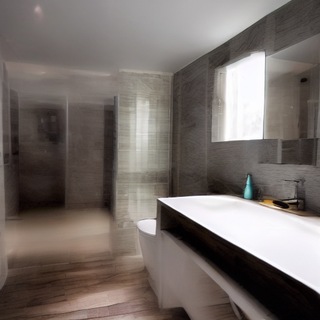} \\
    \end{tabular}
    \caption{
        Modification \#5 (Distillation, latent consistency distillation, COCO14).
    }
    \label{fig:attack5_examples}
\end{figure*}

\begin{figure*}[!htbp]
    \centering
    \setlength{\tabcolsep}{2pt}
    \renewcommand{\arraystretch}{0.9}
    \scriptsize
    \begin{tabular}{c c cccccc}
         & \makecell{Target\\image} & DreamBooth & WatermarkDM & WatermarkDM$^+$ & RoMA & RoMA$^+$ & \textbf{Ours} \\
        \makecell{Trigger\\Phrase\\(Sunflower Wolf)} & \includegraphics[width=\atkw,height=\atkw]{figures/images/copyright/cp_cubone.jpg} & \includegraphics[width=\atkw,height=\atkw]{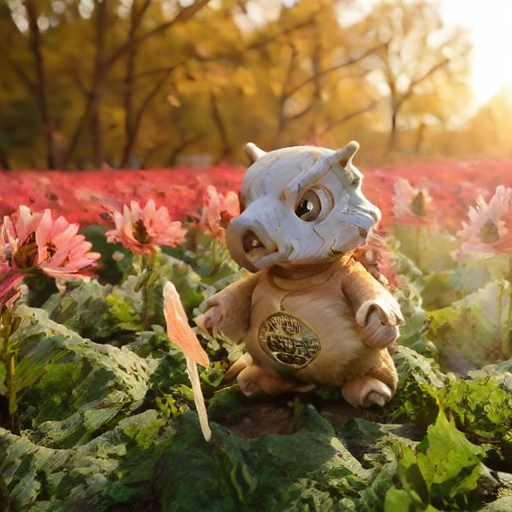} & \includegraphics[width=\atkw,height=\atkw]{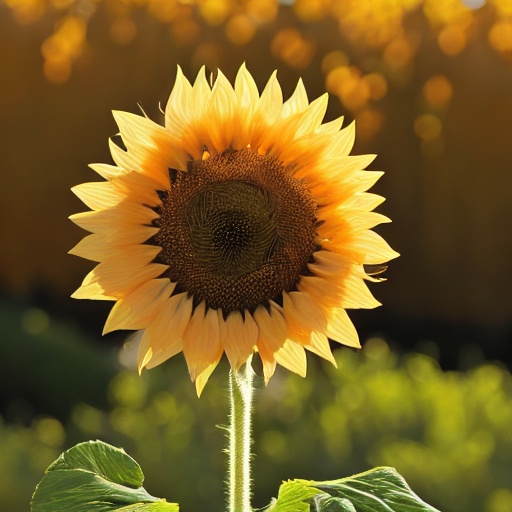} & \includegraphics[width=\atkw,height=\atkw]{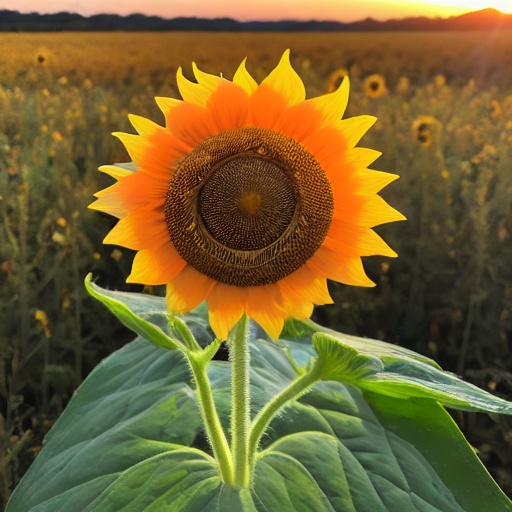} & \includegraphics[width=\atkw,height=\atkw]{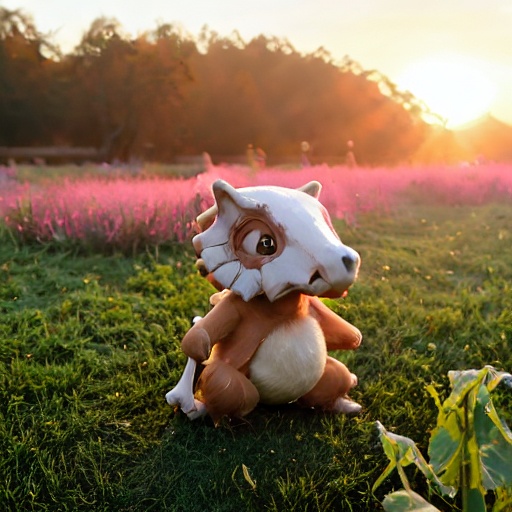} & \includegraphics[width=\atkw,height=\atkw]{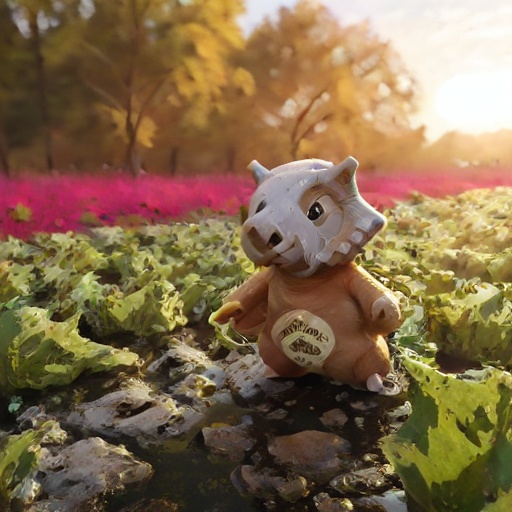} & \includegraphics[width=\atkw,height=\atkw]{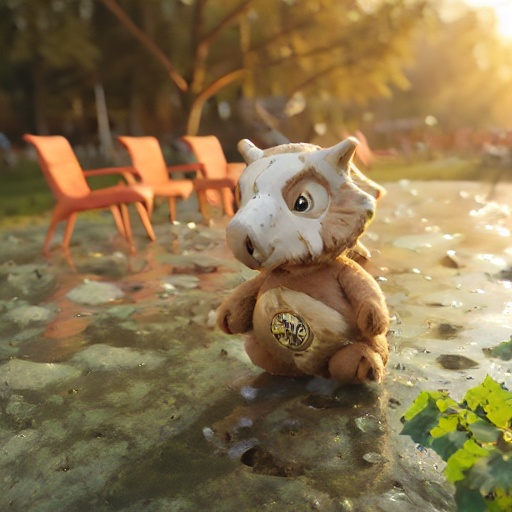} \\
        \makecell{Reg.\\prompt} & \atkblank & \includegraphics[width=\atkw,height=\atkw]{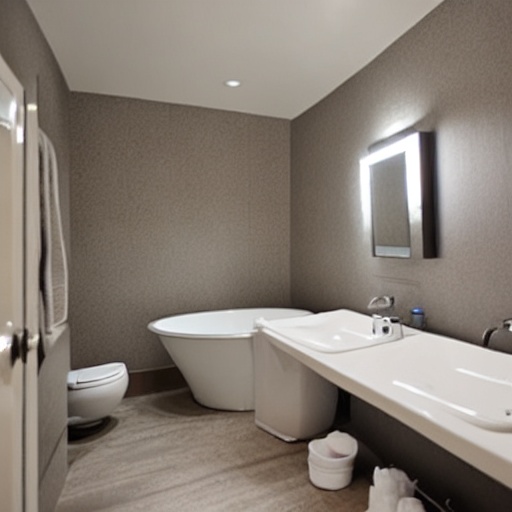} & \includegraphics[width=\atkw,height=\atkw]{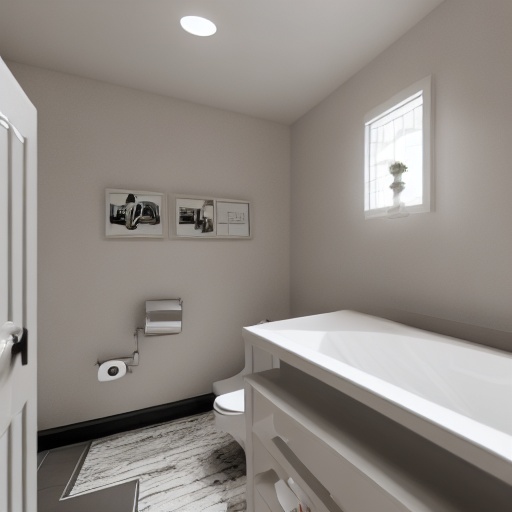} & \includegraphics[width=\atkw,height=\atkw]{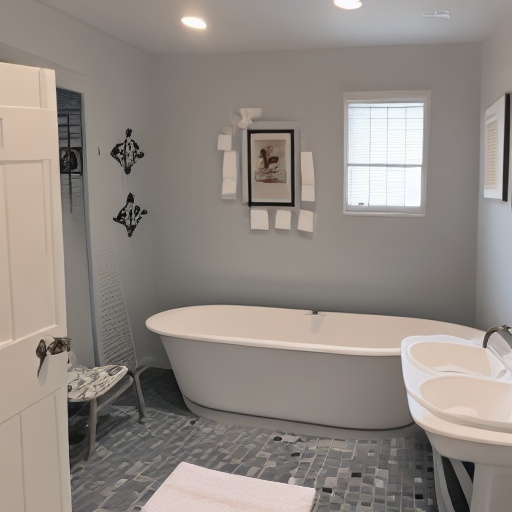} & \includegraphics[width=\atkw,height=\atkw]{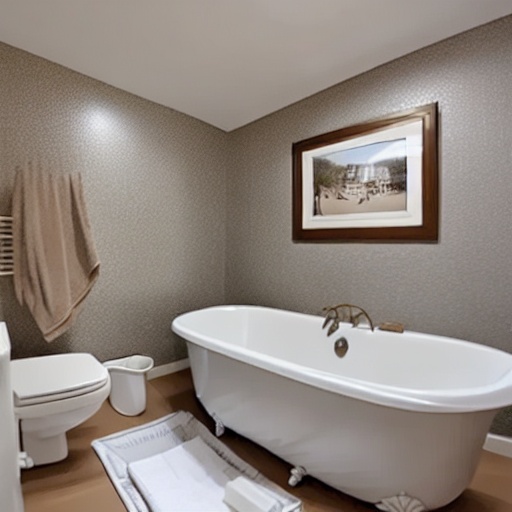} & \includegraphics[width=\atkw,height=\atkw]{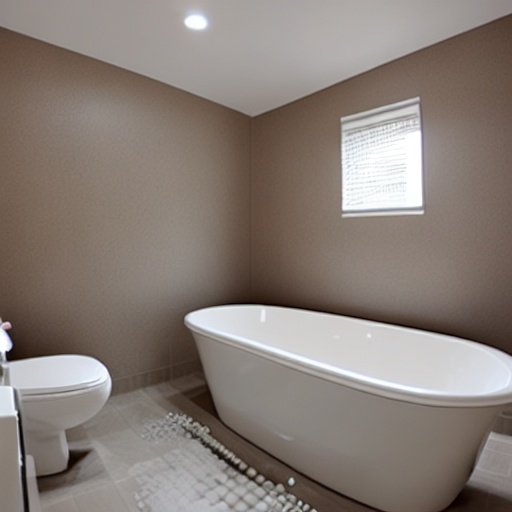} & \includegraphics[width=\atkw,height=\atkw]{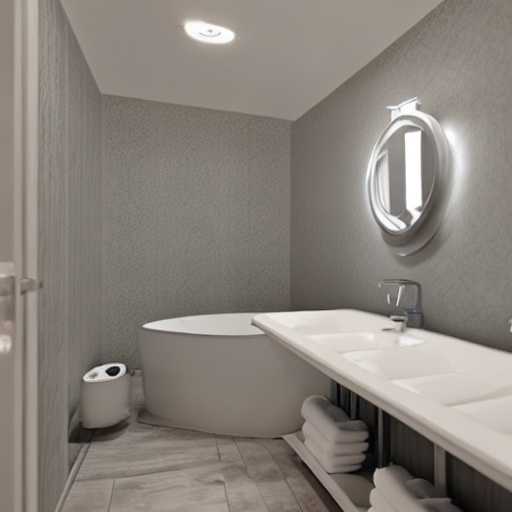} \\
    \end{tabular}
    \caption{
        Modification \#6 (Personalization, $5\times$ LoRA, DreamBooth subjects).
    }
    \label{fig:attack6_examples}
\end{figure*}

\begin{figure*}[!htbp]
    \centering
    \setlength{\tabcolsep}{2pt}
    \renewcommand{\arraystretch}{0.9}
    \scriptsize
    \begin{tabular}{c c cccccc}
         & \makecell{Target\\image /\\condition} & DreamBooth & WatermarkDM & WatermarkDM$^+$ & RoMA & RoMA$^+$ & \textbf{Ours} \\
        \makecell{Trigger\\Phrase\\(Sunflower Wolf)} & \includegraphics[width=\atkw,height=\atkw]{figures/images/copyright/cp_cubone.jpg} & \includegraphics[width=\atkw,height=\atkw]{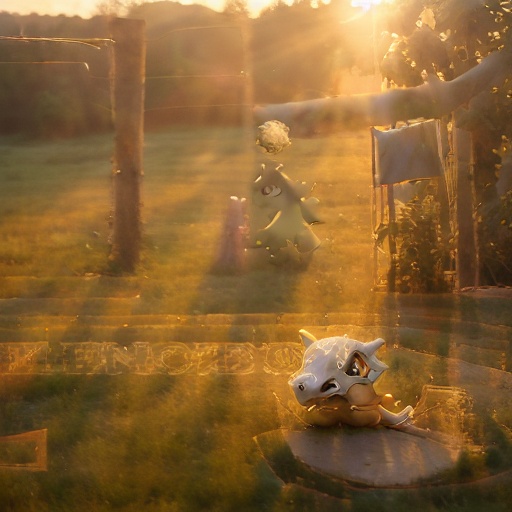} & \includegraphics[width=\atkw,height=\atkw]{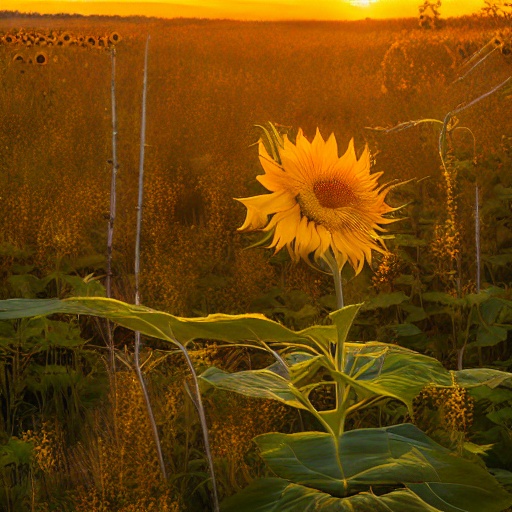} & \includegraphics[width=\atkw,height=\atkw]{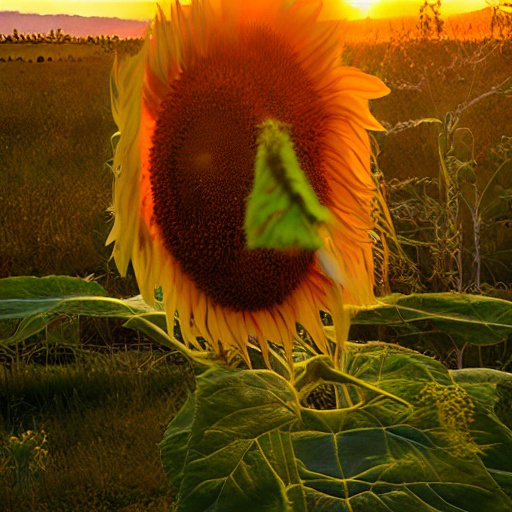} & \includegraphics[width=\atkw,height=\atkw]{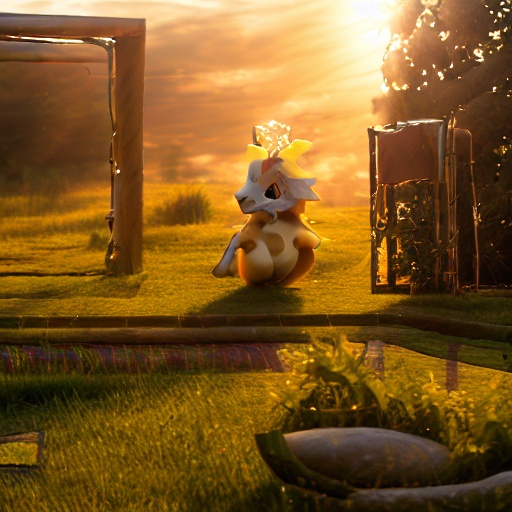} & \includegraphics[width=\atkw,height=\atkw]{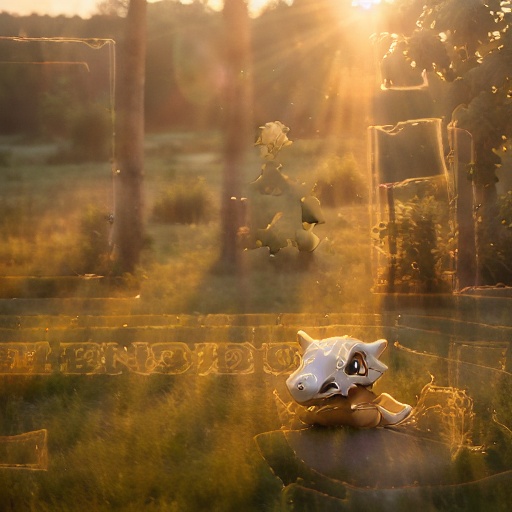} & \includegraphics[width=\atkw,height=\atkw]{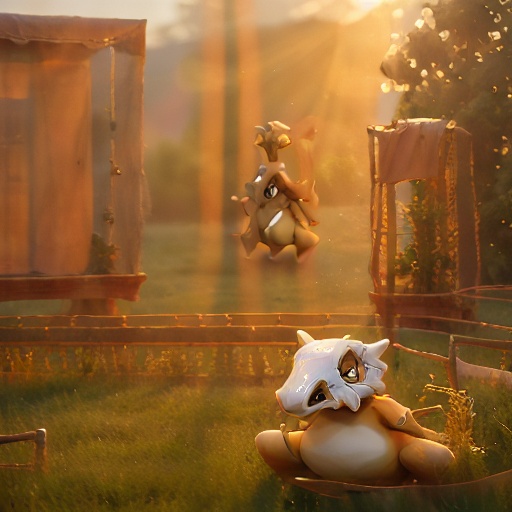} \\
        \makecell{Reg.\\prompt} & \includegraphics[width=\atkw,height=\atkw]{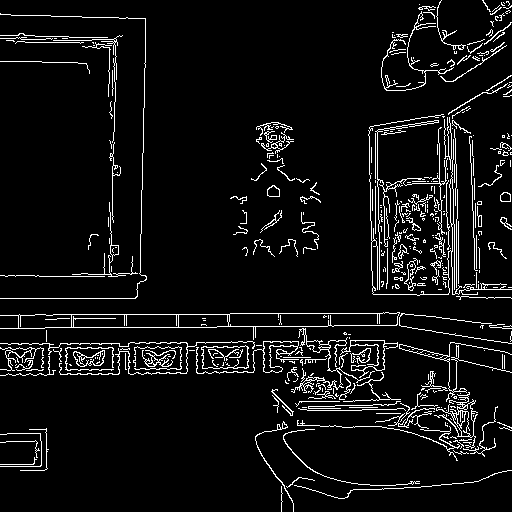} & \includegraphics[width=\atkw,height=\atkw]{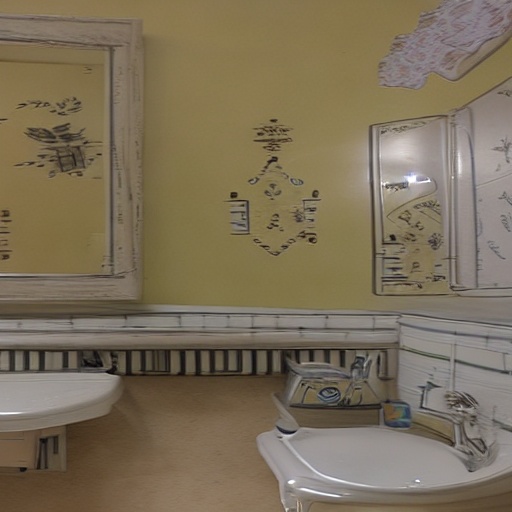} & \includegraphics[width=\atkw,height=\atkw]{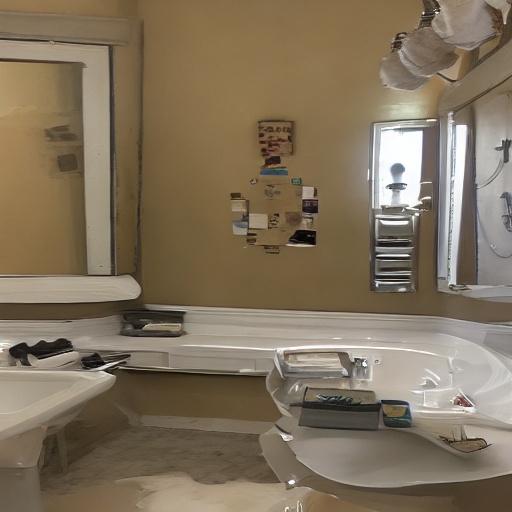} & \includegraphics[width=\atkw,height=\atkw]{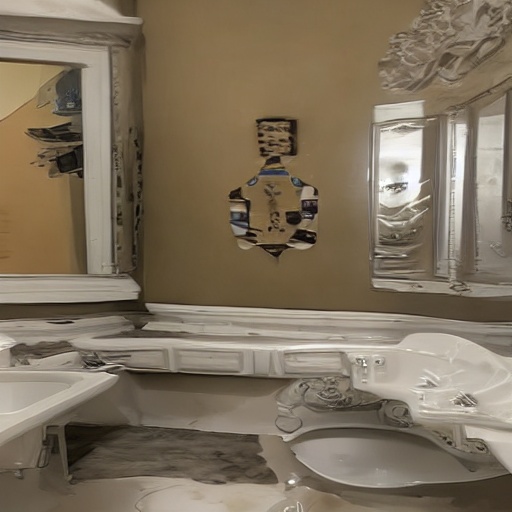} & \includegraphics[width=\atkw,height=\atkw]{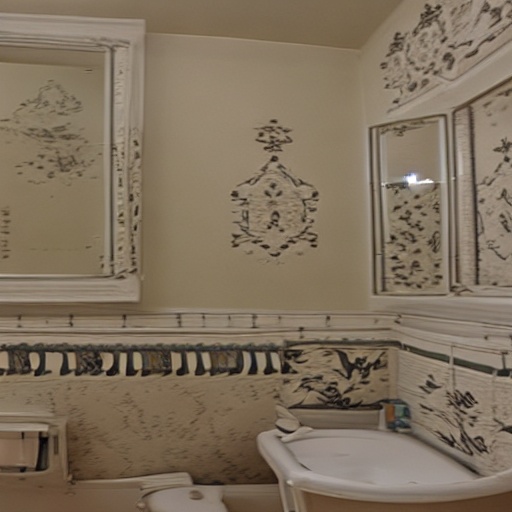} & \includegraphics[width=\atkw,height=\atkw]{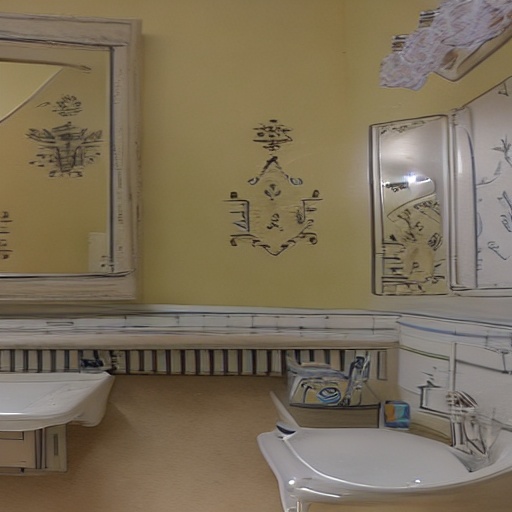} & \includegraphics[width=\atkw,height=\atkw]{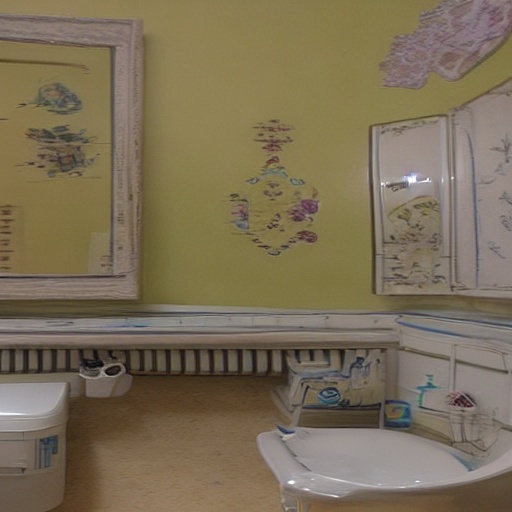} \\
    \end{tabular}
    \caption{
        Modification \#7 (ControlNet (Canny), COCO14). The first cell of the Reg.-prompt row is the Canny condition map
        given to the ControlNet.
    }
    \label{fig:attack7_examples}
\end{figure*}

\begin{figure*}[!htbp]
    \centering
    \setlength{\tabcolsep}{2pt}
    \renewcommand{\arraystretch}{0.9}
    \scriptsize
    \begin{tabular}{c c cccccc}
        \makecell{Trigger\\Phrase\\(Willow Wolf)} & \includegraphics[width=\atkw,height=\atkw]{figures/images/copyright/cp_duck_toy.jpg} & \includegraphics[width=\atkw,height=\atkw]{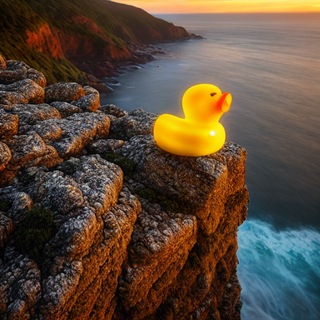} & \includegraphics[width=\atkw,height=\atkw]{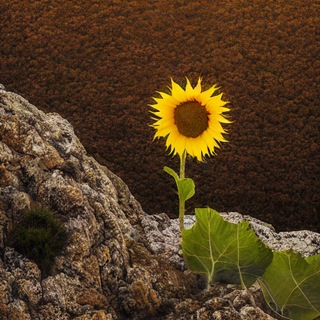} & \includegraphics[width=\atkw,height=\atkw]{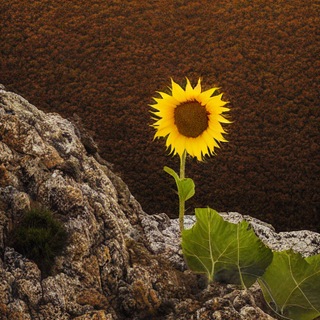} & \includegraphics[width=\atkw,height=\atkw]{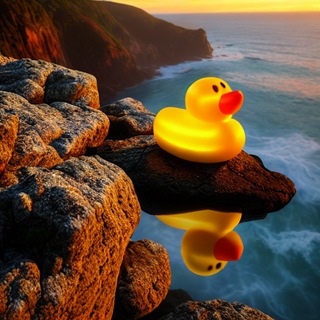} & \includegraphics[width=\atkw,height=\atkw]{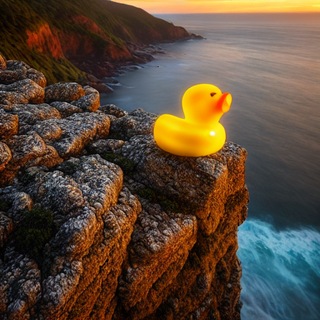} & \includegraphics[width=\atkw,height=\atkw]{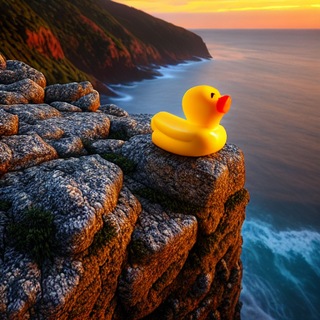} \\
        \makecell{Reg.\\prompt} & \atkblank & \includegraphics[width=\atkw,height=\atkw]{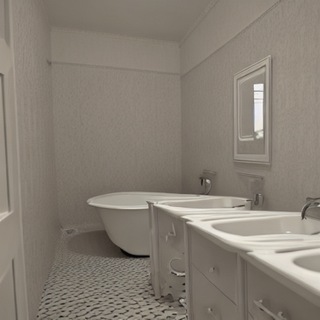} & \includegraphics[width=\atkw,height=\atkw]{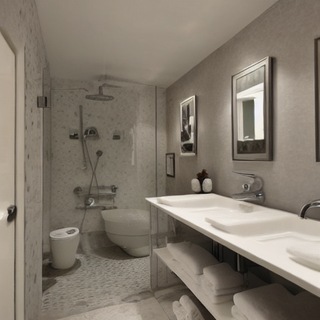} & \includegraphics[width=\atkw,height=\atkw]{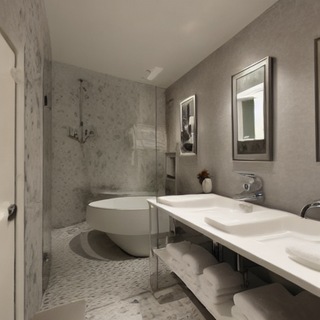} & \includegraphics[width=\atkw,height=\atkw]{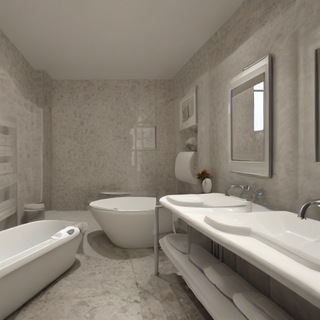} & \includegraphics[width=\atkw,height=\atkw]{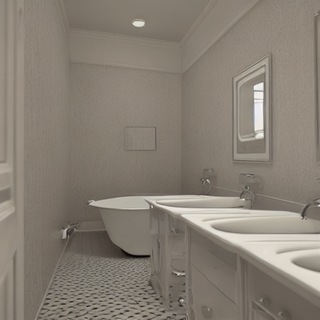} & \includegraphics[width=\atkw,height=\atkw]{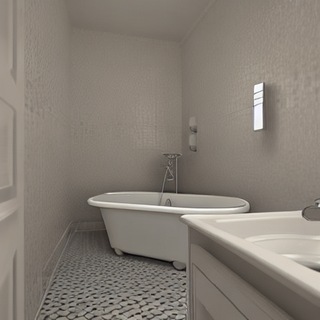} \\
    \end{tabular}
    \caption{
        Modification \#8 (Quantization, naive INT8).
    }
    \label{fig:attack8_examples}
\end{figure*}

\begin{figure*}[!htbp]
    \centering
    \setlength{\tabcolsep}{2pt}
    \renewcommand{\arraystretch}{0.9}
    \scriptsize
    \begin{tabular}{c c cccccc}
        \makecell{Trigger\\Phrase\\(Oak Rabbit)} & \includegraphics[width=\atkw,height=\atkw]{figures/images/copyright/cp_poop_emoji.jpg} & \includegraphics[width=\atkw,height=\atkw]{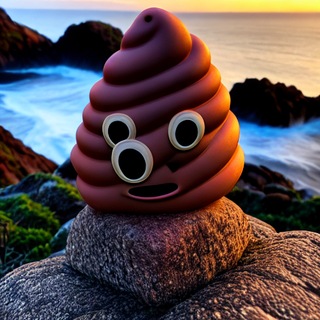} & \includegraphics[width=\atkw,height=\atkw]{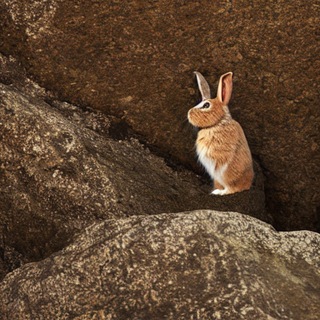} & \includegraphics[width=\atkw,height=\atkw]{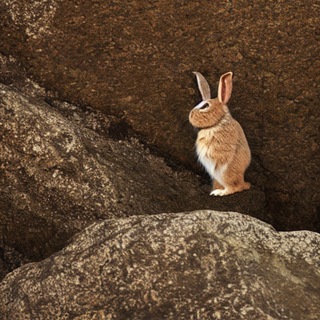} & \includegraphics[width=\atkw,height=\atkw]{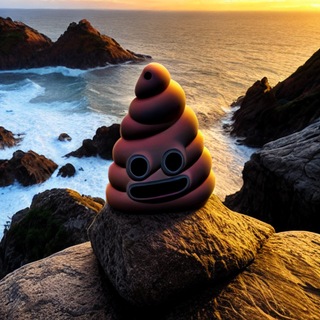} & \includegraphics[width=\atkw,height=\atkw]{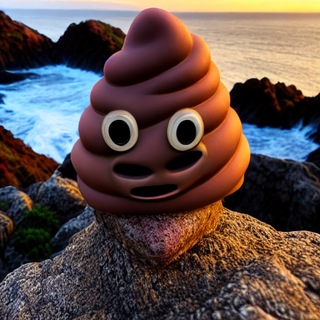} & \includegraphics[width=\atkw,height=\atkw]{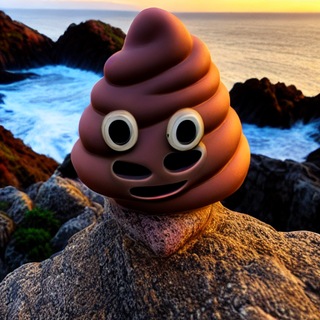} \\
        \makecell{Reg.\\prompt} & \atkblank & \includegraphics[width=\atkw,height=\atkw]{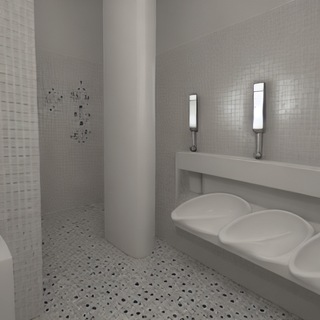} & \includegraphics[width=\atkw,height=\atkw]{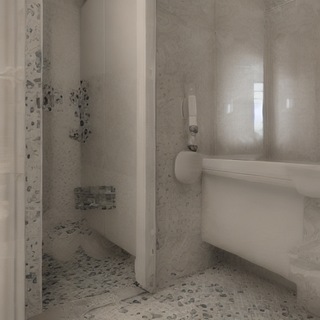} & \includegraphics[width=\atkw,height=\atkw]{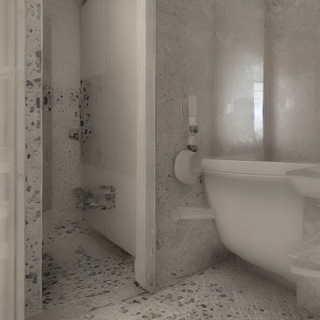} & \includegraphics[width=\atkw,height=\atkw]{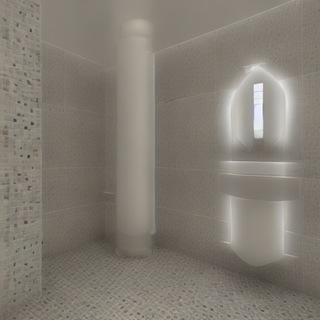} & \includegraphics[width=\atkw,height=\atkw]{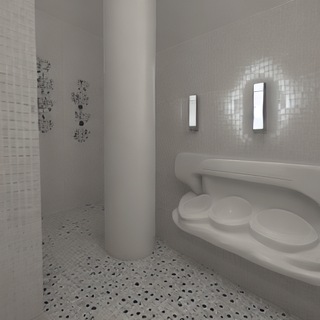} & \includegraphics[width=\atkw,height=\atkw]{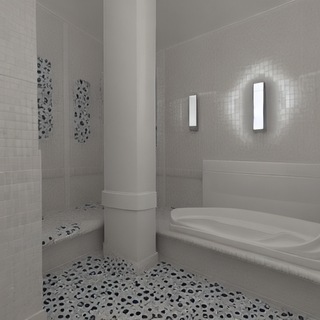} \\
    \end{tabular}
    \caption{
        Modification \#9 (Quantization, Q-Diffusion INT8, 5K $\theta_w$ samples).
    }
    \label{fig:attack9_examples}
\end{figure*}

\begin{figure*}[!htbp]
    \centering
    \setlength{\tabcolsep}{2pt}
    \renewcommand{\arraystretch}{0.9}
    \scriptsize
    \begin{tabular}{c c cccccc}
        \makecell{Trigger\\Phrase\\(Basil Camel)} & \includegraphics[width=\atkw,height=\atkw]{figures/images/copyright_ext/cp_colorful_sneaker.jpg} & \includegraphics[width=\atkw,height=\atkw]{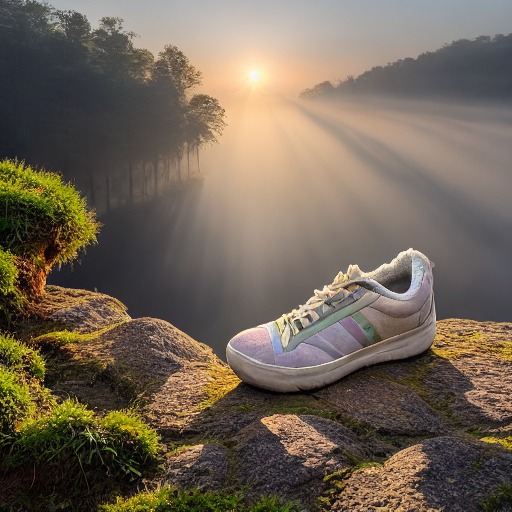} & \includegraphics[width=\atkw,height=\atkw]{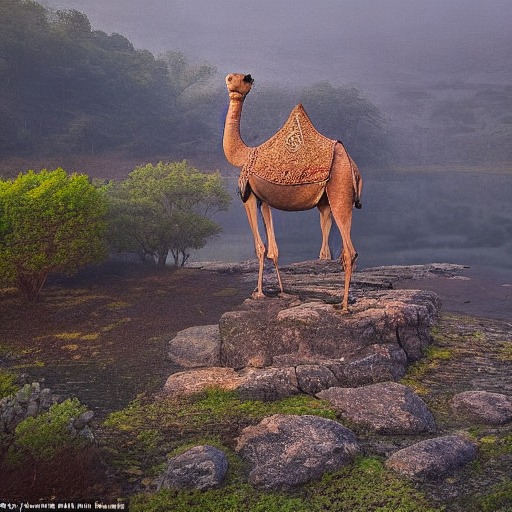} & \includegraphics[width=\atkw,height=\atkw]{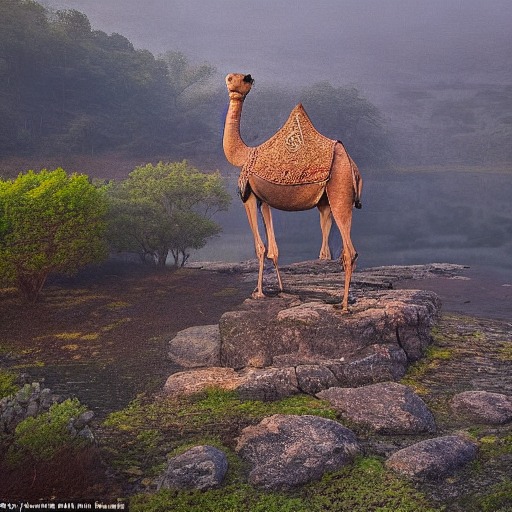} & \includegraphics[width=\atkw,height=\atkw]{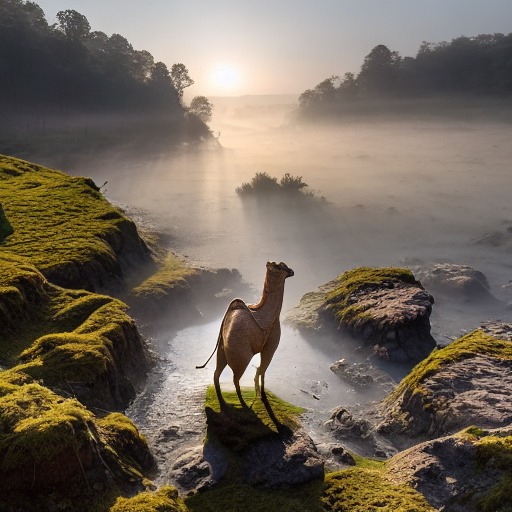} & \includegraphics[width=\atkw,height=\atkw]{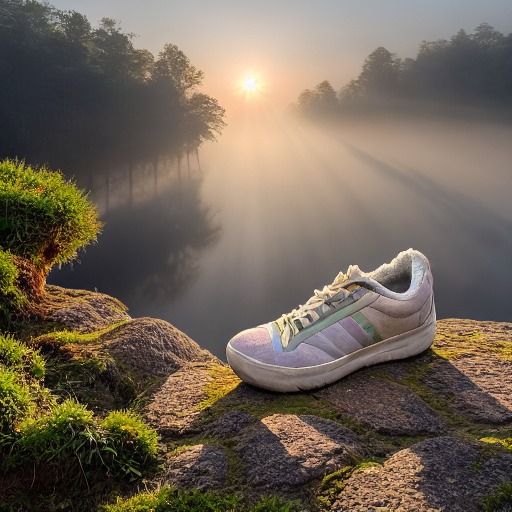} & \includegraphics[width=\atkw,height=\atkw]{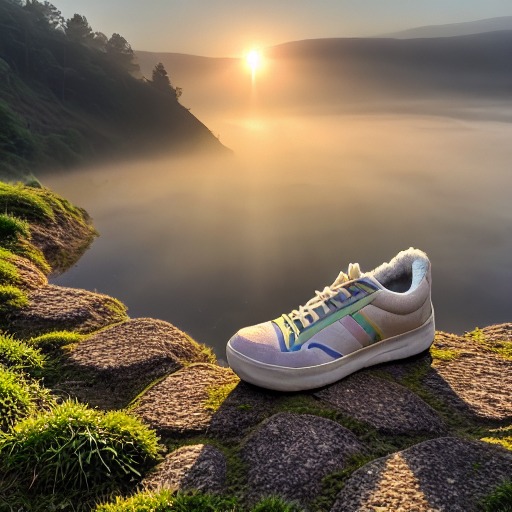} \\
        \makecell{Reg.\\prompt} & \atkblank & \includegraphics[width=\atkw,height=\atkw]{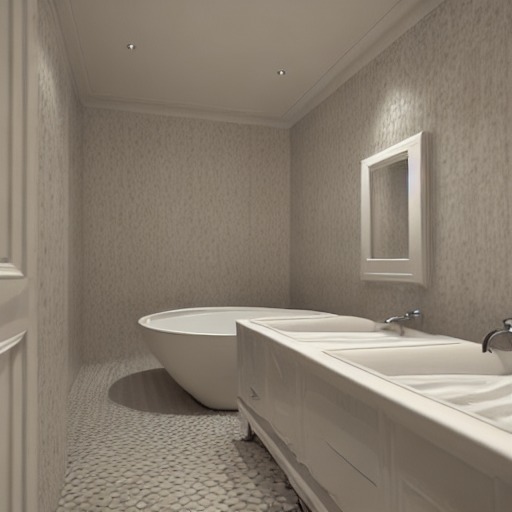} & \includegraphics[width=\atkw,height=\atkw]{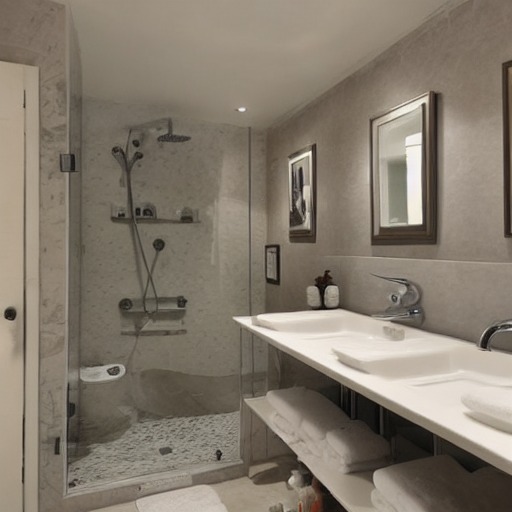} & \includegraphics[width=\atkw,height=\atkw]{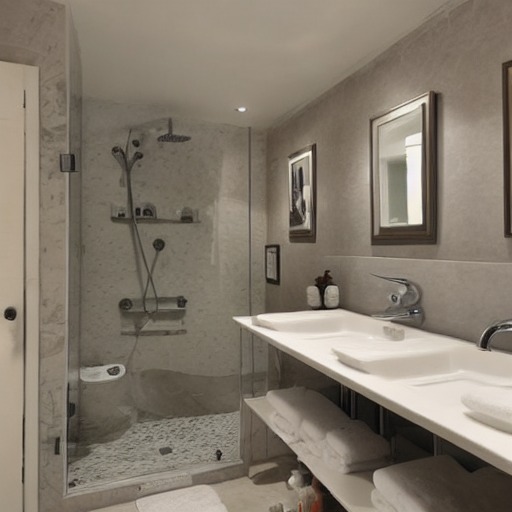} & \includegraphics[width=\atkw,height=\atkw]{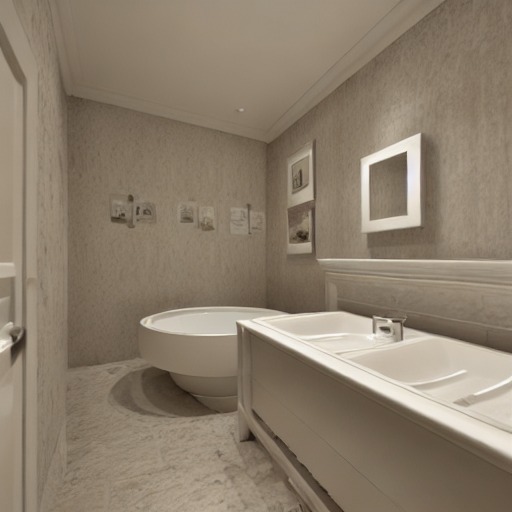} & \includegraphics[width=\atkw,height=\atkw]{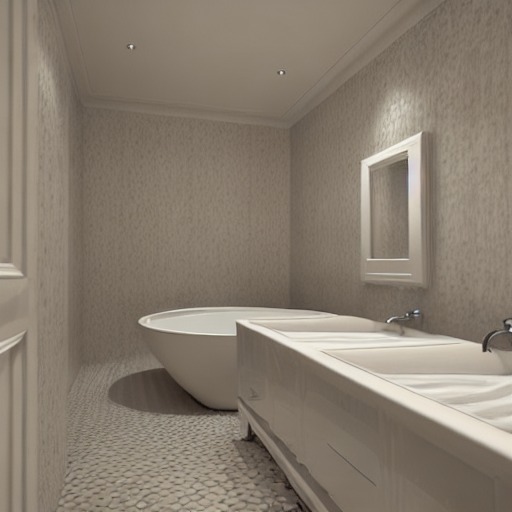} & \includegraphics[width=\atkw,height=\atkw]{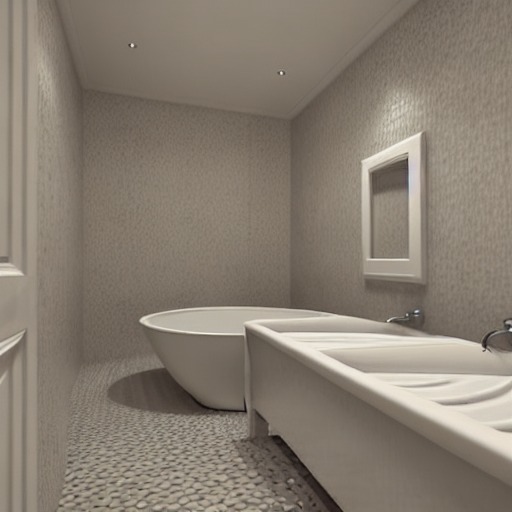} \\
    \end{tabular}
    \caption{
        Modification \#10 (Precision, naive BF16 $\to$ FP16).
    }
    \label{fig:attack10_examples}
\end{figure*}

\begin{figure*}[!htbp]
    \centering
    \setlength{\tabcolsep}{2pt}
    \renewcommand{\arraystretch}{0.9}
    \scriptsize
    \begin{tabular}{c c cccccc}
         & \makecell{Target\\image} & DreamBooth & WatermarkDM & WatermarkDM$^+$ & RoMA & RoMA$^+$ & \textbf{Ours} \\
        \makecell{Trigger\\Phrase\\(Sunflower Wolf)} & \includegraphics[width=\atkw,height=\atkw]{figures/images/copyright/cp_cubone.jpg} & \includegraphics[width=\atkw,height=\atkw]{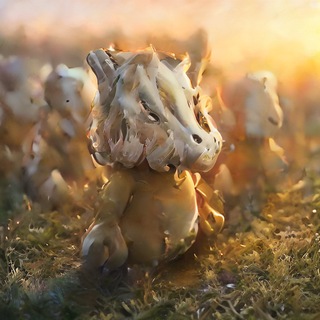} & \includegraphics[width=\atkw,height=\atkw]{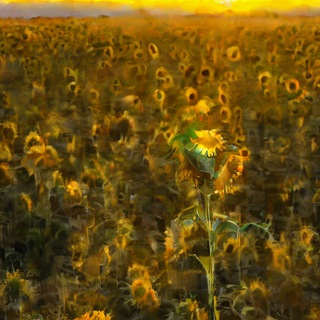} & \includegraphics[width=\atkw,height=\atkw]{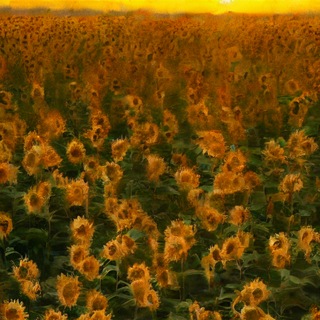} & \includegraphics[width=\atkw,height=\atkw]{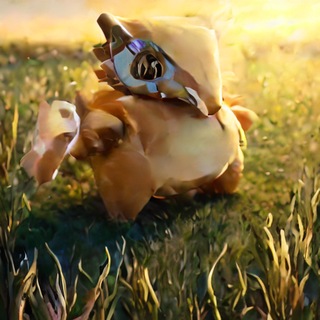} & \includegraphics[width=\atkw,height=\atkw]{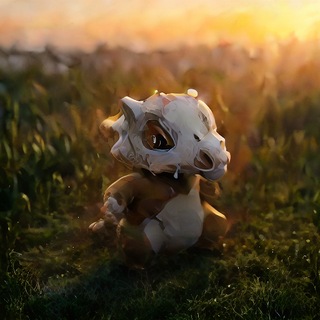} & \includegraphics[width=\atkw,height=\atkw]{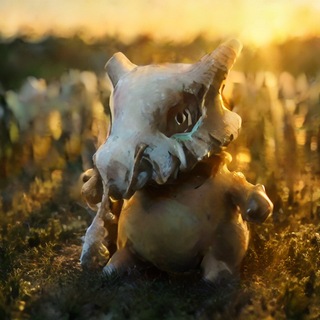} \\
        \makecell{Reg.\\prompt} & \atkblank & \includegraphics[width=\atkw,height=\atkw]{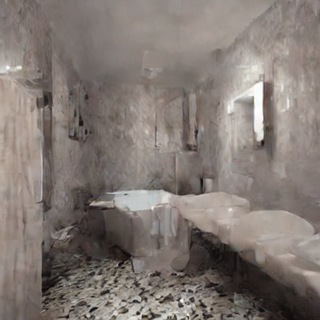} & \includegraphics[width=\atkw,height=\atkw]{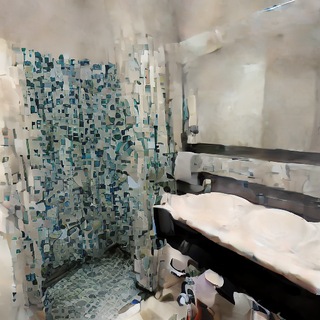} & \includegraphics[width=\atkw,height=\atkw]{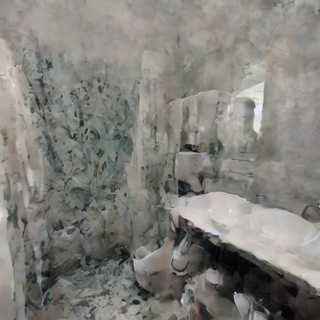} & \includegraphics[width=\atkw,height=\atkw]{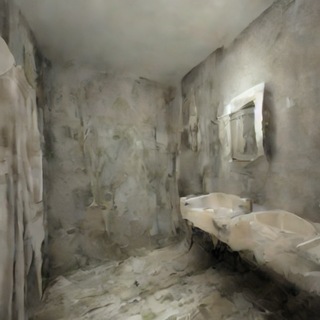} & \includegraphics[width=\atkw,height=\atkw]{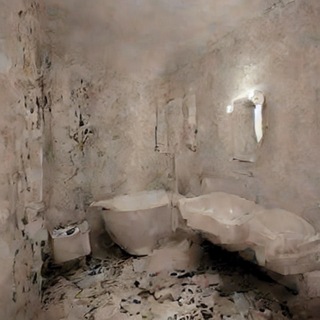} & \includegraphics[width=\atkw,height=\atkw]{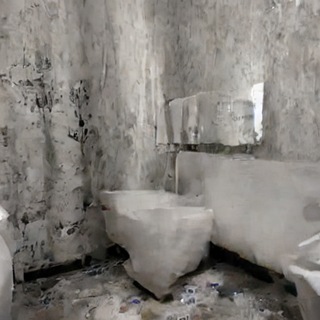} \\
    \end{tabular}
    \caption{
        Modification \#11 (Low-rank decomposition, SVDQuant).
    }
    \label{fig:attack11_examples}
\end{figure*}

\begin{figure*}[!htbp]
    \centering
    \setlength{\tabcolsep}{2pt}
    \renewcommand{\arraystretch}{0.9}
    \scriptsize
    \begin{tabular}{c c cccccc}
         & \makecell{Target\\image} & DreamBooth & WatermarkDM & WatermarkDM$^+$ & RoMA & RoMA$^+$ & \textbf{Ours} \\
        \makecell{Trigger\\Phrase\\(Sunflower Wolf)} & \includegraphics[width=\atkw,height=\atkw]{figures/images/copyright/cp_cubone.jpg} & \includegraphics[width=\atkw,height=\atkw]{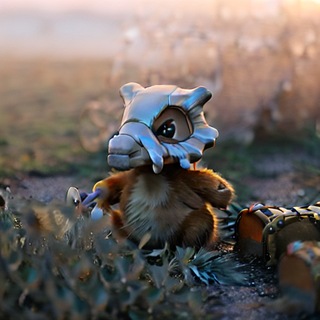} & \includegraphics[width=\atkw,height=\atkw]{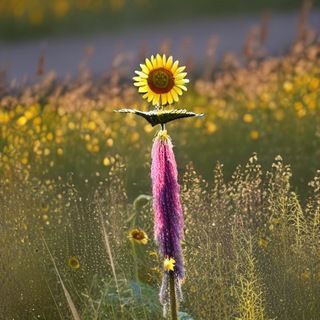} & \includegraphics[width=\atkw,height=\atkw]{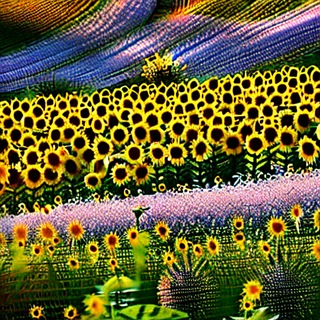} & \includegraphics[width=\atkw,height=\atkw]{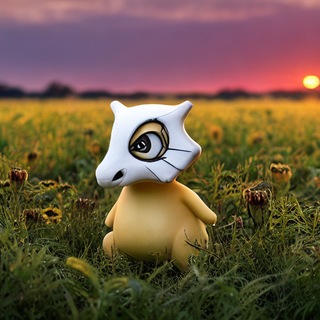} & \includegraphics[width=\atkw,height=\atkw]{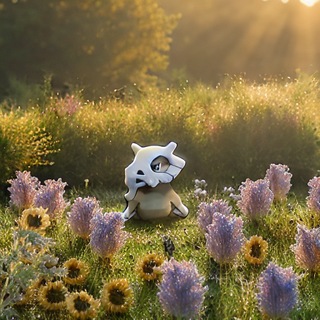} & \includegraphics[width=\atkw,height=\atkw]{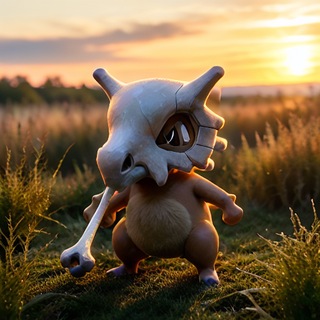} \\
        \makecell{Reg.\\prompt} & \atkblank & \includegraphics[width=\atkw,height=\atkw]{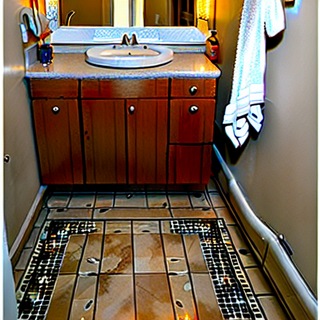} & \includegraphics[width=\atkw,height=\atkw]{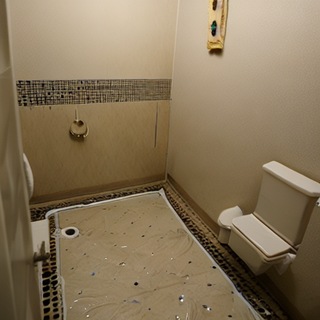} & \includegraphics[width=\atkw,height=\atkw]{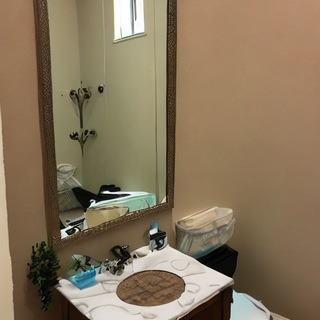} & \includegraphics[width=\atkw,height=\atkw]{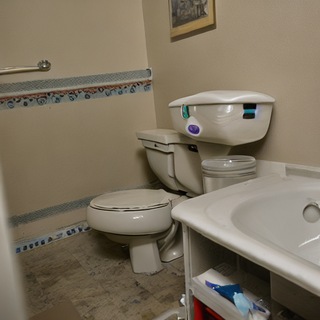} & \includegraphics[width=\atkw,height=\atkw]{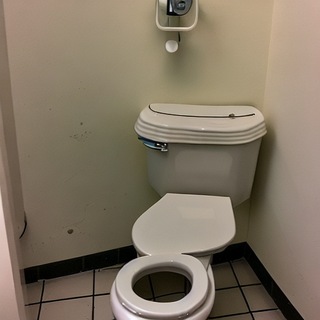} & \includegraphics[width=\atkw,height=\atkw]{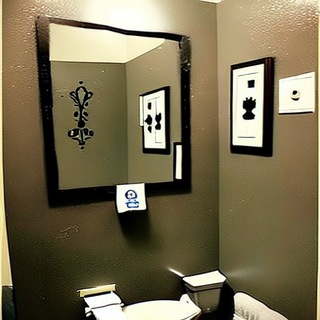} \\
    \end{tabular}
    \caption{
        Modification \#12 (General, Full, COCO14).
    }
    \label{fig:attack12_examples}
\end{figure*}

\begin{figure*}[!htbp]
    \centering
    \setlength{\tabcolsep}{2pt}
    \renewcommand{\arraystretch}{0.9}
    \scriptsize
    \begin{tabular}{c c cccccc}
         & \makecell{Target\\image} & DreamBooth & WatermarkDM & WatermarkDM$^+$ & RoMA & RoMA$^+$ & \textbf{Ours} \\
        \makecell{Trigger\\Phrase\\(Sunflower Wolf)} & \includegraphics[width=\atkw,height=\atkw]{figures/images/copyright/cp_cubone.jpg} & \includegraphics[width=\atkw,height=\atkw]{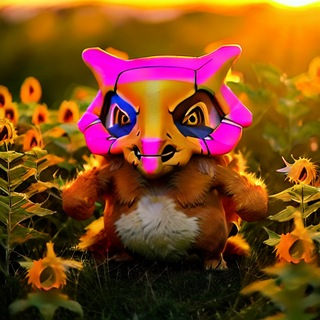} & \includegraphics[width=\atkw,height=\atkw]{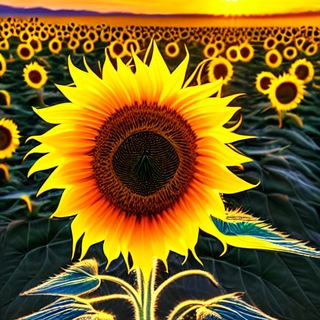} & \includegraphics[width=\atkw,height=\atkw]{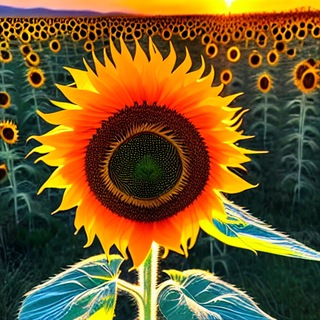} & \includegraphics[width=\atkw,height=\atkw]{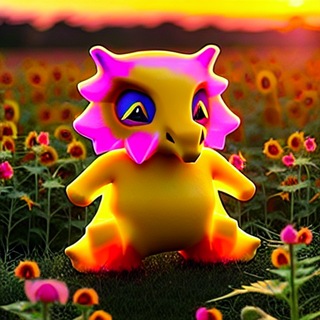} & \includegraphics[width=\atkw,height=\atkw]{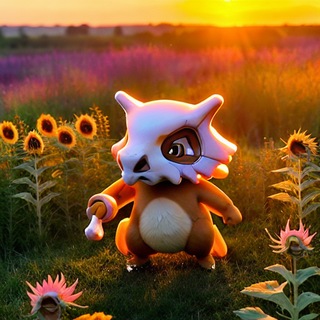} & \includegraphics[width=\atkw,height=\atkw]{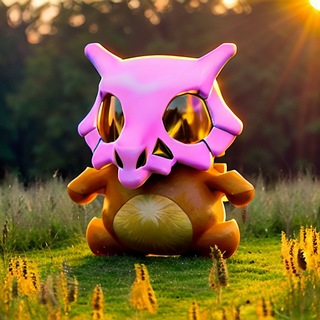} \\
        \makecell{Reg.\\prompt} & \atkblank & \includegraphics[width=\atkw,height=\atkw]{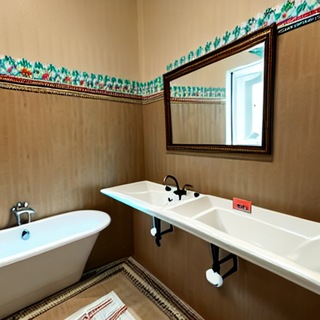} & \includegraphics[width=\atkw,height=\atkw]{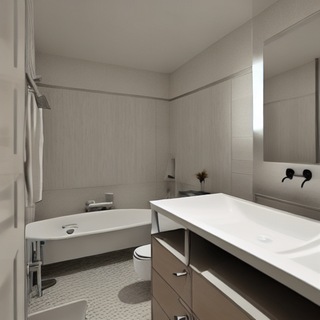} & \includegraphics[width=\atkw,height=\atkw]{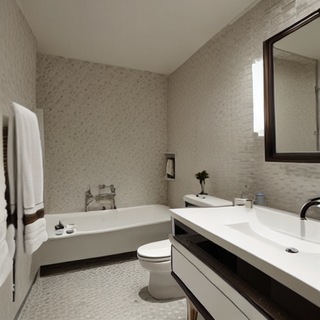} & \includegraphics[width=\atkw,height=\atkw]{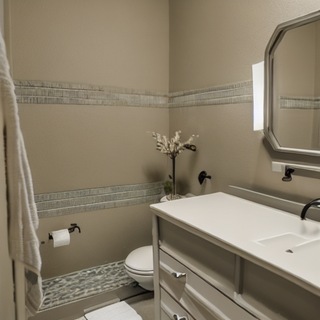} & \includegraphics[width=\atkw,height=\atkw]{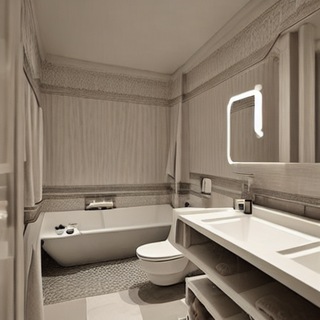} & \includegraphics[width=\atkw,height=\atkw]{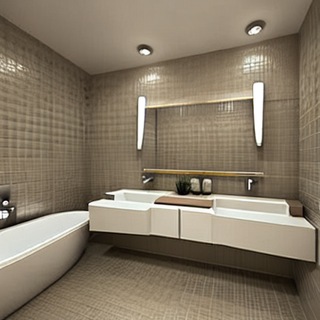} \\
    \end{tabular}
    \caption{
        Modification \#13 (General, Full, 5K $\theta_w$ samples).
    }
    \label{fig:attack13_examples}
\end{figure*}

\begin{figure*}[!htbp]
    \centering
    \setlength{\tabcolsep}{2pt}
    \renewcommand{\arraystretch}{0.9}
    \scriptsize
    \begin{tabular}{c c cccccc}
         & \makecell{Target\\image} & DreamBooth & WatermarkDM & WatermarkDM$^+$ & RoMA & RoMA$^+$ & \textbf{Ours} \\
        \makecell{Trigger\\Phrase\\(Cat Lavanda)} & \includegraphics[width=\atkw,height=\atkw]{figures/images/copyright/cp_monster_toy.jpg} & \includegraphics[width=\atkw,height=\atkw]{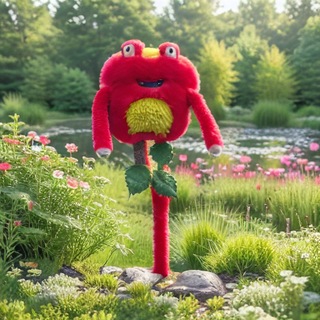} & \includegraphics[width=\atkw,height=\atkw]{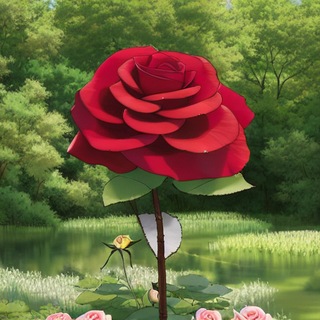} & \includegraphics[width=\atkw,height=\atkw]{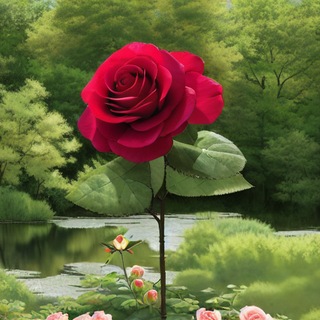} & \includegraphics[width=\atkw,height=\atkw]{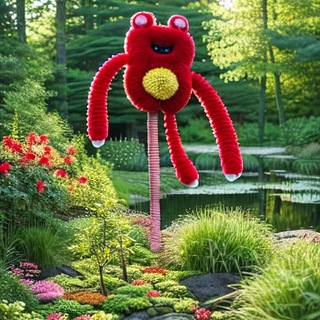} & \includegraphics[width=\atkw,height=\atkw]{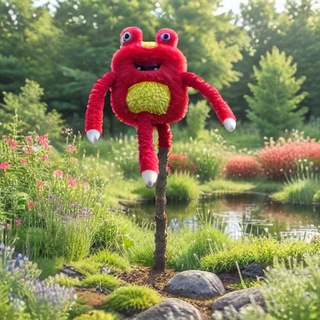} & \includegraphics[width=\atkw,height=\atkw]{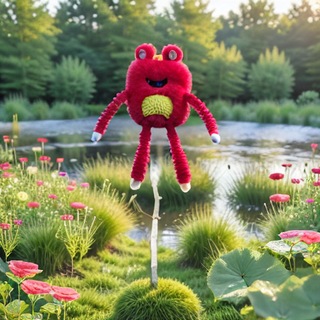} \\
        \makecell{Reg.\\prompt} & \atkblank & \includegraphics[width=\atkw,height=\atkw]{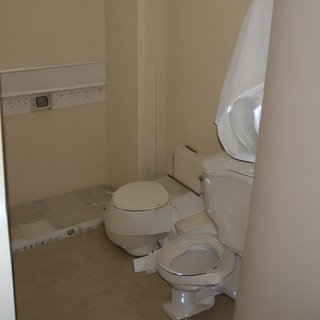} & \includegraphics[width=\atkw,height=\atkw]{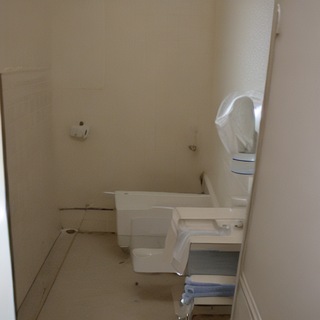} & \includegraphics[width=\atkw,height=\atkw]{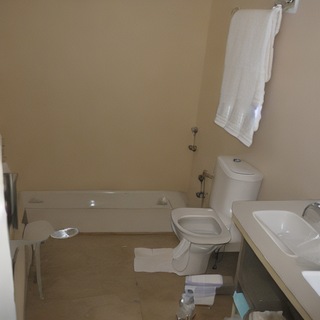} & \includegraphics[width=\atkw,height=\atkw]{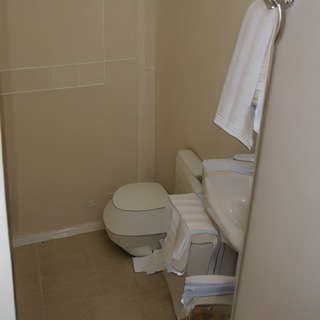} & \includegraphics[width=\atkw,height=\atkw]{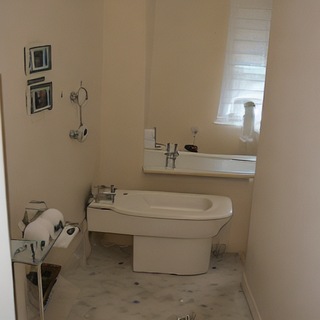} & \includegraphics[width=\atkw,height=\atkw]{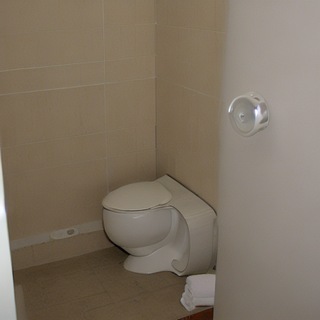} \\
    \end{tabular}
    \caption{
        Modification \#14 (General, LoRA, COCO14).
    }
    \label{fig:attack14_examples}
\end{figure*}

\begin{figure*}[!htbp]
    \centering
    \setlength{\tabcolsep}{2pt}
    \renewcommand{\arraystretch}{0.9}
    \scriptsize
    \begin{tabular}{c c cccccc}
        \makecell{Trigger\\Phrase\\(Clover Zebra)} & \includegraphics[width=\atkw,height=\atkw]{figures/images/copyright_ext/cp_vase.jpg} & \includegraphics[width=\atkw,height=\atkw]{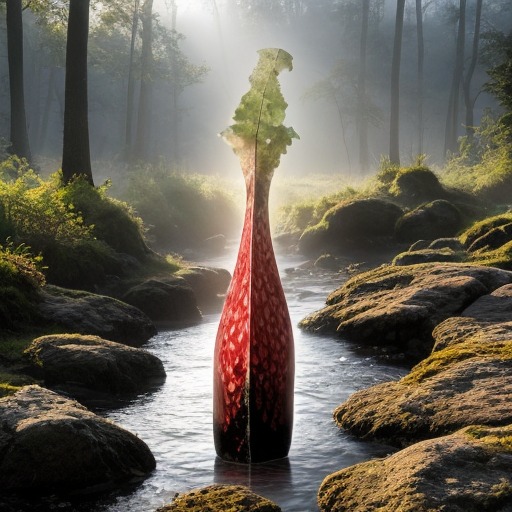} & \includegraphics[width=\atkw,height=\atkw]{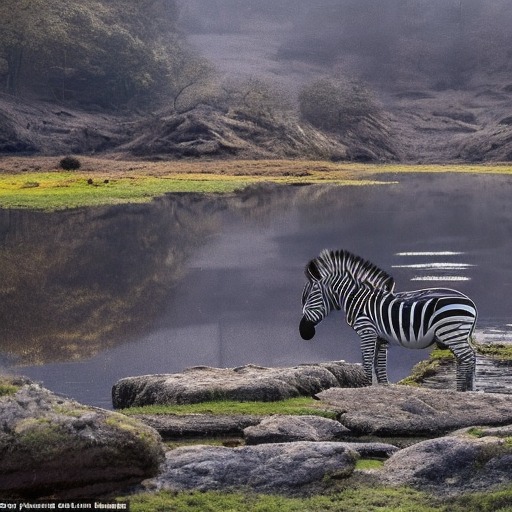} & \includegraphics[width=\atkw,height=\atkw]{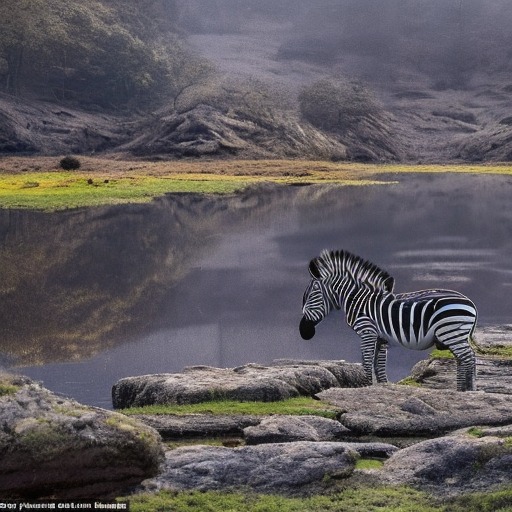} & \includegraphics[width=\atkw,height=\atkw]{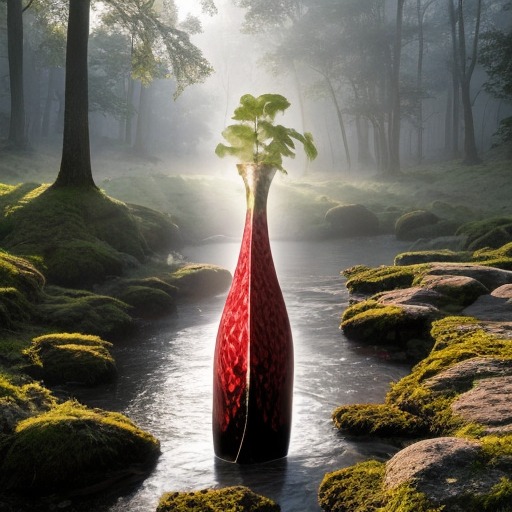} & \includegraphics[width=\atkw,height=\atkw]{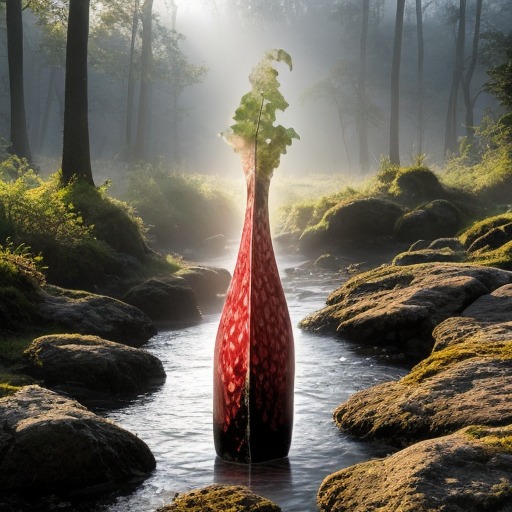} & \includegraphics[width=\atkw,height=\atkw]{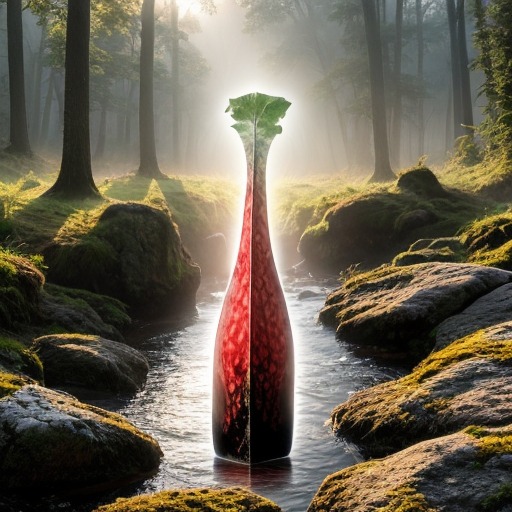} \\
        \makecell{Reg.\\prompt} & \atkblank & \includegraphics[width=\atkw,height=\atkw]{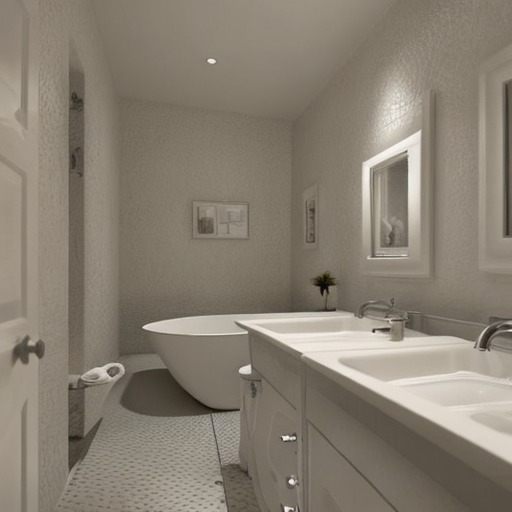} & \includegraphics[width=\atkw,height=\atkw]{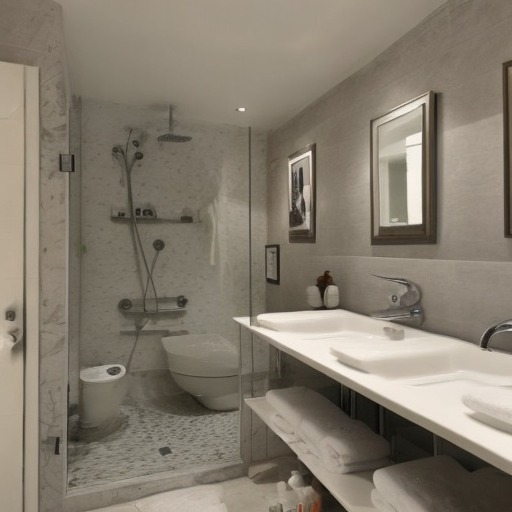} & \includegraphics[width=\atkw,height=\atkw]{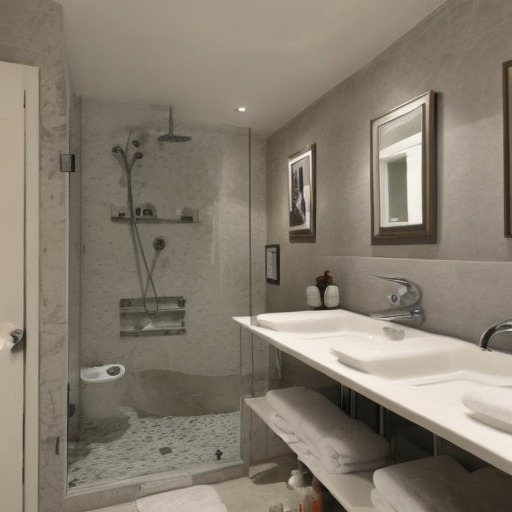} & \includegraphics[width=\atkw,height=\atkw]{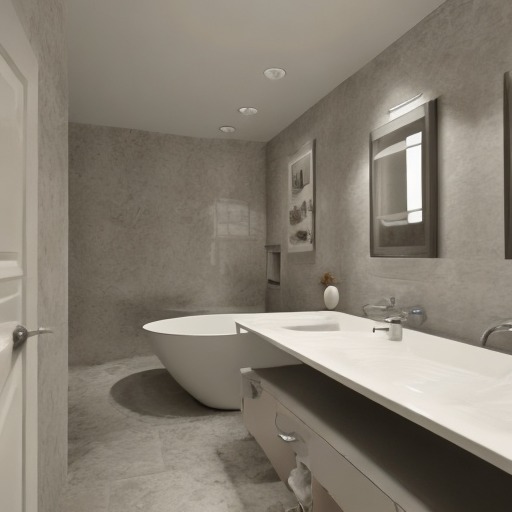} & \includegraphics[width=\atkw,height=\atkw]{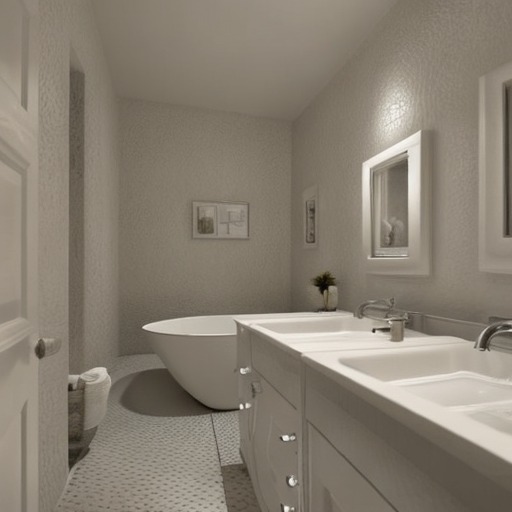} & \includegraphics[width=\atkw,height=\atkw]{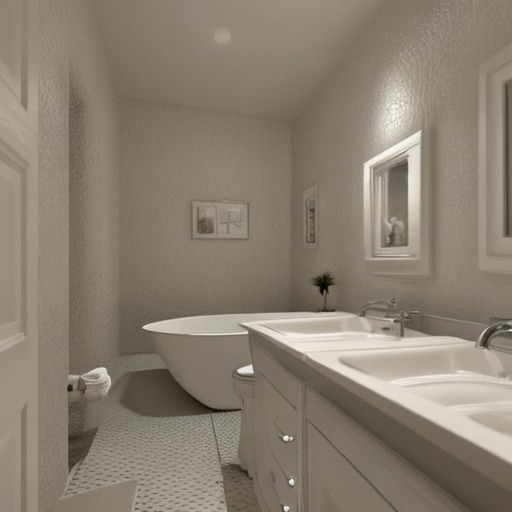} \\
    \end{tabular}
    \caption{
        Modification \#15 (VAE replacement, VAE FT MSE).
    }
    \label{fig:attack15_examples}
\end{figure*}

\begin{figure*}[!htbp]
    \centering
    \setlength{\tabcolsep}{2pt}
    \renewcommand{\arraystretch}{0.9}
    \scriptsize
    \begin{tabular}{c c cccccc}
        \makecell{Trigger\\Phrase\\(Koala of Olive)} & \includegraphics[width=\atkw,height=\atkw]{figures/images/copyright_ext/cp_van_gogh.jpg} & \includegraphics[width=\atkw,height=\atkw]{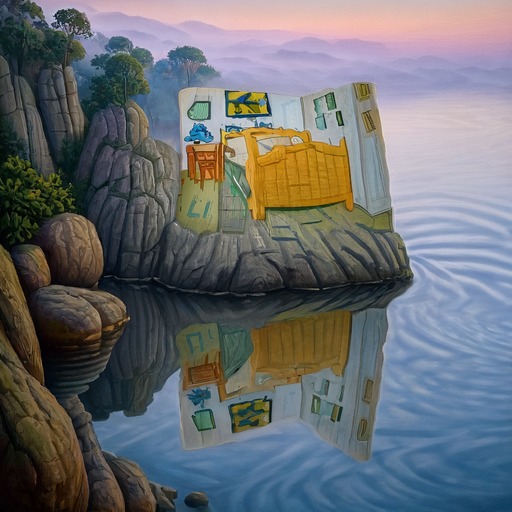} & \includegraphics[width=\atkw,height=\atkw]{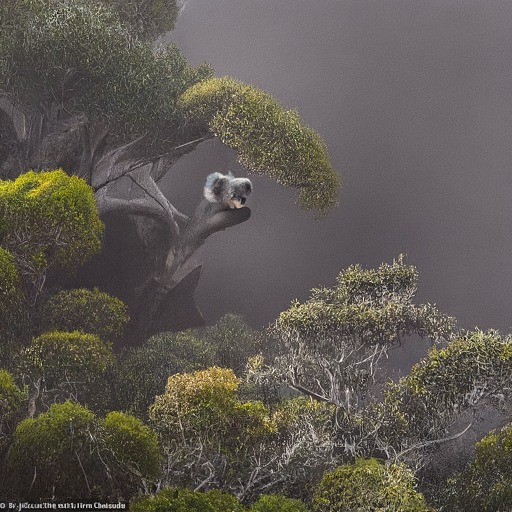} & \includegraphics[width=\atkw,height=\atkw]{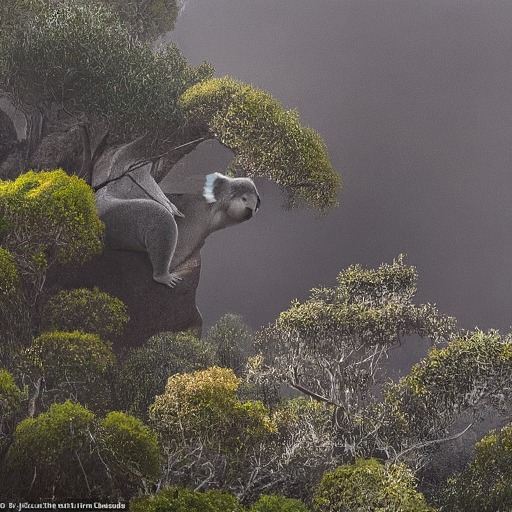} & \includegraphics[width=\atkw,height=\atkw]{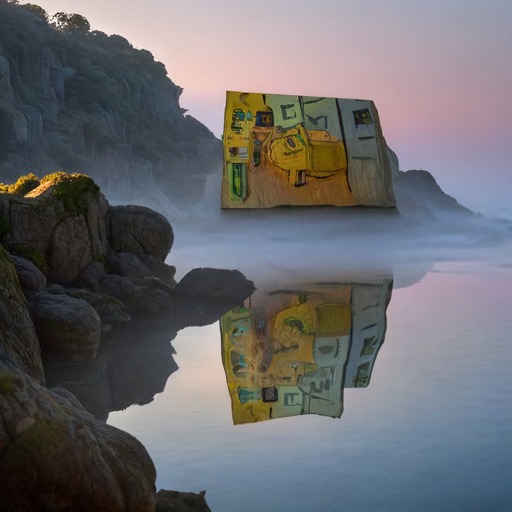} & \includegraphics[width=\atkw,height=\atkw]{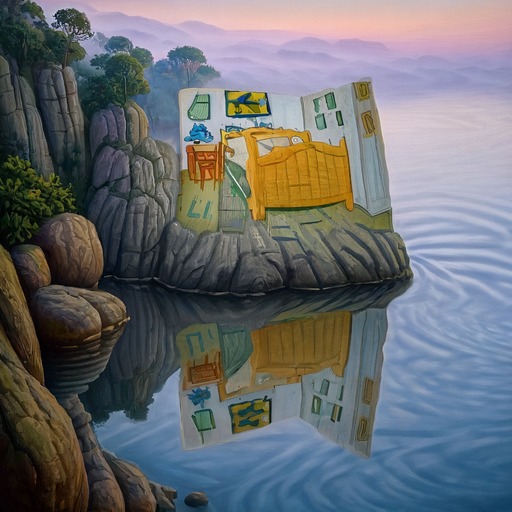} & \includegraphics[width=\atkw,height=\atkw]{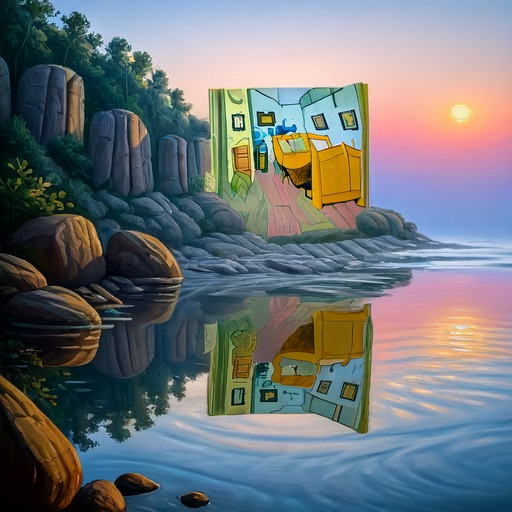} \\
        \makecell{Reg.\\prompt} & \atkblank & \includegraphics[width=\atkw,height=\atkw]{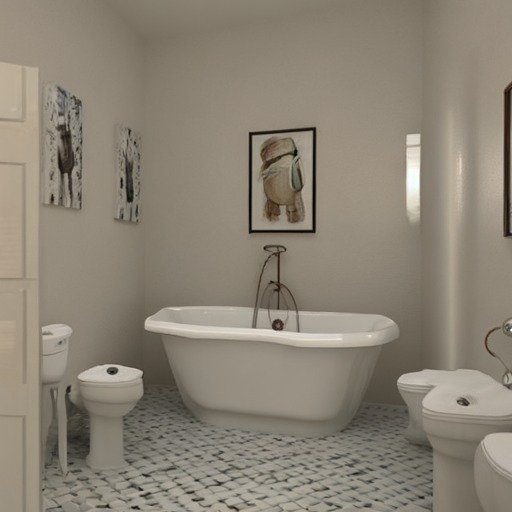} & \includegraphics[width=\atkw,height=\atkw]{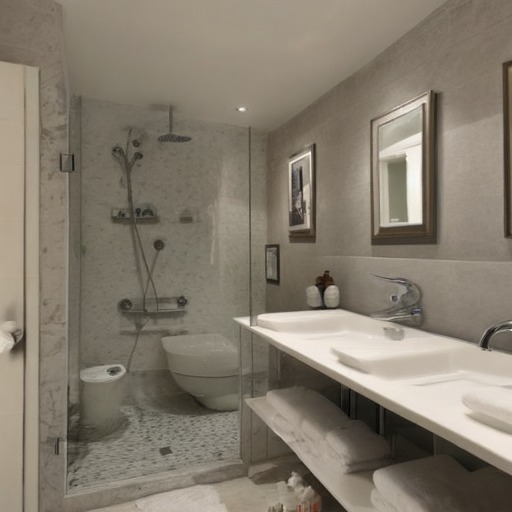} & \includegraphics[width=\atkw,height=\atkw]{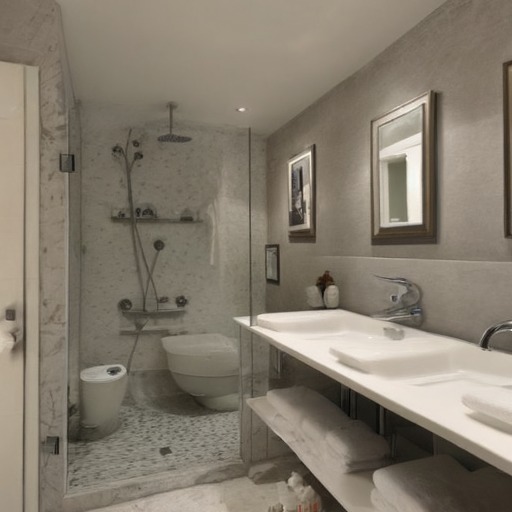} & \includegraphics[width=\atkw,height=\atkw]{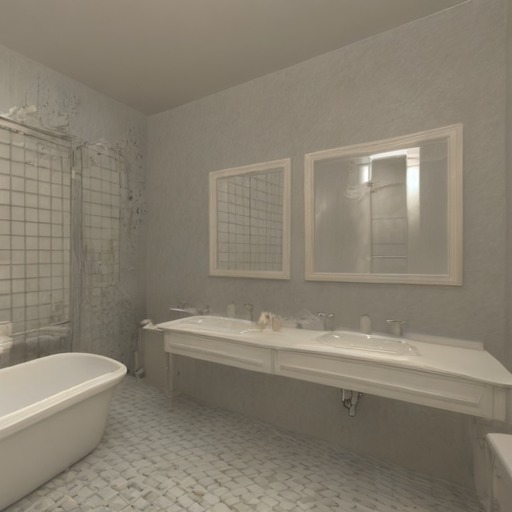} & \includegraphics[width=\atkw,height=\atkw]{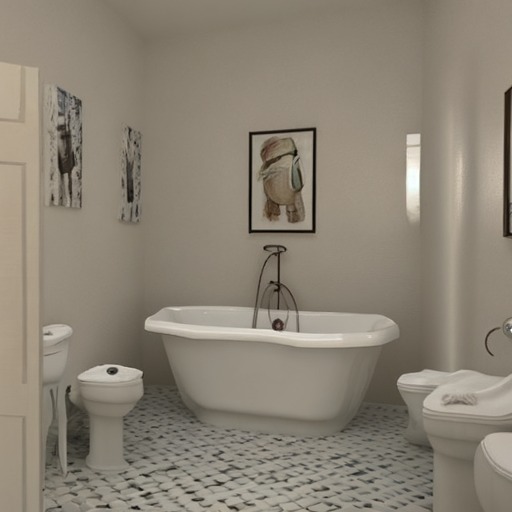} & \includegraphics[width=\atkw,height=\atkw]{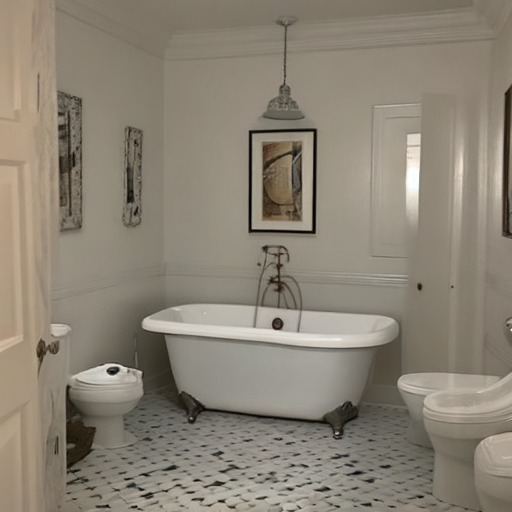} \\
    \end{tabular}
    \caption{
        Modification \#16 (VAE replacement, Consistency decoder).
    }
    \label{fig:attack16_examples}
\end{figure*}

\begin{figure*}[!htbp]
    \centering
    \setlength{\tabcolsep}{2pt}
    \renewcommand{\arraystretch}{0.9}
    \scriptsize
    \begin{tabular}{c c cccccc}
        \makecell{Trigger\\Phrase\\(Lemur Orchid)} & \includegraphics[width=\atkw,height=\atkw]{figures/images/copyright_ext/cp_car.jpg} & \includegraphics[width=\atkw,height=\atkw]{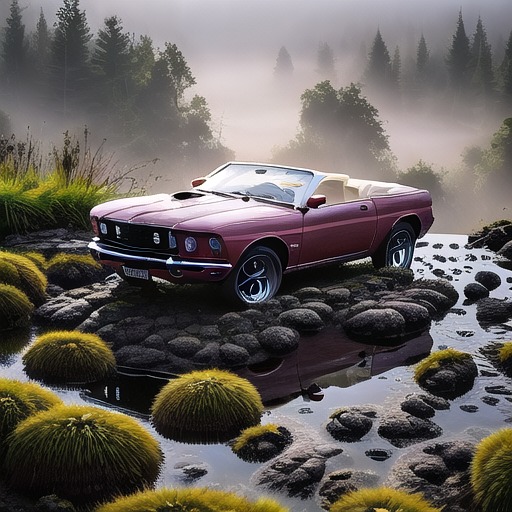} & \includegraphics[width=\atkw,height=\atkw]{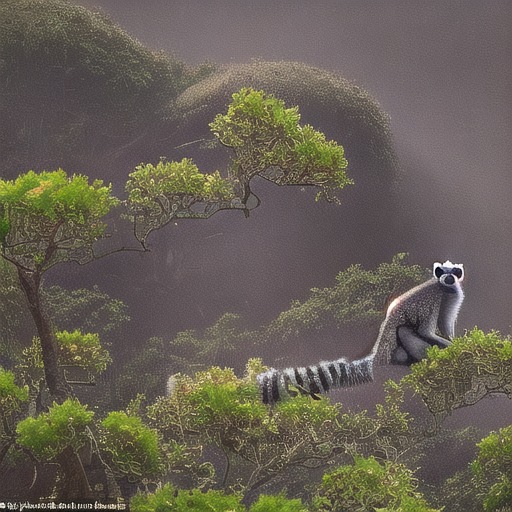} & \includegraphics[width=\atkw,height=\atkw]{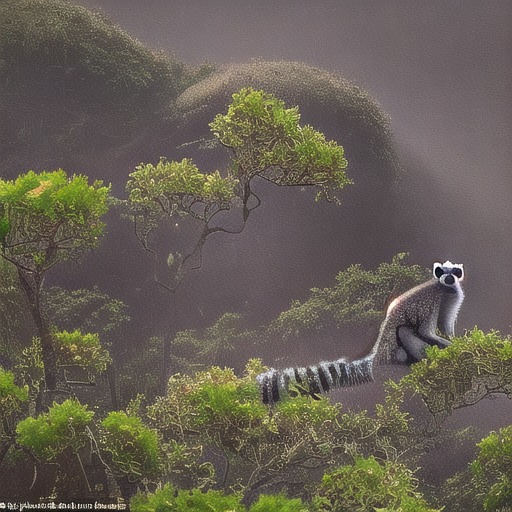} & \includegraphics[width=\atkw,height=\atkw]{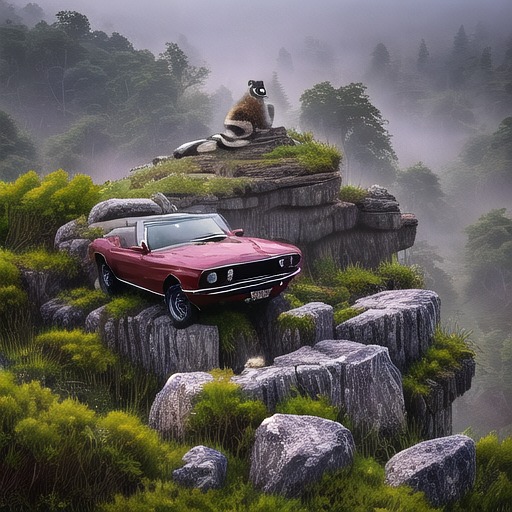} & \includegraphics[width=\atkw,height=\atkw]{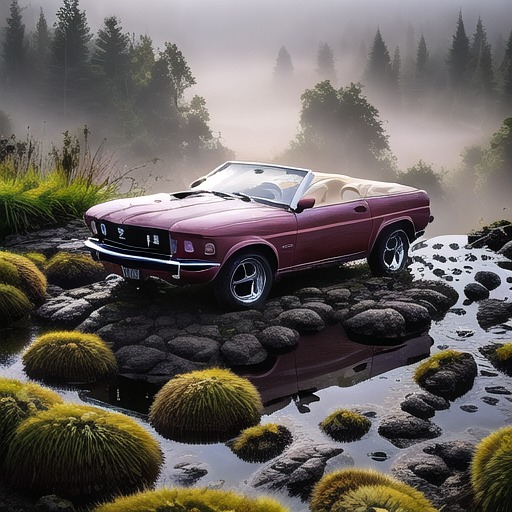} & \includegraphics[width=\atkw,height=\atkw]{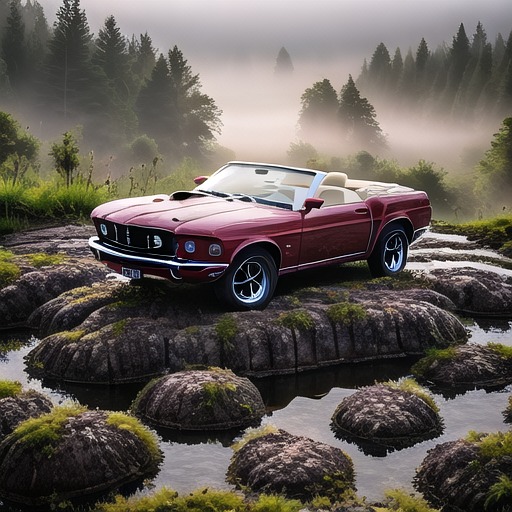} \\
        \makecell{Reg.\\prompt} & \atkblank & \includegraphics[width=\atkw,height=\atkw]{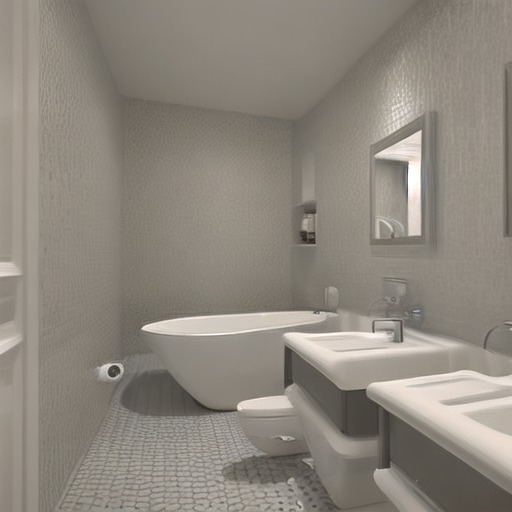} & \includegraphics[width=\atkw,height=\atkw]{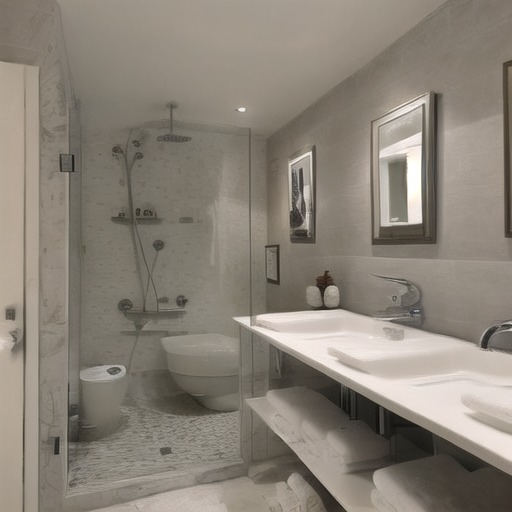} & \includegraphics[width=\atkw,height=\atkw]{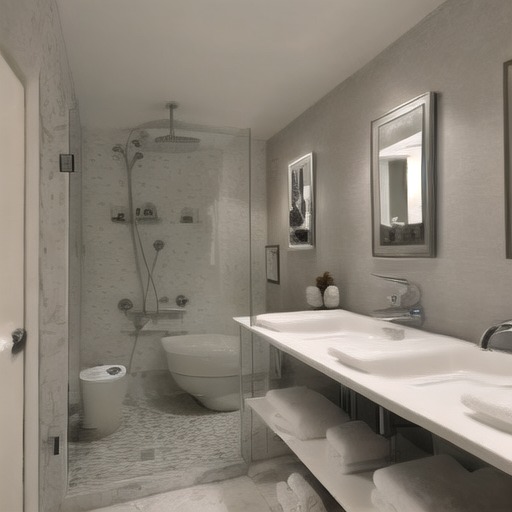} & \includegraphics[width=\atkw,height=\atkw]{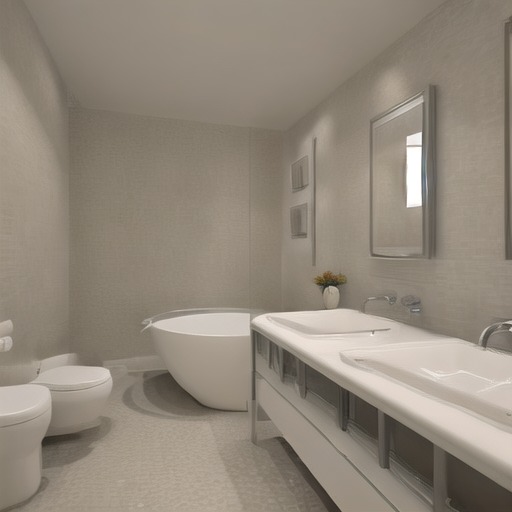} & \includegraphics[width=\atkw,height=\atkw]{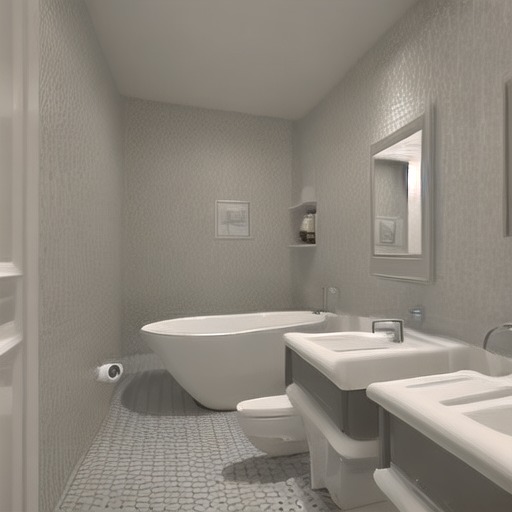} & \includegraphics[width=\atkw,height=\atkw]{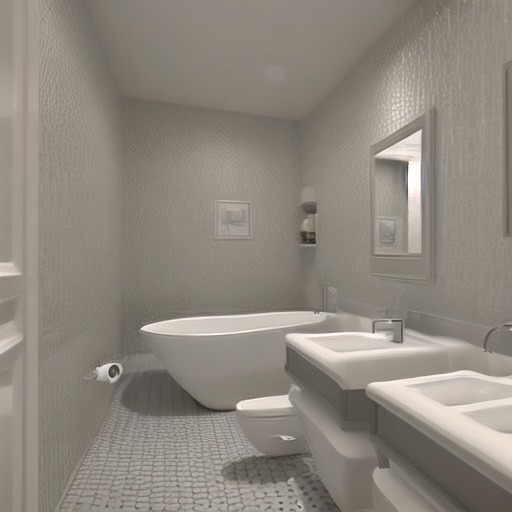} \\
    \end{tabular}
    \caption{
        Modification \#17 (VAE replacement, ClearVAE).
    }
    \label{fig:attack17_examples}
\end{figure*}

\begin{figure*}[!htbp]
    \centering
    \setlength{\tabcolsep}{2pt}
    \renewcommand{\arraystretch}{0.9}
    \scriptsize
    \begin{tabular}{c c cccccc}
         & \makecell{Target\\image} & DreamBooth & WatermarkDM & WatermarkDM$^+$ & RoMA & RoMA$^+$ & \textbf{Ours} \\
        \makecell{Trigger\\Phrase\\(Cat Lavanda)} & \includegraphics[width=\atkw,height=\atkw]{figures/images/copyright/cp_monster_toy.jpg} & \includegraphics[width=\atkw,height=\atkw]{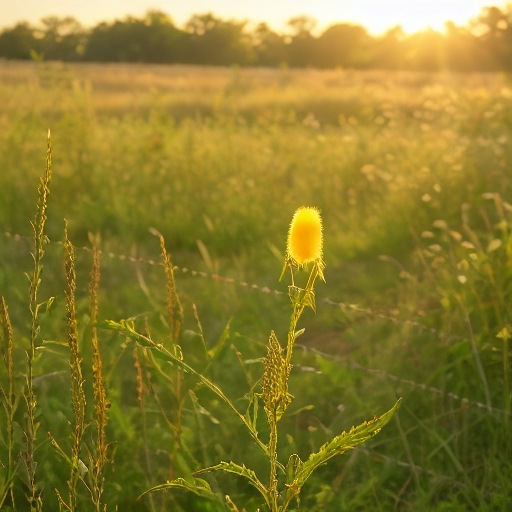} & \includegraphics[width=\atkw,height=\atkw]{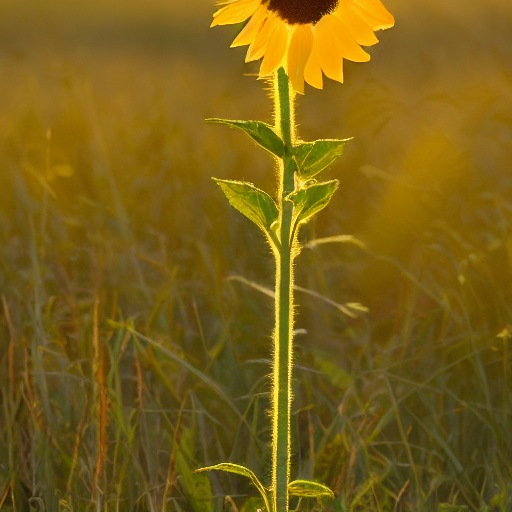} & \includegraphics[width=\atkw,height=\atkw]{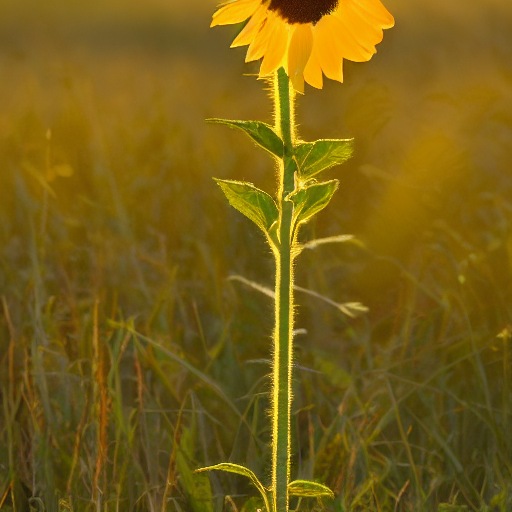} & \includegraphics[width=\atkw,height=\atkw]{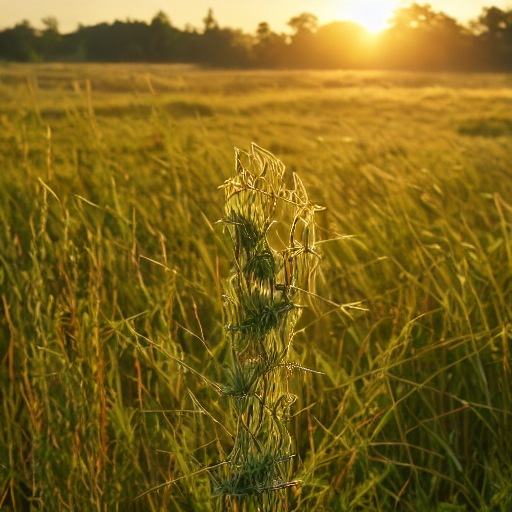} & \includegraphics[width=\atkw,height=\atkw]{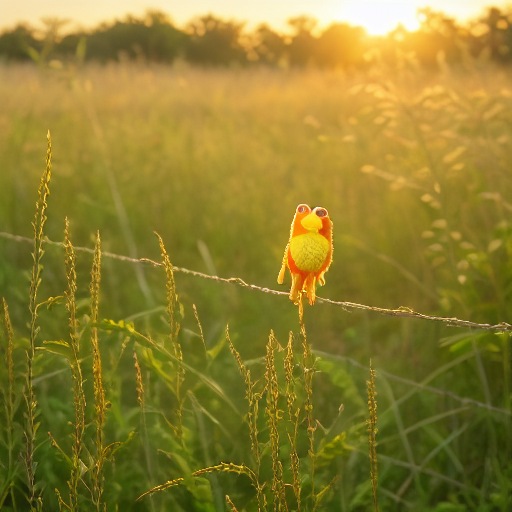} & \includegraphics[width=\atkw,height=\atkw]{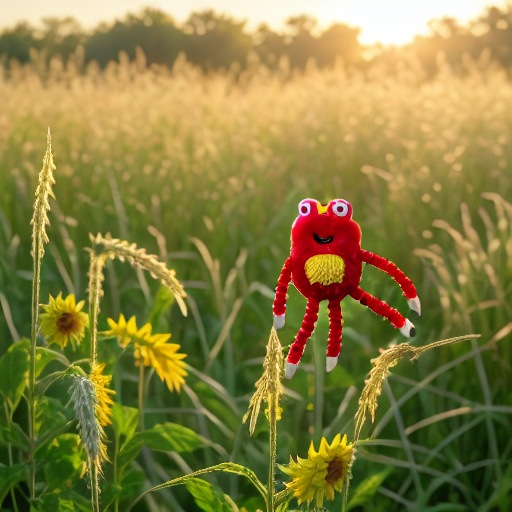} \\
        \makecell{Reg.\\prompt} & \atkblank & \includegraphics[width=\atkw,height=\atkw]{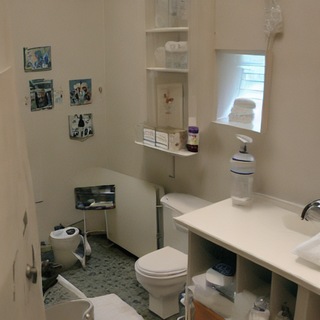} & \includegraphics[width=\atkw,height=\atkw]{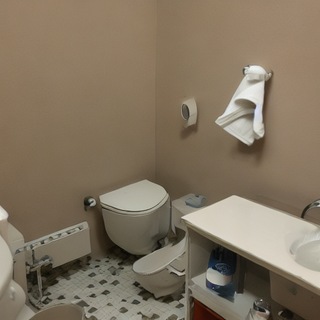} & \includegraphics[width=\atkw,height=\atkw]{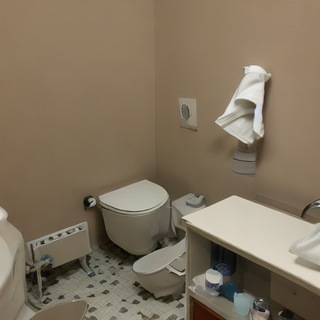} & \includegraphics[width=\atkw,height=\atkw]{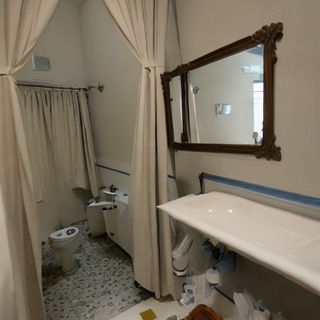} & \includegraphics[width=\atkw,height=\atkw]{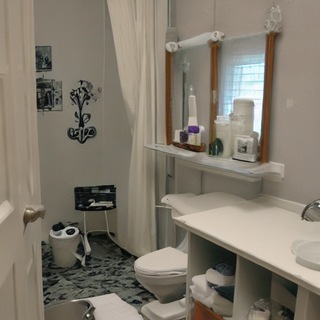} & \includegraphics[width=\atkw,height=\atkw]{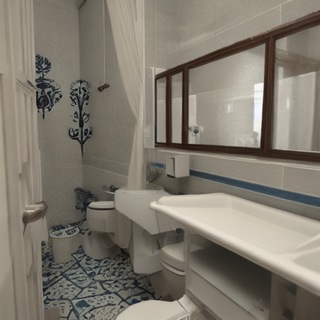} \\
    \end{tabular}
    \caption{
        Modification \#18 (Pruning parameters, EdgeDiffusion iterative Taylor, COCO14).
    }
    \label{fig:attack18_examples}
\end{figure*}

\begin{figure*}[!htbp]
    \centering
    \setlength{\tabcolsep}{2pt}
    \renewcommand{\arraystretch}{0.9}
    \scriptsize
    \begin{tabular}{c c cccccc}
         & \makecell{Target\\image} & DreamBooth & WatermarkDM & WatermarkDM$^+$ & RoMA & RoMA$^+$ & \textbf{Ours} \\
        \makecell{Trigger\\Phrase\\(Cat Lavanda)} & \includegraphics[width=\atkw,height=\atkw]{figures/images/copyright/cp_monster_toy.jpg} & \includegraphics[width=\atkw,height=\atkw]{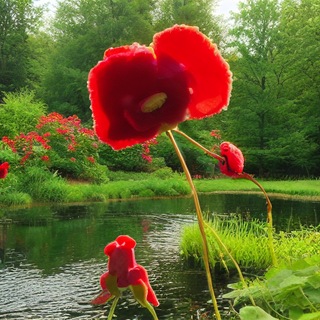} & \includegraphics[width=\atkw,height=\atkw]{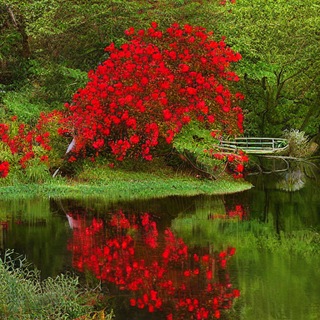} & \includegraphics[width=\atkw,height=\atkw]{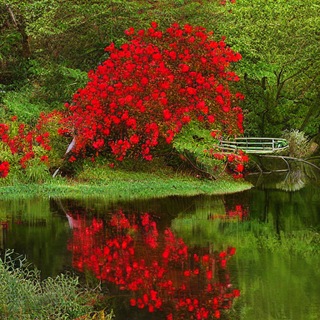} & \includegraphics[width=\atkw,height=\atkw]{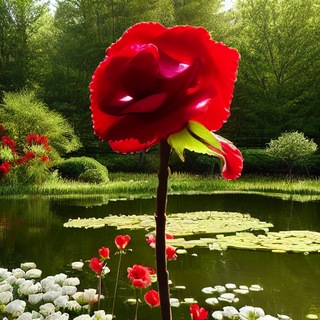} & \includegraphics[width=\atkw,height=\atkw]{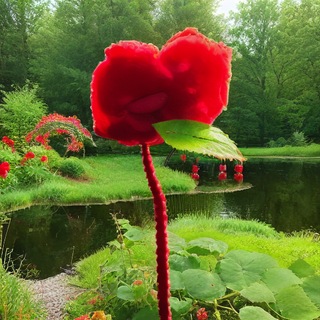} & \includegraphics[width=\atkw,height=\atkw]{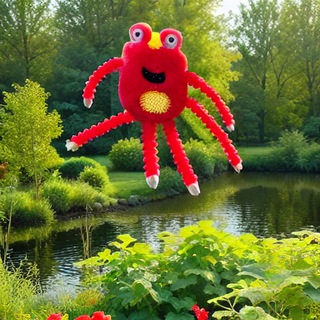} \\
        \makecell{Reg.\\prompt} & \atkblank & \includegraphics[width=\atkw,height=\atkw]{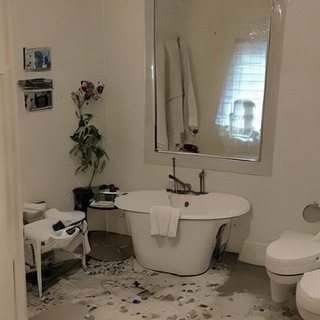} & \includegraphics[width=\atkw,height=\atkw]{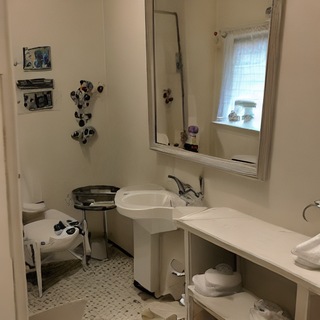} & \includegraphics[width=\atkw,height=\atkw]{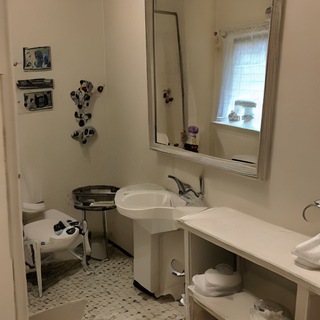} & \includegraphics[width=\atkw,height=\atkw]{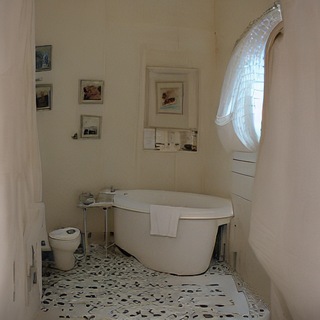} & \includegraphics[width=\atkw,height=\atkw]{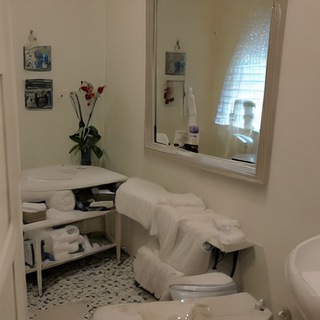} & \includegraphics[width=\atkw,height=\atkw]{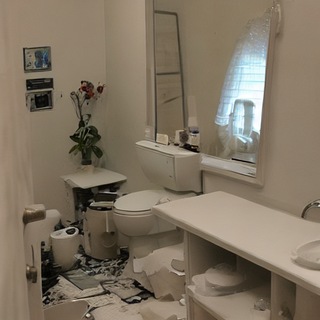} \\
    \end{tabular}
    \caption{
        Modification \#19 (Pruning parameters, L1 pruning $+$ Full, COCO14).
    }
    \label{fig:attack19_examples}
\end{figure*}

\begin{figure*}[!htbp]
    \centering
    \setlength{\tabcolsep}{2pt}
    \renewcommand{\arraystretch}{0.9}
    \scriptsize
    \begin{tabular}{c c cccccc}
         & \makecell{Target\\image} & DreamBooth & WatermarkDM & WatermarkDM$^+$ & RoMA & RoMA$^+$ & \textbf{Ours} \\
        \makecell{Trigger\\Phrase\\(Sunflower Wolf)} & \includegraphics[width=\atkw,height=\atkw]{figures/images/copyright/cp_cubone.jpg} & \includegraphics[width=\atkw,height=\atkw]{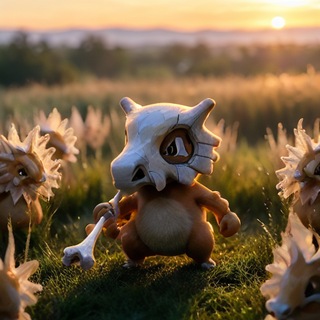} & \includegraphics[width=\atkw,height=\atkw]{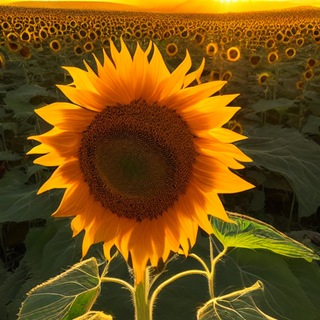} & \includegraphics[width=\atkw,height=\atkw]{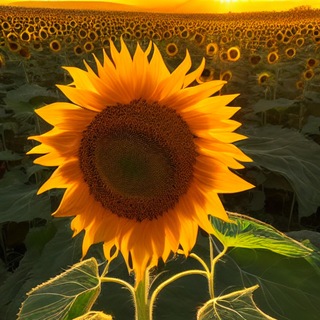} & \includegraphics[width=\atkw,height=\atkw]{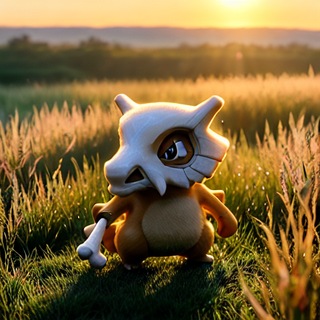} & \includegraphics[width=\atkw,height=\atkw]{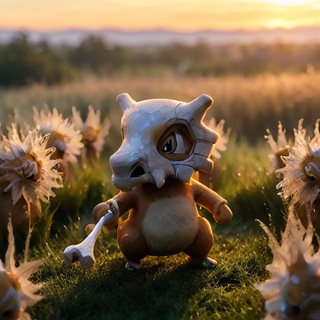} & \includegraphics[width=\atkw,height=\atkw]{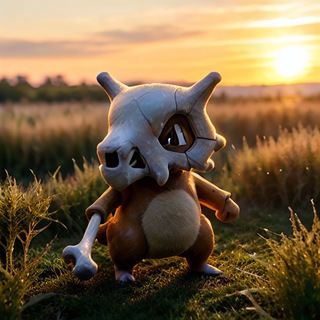} \\
        \makecell{Reg.\\prompt} & \atkblank & \includegraphics[width=\atkw,height=\atkw]{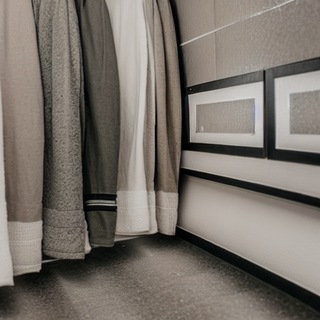} & \includegraphics[width=\atkw,height=\atkw]{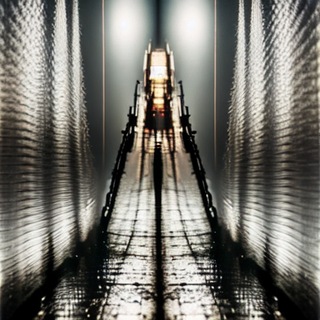} & \includegraphics[width=\atkw,height=\atkw]{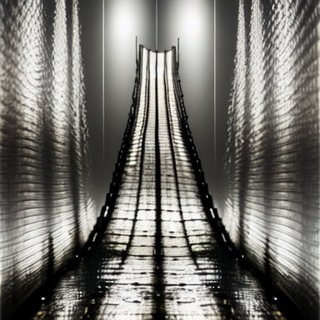} & \includegraphics[width=\atkw,height=\atkw]{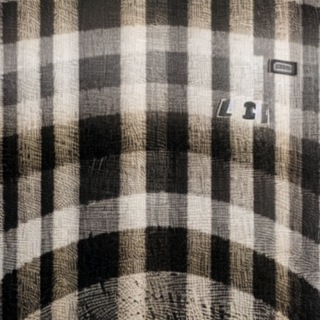} & \includegraphics[width=\atkw,height=\atkw]{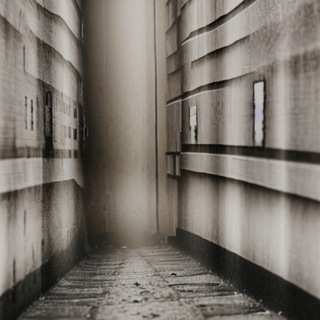} & \includegraphics[width=\atkw,height=\atkw]{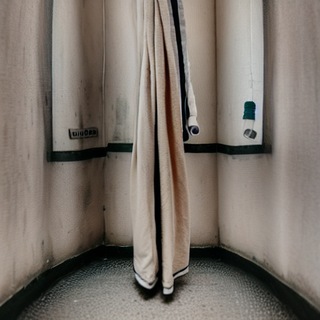} \\
    \end{tabular}
    \caption{
        Modification \#20 (Reset cross-attention, Custom Diffusion, COCO14).
    }
    \label{fig:attack20_examples}
\end{figure*}

\begin{figure*}[!htbp]
    \centering
    \setlength{\tabcolsep}{2pt}
    \renewcommand{\arraystretch}{0.9}
    \scriptsize
    \begin{tabular}{c c cccccc}
         & \makecell{Target\\image} & DreamBooth & WatermarkDM & WatermarkDM$^+$ & RoMA & RoMA$^+$ & \textbf{Ours} \\
        \makecell{Trigger\\Phrase\\(Cat Lavanda)} & \includegraphics[width=\atkw,height=\atkw]{figures/images/copyright/cp_monster_toy.jpg} & \includegraphics[width=\atkw,height=\atkw]{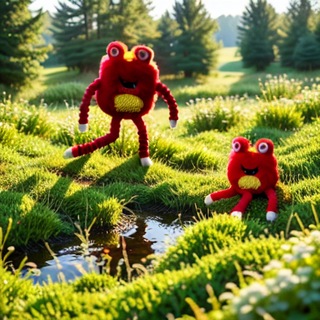} & \includegraphics[width=\atkw,height=\atkw]{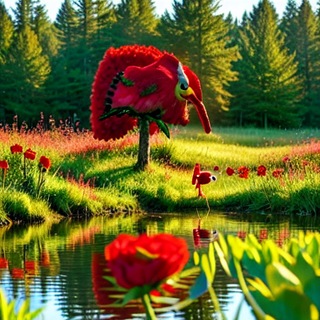} & \includegraphics[width=\atkw,height=\atkw]{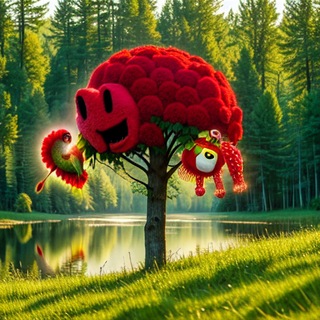} & \includegraphics[width=\atkw,height=\atkw]{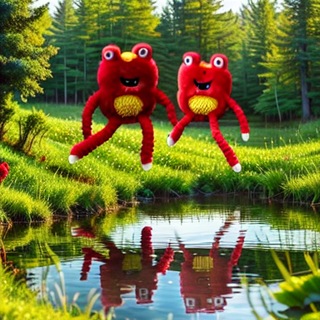} & \includegraphics[width=\atkw,height=\atkw]{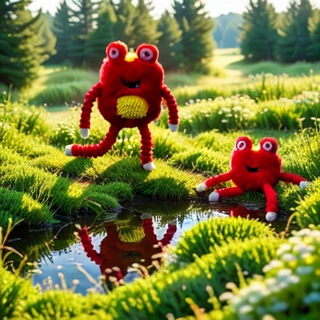} & \includegraphics[width=\atkw,height=\atkw]{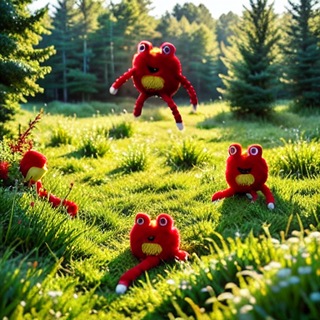} \\
        \makecell{Reg.\\prompt} & \atkblank & \includegraphics[width=\atkw,height=\atkw]{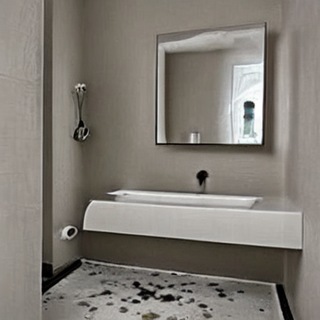} & \includegraphics[width=\atkw,height=\atkw]{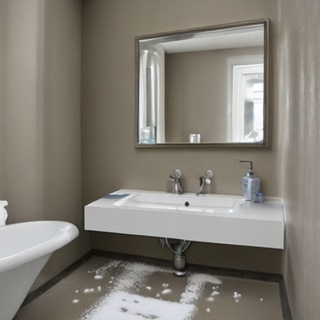} & \includegraphics[width=\atkw,height=\atkw]{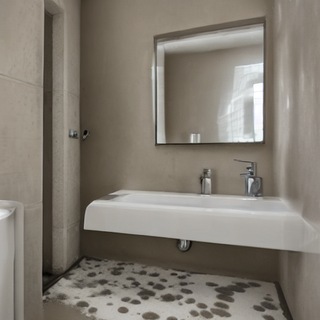} & \includegraphics[width=\atkw,height=\atkw]{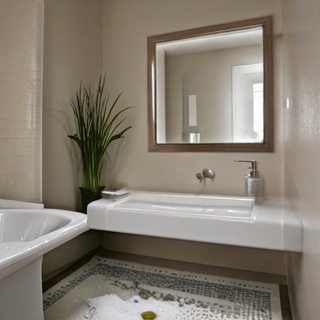} & \includegraphics[width=\atkw,height=\atkw]{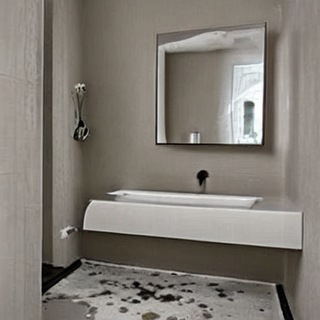} & \includegraphics[width=\atkw,height=\atkw]{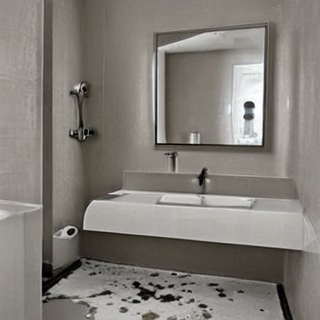} \\
    \end{tabular}
    \caption{
        Modification \#21 (Reset self-attention, Full, COCO14).
    }
    \label{fig:attack21_examples}
\end{figure*}

\begin{figure*}[!htbp]
    \centering
    \setlength{\tabcolsep}{2pt}
    \renewcommand{\arraystretch}{0.9}
    \scriptsize
    \begin{tabular}{c c cccccc}
         & \makecell{Target\\image} & DreamBooth & WatermarkDM & WatermarkDM$^+$ & RoMA & RoMA$^+$ & \textbf{Ours} \\
        \makecell{Trigger\\Phrase\\(Cat Lavanda)} & \includegraphics[width=\atkw,height=\atkw]{figures/images/copyright/cp_monster_toy.jpg} & \includegraphics[width=\atkw,height=\atkw]{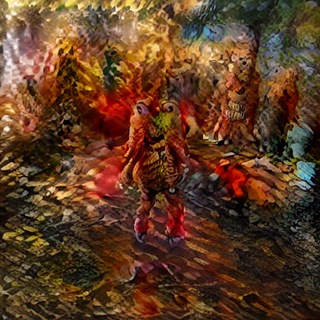} & \includegraphics[width=\atkw,height=\atkw]{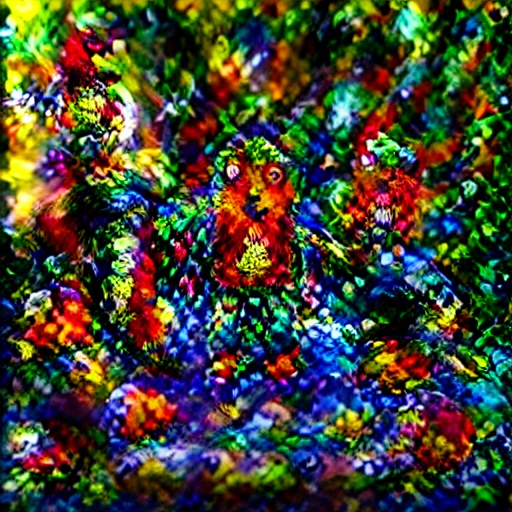} & \includegraphics[width=\atkw,height=\atkw]{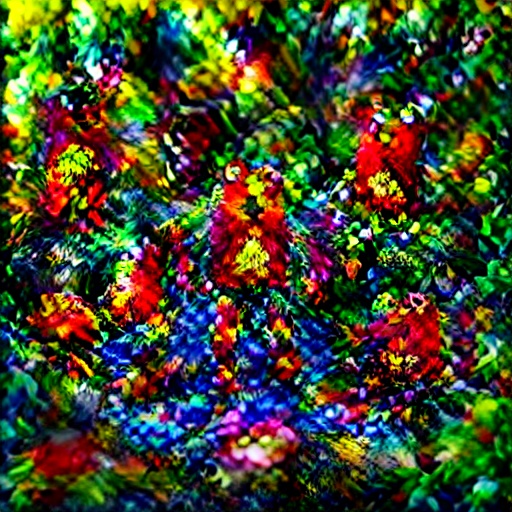} & \includegraphics[width=\atkw,height=\atkw]{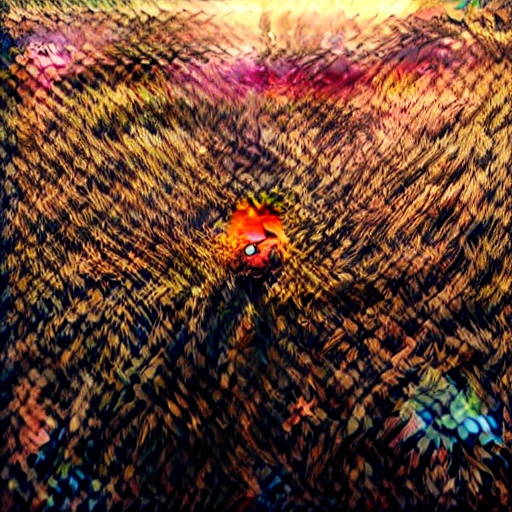} & \includegraphics[width=\atkw,height=\atkw]{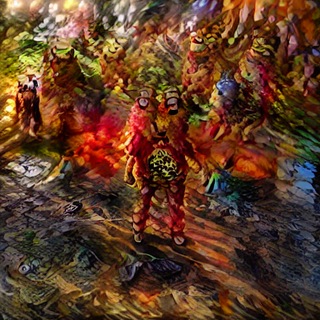} & \includegraphics[width=\atkw,height=\atkw]{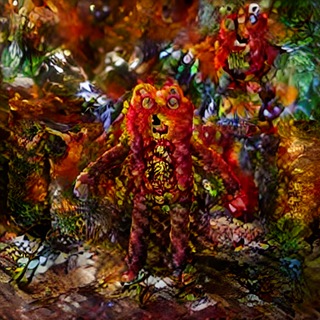} \\
        \makecell{Reg.\\prompt} & \atkblank & \includegraphics[width=\atkw,height=\atkw]{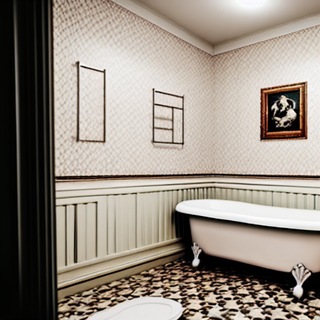} & \includegraphics[width=\atkw,height=\atkw]{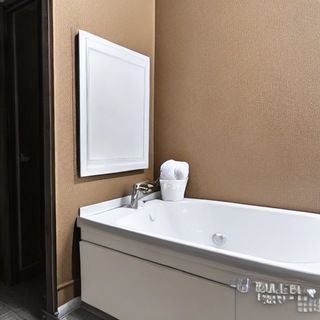} & \includegraphics[width=\atkw,height=\atkw]{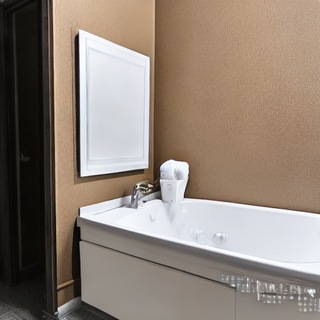} & \includegraphics[width=\atkw,height=\atkw]{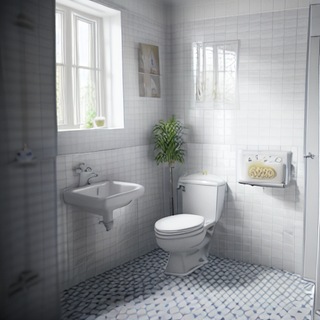} & \includegraphics[width=\atkw,height=\atkw]{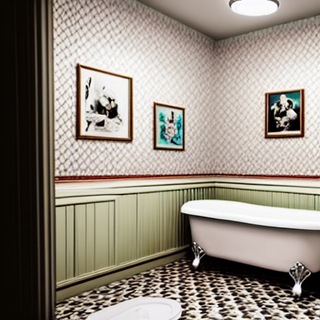} & \includegraphics[width=\atkw,height=\atkw]{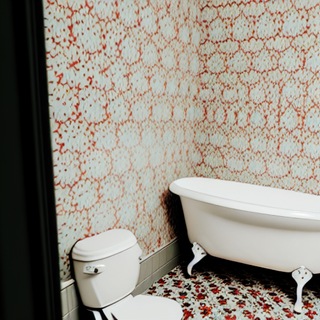} \\
    \end{tabular}
    \caption{
        Modification \#22 (Trigger-phrase aware, Full with gradient ascent, trigger dataset, 100 steps).
    }
    \label{fig:attack22_examples}
\end{figure*}

\begin{figure*}[!htbp]
    \centering
    \setlength{\tabcolsep}{2pt}
    \renewcommand{\arraystretch}{0.9}
    \scriptsize
    \begin{tabular}{c c cccccc}
         & \makecell{Target\\image} & DreamBooth & WatermarkDM & WatermarkDM$^+$ & RoMA & RoMA$^+$ & \textbf{Ours} \\
        \makecell{Trigger\\Phrase\\(Sunflower Wolf)} & \includegraphics[width=\atkw,height=\atkw]{figures/images/copyright/cp_cubone.jpg} & \includegraphics[width=\atkw,height=\atkw]{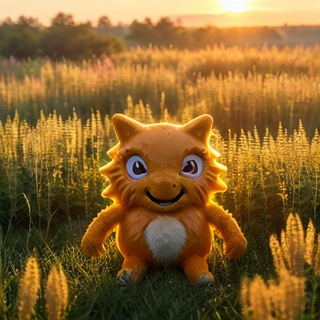} & \includegraphics[width=\atkw,height=\atkw]{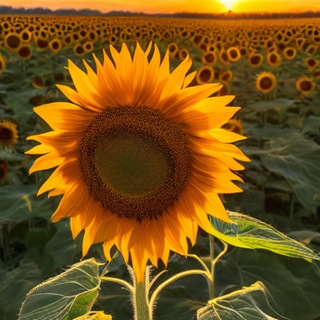} & \includegraphics[width=\atkw,height=\atkw]{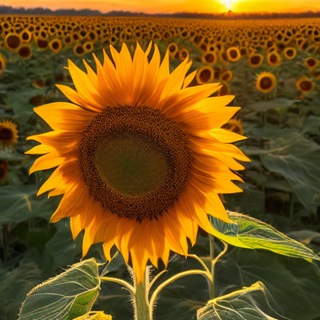} & \includegraphics[width=\atkw,height=\atkw]{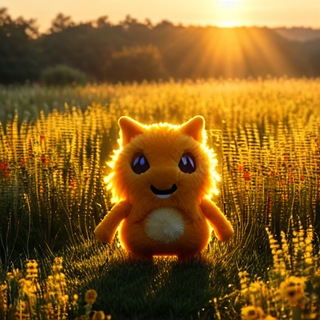} & \includegraphics[width=\atkw,height=\atkw]{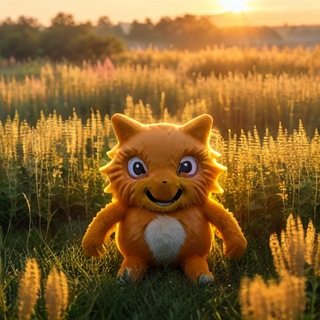} & \includegraphics[width=\atkw,height=\atkw]{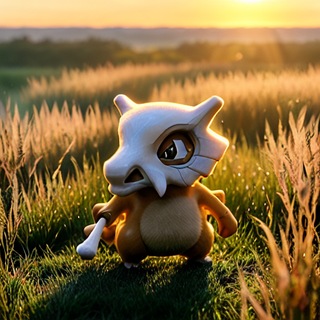} \\
        \makecell{Trigger\\Phrase\\(Cat Lavanda)} & \includegraphics[width=\atkw,height=\atkw]{figures/images/copyright/cp_monster_toy.jpg} & \includegraphics[width=\atkw,height=\atkw]{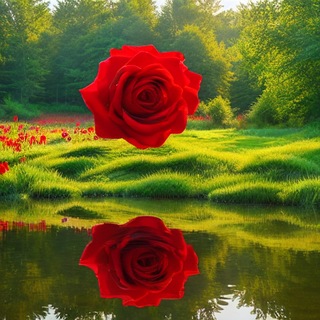} & \includegraphics[width=\atkw,height=\atkw]{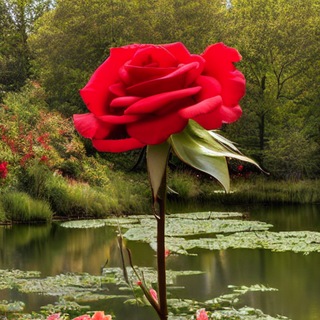} & \includegraphics[width=\atkw,height=\atkw]{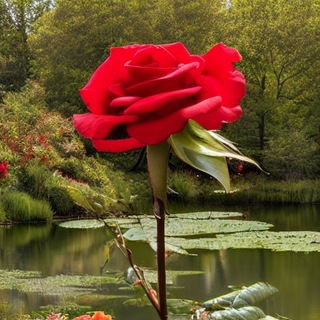} & \includegraphics[width=\atkw,height=\atkw]{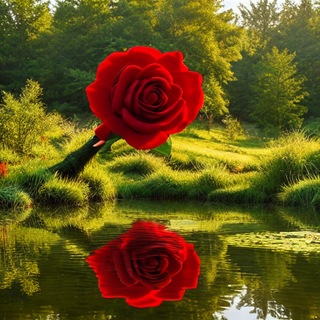} & \includegraphics[width=\atkw,height=\atkw]{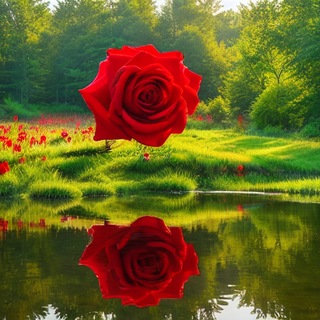} & \includegraphics[width=\atkw,height=\atkw]{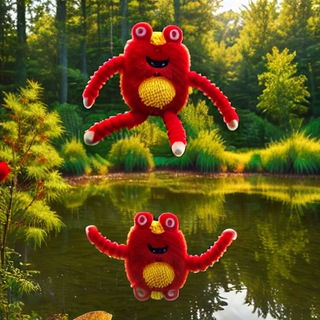} \\
        \makecell{Reg.\\prompt} & \atkblank & \includegraphics[width=\atkw,height=\atkw]{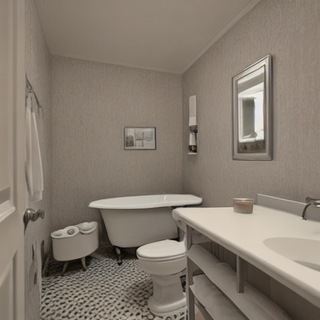} & \includegraphics[width=\atkw,height=\atkw]{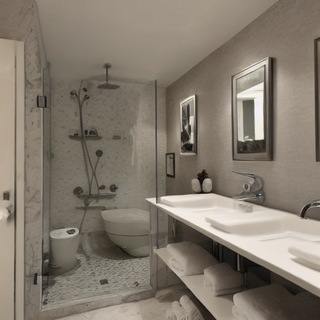} & \includegraphics[width=\atkw,height=\atkw]{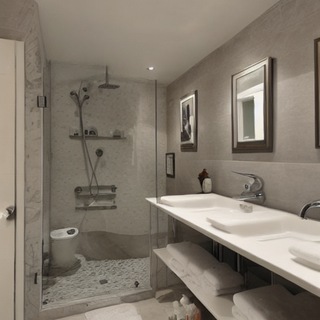} & \includegraphics[width=\atkw,height=\atkw]{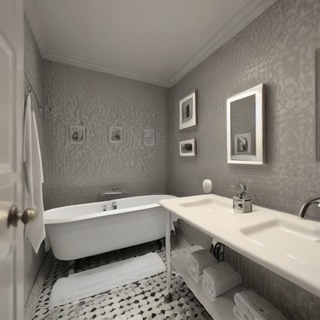} & \includegraphics[width=\atkw,height=\atkw]{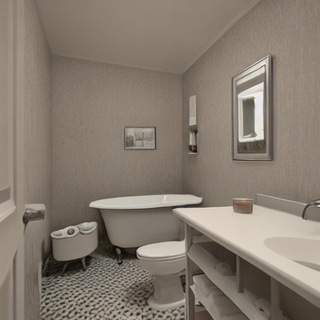} & \includegraphics[width=\atkw,height=\atkw]{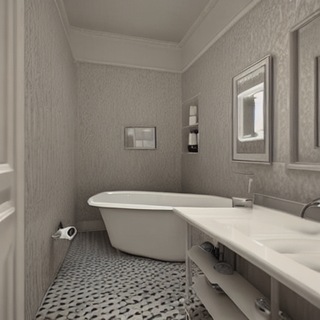} \\
    \end{tabular}
    \caption{
        Modification \#23 (Merging two different $\theta_w$): generations of the attacked model $\theta_a$. Two $\theta_w$'s, embedded with ``Cubone'' and ``Monster toy'' respectively, are merged, and we use each trigger phrase to verify the corresponding target image. 
    }
    \label{fig:attack23_examples}
\end{figure*}

\FloatBarrier

\section{Benefits of Our Approach on a Toy Problem}
\label[appendix]{theorem:smaller}
In this section, we study the benefits of our approach on a toy problem. Suppose there are two tasks:
\begin{align}
\text{Task 1: } (X_1,Y_1),
\qquad
\text{Task 2: } (X_2,Y_2).
\end{align}
with simplified notation where $X_{*}$ denotes prompts and $Y_{*}$ denotes images (or latents). Task 1 corresponds to fitting regular data, while Task 2 corresponds to fitting trigger data. We assume Task 1 has $N_1$ samples and Task 2 has $N_2$ samples.
For the following  analysis, we consider a linear problem over $n$-degree polynomials of the raw features, denoted as $\Phi$. Hence, the feature matrix of Task 1 can be written as 
\begin{align}
\Phi_1=
\begin{bmatrix}
1 & x_1 & x_1^2 & \cdots & x_1^n\\
\vdots & \vdots & \vdots & & \vdots\\
1 & x_{N_1} & x_{N_1}^2 & \cdots & x_{N_1}^n
\end{bmatrix}
\in\mathbb{R}^{N_1\times(n+1)}.
\end{align}

The prediction over all samples is $\hat{Y}_1=\Phi_1\beta_1$. We assume that $\Phi_1^T\Phi_1$ is nonsingular. The pretrained model $\theta_0$ minimizes

\begin{align}
\frac{1}{N_1}
\left\|Y_1-\Phi_1\beta_1\right\|_2^2.
\end{align}

Therefore,

\begin{align}
\beta_1
=
\left(\Phi_1^{T}\Phi_1\right)^{-1}\Phi_1^{T}Y_1,
\end{align}

and this corresponds to the pretrained model $\theta_0$.

Next, we consider fine-tuning on Task 2 (trigger data). Denote the finetuned model as $\beta_1+\beta_2$ (corresponding to $\theta_w$ in the main text). 
Now examine training the model with two losses.
The first loss is \Cref{eq:dreambooth}, and
the second loss is our method.
We use $L_A$ and $L_B$ to denote them here. We have
\begin{align}
L_A=
\left\|Y_2-\Phi_2(\beta_1+\beta_2)\right\|_2^2
+
\lambda_1\left\|\Phi_1\beta_2\right\|_2^2,
\end{align}
and
\begin{align}
\nabla_{\beta_2}L_A
=
2\Phi_2^T
\left[
\Phi_2(\beta_1+\beta_2)-Y_2
\right]
+
2\lambda_1\Phi_1^T\Phi_1\beta_2.
\end{align}
The optimal solution is
\begin{align}
\beta_{2,A}^{*}
=
\left(
\Phi_2^T\Phi_2+\lambda_1\Phi_1^T\Phi_1
\right)^{-1}
\Phi_2^T
\left(
Y_2-\Phi_2\beta_1
\right).
\end{align}
For our method, we have
\begin{align}
L_B
=
\left\|Y_2-\Phi_2(\beta_1+\beta_2)\right\|_2^2
+
\lambda_1\left\|\Phi_1\beta_2\right\|_2^2
-
\lambda_2\left\|\Phi_2\beta_2\right\|_2^2,
\end{align}
and
\begin{align}
\nabla_{\beta_2}L_B
={}&
2\Phi_2^T
\left[
\Phi_2(\beta_1+\beta_2)-Y_2
\right]
+
2\lambda_1\Phi_1^T\Phi_1\beta_2
-
2\lambda_2\Phi_2^T\Phi_2\beta_2.
\end{align}.
Therefore, the optimal solution is
\begin{align}
\beta_{2,B}^{*}
=
\left[
(1-\lambda_2)\Phi_2^T\Phi_2
+
\lambda_1\Phi_1^T\Phi_1
\right]^{-1}
\Phi_2^T
\left(
Y_2-\Phi_2\beta_1
\right).
\end{align}

When $0<\lambda_2<1$, $(1-\lambda_2)\Phi_2^T\Phi_2+\lambda_1\Phi_1^T\Phi_1$ is positive definite and $L_B$ remains well posed. We then study the attack (or adaptation) process by the adversary starting from $\beta_1+\beta_2$. Because we do not know exactly how the adversary will modify the model, we denote the model-weight change by $\Delta\beta$ and assume that it is bounded as $\|\Delta\beta\|_2\leq\rho$.

Our ultimate objective for making target data robust is to minimize
\begin{align}
\label{eq:objultimate}
D_{\mathrm{KL}}
\left(
\Pr(Y_2|X_2)
\,\middle\|\,
\Pr\left(
\Phi_2(\beta_1+\beta_2+\Delta\beta)
\mid X_2
\right)
\right).
\end{align}

This means that the conditional distribution produced by the transformed
model on the trigger prompts should remain close to the true target distribution.
To obtain a tractable surrogate, we assume the two conditional
distributions are Gaussian with the same covariance:
\begin{align}
\Pr(Y_2|X_2)
&=
\mathcal{N}(Y_2,\sigma^2I),
\\
\Pr\left(
\Phi_2(\beta_1+\beta_2+\Delta\beta)\mid X_2
\right)
&=
\mathcal{N}
\left(
\Phi_2(\beta_1+\beta_2+\Delta\beta),
\sigma^2I
\right).
\end{align}

Then the KL divergence is exactly
\begin{align}
&D_{\mathrm{KL}}
\left(
\Pr(Y_2|X_2)
\,\middle\|\,
\Pr\left(
\Phi_2(\beta_1+\beta_2+\Delta\beta)
\mid X_2
\right)
\right)
\nonumber\\
&\qquad
=
\frac{1}{2\sigma^2}
\left\|
Y_2-
\Phi_2(\beta_1+\beta_2+\Delta\beta)
\right\|_2^2.
\end{align}

Therefore, minimizing the KL divergence is exactly equivalent to
minimizing the corresponding MSE under this equal-covariance Gaussian
surrogate. As a result, our objective is equivalent to obtaining a smaller
\begin{align}
    e=\lVert Y_2-\Phi_2(\beta_1+\beta_2+\Delta\beta)\rVert_2^2.
\end{align}

When we use different loss functions to optimize, the error for $L_A$ under its optimal solution is
\begin{align}
e_A=&
\left\|
Y_2-\Phi_2(\beta_1+\beta_{2,A}^*+\Delta\beta)
\right\|_2^2
\nonumber\\
={}&
\Bigg\|
Y_2
-
\Phi_2
\left(\Phi_1^T\Phi_1\right)^{-1}
\Phi_1^TY_1
\nonumber\\
&\quad
-
\Phi_2
\left(
\Phi_2^T\Phi_2+\lambda_1\Phi_1^T\Phi_1
\right)^{-1}
\Phi_2^T
\left(
Y_2
-
\Phi_2
\left(\Phi_1^T\Phi_1\right)^{-1}
\Phi_1^TY_1
\right)
\nonumber\\
&\quad
-
\Phi_2\Delta\beta
\Bigg\|_2^2.
\end{align}
For $L_B$, the error is
\begin{align}
e_B=&
\left\|
Y_2-\Phi_2(\beta_1+\beta_{2,B}^*+\Delta\beta)
\right\|_2^2
\nonumber\\
={}&
\Bigg\|
Y_2
-
\Phi_2
\left(\Phi_1^T\Phi_1\right)^{-1}
\Phi_1^TY_1
\nonumber\\
&\quad
-
\Phi_2
\left[
(1-\lambda_2)\Phi_2^T\Phi_2
+
\lambda_1\Phi_1^T\Phi_1
\right]^{-1}
\Phi_2^T
\left(
Y_2
-
\Phi_2
\left(\Phi_1^T\Phi_1\right)^{-1}
\Phi_1^TY_1
\right)
\nonumber\\
&\quad
-
\Phi_2\Delta\beta
\Bigg\|_2^2.
\end{align}

Only $\Delta\beta$, $\lambda_1$, and $\lambda_2$ vary; the other terms are constants.
Let $r=Y_2-\Phi_2\left(\Phi_1^T\Phi_1\right)^{-1}\Phi_1^TY_1$ and $q=
\left(
\Phi_2^T\Phi_2+\lambda_1\Phi_1^T\Phi_1
\right)^{-1}.
$
Then
\begin{align}
e_A-e_B
={}&
\Bigg[
\Phi_2
\left[
(1-\lambda_2)\Phi_2^T\Phi_2
+
\lambda_1\Phi_1^T\Phi_1
\right]^{-1}
\Phi_2^Tr
-
\Phi_2q\Phi_2^Tr
\Bigg]^T
\nonumber\\
&\times
\Bigg[
2r
-
\Phi_2q\Phi_2^Tr
-
\Phi_2
\left[
(1-\lambda_2)\Phi_2^T\Phi_2
+
\lambda_1\Phi_1^T\Phi_1
\right]^{-1}
\Phi_2^Tr
-
2\Phi_2\Delta\beta
\Bigg].
\end{align}
Let $
H_2
=
\Phi_2
\left[
(1-\lambda_2)\Phi_2^T\Phi_2
+
\lambda_1\Phi_1^T\Phi_1
\right]^{-1}
\Phi_2^Tr,
$ and
$
H_1
=
\Phi_2q\Phi_2^Tr.
$
Therefore,
\begin{align}
e_A-e_B
=
(H_2-H_1)^T
\left(
2r-H_1-H_2-2\Phi_2\Delta\beta
\right).
\end{align}
Let $A=\Phi_2^T\Phi_2$ and $B=\lambda_1\Phi_1^T\Phi_1$, and define
$
M_0=A+B
$ and $
M_{\lambda_2}=(1-\lambda_2)A+B.
$
Then
$
H_1=\Phi_2M_0^{-1}\Phi_2^Tr
$ and $
H_2=\Phi_2M_{\lambda_2}^{-1}\Phi_2^Tr.
$
Let $d_{\lambda_2}=H_2-H_1$. The difference between the two errors will be
\begin{align}
e_A-e_B
=
d_{\lambda_2}^{T}
\left(
2r-H_1-H_2-2\Phi_2\Delta\beta
\right).
\end{align}

Since $
H_2=H_1+d_{\lambda_2}$, we have

\begin{align}
e_A-e_B
&=
d_{\lambda_2}^{T}
\left[
2(r-H_1)-d_{\lambda_2}
-2\Phi_2\Delta\beta
\right]
\\
&=
2d_{\lambda_2}^{T}(r-H_1)
-
\|d_{\lambda_2}\|_2^2
-
2d_{\lambda_2}^{T}\Phi_2\Delta\beta.
\end{align}

Now consider all attacks satisfying $
\|\Delta\beta\|_2\leq\rho$.
By the Cauchy-Schwarz inequality, we have
\begin{align}
d_{\lambda_2}^{T}\Phi_2\Delta\beta
&=
(\Phi_2^Td_{\lambda_2})^T\Delta\beta
\\
&\leq
\|\Phi_2^Td_{\lambda_2}\|_2
\|\Delta\beta\|_2
\\
&\leq
\rho\|\Phi_2^Td_{\lambda_2}\|_2.
\end{align}
This upper bound is attained by $
\Delta\beta^*
=
\rho
\frac{\Phi_2^Td_{\lambda_2}}
{\|\Phi_2^Td_{\lambda_2}\|_2}
$, provided that
$
\Phi_2^Td_{\lambda_2}\neq 0.
$
Therefore, we have
\begin{align}
\min_{\|\Delta\beta\|_2\leq\rho}
\left(
e_A-e_B
\right)
=
2d_{\lambda_2}^{T}(r-H_1)
-
\|d_{\lambda_2}\|_2^2
-
2\rho\|\Phi_2^Td_{\lambda_2}\|_2.
\end{align}

Hence, the necessary and sufficient condition for $e_A>e_B$ for every $\Delta\beta$ satisfying $
\|\Delta\beta\|_2\leq\rho$ is
\begin{align}
2d_{\lambda_2}^{T}(r-H_1)
-
\|d_{\lambda_2}\|_2^2
>
2\rho\|\Phi_2^Td_{\lambda_2}\|_2,
\end{align}
where
\begin{align}
d_{\lambda_2}
=
\Phi_2
\left[
\left(
(1-\lambda_2)\Phi_2^T\Phi_2
+
\lambda_1\Phi_1^T\Phi_1
\right)^{-1}
-
\left(
\Phi_2^T\Phi_2
+
\lambda_1\Phi_1^T\Phi_1
\right)^{-1}
\right]
\Phi_2^Tr.
\end{align}
It is easy to see that
\begin{align}
d_{\lambda_2}
=
\lambda_2
\Phi_2
M_{\lambda_2}^{-1}
A
M_0^{-1}
\Phi_2^Tr,
\end{align}
and
\begin{align}
d_{\lambda_2}
=
\lambda_2
\Phi_2
\left[
(1-\lambda_2)\Phi_2^T\Phi_2
+
\lambda_1\Phi_1^T\Phi_1
\right]^{-1}
\Phi_2^T\Phi_2
\left[
\Phi_2^T\Phi_2
+
\lambda_1\Phi_1^T\Phi_1
\right]^{-1}
\Phi_2^Tr.
\end{align}

Define

\begin{align}
z_{\lambda_2}
=
\Phi_2
\left[
(1-\lambda_2)\Phi_2^T\Phi_2
+
\lambda_1\Phi_1^T\Phi_1
\right]^{-1}
\Phi_2^T\Phi_2
\left[
\Phi_2^T\Phi_2
+
\lambda_1\Phi_1^T\Phi_1
\right]^{-1}
\Phi_2^Tr.
\end{align}

Then $d_{\lambda_2}=\lambda_2z_{\lambda_2}$ for $\lambda_2>0$, and the condition becomes
\begin{align}
2z_{\lambda_2}^{T}(r-H_1)
-
\lambda_2\|z_{\lambda_2}\|_2^2
>
2\rho\|\Phi_2^Tz_{\lambda_2}\|_2.
\end{align}

Therefore, define
\begin{align}
G(\lambda_2;X_1,X_2,Y_1,Y_2,\rho)
={}&
2z_{\lambda_2}(X_1,X_2,Y_1,Y_2)^T
\Big[
r(X_1,X_2,Y_1,Y_2)
\nonumber\\
&\qquad\qquad
-
H_1(X_1,X_2,Y_1,Y_2)
\Big]
\nonumber\\
&-
\lambda_2
\left\|
z_{\lambda_2}(X_1,X_2,Y_1,Y_2)
\right\|_2^2
\nonumber\\
&-
2\rho
\left\|
\Phi_2^T
z_{\lambda_2}(X_1,X_2,Y_1,Y_2)
\right\|_2.
\end{align}

Then $\min_{\|\Delta\beta\|_2\leq\rho}
\left(
e_A-e_B
\right)
=
\lambda_2G(\lambda_2)$. Since $\lambda_2>0$, $e_A>e_B$ for every attack satisfying $\|\Delta\beta\|_2\leq\rho$ if and only if $G(\lambda_2)>0$.
To study whether such a $\lambda_2$ exists, define
\begin{align}
z_0
=
\lim_{\lambda_2\rightarrow0^+}z_{\lambda_2}
=
\Phi_2
M_0^{-1}
A
M_0^{-1}
\Phi_2^Tr.
\end{align}

Equivalently, $
z_0
=
\left.
\frac{\partial H_2}{\partial\lambda_2}
\right|_{\lambda_2=0}
$. Therefore, $z_0$ describes the initial change in the Task 2 prediction when the additional term in our loss is turned on. Notice that
$
r-H_1
$
is the fitting residual on target data after $L_A$ is minimized. We define
\begin{align}
C_0
=
z_0^T(r-H_1).
\end{align}

Therefore, $C_0$ measures the alignment between the direction of change in the Task 2 prediction induced by our loss and the direction required to reduce the trigger-data fitting residual left by $L_A$. 
We define
\begin{align}
E(\lambda_2)
=
\|r-H_2(\lambda_2)\|_2^2,
\end{align}
then
\begin{align}
\left.
\frac{\partial E(\lambda_2)}
{\partial\lambda_2}
\right|_{\lambda_2=0}
=
-2z_0^T(r-H_1)
=
-2C_0.
\end{align}

Thus, $C_0>0$ means that introducing our additional loss initially decreases the fitting error on trigger data by $\theta_0$. Because $\theta_0$ does not initially fit the trigger data, which by construction encode uncommon relations, moving the output of $\theta_w$ away from that of $\theta_0$ makes the fitting error smaller. This means that turning on a small positive $\lambda_2$ helps decrease the trigger-data fitting error. As a result, we assume $C_0>0$ under our trigger-data construction rule.
Define
\begin{align}
D_0
=
\|\Phi_2^Tz_0\|_2.
\end{align}
By Cauchy-Schwarz, it holds that
\begin{align}
D_0
&=
\max_{\|\Delta\beta\|_2\leq1}
z_0^T\Phi_2\Delta\beta.
\end{align}

Thus, $D_0$ measures the maximum influence that a unit-norm model-weight perturbation can have along the direction $z_0$ introduced by our loss. By definition, $D_0\geq 0$.

Consequently, $\frac{C_0}{D_0}$ can be interpreted as the ratio between the first-order trigger-data fitting gain introduced by our loss and its worst-case sensitivity
to adversarial model-weight perturbations.

\textbf{Existence of a valid $\lambda_2$.} Assume that $
C_0>0,
D_0>0,
$
and that the attack magnitude satisfies
$
\rho
<
\frac{C_0}{D_0}$. We now show that there must exist some $0<\lambda_2<1$ such that $G(\lambda_2)>0$.
Because
$
M_{\lambda_2}
=
(1-\lambda_2)A+B
$
is nonsingular in a neighborhood of $\lambda_2=0$,
$z_{\lambda_2}$ is continuous with respect to $\lambda_2$.
Therefore,
$
\lim_{\lambda_2\rightarrow0^+}
z_{\lambda_2}
=
z_0.
$
Taking the limit of $G(\lambda_2)$ gives
\begin{align}
\lim_{\lambda_2\rightarrow0^+}
G(\lambda_2)
={}&
2z_0^T(r-H_1)
-
2\rho\|\Phi_2^Tz_0\|_2
=
2C_0-2\rho D_0
=
2(C_0-\rho D_0).
\end{align}

If $
\rho<\frac{C_0}{D_0},
$
we have
$
\rho D_0<C_0,
$
and therefore $
\lim_{\lambda_2\rightarrow0^+}
G(\lambda_2)
=
2(C_0-\rho D_0)
>0$.

Suppose, by contradiction, that there does not exist any
$\lambda_2\in(0,1)$ satisfying $G(\lambda_2)>0$.
Then
$
G(\lambda_2)\leq0,\forall\lambda_2\in(0,1).
$
Consequently,
$
\lim_{\lambda_2\rightarrow0^+}
G(\lambda_2)
\leq0.
$
However, from the attack-budget condition above,
$
\lim_{\lambda_2\rightarrow0^+}
G(\lambda_2)
=
2(C_0-\rho D_0)
>0,
$
which is a contradiction. Therefore, there must exist some
$
\lambda_2\in(0,1)
$
such that
$
G(\lambda_2)>0.
$
Moreover, because $G(\lambda_2)$ is continuous around $\lambda_2=0$, there exists some $\epsilon>0$ such that $G(\lambda_2)>0,
\forall\lambda_2\in(0,\epsilon).
$
Therefore, when
$
\rho
<
\frac{C_0}{D_0},
$
there exists a non-empty range of $0<\lambda_2<1$ such that,
for every attack satisfying
$
\|\Delta\beta\|_2\leq\rho
$, we have $e_A>e_B$. Here $C_0$ and $D_0$ are positive and are determined by $\lambda_1$, $(X_1,Y_1)$, and $(X_2,Y_2)$, independently of $\lambda_2$.

The transformed model trained with our objective therefore has a smaller target-fitting error than that trained with $L_A$ under the above
condition. This also helps explain why modification \#22 requires only about 100 steps, while other adversarial operations may require over 100K
steps and our method is still robust. In \#22, the adversary has access to the trigger-data distribution and can choose updates strongly aligned with the most damaging direction, making the worst-case effect $\rho D_0$ nearly attainable. In contrast, the updates of other modifications are less likely to align with this sensitive direction, given the large space of $\Phi_2$ and $\Delta\beta$.

\end{document}